\documentclass[10pt,twocolumn,letterpaper]{article}

\usepackage[pagenumbers]{wacv} % To force page numbers, e.g. for an arXiv version

\definecolor{wacvblue}{rgb}{0.21,0.49,0.74}
\usepackage[pagebackref,breaklinks,colorlinks,allcolors=wacvblue]{hyperref}

\usepackage[linesnumbered,ruled,vlined]{algorithm2e}
\SetKwInput{StepOne}{Step 1}
\SetKwInput{StepTwo}{Step 2}
\usepackage{multirow}
\usepackage{comment}

\usepackage{amsthm}
\usepackage{wrapfig}
\usepackage{placeins}
\usepackage{xcolor}%
\usepackage{subcaption}
\usepackage{bbm}
\usepackage{graphicx} % for \resizebox
\usepackage{booktabs} % for \toprule \midrule \bottomrule
\usepackage{makecell} % for \makecell
\usepackage{float}
\usepackage{longtable}
\usepackage{balance}
\newtheorem{proposition}{Proposition}

\def\wacvPaperID{1571} % *** Enter the WACV Paper ID here
\def\confName{WACV}
\def\confYear{2027}

\title{Source-Free Class Relearning: Diagnosing Forgetting in Class Unlearning}

\author{
Zahra Dehghani$^{1,2,3,4}$ \qquad
Pablo Piantanida$^{2,3,4,5}$ \qquad
Mohammadhadi Shateri$^{1,2,3}$ \\%[1pt]
$^{1}$LIVIA, $^{2}$ILLS, $^{3}$ÉTS Montreal,
$^{4}$Mila - Quebec AI Institute,\\
$^{5}$CNRS, CentraleSupélec - Université Paris-Saclay \\%[1pt]
{\tt\small
zahra.dehghani-tafti.1@ens.etsmtl.ca %\quad
}
}

\begin{document}
\maketitle
\enlargethispage{1.1\baselineskip}
\begin{abstract}
    Class unlearning aims to remove a model's ability to recognize designated forget classes while preserving performance on retain classes. However, low forget accuracy after unlearning does not necessarily mean the class structure has been erased. Some approximate unlearning methods can alter classifier decision boundaries while leaving recoverable structure in the feature representation. Prior work has shown that forget classes can be recovered, but existing approaches require real forget or retain samples, auxiliary data, or reference checkpoints. We instead study class relearning in a strictly source-free setting, asking whether a forget class can be recovered through a classifier-head update using only the released unlearned model. Our approach rests on a theoretical analysis establishing a sufficient alignment condition under which a single gradient step on a synthetic probe set increases the expected logit margin of the forget class. Building on this, we propose a white-box \textbf{Source-Free Relearning Audit (SFRA)}%
    \footnote{Code: \url{https://github.com/Yasaman-dt/SFRA}},
    which generates candidate embeddings in representation space and uses model-guided confidence filtering to construct high-confidence retain probes and low-confidence boundary-adjacent probes that are relabelled as the forget class. Gaussian sampling and Softmax confidence are used by default, while ablations with alternative proposal distributions and uncertainty criteria show that recoverability is not specific to these choices. To quantify recoverability, we introduce the Relearning Score~($\mathrm{RS}$), which jointly measures forget-class recovery and retain-accuracy preservation, and report class-matched $\Delta\mathrm{RS}$ relative to a retrained reference. Experiments on CIFAR-10, CIFAR-100, and TinyImageNet with ResNet-18, ViT-B/16, and Swin-T show that several state-of-the-art unlearning methods exhibit substantial source-free recoverability, and that for a subset of methods this recoverability exceeds the matched retrained reference. These results support SFRA as a practical diagnostic of post-unlearning recoverability without claiming reconstruction of the real forget training samples. 
    
\end{abstract}

\begin{figure}
    \centering
    \includegraphics[width=0.9\linewidth]
    {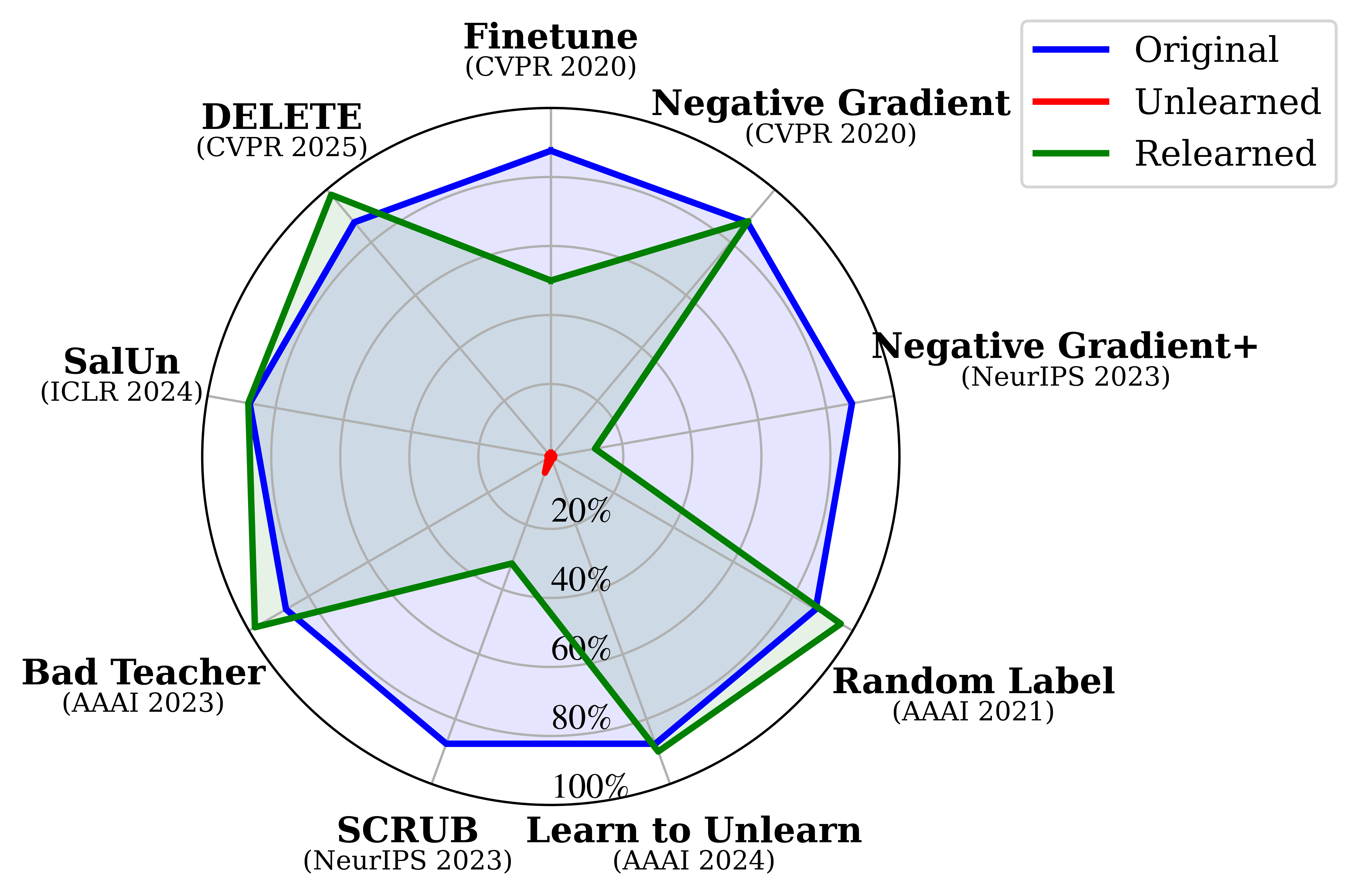}
    \caption{Comparison of our proposed SFRA applied to state-of-the-art unlearning methods on TinyImageNet with a \textbf{ViT-B/16} backbone and class~$160$ as the forget class. Forget accuracy is shown before unlearning (Original), after unlearning (Unlearned), and after SFRA (Relearned). Some methods remain robust to relearning, whereas others exhibit substantial recoverability.}
    \label{fig:motiv}
    \vspace{-0.4cm}
\end{figure}
    
\section{Introduction}
\label{sec:intro}

Class unlearning arises in many practical settings, including face recognition, backdoor defense, data-poisoning mitigation, semantic segmentation, and bias removal~\cite{chen2023boundary,liu2022backdoor,zhou2025decoupled,kurmanji2023towards}. Formally, it requires a model to selectively forget one or more designated classes~\cite{tarun2023fast,chundawat2023can,foster2024fast,kodge2024deep,zhou2025decoupled,zhang2025toward,wang2025efficient,dehghani2026universal,chundawat2023zero,ahmed2025towards}. This is a specific instance of \emph{machine unlearning}~\cite{bourtoule2021machine}—the process of forgetting specified data from a trained model~\cite{cao2015towards,ginart2019making}. Exact unlearning retrains a model from scratch on the retain set~\cite{bourtoule2021machine,graves2021amnesiac}, which is conceptually simple but computationally prohibitive. Hence, practical methods rely on approximate unlearning~\cite{sekhari2021remember,neel2021descent,foster2024fast,golatkar2020eternal,izzo2021approximate}, which aims to remove the influence of the designated data without full retraining~\cite{nguyen2025survey,shaik2024exploring}.

Class unlearning methods may match a retrained model on simple metrics such as forget-set accuracy, but whether they durably erase the effect of forget samples remains unclear~\cite{hayes2025inexact}. In fact, verifying that unlearning has eliminated the forget set’s influence is an open problem in its own right~\cite{triantafillou2024we}. From a privacy standpoint, recent works~\cite{guo2019certified,neel2021descent,chen2021machine,carlini2022privacy} have shown that unlearning can inadvertently expose information about the forget samples. Some approximate unlearning methods can suppress the forget class at the classifier level by altering decision boundaries without fully eliminating class-separable structure in representation space. As a result, a forget class can sometimes be relearned after a small amount of post–unlearning adaptation; accuracy on the forget class rebounds while retain performance stays nearly unchanged. This vulnerability questions how deep forgetting truly is and whether internal features still carry recoverable structure. 

We uncover and quantify this in a strictly source-free setting (Fig.~\ref{fig:motiv}). We generate synthetic probes in the classifier-input representation space and use the classifier head to assign them to class-specific regions. If a recoverable decision structure associated with a forget class remains after unlearning, these probes can be used to identify and reactivate that structure. 
We treat these uncertain samples as boundary probes, since uncertainty concentrates near decision surfaces; relabeling these boundary probes as the forget class provides a source-free supervision signal to test whether a forget boundary can be re-formed without data. These uncertain samples may lie near residual decision boundaries associated with the forget class. By updating the model using only these carefully selected synthetic points, we reveal class relearning without any access to the original data.

Class relearning under these constraints suggests that some state-of-the-art unlearning methods may leave residual recoverable structure associated with the forget class. Rather than proving training-data memorization, relearning indicates that the post-unlearning representation can still support re-separation of the forget class under a lightweight source-free update. To quantify class relearning, we introduce the Relearning Score~($\mathrm{RS}$), which combines retain-stability and forget-recoverability into a compact metric. A high $\mathrm{RS}$ indicates substantial \emph{absolute} source-free recoverability, i.e., the forget class can be re-separated through a classifier-head update while largely preserving retain-class performance.  $\mathrm{RS}$ alone does not distinguish residual structure associated with prior exposure to the forget class from generic transferability of a representation learned without that class. We therefore use a matched retrained reference model as a control and report $\Delta\mathrm{RS}$, the excess recoverability over this reference. Positive $\Delta\mathrm{RS}$ provides stronger evidence that an unlearned model remains more recoverable than a model that never observed the forget class. 

Our main contributions are summarized as follows:
\begin{itemize}
    \item We provide a theoretical motivation by establishing a sufficient alignment condition under which synthetic probs can increase forget-class margin, providing a principled basis for source-free relearning. Building on this result, we introduce a practical Source-Free Relearning Audit (SFRA) to determine whether forget-class structure remains recoverable after class unlearning. Successful relearning indicates that the post-unlearning feature-space geometry retains structure capable of supporting re-separation of the forget class.
    \item We propose Relearning Score ($\mathrm{RS}$), a scalar metric jointly measuring retain stability and forget class recoverability, and pair it with $\Delta\mathrm{RS}$ relative to a matched retrained reference to distinguish absolute recoverability from excess recoverability beyond generic representation transfer.
    \item We conduct an extensive analysis, applying our relearning method to the state-of-the-art machine unlearning methods across three diverse datasets (CIFAR-10, CIFAR-100, TinyImageNet) and three backbone architectures (ResNet-18, ViT-B/16, and Swin-T).
    
\end{itemize}

\section{Related Work}
\label{sec:relatedwork}

\textbf{Class Unlearning.}
Class unlearning aims to remove a designated \textit{forget class} while preserving performance on the remaining \textit{retain classes}. Approximate approaches include Finetune~\cite{golatkar2020eternal}, Negative Gradient~\cite{golatkar2020eternal}, Negative Gradient+~\cite{kurmanji2023towards}, SCRUB~\cite{kurmanji2023towards}, Random Label~\cite{hayase2020selective}, Boundary Shrink~\cite{chen2023boundary}, Learn to Unlearn~\cite{cha2024learning}, Bad Teacher~\cite{chundawat2023can}, SalUn~\cite{fan2023salun}, and DELETE~\cite{zhou2025decoupled}.

\noindent \textbf{Privacy Risks and Auditing in Machine Unlearning.}
Successful unlearning under conventional evaluation metrics does not necessarily eliminate privacy risks, as unlearned models may remain vulnerable to membership inference attacks or provide a false sense of privacy under insufficient evaluation~\cite{chen2021machine,hayes2025inexact}. Moreover, recent work highlights privacy risks beyond the forget set and the role of shared information between forget and retain data in post-unlearning privacy~\cite{fu2026revisiting,ye2025data,carlini2022privacy}. These concerns further motivate stronger post-unlearning auditing, including the recoverability-based perspective considered next.

\noindent \textbf{Post-unlearning Recoverability and Relearning.}
A growing body of research shows that unlearned knowledge may remain dormant and can be reactivated through finetuning, model tampering, or in-context reintroduction~\cite{hu2024jogging,deeb2024unlearning,siddiqui2025dormant,shumailov2024ununlearning}. In diffusion models, unlearned concepts can re-emerge under finetuning, even on unrelated prompts~\cite{george2025illusion}, while in LLMs, finetuning or lightweight modifications to weights or activations can restore removed knowledge, capabilities, or safeguards~\cite{lynch2024eight,deeb2024unlearning,hu2024jogging,lucki2024adversarial,qi2024fine,lermen2023lora,tamirisa2025tamper,che2025model}. 
In vision classifiers, Siddiqui et al.~\cite{siddiqui2025dormant} show that example-level unlearning can be reversed by finetuning solely on retain data, without access to forget samples. Ha et al.~\cite{ha2025unlearning} introduce the Prototypical Relearning Attack (PRA), which uses a few real forget class samples to construct class prototypes and restore forget class classifier weights, together with Spotter, a defense that disperses forget class representations. Inversion-based analyses similarly reconstruct features or infer labels by comparing the original and unlearned checkpoints~\cite{hu2024learn}. These approaches demonstrate important post-unlearning vulnerabilities, but they remain source-dependent or reference-dependent. In contrast, our SFRA constructs synthetic probes using only the released unlearned model, without access to forget, retain, auxiliary, or surrogate data or the original checkpoint.

\noindent \textbf{Representation-level Audits of Unlearning.}
Studies show that output-level unlearning metrics can coexist with recoverable information in model representations~\cite{seo2025revisiting, kim2026we,gao2026an,jeon2026information}. Seo et al.~\cite{seo2025revisiting} show that classifier-level changes can yield favorable unlearning metrics while substantial information remains recoverable from the representation. Kim et al.~\cite{kim2026we} evaluate residual information using logit-gap and representation-similarity measures, while Gao et al.~\cite{gao2026an} show that low forget class accuracy can reflect feature--classifier misalignment and recover the forget class using source-dependent linear probing and nearest-class-center evaluation. Jeon et al.~\cite{jeon2026information} measure residual forget-related information in intermediate layers using mutual information and assess recoverability through head retraining with labeled data. Related approaches use representation-level verification, feature restoration, data-dependent probes, and correlation-based auditing ~\cite{cosma2026ruler,jang2026suppression,yu2026can,pawelczyk2025machine}.

An important distinction from source-dependent relearning and representation-level auditing methods concerns the interpretation of successful recovery. Because these methods use labeled forget class samples to guide the classifier, they may construct a new forget class decision region even when limited recoverable structure remains after unlearning. In contrast, our audit uses no real forget, retain, auxiliary, or surrogate samples, nor the original pre-unlearning checkpoint, during probe construction or classifier-head updating. Consequently, when the relearned classifier correctly recognizes held-out forget samples, the recovery indicates that the post-unlearning representation itself retains structure capable of supporting re-separation of the forget class. This does not establish memorization of the real training samples, but instead provides evidence of source-free recoverability from the released unlearned model.

\section{Methodology}
\label{method}
In this section, we establish notation, define the problem setup, and present our Source-Free Relearning Audit (SFRA). We assume white-box access to the unlearned model and knowledge of the forget class identity, but no access to forget samples, retain samples, auxiliary or surrogate data, or the original pre-unlearning model. Real labeled samples are used only for post-hoc evaluation.

\subsection{Notations and Problem Setup}
We are given a $C$-class classifier that has undergone a class unlearning procedure using an off-the-shelf algorithm. We assume that the released unlearned model either retains the original $C$-dimensional output space or is extended with an output logit for any removed forget class. Formally, let the unlearned model be $\Phi_{un} = h \circ e : \mathcal{X} \to \mathbb{R}^{C}$ where $e : \mathcal{X} \to \mathbb{R}^{d}$ denotes the feature extractor mapping an input data $x \in \mathcal{X}$ to a latent representation 
$z = e(x)\in \mathbb{R}^{d}$, and $h: \mathbb{R}^{d} \to \mathbb{R}^{C}$ is a classification head parameterized by $\theta = (W, b)$ with $h(z) = W z + b \in \mathbb{R}^{C}$ where $\quad W \in \mathbb{R}^{C \times d}$ and $b \in \mathbb{R}^{C}$. The label space $\mathcal{Y}= \{1, \ldots, C\}$ is partitioned into two disjoint subsets including retain classes $\mathcal{Y}_r$ and forget classes $\mathcal{Y}_f$ with $\mathcal{Y} = \mathcal{Y}_r \cup \mathcal{Y}_f$. Given the unlearned classifier $\Phi_{un}$, we investigate whether it still contains recoverable structure corresponding to the forget classes. Without loss of generality, we focus on the single-class unlearning
case, with $\mathcal{Y}_f=\{c_f\}$ and $\mathcal{Y}_r = \mathcal{Y} \setminus \{c_f\}$. We describe the extension to multi-class in App.~\ref{app:additional_results}. We further examine a restrictive setting where the forget class output weight is unavailable and show that SFRA remains effective in App.~\ref{app:forget_row_ablation}.%Our formulation assumes that the released model preserves an output logit for each designated forget class. 

%\subsection{Proposed Methodology}
\subsection{Source-Free Relearning Audit (SFRA)}
For analytical clarity, we first describe relearning by updating the classifier head $h: \mathbb{R}^{d} \to \mathbb{R}^{C}$, while keeping the feature extractor $e: \mathcal{X} \to \mathbb{R}^{d}$ fixed. Our goal is to probe whether the unlearned model $\Phi_{un} = h \circ e: \mathcal{X} \to \mathbb{R}^{C}$ still encodes recoverable information about the forget class $c_f$ in its feature space, even when no source data are available. To this end, we construct two sets of synthetic probs in the classifier’s input space $\mathbb{R}^d$, including the synthetic retain set \(\mathcal{S}_r\) and the synthetic forget set \(\mathcal{S}_f\). Each synthetic prob $s\in\mathbb{R}^d$ is sampled at random and then labeled according to the classifier head's prediction. Specifically,
we compute
$
p(s)=\operatorname{softmax}(h(s)),
\hat{y}(s)=\arg\max_{c\in\mathcal{Y}}p_c(s),
$
where $p_c(s)$ denotes the predicted probability of class $c$. 
The retain set \(\mathcal{S}_r\) contains embeddings confidently assigned to retain classes, while the forget set $\mathcal{S}_f$ contains embeddings associated with low-confidence predictions. These embeddings are generated purely in feature space—without any access to the original training data or any surrogate dataset. The complete procedure for constructing $\mathcal{S}_r$ and $\mathcal{S}_f$ is provided in Alg.~\ref{alg:alg1}. In the original model, samples from the forget class $c_f$ typically occupy a characteristic region in the representation space and are mapped to $c_f$ by the classifier head. Recent analyses suggest that several approximate unlearning methods can exhibit superficial forgetting, achieved primarily by altering the final classifier or decision boundary (suppressing $c_f$ logits), while leaving intermediate representations largely similar to the original model~\cite{kim2026we,lee2026erase,le2026pour,gao2026an}. In this case, residual geometric structure for $c_f$ may persist in the embedding space even if the current head assigns nearby directions to retain classes. Our low-confidence selection used to form $\mathcal{S}_f$ is therefore used as a boundary probe, not as an estimator of the forget class data distribution. In discriminative classifiers, low confidence tends to concentrate near decision boundaries; hence, mining low-confidence synthetic probs targets boundary-adjacent regions that are particularly informative about the current separating surface. Importantly, we do not claim these probes are more likely to be true forget embeddings than any other class. Instead, we relabel them as $c_f$ to test whether the frozen post-unlearning representation retains sufficient structure to support forget class recovery through a lightweight head update.

\noindent\textbf{Definition of Source-Free Class Relearning.}
Let $\Phi_{un}$ be a model after unlearning class $c_f$, with retain and forget accuracies $\mathcal{A}^{un}_r$ and $\mathcal{A}^{un}_f$. A source-free update produces $\Phi_{re}$ using only $\Phi_{un}$ and $c_f$, without real forget, retain, auxiliary, or surrogate data, or a pre-unlearning checkpoint. For $\epsilon_r\geq0$ and $\delta_f>0$, the update achieves $(\epsilon_r,\delta_f)$-source-free class relearning if
$\mathcal{A}^{re}_f-\mathcal{A}^{un}_f\geq\delta_f$ and
$\mathcal{A}^{un}_r-\mathcal{A}^{re}_r\leq\epsilon_r$.
Thus, forget accuracy increases by at least $\delta_f$, while retain accuracy decreases by at most $\epsilon_r$. Rather than fixing thresholds, we quantify source-free relearnability using the Relearning Score ($\mathrm{RS}$) introduced in Sec.~\ref{sec:experimental_setup}.

\noindent\textbf{Analysis of Forget-Class Margin Improvement.}
To analyze source-free improvement of $c_f$ recognition, let
$z\sim\mathcal{E}_f$ be a real forget-class embedding and, for
$j\in\mathcal{Y}_r$, define
\begin{equation}
m_{c_f,j}(z) = (w_{c_f} - w_j)^{\top} z,
\label{eq:margin}
\end{equation}
where $w_c$ denotes the $c$-th row of $W$, and for simplicity, the term associated with the biases is ignored. An increase in $m_{c_f,j}(z)$ after an update indicates an initial movement toward improved separation of $c_f$ from class $j$; however, it does not by itself guarantee successful final relearning.

\begin{proposition}[Sufficient condition for expected margin increase]
\label{prop:margin_increase}
Let $\mathcal{E}_f$ denote the unknown distribution of real embeddings from the forget class $c_f$, with mean $\mu_{\mathcal{E}_f}$ and $\mathcal{S}_f$ be a synthetic forget set constructed from an arbitrary distribution, with mean $\mu_{\mathcal{S}_f}$. We emphasize that $\mathcal{S}_f$ is not required to approximate the full distribution of the real forget samples. Consider the contribution of the synthetic forget set $\mathcal{S}_f$ to a single classifier-head update. For any retain class $j \in \mathcal{Y}_r$, the expected pairwise logit margin of the forget class on real forget embeddings increases if
\begin{equation}
    \mu_{\mathcal{S}_f}^{\top}\mu_{\mathcal{E}_f} + r_j > 0,
\end{equation}
where $r_j$ denotes the coefficient-approximation residual defined in App.~\ref{app:proof}. 
A sufficient condition is
\begin{equation}
    \mu_{\mathcal{S}_f}^{\top}\mu_{\mathcal{E}_f} > |r_j|.
\end{equation}
When $r_j \approx 0$, this condition reduces to positive alignment between synthetic and real forget class mean embeddings. We empirically assess this approximation in App.~\ref{app:margin_approximation}, where the approximate mean-alignment expression shows strong aggregate agreement with the corresponding exact weighted expression in the evaluated setting.
\end{proposition}
A complete proof of Prop.~\ref{prop:margin_increase} is provided in App.~\ref{app:proof}.  
We stress that Prop.~\ref{prop:margin_increase} is deliberately local: it characterizes the contribution of the synthetic-forget loss to one classifier-head gradient step. The complete SFRA update also contains the synthetic-retain loss and is optimized iteratively; accordingly, the proposition supplies a sufficient geometric mechanism, not a necessary-and-sufficient theory of final $\mathrm{RS}$ . Prop.~\ref{prop:margin_increase} shows that the contribution of the synthetic forget set to the expected margin change is governed by the alignment between the induced synthetic update direction and the residual real forget class representation. Importantly, the synthetic forget set is not required to reproduce the real forget class distribution. Rather, relearning is supported when the update induced by the selected synthetic probs is sufficiently aligned with the residual forget class representation. In particular, a sufficient condition for increasing the expected forget class margin is that the synthetic forget-set mean is positively aligned with the real forget class mean by an amount that dominates the approximation residual. Conversely, non-positive alignment does not imply that relearning is impossible; it only means that this sufficient guarantee no longer holds. Thus, the theoretical result is not tied to a specific proposal distribution, such as Gaussian or uniform sampling, but instead depends on the geometry of the synthetic set obtained after model-guided filtering. Synthetic--real alignment across unlearning methods and the retrained reference is analyzed in App.~\ref{app:alignment_analysis}.

\noindent\textbf{Construction of Synthetic Retain and Forget Probes.}
Because real forget embeddings are unavailable in the source-free setting, the alignment condition cannot be evaluated directly. Alg.~\ref{alg:alg1} therefore uses the unlearned classifier to implicitly search for useful update directions. Specifically, embeddings assigned to retain classes with low confidence are selected as boundary-adjacent synthetic forget probes, whereas high-confidence embeddings form the synthetic retain set $\mathcal{S}_r$ and help limit degradation in retain-class performance. The low-confidence probes are not assumed to approximate the real forget class distribution; rather, they provide a practical source-free surrogate for testing whether the forget class decision region can be reconstructed from the residual geometry of the unlearned model. We empirically examine this confidence pattern in App.~\ref{app:forget_assignment_confidence}, %\ref{app:forget_assignment_confidence},
where real forget class samples assigned to retain classes receive substantially lower confidence than correctly classified retain samples. This observation is consistent with our use of low-confidence synthetic assignments as boundary probes.

The Gaussian proposal is not intended to approximate the support of true classifier-input features. In particular, for architectures whose classifier-input representation is constrained---for example, non-negative post-ReLU features---samples from $\mathcal{N}(0,I_d)$ are generally off-manifold. SFRA uses this distribution only as a broad source of candidate directions; the released classifier subsequently performs model-guided filtering and selects the boundary-adjacent probes used for the update. Although Softmax confidence and Gaussian sampling are the defaults, alternative uncertainty criteria and proposal distributions preserve recoverability trends (Apps.~\ref{app:uncertainty_metric} and~\ref{app:distribution_ablation}), indicating that model-guided probe selection, rather than either choice, is key to the audit.

\begin{algorithm}[t]
\caption{Single-Class SFRA}
\label{alg:alg1}

\KwIn{
Unlearned classifier $\Phi_{un}=h\circ e$;
retain-set $\mathcal{Y}_r$;
forget-set $\mathcal{Y}_f=\{c_f\}$;
accepted probs per retain class $N$;
selected probs $M$, where $2M\leq N$;
relearning loss $\mathcal{L}_{\mathrm{re}}$;
number of relearning steps $T$;
learning rate $\eta$.
}
\KwOut{Relearned classifier $\Phi_{re}=h'\circ e$.}

\BlankLine
\StepOne{Synthetic Prob Generation}

Initialize retain set $\mathcal{S}_r\gets\emptyset$
and forget set $\mathcal{S}_f\gets\emptyset$\;

\For{each retain class $k\in\mathcal{Y}_r$}{
    Initialize the candidate pool
    $\mathcal{P}_k\gets\emptyset$\;

    \While{$|\mathcal{P}_k|<N$}{
        Sample an embedding
        $s\sim\mathcal{N}(0,I_d)$\;

        Compute class probabilities
        $p(s)=\operatorname{softmax}(h(s))$\;

        \If{$\arg\max_{c\in\mathcal{Y}}p_c(s)=k$}{
            Append $(s,p_k(s))$ to $\mathcal{P}_k$\;
        }
    }

    Sort $\mathcal{P}_k$ in descending order of $p_k(s)$\;

    \textbf{Retain selection:}
    append the $M$ probs with the highest $p_k(s)$
    to $\mathcal{S}_r$, with label $k$\;

    \textbf{Forget selection:}
    append the $M$ probs with the lowest $p_k(s)$
    to $\mathcal{S}_f$, relabeled as $c_f$\;
}

\BlankLine
\StepTwo{Class Relearning}

Combine sets
$\mathcal{S}=\mathcal{S}_r\cup\mathcal{S}_f$\;

Freeze the feature extractor $e$
and update only the classifier head $h$\;

\For{$t=1,\ldots,T$}{
    Sample a mini-batch
    $\mathcal{B}\subseteq\mathcal{S}$\;

    Compute
    $\mathcal{L}_{\mathrm{re}}
    =-\frac{1}{|\mathcal{B}|}
    \sum_{(s,y)\in\mathcal{B}}
    \log p_y(s;\theta)$\;
    
    %Update $h$ by minimizing    $\mathcal{L}_{\mathrm{re}}$ on $\mathcal{B}$;
    Update
    $\theta\gets\theta-\eta
    \nabla_\theta\mathcal{L}_{\mathrm{re}}$\;
}

\Return{relearned model $\Phi_{re}=h'\circ e$}\;

\end{algorithm}

\vspace{-0.2cm}

\section{Experiments}
\label{sec:experiments}

\subsection{Experimental Setup}
\label{sec:experimental_setup}

\textbf{Models.}
We evaluate ResNet-18~\cite{he2016deep},
ViT-B/16~\cite{dosovitskiy2021an}, and Swin-T~\cite{liu2021swin}, with the latter two reported in App.~\ref{app:additional_results}.

\noindent\textbf{Datasets.} We evaluate on CIFAR-10 \cite{krizhevsky2009learning}, CIFAR-100 \cite{krizhevsky2009learning}, and TinyImageNet \cite{le2015tiny}. CIFAR-10 and CIFAR-100 contain $60K$ images at $32{\times}32$, with $50K$ training and $10K$ test examples, spanning $10$ and $100$ categories, respectively. TinyImageNet has $110K$ images at $64{\times}64$ across $200$ classes, split into $100K$ training and $10K$ test samples.

\noindent\textbf{Baselines.}
We evaluate our proposed Source-Free Relearning Audit (SFRA) on ten class unlearning methods: Finetune \cite{golatkar2020eternal}, Negative Gradient \cite{golatkar2020eternal}, Negative Gradient+ \cite{kurmanji2023towards}, Random Label \cite{hayase2020selective}, Boundary Shrink \cite{chen2023boundary}, Learn to Unlearn \cite{cha2024learning}, SCRUB \cite{kurmanji2023towards}, Bad Teacher \cite{chundawat2023can}, SalUn \cite{fan2023salun}, and DELETE \cite{zhou2025decoupled}. Brief descriptions of them are provided in App.~\ref{app:additional_results}.
For comparison with SOTA, we include the Prototypical Relearning Attack (PRA)~\cite{ha2025unlearning}, which uses five real forget class samples to construct a class prototype and restore the forget class classifier. To ensure a fair comparison, evaluations are done using the same unlearned checkpoints and forget class splits. In contrast to the PRA as a source-dependent diagnostic method, our source-free audit requires no real forget, retain, auxiliary, or surrogate data.

\begin{table*}[t]
\centering
\caption{Comparison of unlearning methods using our proposed SFRA and the source-dependent PRA baseline on ResNet-18 models under single-class unlearning across three datasets. For all model variants, retain accuracy $\mathcal{A}^t_r$ is reported as the mean $\pm$ standard deviation across forget classes, while forget accuracy $\mathcal{A}^t_f$ is reported as $(\min,\mathrm{mean},\max)$. $\mathrm{RS}$ and $\Delta\mathrm{RS}$ are independently reported as maxima across forget classes. Within each dataset, the highest and second-highest $\mathrm{RS}$ and $\Delta\mathrm{RS}$ values are shown in \textbf{bold} and \underline{underlined}, respectively.}
\label{tab:ResNet-18_single_class_all_datasets}
\vspace{-0.3cm}
\resizebox{\textwidth}{!}{%
\begin{tabular}{c|c|cccc|cccc|cccc}
\toprule
\toprule
\multirow{2}{*}{Unlearning Method} & \multirow{2}{*}{Model Variant} & \multicolumn{4}{c}{\textbf{CIFAR-10}} & \multicolumn{4}{|c|}{\textbf{CIFAR-100}} & \multicolumn{4}{c}{\textbf{TinyImageNet}} \\
 &  & $\mathcal{A}^{t}_{r}(\%)$ & $\mathcal{A}^{t}_{f}(\%)$ & $\mathrm{RS}$ & $\Delta\mathrm{RS}$ & $\mathcal{A}^{t}_{r}(\%)$ & $\mathcal{A}^{t}_{f}(\%)$ & $\mathrm{RS}$ & $\Delta\mathrm{RS}$ & $\mathcal{A}^{t}_{r}(\%)$ & $\mathcal{A}^{t}_{f}(\%)$ & $\mathrm{RS}$ & $\Delta\mathrm{RS}$\\
\midrule
\midrule
Original & Original & $94.65{\scriptstyle\,\pm\,0.31}$ & $(89.00,94.65,97.90)$ & - & - & $79.96{\scriptstyle\,\pm\,0.09}$ & $(63.00,79.60,92.00)$ & - & - & $71.40{\scriptstyle\,\pm\,0.07}$ & $(54.00,70.20,88.00)$ & - & - \\
\midrule
\midrule
\multirow{3}{*}{Retrained} & Unlearned & $95.19{\scriptstyle\,\pm\,0.58}$ & $(0.00,0.00,0.00)$ &  &  & $80.19{\scriptstyle\,\pm\,0.24}$ & $(0.00,0.00,0.00)$ &  &  & $70.83{\scriptstyle\,\pm\,0.53}$ & $(0.00,0.00,0.00)$ &  &  \\
 & PRA \cite{ha2025unlearning} & $95.05{\scriptstyle\,\pm\,0.60}$ & $(2.90,10.51,17.80)$ & $0.30$ & - & $78.43{\scriptstyle\,\pm\,0.86}$ & $(21.00,39.60,69.00)$ & $0.81$ & - & $71.26{\scriptstyle\,\pm\,0.67}$ & $(0.00,4.20,42.00)$ & $0.59$ & - \\
 & SFRA (ours) & $92.42{\scriptstyle\,\pm\,0.87}$ & $(12.20,32.72,54.10)$ & $0.69$ & - & $72.57{\scriptstyle\,\pm\,0.40}$ & $(26.00,46.70,80.00)$ & $0.86$ & - & $64.83{\scriptstyle\,\pm\,0.47}$ & $(22.00,39.40,50.00)$ & $0.65$ & - \\
\midrule
\multirow{3}{*}{Finetune \cite{golatkar2020eternal}} & Unlearned & $94.73{\scriptstyle\,\pm\,0.60}$ & $(0.00,0.00,0.00)$ &  &  & $79.13{\scriptstyle\,\pm\,1.03}$ & $(0.00,0.10,1.00)$ &  &  & $66.65{\scriptstyle\,\pm\,0.81}$ & $(0.00,0.00,0.00)$ &  &  \\
 & PRA \cite{ha2025unlearning} & $94.68{\scriptstyle\,\pm\,0.59}$ & $(0.70,4.76,8.10)$ & $0.15$ & $-0.01$ & $78.36{\scriptstyle\,\pm\,1.13}$ & $(10.00,26.30,49.00)$ & $0.66$ & $+0.08$ & $63.98{\scriptstyle\,\pm\,1.09}$ & $(12.00,47.60,92.00)$ & $\underline{0.93}$ & $\underline{+0.93}$ \\
 & SFRA (ours) & $93.26{\scriptstyle\,\pm\,0.82}$ & $(12.90,31.98,54.20)$ & $0.70$ & $+0.18$ & $76.71{\scriptstyle\,\pm\,1.82}$ & $(10.00,33.50,67.00)$ & $0.80$ & $+0.25$ & $60.84{\scriptstyle\,\pm\,0.92}$ & $(22.00,37.80,64.00)$ & $0.76$ & $+0.14$ \\
\midrule
\multirow{3}{*}{Negative Gradient \cite{golatkar2020eternal}} & Unlearned & $90.12{\scriptstyle\,\pm\,1.50}$ & $(5.30,7.52,11.40)$ &  &  & $72.63{\scriptstyle\,\pm\,2.39}$ & $(0.00,1.40,8.00)$ &  &  & $66.35{\scriptstyle\,\pm\,1.79}$ & $(0.00,0.40,2.00)$ &  &  \\
 & PRA \cite{ha2025unlearning} & $89.55{\scriptstyle\,\pm\,1.64}$ & $(16.40,24.95,35.50)$ & $0.45$ & $+0.39$ & $71.79{\scriptstyle\,\pm\,2.10}$ & $(22.00,41.30,65.00)$ & $0.78$ & $+0.11$ & $65.96{\scriptstyle\,\pm\,1.72}$ & $(36.00,50.60,72.00)$ & $0.83$ & $+0.82$ \\
 & SFRA (ours) & $89.20{\scriptstyle\,\pm\,1.84}$ & $(47.10,56.76,68.40)$ & $0.76$ & $+0.55$ & $67.14{\scriptstyle\,\pm\,2.51}$ & $(51.00,64.90,82.00)$ & $0.88$ & $+0.33$ & $60.09{\scriptstyle\,\pm\,1.55}$ & $(4.00,33.60,66.00)$ & $0.77$ & $+0.18$ \\
\midrule
\multirow{3}{*}{Negative Gradient+ \cite{kurmanji2023towards}} & Unlearned & $88.56{\scriptstyle\,\pm\,1.61}$ & $(0.00,0.01,0.10)$ &  &  & $74.33{\scriptstyle\,\pm\,3.27}$ & $(0.00,0.20,2.00)$ &  &  & $69.67{\scriptstyle\,\pm\,0.95}$ & $(0.00,1.20,6.00)$ &  &  \\
 & PRA \cite{ha2025unlearning} & $88.54{\scriptstyle\,\pm\,1.63}$ & $(0.20,1.80,5.00)$ & $0.10$ & $-0.03$ & $73.26{\scriptstyle\,\pm\,3.05}$ & $(41.00,52.10,72.00)$ & $0.83$ & $+0.26$ & $69.55{\scriptstyle\,\pm\,0.88}$ & $(0.00,37.00,78.00)$ & $0.86$ & $+0.86$ \\
 & SFRA (ours) & $86.32{\scriptstyle\,\pm\,2.41}$ & $(11.00,23.07,32.60)$ & $0.49$ & $+0.11$ & $67.72{\scriptstyle\,\pm\,3.22}$ & $(52.00,69.30,80.00)$ & $0.86$ & $+0.43$ & $64.82{\scriptstyle\,\pm\,1.38}$ & $(0.00,30.60,50.00)$ & $0.65$ & $+0.09$ \\
\midrule
\multirow{3}{*}{Random Label \cite{hayase2020selective}} & Unlearned & $92.02{\scriptstyle\,\pm\,1.10}$ & $(8.70,12.84,16.70)$ &  &  & $69.45{\scriptstyle\,\pm\,5.65}$ & $(0.00,2.70,5.00)$ &  &  & $67.25{\scriptstyle\,\pm\,1.62}$ & $(0.00,0.80,2.00)$ &  &  \\
 & PRA \cite{ha2025unlearning} & $91.43{\scriptstyle\,\pm\,1.32}$ & $(33.80,41.54,54.70)$ & $0.58$ & $+0.52$ & $68.44{\scriptstyle\,\pm\,5.22}$ & $(39.00,58.20,77.00)$ & $0.84$ & $+0.35$ & $66.69{\scriptstyle\,\pm\,1.59}$ & $(38.00,59.20,90.00)$ & $\underline{0.93}$ & $\underline{+0.93}$ \\
 & SFRA (ours) & $91.08{\scriptstyle\,\pm\,1.23}$ & $(63.60,71.57,80.40)$ & $0.80$ & $+0.58$ & $63.39{\scriptstyle\,\pm\,5.50}$ & $(54.00,81.90,92.00)$ & $0.92$ & $+0.51$ & $61.45{\scriptstyle\,\pm\,1.76}$ & $(22.00,49.60,74.00)$ & $0.82$ & $+0.25$ \\
\midrule
\multirow{3}{*}{Boundary Shrink \cite{chen2023boundary}} & Unlearned & $92.03{\scriptstyle\,\pm\,1.18}$ & $(8.50,12.98,16.90)$ &  &  & $69.31{\scriptstyle\,\pm\,5.57}$ & $(0.00,2.60,5.00)$ &  &  & $63.33{\scriptstyle\,\pm\,2.61}$ & $(0.00,3.40,10.00)$ &  &  \\
 & PRA \cite{ha2025unlearning} & $91.45{\scriptstyle\,\pm\,1.41}$ & $(33.10,40.85,55.10)$ & $0.58$ & $+0.52$ & $68.28{\scriptstyle\,\pm\,5.13}$ & $(39.00,58.20,76.00)$ & $0.84$ & $+0.37$ & $62.73{\scriptstyle\,\pm\,2.62}$ & $(18.00,48.20,76.00)$ & $0.79$ & $+0.79$ \\
 & SFRA (ours) & $91.10{\scriptstyle\,\pm\,1.57}$ & $(60.30,70.48,80.90)$ & $0.80$ & $+0.58$ & $63.48{\scriptstyle\,\pm\,5.37}$ & $(55.00,82.00,91.00)$ & $0.92$ & $+0.51$ & $58.19{\scriptstyle\,\pm\,2.02}$ & $(12.00,24.80,54.00)$ & $0.62$ & $+0.02$ \\
\midrule
\multirow{3}{*}{Learn to Unlearn \cite{cha2024learning}} & Unlearned & $90.34{\scriptstyle\,\pm\,1.44}$ & $(5.70,8.70,13.60)$ &  &  & $73.45{\scriptstyle\,\pm\,1.50}$ & $(0.00,0.30,3.00)$ &  &  & $66.83{\scriptstyle\,\pm\,1.39}$ & $(0.00,0.80,2.00)$ &  &  \\
 & PRA \cite{ha2025unlearning} & $89.90{\scriptstyle\,\pm\,1.54}$ & $(15.80,24.78,31.60)$ & $0.35$ & $+0.23$ & $72.14{\scriptstyle\,\pm\,1.38}$ & $(52.00,64.00,73.00)$ & $0.84$ & $+0.41$ & $66.45{\scriptstyle\,\pm\,1.37}$ & $(36.00,54.20,72.00)$ & $0.84$ & $+0.81$ \\
 & SFRA (ours) & $89.39{\scriptstyle\,\pm\,1.56}$ & $(44.40,58.87,63.70)$ & $0.71$ & $+0.47$ & $66.97{\scriptstyle\,\pm\,1.35}$ & $(41.00,77.70,94.00)$ & $0.95$ & $+0.54$ & $60.54{\scriptstyle\,\pm\,1.44}$ & $(10.00,38.00,66.00)$ & $0.77$ & $+0.15$ \\
\midrule
\multirow{3}{*}{SCRUB \cite{kurmanji2023towards}} & Unlearned & $94.30{\scriptstyle\,\pm\,0.67}$ & $(0.00,0.00,0.00)$ &  &  & $71.13{\scriptstyle\,\pm\,5.16}$ & $(0.00,1.00,3.00)$ &  &  & $67.73{\scriptstyle\,\pm\,1.29}$ & $(0.00,0.00,0.00)$ &  &  \\
 & PRA \cite{ha2025unlearning} & $94.30{\scriptstyle\,\pm\,0.66}$ & $(0.00,1.39,6.60)$ & $0.12$ & $-0.06$ & $70.21{\scriptstyle\,\pm\,4.81}$ & $(31.00,47.20,73.00)$ & $0.83$ & $+0.27$ & $67.54{\scriptstyle\,\pm\,1.28}$ & $(38.00,59.20,88.00)$ & $\underline{0.93}$ & $+0.89$ \\
 & SFRA (ours) & $91.05{\scriptstyle\,\pm\,1.31}$ & $(20.50,33.18,47.10)$ & $0.64$ & $+0.12$ & $65.12{\scriptstyle\,\pm\,5.00}$ & $(57.00,75.00,87.00)$ & $0.89$ & $+0.43$ & $61.84{\scriptstyle\,\pm\,1.36}$ & $(12.00,35.80,80.00)$ & $0.86$ & $+0.28$ \\
\midrule
\multirow{3}{*}{Bad Teacher \cite{chundawat2023can}} & Unlearned & $92.36{\scriptstyle\,\pm\,5.40}$ & $(0.00,1.13,10.30)$ &  &  & $79.80{\scriptstyle\,\pm\,0.13}$ & $(0.00,0.00,0.00)$ &  &  & $71.14{\scriptstyle\,\pm\,0.14}$ & $(0.00,3.40,18.00)$ &  &  \\
 & PRA \cite{ha2025unlearning} & $92.25{\scriptstyle\,\pm\,5.37}$ & $(3.50,70.86,95.80)$ & \textbf{\boldmath $0.98$} & $\mathbf{+0.92}$ & $77.73{\scriptstyle\,\pm\,1.51}$ & $(75.00,88.30,100.00)$ & \textbf{\boldmath $1.00$} & $\mathbf{+0.59}$ & $70.31{\scriptstyle\,\pm\,0.38}$ & $(70.00,86.20,98.00)$ & \textbf{\boldmath $0.99$} & $\mathbf{+0.99}$ \\
 & SFRA (ours) & $88.15{\scriptstyle\,\pm\,5.67}$ & $(76.90,96.24,99.80)$ & \textbf{\boldmath $0.98$} & $+0.77$ & $73.74{\scriptstyle\,\pm\,1.64}$ & $(93.00,97.10,100.00)$ & $\underline{0.99}$ & $\underline{+0.58}$ & $65.37{\scriptstyle\,\pm\,1.73}$ & $(0.00,44.20,80.00)$ & $0.86$ & $+0.41$ \\
\midrule
\multirow{3}{*}{SalUn \cite{fan2023salun}} & Unlearned & $94.21{\scriptstyle\,\pm\,0.69}$ & $(5.50,8.41,14.50)$ &  &  & $77.17{\scriptstyle\,\pm\,0.54}$ & $(0.00,0.00,0.00)$ &  &  & $69.86{\scriptstyle\,\pm\,0.41}$ & $(0.00,0.00,0.00)$ &  &  \\
 & PRA \cite{ha2025unlearning} & $93.53{\scriptstyle\,\pm\,0.69}$ & $(24.00,44.35,66.20)$ & $0.68$ & $+0.43$ & $73.43{\scriptstyle\,\pm\,1.07}$ & $(15.00,41.10,72.00)$ & $0.82$ & $+0.11$ & $69.65{\scriptstyle\,\pm\,0.49}$ & $(40.00,57.80,82.00)$ & $0.90$ & $+0.90$ \\
 & SFRA (ours) & $89.56{\scriptstyle\,\pm\,0.69}$ & $(69.10,80.92,89.40)$ & $0.87$ & $+0.55$ & $71.53{\scriptstyle\,\pm\,1.28}$ & $(18.00,35.70,53.00)$ & $0.68$ & $+0.08$ & $66.35{\scriptstyle\,\pm\,1.00}$ & $(6.00,22.80,52.00)$ & $0.68$ & $+0.14$ \\
\midrule
\multirow{3}{*}{DELETE \cite{zhou2025decoupled}} & Unlearned & $94.97{\scriptstyle\,\pm\,0.59}$ & $(0.00,0.00,0.00)$ &  &  & $77.28{\scriptstyle\,\pm\,1.81}$ & $(0.00,0.00,0.00)$ &  &  & $69.28{\scriptstyle\,\pm\,1.84}$ & $(0.00,0.20,2.00)$ &  &  \\
 & PRA \cite{ha2025unlearning} & $92.60{\scriptstyle\,\pm\,1.54}$ & $(59.10,75.94,86.60)$ & $0.92$ & $\underline{+0.85}$ & $73.53{\scriptstyle\,\pm\,1.62}$ & $(19.00,55.00,85.00)$ & $0.89$ & $+0.36$ & $68.74{\scriptstyle\,\pm\,1.78}$ & $(24.00,48.40,76.00)$ & $0.86$ & $+0.81$ \\
 & SFRA (ours) & $90.82{\scriptstyle\,\pm\,0.91}$ & $(89.00,95.19,97.70)$ & $\underline{0.97}$ & $+0.75$ & $70.05{\scriptstyle\,\pm\,1.36}$ & $(33.00,61.40,86.00)$ & $0.89$ & $+0.37$ & $63.37{\scriptstyle\,\pm\,1.59}$ & $(8.00,33.80,80.00)$ & $0.86$ & $+0.21$ \\
\midrule
\bottomrule
\end{tabular}
}
\vspace{-0.3cm}
\end{table*}

\noindent\textbf{Evaluation Metrics.}
We evaluate class relearning using retain test accuracy ($\mathcal{A}^{t}_r$), forget test accuracy ($\mathcal{A}^{t}_f$), and the proposed Relearning Score ($\mathrm{RS}$). A meaningful relearning outcome must satisfy two conditions: the forget class should become more recognizable after relearning, while the performance on the retain classes should remain stable.
Forget class accuracy alone is insufficient because it can be increased by excessively expanding the forget class decision region, causing retain samples to be misclassified as the forget class. $\mathrm{RS}$ is therefore designed to reward forget class recovery only when it is achieved without substantial degradation of retain class performance. We first define the retain-preservation and forget-recovery terms as
\begin{align}
R_r &=
1-\max\!\left(
0,\,
\mathcal{A}^{t\text{-}un}_r
-
\mathcal{A}^{t\text{-}re}_r
\right),
\label{eq:retain-preservation}
\\
R_f &=
\max\!\left(
0,\,
\mathcal{A}^{t\text{-}re}_f
-
\mathcal{A}^{t\text{-}un}_f
\right).
\label{eq:forget-recovery}
\end{align}
All accuracies are normalized to $[0,1]$. Here, $\mathcal{A}^{t\text{-}un}_r$ and $\mathcal{A}^{t\text{-}un}_f$ denote the retain class and forget class accuracies after unlearning, while $\mathcal{A}^{t\text{-}re}_r$ and $\mathcal{A}^{t\text{-}re}_f$ denote the accuracies after relearning. The retain-preservation term $R_r$ starts from one and decreases according to the loss in retain accuracy caused by relearning. If retain accuracy is preserved or improved, $R_r=1$; improvements above the post-unlearning accuracy are not rewarded because they do not provide additional evidence of forget class recoverability. The forget-recovery term $R_f$ measures only the increase in forget class accuracy relative to its unlearning value. Consequently, a model receives no recovery credit when forget class accuracy remains unchanged or decreases. We combine the two terms using their harmonic mean:
\begin{equation}
\mathrm{RS} =
\begin{cases}
\displaystyle \frac{2R_rR_f}{R_r+R_f},
& \hspace{1cm}\text{if}\hspace{0.5 cm} R_r+R_f>0 \\[4pt]
0,
& \hspace{1cm} \text{otherwise}
\end{cases}
\label{eq:rs}
\end{equation}
The harmonic mean is appropriate because both retain preservation and forget recovery are necessary for relearning. $\mathrm{RS}$ is a graded diagnostic rather than a hard certificate of retain stability: because $R_r$ decreases linearly with the absolute retain-accuracy drop, a non-negligible drop can still yield an $R_r$ close to one. We therefore report the pre/post retain accuracies alongside $\mathrm{RS}$ in all main tables and interpret $\mathrm{RS}$ jointly with $\mathcal{A}^{t\text{-}re}_r$, rather than using $\mathrm{RS}$ alone to certify that utility is unchanged. The recoverability--utility trade-off is analyzed in App.~\ref{app:retain_forget_tradeoff}. Unlike an arithmetic mean, it is dominated by the smaller component and therefore prevents strong performance in one term from compensating for poor performance in the other. For example, high forget recovery accompanied by severe retain degradation produces a low $\mathrm{RS}$, rather than being considered successful relearning. Similarly, $\mathrm{RS}$ is zero when no forget class recovery occurs, regardless of how well retain accuracy is preserved. Thus, $\mathrm{RS}\in[0,1]$, where a high value indicates that the forget class can be substantially recovered while retain class performance remains largely unchanged. In contrast, a low value may result from weak forget class recovery, substantial retain degradation, or both. For each method $m$, audit variant $v$, and forget class $c$, we define $\Delta\mathrm{RS}^{(v)}_{m,c}= \mathrm{RS}^{(v)}_{m,c} - \mathrm{RS}^{(v)}_{\mathrm{retrained},c}$, where the retrained reference is matched to the same forget class. Positive (negative) values indicate greater (lower) recoverability than the matched retrained control. We therefore use $\mathrm{RS}$ and $\Delta\mathrm{RS}$ for different questions: $\mathrm{RS}$ measures absolute source-free recoverability, while $\Delta\mathrm{RS}$ measures excess recoverability beyond a model that never observed the forget class. A high $\mathrm{RS}$ with $\Delta\mathrm{RS}\approx0$ should not be interpreted as evidence of forget-specific residual structure.

%\vspace{-0.4cm}
\noindent \textbf{Settings.} Following prior class-unlearning evaluation~\cite{bonato2024retain}, we evaluate all $10$ CIFAR-10 classes and $10$ designated classes
for CIFAR-100 and TinyImageNet. The forget class selections, embedding-generation settings, and computational costs are reported in Apps.~\ref{app:additional_results}, \ref{app:hyperparameters}, and
\ref{app:computational_efficiency}, respectively.

\subsection{Results}

Table~\ref{tab:ResNet-18_single_class_all_datasets} 
summarizes single-class unlearning and relearning results on CIFAR-10, CIFAR-100, and TinyImageNet using ResNet-18 backbone. Since SFRA is designed as a worst-case diagnostic, we report the maximum $\mathrm{RS}$ and maximum $\Delta\mathrm{RS}$ across forget classes independently. These maxima need not correspond to the same forget class. Maximum $\mathrm{RS}$ measures the strongest absolute source-free recoverability, whereas maximum $\Delta\mathrm{RS}$ measures the strongest excess recoverability relative to the matched retrained reference. Accordingly, our strongest evidence of residual post-unlearning recoverability is provided by positive $\Delta\mathrm{RS}$; cases with high $\mathrm{RS}$ but small $\Delta\mathrm{RS}$ are interpreted as generic relearnability rather than as recovery uniquely attributable to prior forget class training. For CIFAR-10 with a ResNet-18 backbone, for instance, Bad Teacher and DELETE achieve the highest $\mathrm{RS}$, indicating substantial absolute source-free recoverability, while their positive $\Delta\mathrm{RS}$ further indicates greater recoverability than the matched retrained reference. In contrast, SCRUB and Negative Gradient+ yield among the lowest $\mathrm{RS}$ and $\Delta\mathrm{RS}$ values, reflecting weaker relearning. Results across backbones and datasets, including full per-class results, are in Apps.~\ref{app:additional_results} and~\ref{app:linear_separability}.
%the forget class geometry is still present in the modeland can be reinstated from synthetic queries alone, whereas

\begin{figure}[h]
    \centering
    \includegraphics[width=0.83\linewidth]{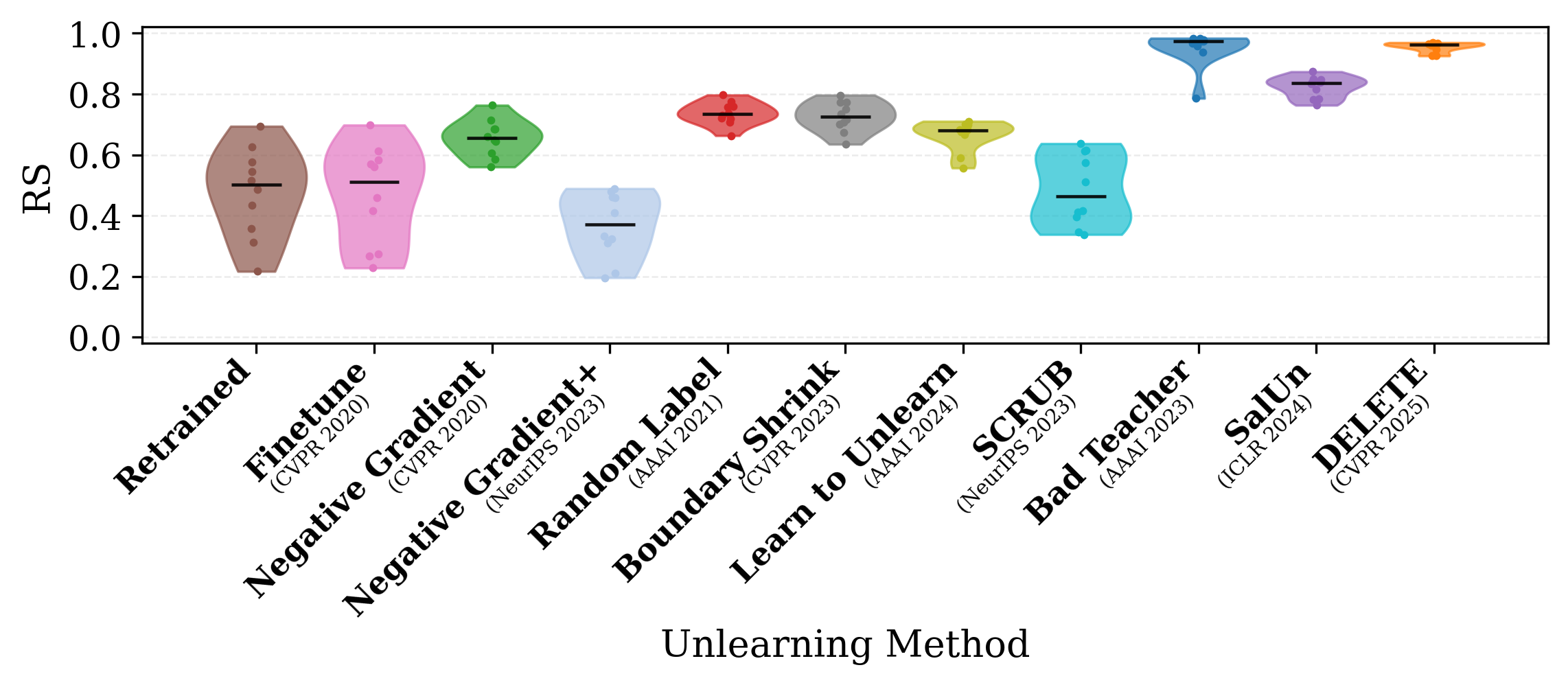}
    \vspace{-0.3cm}
    \caption{$\mathrm{RS}$ distribution across forget classes for CIFAR-10 dataset with ResNet-18 Backbone. Each violin represents an unlearning method, with $\mathrm{RS}$ values across forget classes. Markers and thick bars denote the median and interquartile range (IQR).}
    \label{fig:rs_violin_plot_resnet18}
    \vspace{-0.2cm}
\end{figure}

\noindent \textbf{$\mathbf{RS}$ distribution across forget classes.}
Fig.~\ref{fig:rs_violin_plot_resnet18} reports the distribution of $\mathrm{RS}$ across forget classes for CIFAR-10 with ResNet-18, showing whether relearning is systematic across classes or driven by a few highly vulnerable cases. The results show that methods such as Bad Teacher and DELETE exhibit consistently high $\mathrm{RS}$ across forget classes, indicating systematic relearning, whereas methods such as Negative Gradient+ and SCRUB show lower $\mathrm{RS}$ and are more resistant to relearning. $\mathrm{RS}$ distribution results for all datasets and backbones are provided in App.~\ref{app:rs_distribution}.
\vspace{-0.3cm}

\begin{figure}[H]
    \centering
    \includegraphics[width=0.83\linewidth]{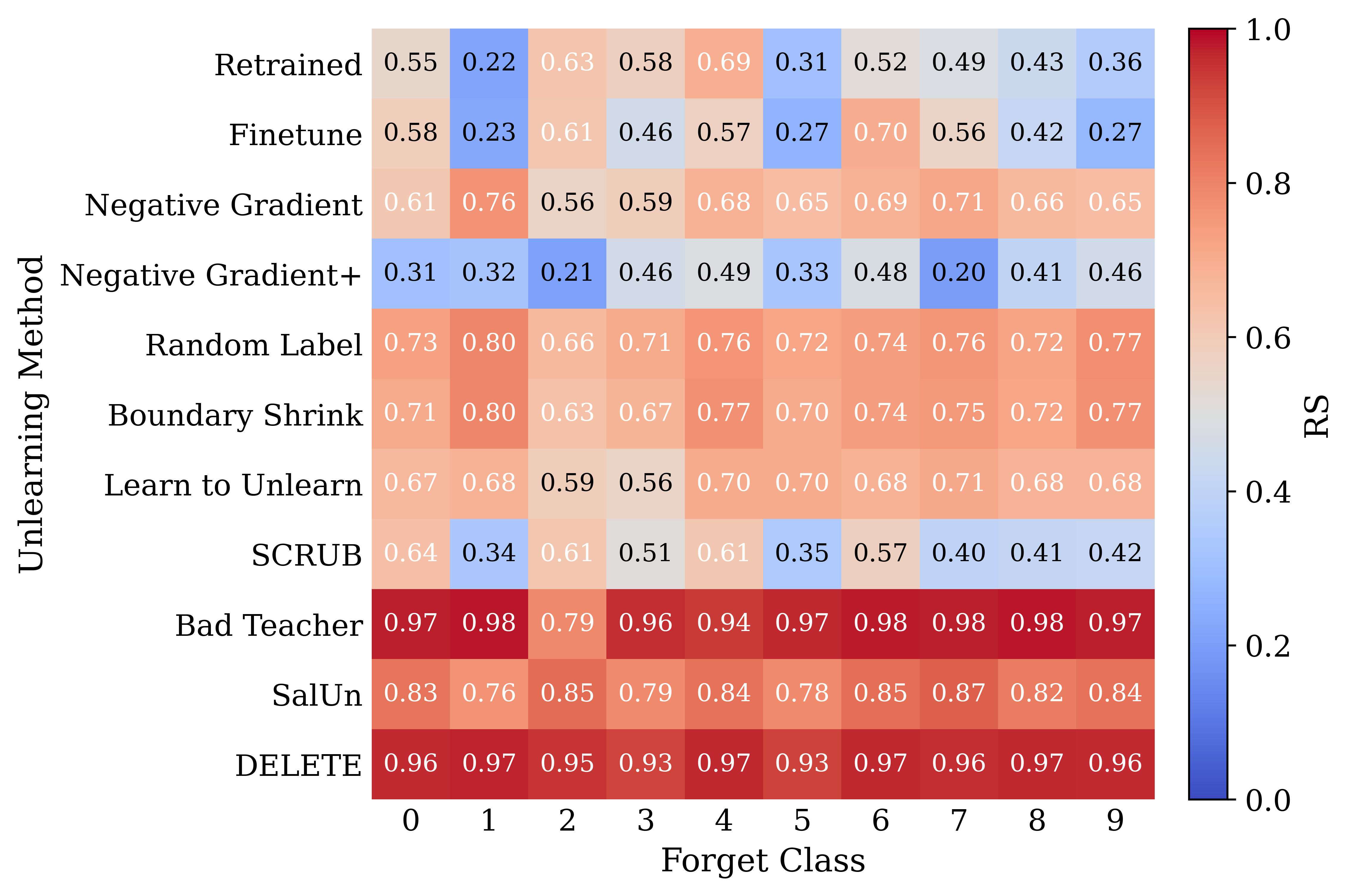}
     \vspace{-0.3cm}
     \caption{Per-class $\mathrm{RS}$ heatmaps for CIFAR-10 dataset and ResNet-18 backbone, comparing unlearning methods. Rows correspond to unlearning methods and columns to forget classes; each cell reports the $\mathrm{RS}$ obtained when the corresponding class is designated for forgetting and color intensity encodes the $\mathrm{RS}$ magnitude.}
    \label{fig:rs_heatmap_plot_resnet18}
    \vspace{-0.3cm}
\end{figure}

\noindent \textbf{Per-class $\mathbf{RS}$ heatmaps.}
Fig.~\ref{fig:rs_heatmap_plot_resnet18} reports per-class $\mathrm{RS}$ heatmaps for CIFAR-10 with ResNet-18, showing class-specific relearning behavior across unlearning methods and complementing the aggregate worst-case $\mathrm{RS}$ reported in Table~\ref{tab:ResNet-18_single_class_all_datasets}. For each forget class, this heatmap allows us to compare unlearning methods and identify which method is most robust to relearning, with lower $\mathrm{RS}$ , and which method is most vulnerable, with higher $\mathrm{RS}$. Additional heatmaps for all datasets and backbones are in App.~\ref{app:rs_heatmaps}.
\vspace{-0.4cm}

\begin{figure}[h]
    \centering \includegraphics[width=\linewidth]{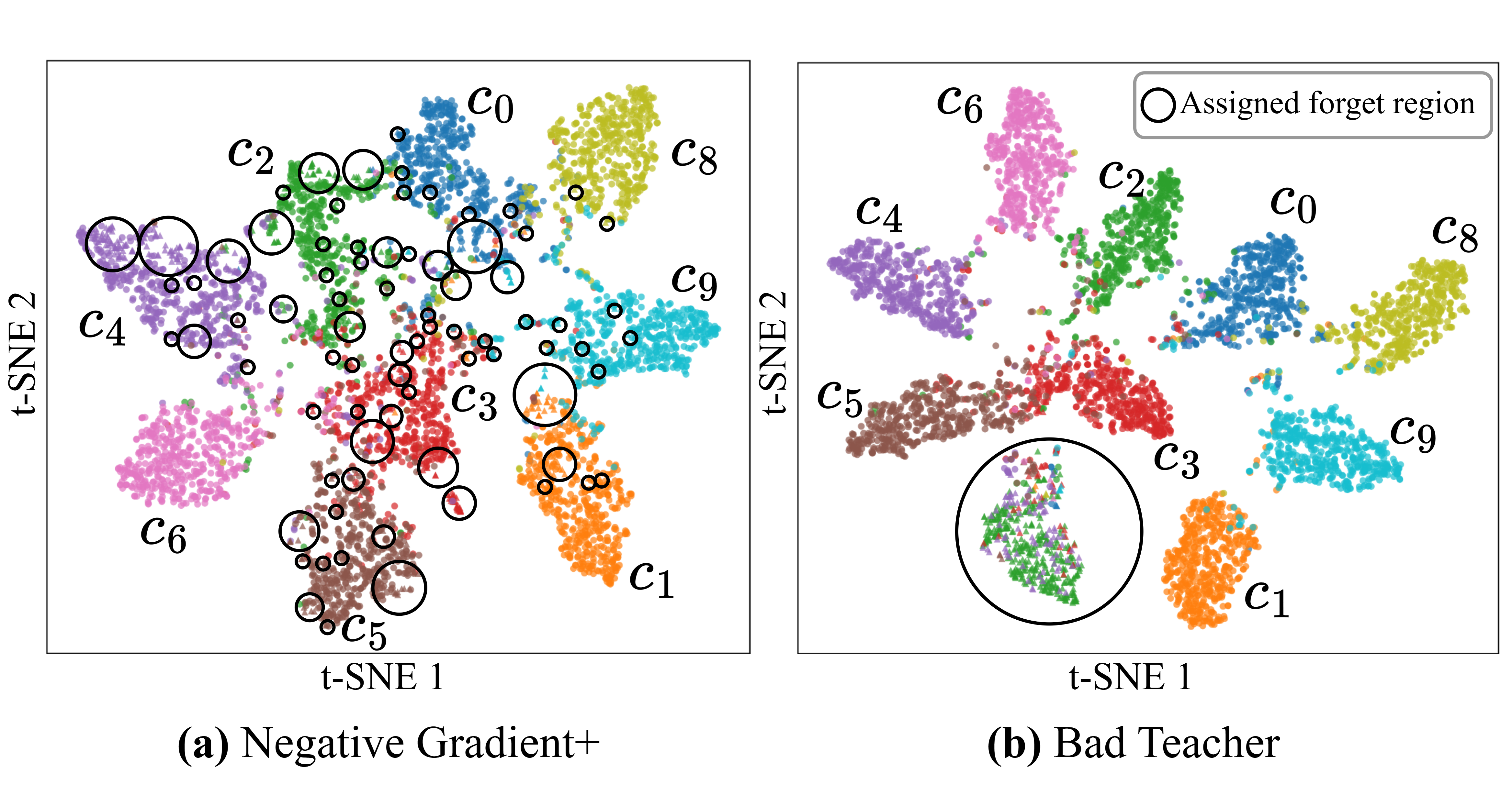}
    \vspace{-0.8cm}
    \caption{Visualization of CIFAR-10 real samples with ResNet-18 (class~$7$ as the forget class): \textbf{(a)}~Negative~Gradient+ and \textbf{(b)}~Bad~Teacher. Black rings highlight the regions assigned to forget embeddings, which are shown as triangles colored by their predicted label, and retain embeddings as circles colored by their class labels. Bad Teacher collapses forget samples into a compact region, whereas Negative Gradient+ scatters them across multiple clusters, disrupting the class geometry and preventing relearning.}
    \label{fig:tsne}
\end{figure}

\noindent \textbf{Geometric evidence and relearning.}
Fig.~\ref{fig:tsne} visualizes the test-set features of a ResNet-18 model after unlearning class~$7$ using Negative Gradient+ and Bad Teacher. As discussed earlier, a high $\mathrm{RS}$ indicates that the representation retains recoverable structure capable of supporting re-separation of the forget class under our source-free head update, whereas a low $\mathrm{RS}$ indicates weaker recoverability under the same audit.
For class~$7$, Negative Gradient+ exhibits limited relearning: forget class accuracy increases from $0.1\%$ to $11.0\%$ following our SFRA, corresponding to an $\mathrm{RS}$ of $0.11$. This low score is consistent with the t-SNE visualization, where the forget class samples appear highly dispersed, suggesting that the method substantially disrupts the feature geometry. In contrast, Bad Teacher shows near-complete relearning: accuracy rises from $0.0\%$ to $99.8\%$ with a corresponding $\mathrm{RS}$ of $0.95$. The visualization corroborates this outcome, the forget class samples remain tightly clustered, indicating that the geometric structure of forget class was largely preserved despite the unlearning step. Together, these results highlight that different unlearning methods produce different effects on the feature space, and that $\mathrm{RS}$ captures this behavior in a quantitative and model-agnostic manner. Additional geometric analyses are provided in App.~\ref{app:geometric_interpretation}, with linear probing in App.~\ref{app:linear_separability} assessing forget-class separability after unlearning.
\vspace{-0.4cm}

\subsection{Ablation Study}
\label{sec:ablation}
\begin{figure}[h]
  \centering
  \includegraphics[width=0.95\columnwidth]{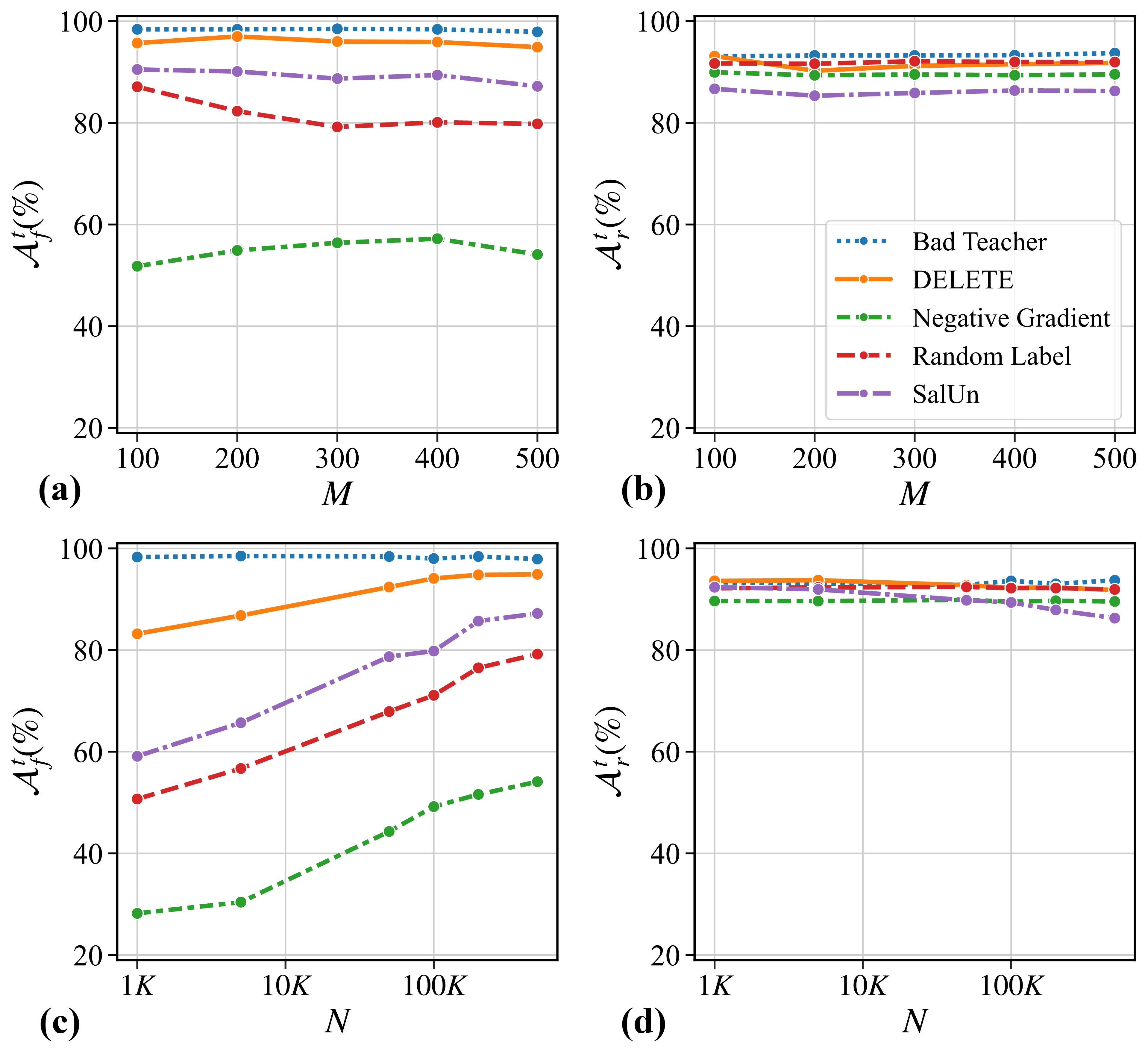}
  \caption{Impact of the number of embeddings on SFRA for CIFAR-10 with a ResNet-18 under single-class unlearning of class~$9$. \textbf{(a,b)} Effect of varying \(M\) with \(N=500K\) on $\mathcal{A}_f^t$ and $\mathcal{A}_r^t$. \textbf{(c,d)} Effect of varying \(N\) with \(M=500\) on $\mathcal{A}_f^t$ and $\mathcal{A}_r^t$.}
  \label{fig:acc_mn_resnet18}
\end{figure}

\noindent\textbf{Sensitivity to the number of synthetic prob.}
We analyze the sensitivity of class relearning to $N$, the number of generated embeddings per retain class, and $M$, the number of selected embeddings. 
For ResNet-18, Fig.~\ref{fig:acc_mn_resnet18} shows that increasing $M$ improves $\mathcal{A}_f^{t}$ until saturation while $\mathcal{A}_r^{t}$ remains within $\pm1\%$; increasing $N$ yields higher $\mathcal{A}_f^{t}$ and more stable $\mathcal{A}_r^{t}$. ViT-B/16 results are provided in App.~\ref{app:sensitivity}. 
%\vspace{-0.4cm}

\noindent\textbf{Multi-class SFRA.}
We extend the proposed procedure to settings with multiple forget classes $\mathcal{Y}_f$. Because a boundary-adjacent synthetic probe does not have a natural forget class label, we use the unlearned classifier scores to partition the selected low-confidence probes among the forget classes, while maintaining balanced assignments and preventing probe reuse. The details and algorithm for multi-class SFRA are provided in App.~\ref{app:additional_results}.
Using this protocol, we evaluate whether class relearning remains effective when multiple classes are forgotten simultaneously. Table~\ref{tab:CIFAR-100_ResNet-18_5v10} reports results on CIFAR-100 with a ResNet-18 backbone under two settings: $5$ forget classes and $10$ forget classes, following the class selection protocol of \cite{zhou2025decoupled}. Multi-class SFRA is more challenging because boundary probes lack a natural forget class identity, introducing class-assignment ambiguity. Consequently, recovery may depend on the separability of the residual forget class representations, which can also explain the varying performance relative to PRA, which constructs class-specific prototypes from real forget samples. The multi-class results are not monotone in the number of forget classes. For example, DELETE yields $\mathrm{RS}=0$ in the 5-class setting but $\mathrm{RS}=0.28$ in the 10-class setting. This does not imply that forgetting becomes intrinsically weaker as more classes are removed; changing $\mathcal{Y}_f$ also changes the synthetic probe-assignment problem and can alter how the greedy partition aligns with the residual geometry of each class. We therefore regard the present multi-class construction as a more fragile diagnostic than the single-class audit, and do not interpret differences across 5 and 10 classes as a monotone measure of unlearning strength. Results for 2-class unlearning and forget class selections for the 2-, 5-, and 10-class settings are in App.~\ref{app:additional_results}.

\begin{table}[t]
\centering
\caption{Comparison of unlearning methods under our proposed SFRA and the source-dependent PRA baseline for 5-class and 10-class unlearning on CIFAR-100 with ResNet-18. Within each forget-set-size setting, the highest and second-highest $\mathrm{RS}$ and $\Delta\mathrm{RS}$ values are shown in \textbf{bold} and \underline{underlined}, respectively.}
\label{tab:CIFAR-100_ResNet-18_5v10}
\vspace{-3mm}
\fontsize{10.5}{10.5}\selectfont
\resizebox{\columnwidth}{!}{%
\begin{tabular}{c|c|cccc|cccc}
\toprule
\toprule
\multirow{2}{*}{Unlearning Method} & \multirow{2}{*}{Model Variant} & \multicolumn{4}{c}{\textbf{5-class}} & \multicolumn{4}{|c}{\textbf{10-class}} \\
 &  & $\mathcal{A}^{t}_{r}(\%)$ & $\mathcal{A}^{t}_{f}(\%)$ & $\mathrm{RS}$ & $\Delta\mathrm{RS}$ & $\mathcal{A}^{t}_{r}(\%)$ & $\mathcal{A}^{t}_{f}(\%)$ & $\mathrm{RS}$ & $\Delta\mathrm{RS}$\\
\midrule
\midrule
Original & Original & {\fontsize{10.5}{10.5}\selectfont $80.03$} & {\fontsize{10.5}{10.5}\selectfont $78.60$} & - & - & {\fontsize{10.5}{10.5}\selectfont $79.80$} & {\fontsize{10.5}{10.5}\selectfont $81.40$} & - & - \\
\midrule
\midrule
\multirow{3}{*}{Retrained} & Unlearned & {\fontsize{10.5}{10.5}\selectfont $77.26$} & {\fontsize{10.5}{10.5}\selectfont $0.00$} & - & - & {\fontsize{10.5}{10.5}\selectfont $77.54$} & {\fontsize{10.5}{10.5}\selectfont $0.00$} & - & - \\
 & PRA \cite{ha2025unlearning} & {\fontsize{10.5}{10.5}\selectfont $76.46$} & {\fontsize{10.5}{10.5}\selectfont $8.20$} & {\fontsize{10.5}{10.5}\selectfont $0.15$} & - & {\fontsize{10.5}{10.5}\selectfont $76.87$} & {\fontsize{10.5}{10.5}\selectfont $10.60$} & {\fontsize{10.5}{10.5}\selectfont $0.19$} & - \\
 & SFRA (ours) & {\fontsize{10.5}{10.5}\selectfont $77.42$} & {\fontsize{10.5}{10.5}\selectfont $0.00$} & {\fontsize{10.5}{10.5}\selectfont $0.00$} & - & {\fontsize{10.5}{10.5}\selectfont $77.72$} & {\fontsize{10.5}{10.5}\selectfont $0.00$} & {\fontsize{10.5}{10.5}\selectfont $0.00$} & - \\
\midrule
\midrule
\multirow{3}{*}{Finetune \cite{golatkar2020eternal}} & Unlearned & {\fontsize{10.5}{10.5}\selectfont $78.73$} & {\fontsize{10.5}{10.5}\selectfont $0.60$} & - & - & {\fontsize{10.5}{10.5}\selectfont $79.71$} & {\fontsize{10.5}{10.5}\selectfont $1.70$} & - & - \\
 & PRA \cite{ha2025unlearning} & {\fontsize{10.5}{10.5}\selectfont $78.18$} & {\fontsize{10.5}{10.5}\selectfont $16.40$} & {\fontsize{10.5}{10.5}\selectfont $0.27$} & {\fontsize{10.5}{10.5}\selectfont $+0.12$} & {\fontsize{10.5}{10.5}\selectfont $79.11$} & {\fontsize{10.5}{10.5}\selectfont $26.80$} & {\fontsize{10.5}{10.5}\selectfont $0.40$} & {\fontsize{10.5}{10.5}\selectfont $+0.21$} \\
 & SFRA (ours) & {\fontsize{10.5}{10.5}\selectfont $70.95$} & {\fontsize{10.5}{10.5}\selectfont $45.40$} & {\fontsize{10.5}{10.5}\selectfont $0.60$} & {\fontsize{10.5}{10.5}\selectfont $\underline{+0.60}$} & {\fontsize{10.5}{10.5}\selectfont $72.00$} & {\fontsize{10.5}{10.5}\selectfont $60.50$} & {\fontsize{10.5}{10.5}\selectfont \textbf{\boldmath $0.72$}} & {\fontsize{10.5}{10.5}\selectfont $\mathbf{+0.72}$} \\
\midrule
\multirow{3}{*}{Negative Gradient \cite{golatkar2020eternal}} & Unlearned & {\fontsize{10.5}{10.5}\selectfont $72.00$} & {\fontsize{10.5}{10.5}\selectfont $0.20$} & - & - & {\fontsize{10.5}{10.5}\selectfont $71.28$} & {\fontsize{10.5}{10.5}\selectfont $3.20$} & - & - \\
 & PRA \cite{ha2025unlearning} & {\fontsize{10.5}{10.5}\selectfont $71.72$} & {\fontsize{10.5}{10.5}\selectfont $14.00$} & {\fontsize{10.5}{10.5}\selectfont $0.24$} & {\fontsize{10.5}{10.5}\selectfont $+0.09$} & {\fontsize{10.5}{10.5}\selectfont $70.80$} & {\fontsize{10.5}{10.5}\selectfont $14.60$} & {\fontsize{10.5}{10.5}\selectfont $0.20$} & {\fontsize{10.5}{10.5}\selectfont $+0.01$} \\
 & SFRA (ours) & {\fontsize{10.5}{10.5}\selectfont $67.73$} & {\fontsize{10.5}{10.5}\selectfont $18.20$} & {\fontsize{10.5}{10.5}\selectfont $0.30$} & {\fontsize{10.5}{10.5}\selectfont $+0.30$} & {\fontsize{10.5}{10.5}\selectfont $67.84$} & {\fontsize{10.5}{10.5}\selectfont $22.30$} & {\fontsize{10.5}{10.5}\selectfont $0.32$} & {\fontsize{10.5}{10.5}\selectfont $+0.32$} \\
\midrule
\multirow{3}{*}{Negative Gradient+ \cite{kurmanji2023towards}} & Unlearned & {\fontsize{10.5}{10.5}\selectfont $76.08$} & {\fontsize{10.5}{10.5}\selectfont $0.20$} & - & - & {\fontsize{10.5}{10.5}\selectfont $75.69$} & {\fontsize{10.5}{10.5}\selectfont $0.00$} & - & - \\
 & PRA \cite{ha2025unlearning} & {\fontsize{10.5}{10.5}\selectfont $75.09$} & {\fontsize{10.5}{10.5}\selectfont $24.40$} & {\fontsize{10.5}{10.5}\selectfont $0.39$} & {\fontsize{10.5}{10.5}\selectfont $+0.24$} & {\fontsize{10.5}{10.5}\selectfont $75.07$} & {\fontsize{10.5}{10.5}\selectfont $10.60$} & {\fontsize{10.5}{10.5}\selectfont $0.19$} & {\fontsize{10.5}{10.5}\selectfont $+0.00$} \\
 & SFRA (ours) & {\fontsize{10.5}{10.5}\selectfont $72.16$} & {\fontsize{10.5}{10.5}\selectfont $14.20$} & {\fontsize{10.5}{10.5}\selectfont $0.24$} & {\fontsize{10.5}{10.5}\selectfont $+0.24$} & {\fontsize{10.5}{10.5}\selectfont $71.51$} & {\fontsize{10.5}{10.5}\selectfont $4.50$} & {\fontsize{10.5}{10.5}\selectfont $0.09$} & {\fontsize{10.5}{10.5}\selectfont $+0.09$} \\
\midrule
\multirow{3}{*}{Random Label \cite{hayase2020selective}} & Unlearned & {\fontsize{10.5}{10.5}\selectfont $72.01$} & {\fontsize{10.5}{10.5}\selectfont $3.00$} & - & - & {\fontsize{10.5}{10.5}\selectfont $72.27$} & {\fontsize{10.5}{10.5}\selectfont $6.60$} & - & - \\
 & PRA \cite{ha2025unlearning} & {\fontsize{10.5}{10.5}\selectfont $71.28$} & {\fontsize{10.5}{10.5}\selectfont $39.20$} & {\fontsize{10.5}{10.5}\selectfont $0.53$} & {\fontsize{10.5}{10.5}\selectfont $+0.38$} & {\fontsize{10.5}{10.5}\selectfont $71.69$} & {\fontsize{10.5}{10.5}\selectfont $28.60$} & {\fontsize{10.5}{10.5}\selectfont $0.36$} & {\fontsize{10.5}{10.5}\selectfont $+0.17$} \\
 & SFRA (ours) & {\fontsize{10.5}{10.5}\selectfont $65.66$} & {\fontsize{10.5}{10.5}\selectfont $40.80$} & {\fontsize{10.5}{10.5}\selectfont $0.54$} & {\fontsize{10.5}{10.5}\selectfont $+0.54$} & {\fontsize{10.5}{10.5}\selectfont $65.11$} & {\fontsize{10.5}{10.5}\selectfont $42.20$} & {\fontsize{10.5}{10.5}\selectfont $0.51$} & {\fontsize{10.5}{10.5}\selectfont $+0.51$} \\
\midrule
\multirow{3}{*}{Learn to Unlearn \cite{cha2024learning}} & Unlearned & {\fontsize{10.5}{10.5}\selectfont $70.92$} & {\fontsize{10.5}{10.5}\selectfont $0.60$} & - & - & {\fontsize{10.5}{10.5}\selectfont $71.29$} & {\fontsize{10.5}{10.5}\selectfont $2.90$} & - & - \\
 & PRA \cite{ha2025unlearning} & {\fontsize{10.5}{10.5}\selectfont $70.22$} & {\fontsize{10.5}{10.5}\selectfont $21.60$} & {\fontsize{10.5}{10.5}\selectfont $0.35$} & {\fontsize{10.5}{10.5}\selectfont $+0.20$} & {\fontsize{10.5}{10.5}\selectfont $70.51$} & {\fontsize{10.5}{10.5}\selectfont $18.20$} & {\fontsize{10.5}{10.5}\selectfont $0.26$} & {\fontsize{10.5}{10.5}\selectfont $+0.07$} \\
 & SFRA (ours) & {\fontsize{10.5}{10.5}\selectfont $64.25$} & {\fontsize{10.5}{10.5}\selectfont $25.80$} & {\fontsize{10.5}{10.5}\selectfont $0.40$} & {\fontsize{10.5}{10.5}\selectfont $+0.40$} & {\fontsize{10.5}{10.5}\selectfont $67.01$} & {\fontsize{10.5}{10.5}\selectfont $20.70$} & {\fontsize{10.5}{10.5}\selectfont $0.30$} & {\fontsize{10.5}{10.5}\selectfont $+0.30$} \\
\midrule
\multirow{3}{*}{SCRUB \cite{kurmanji2023towards}} & Unlearned & {\fontsize{10.5}{10.5}\selectfont $80.07$} & {\fontsize{10.5}{10.5}\selectfont $0.00$} & - & - & {\fontsize{10.5}{10.5}\selectfont $80.20$} & {\fontsize{10.5}{10.5}\selectfont $0.00$} & - & - \\
 & PRA \cite{ha2025unlearning} & {\fontsize{10.5}{10.5}\selectfont $79.62$} & {\fontsize{10.5}{10.5}\selectfont $13.20$} & {\fontsize{10.5}{10.5}\selectfont $0.23$} & {\fontsize{10.5}{10.5}\selectfont $+0.08$} & {\fontsize{10.5}{10.5}\selectfont $79.66$} & {\fontsize{10.5}{10.5}\selectfont $11.30$} & {\fontsize{10.5}{10.5}\selectfont $0.20$} & {\fontsize{10.5}{10.5}\selectfont $+0.01$} \\
 & SFRA (ours) & {\fontsize{10.5}{10.5}\selectfont $80.07$} & {\fontsize{10.5}{10.5}\selectfont $0.00$} & {\fontsize{10.5}{10.5}\selectfont $0.00$} & {\fontsize{10.5}{10.5}\selectfont $+0.00$} & {\fontsize{10.5}{10.5}\selectfont $80.20$} & {\fontsize{10.5}{10.5}\selectfont $0.00$} & {\fontsize{10.5}{10.5}\selectfont $0.00$} & {\fontsize{10.5}{10.5}\selectfont $+0.00$} \\
\midrule
\multirow{3}{*}{Bad Teacher \cite{chundawat2023can}} & Unlearned & {\fontsize{10.5}{10.5}\selectfont $79.25$} & {\fontsize{10.5}{10.5}\selectfont $0.00$} & - & - & {\fontsize{10.5}{10.5}\selectfont $78.31$} & {\fontsize{10.5}{10.5}\selectfont $0.20$} & - & - \\
 & PRA \cite{ha2025unlearning} & {\fontsize{10.5}{10.5}\selectfont $78.64$} & {\fontsize{10.5}{10.5}\selectfont $47.40$} & {\fontsize{10.5}{10.5}\selectfont \textbf{\boldmath $0.64$}} & {\fontsize{10.5}{10.5}\selectfont $+0.49$} & {\fontsize{10.5}{10.5}\selectfont $77.87$} & {\fontsize{10.5}{10.5}\selectfont $14.80$} & {\fontsize{10.5}{10.5}\selectfont $0.25$} & {\fontsize{10.5}{10.5}\selectfont $+0.06$} \\
 & SFRA (ours) & {\fontsize{10.5}{10.5}\selectfont $73.72$} & {\fontsize{10.5}{10.5}\selectfont $27.80$} & {\fontsize{10.5}{10.5}\selectfont $0.43$} & {\fontsize{10.5}{10.5}\selectfont $+0.43$} & {\fontsize{10.5}{10.5}\selectfont $70.93$} & {\fontsize{10.5}{10.5}\selectfont $14.50$} & {\fontsize{10.5}{10.5}\selectfont $0.25$} & {\fontsize{10.5}{10.5}\selectfont $+0.25$} \\
\midrule
\multirow{3}{*}{SalUn \cite{fan2023salun}} & Unlearned & {\fontsize{10.5}{10.5}\selectfont $78.82$} & {\fontsize{10.5}{10.5}\selectfont $4.40$} & - & - & {\fontsize{10.5}{10.5}\selectfont $79.44$} & {\fontsize{10.5}{10.5}\selectfont $2.60$} & - & - \\
 & PRA \cite{ha2025unlearning} & {\fontsize{10.5}{10.5}\selectfont $78.05$} & {\fontsize{10.5}{10.5}\selectfont $43.40$} & {\fontsize{10.5}{10.5}\selectfont $0.56$} & {\fontsize{10.5}{10.5}\selectfont $+0.41$} & {\fontsize{10.5}{10.5}\selectfont $78.63$} & {\fontsize{10.5}{10.5}\selectfont $43.60$} & {\fontsize{10.5}{10.5}\selectfont $0.58$} & {\fontsize{10.5}{10.5}\selectfont $+0.39$} \\
 & SFRA (ours) & {\fontsize{10.5}{10.5}\selectfont $76.27$} & {\fontsize{10.5}{10.5}\selectfont $52.20$} & {\fontsize{10.5}{10.5}\selectfont $\underline{0.64}$} & {\fontsize{10.5}{10.5}\selectfont $\mathbf{+0.64}$} & {\fontsize{10.5}{10.5}\selectfont $73.29$} & {\fontsize{10.5}{10.5}\selectfont $52.60$} & {\fontsize{10.5}{10.5}\selectfont $\underline{0.65}$} & {\fontsize{10.5}{10.5}\selectfont $\underline{+0.65}$} \\
\midrule
\multirow{3}{*}{DELETE \cite{zhou2025decoupled}} & Unlearned & {\fontsize{10.5}{10.5}\selectfont $80.08$} & {\fontsize{10.5}{10.5}\selectfont $0.00$} & - & - & {\fontsize{10.5}{10.5}\selectfont $80.56$} & {\fontsize{10.5}{10.5}\selectfont $0.00$} & - & - \\
 & PRA \cite{ha2025unlearning} & {\fontsize{10.5}{10.5}\selectfont $79.37$} & {\fontsize{10.5}{10.5}\selectfont $39.00$} & {\fontsize{10.5}{10.5}\selectfont $0.56$} & {\fontsize{10.5}{10.5}\selectfont $+0.41$} & {\fontsize{10.5}{10.5}\selectfont $79.73$} & {\fontsize{10.5}{10.5}\selectfont $44.70$} & {\fontsize{10.5}{10.5}\selectfont $0.62$} & {\fontsize{10.5}{10.5}\selectfont $+0.42$} \\
 & SFRA (ours) & {\fontsize{10.5}{10.5}\selectfont $80.08$} & {\fontsize{10.5}{10.5}\selectfont $0.00$} & {\fontsize{10.5}{10.5}\selectfont $0.00$} & {\fontsize{10.5}{10.5}\selectfont $+0.00$} & {\fontsize{10.5}{10.5}\selectfont $73.93$} & {\fontsize{10.5}{10.5}\selectfont $16.50$} & {\fontsize{10.5}{10.5}\selectfont $0.28$} & {\fontsize{10.5}{10.5}\selectfont $+0.28$} \\
\midrule
\bottomrule
\end{tabular}
}
\vspace{-0.3cm}
\end{table}

\section{Conclusion}
\label{sec:conclusion}

We introduced a Source-Free Relearning Audit~(SFRA) that uses synthetic feature-space probes and a lightweight classifier-head update to assess post-unlearning recoverability. Across multiple datasets, backbones, and unlearning methods, our results show that low forget class accuracy can coexist with substantial source-free relearning. We quantify absolute recoverability using the Relearning Score~($\mathrm{RS}$), which jointly measures forget class recovery and retain class preservation, and use $\Delta\mathrm{RS}$ relative to a matched retrained reference to quantify excess recoverability beyond generic representation transfer. This distinction is central to our interpretation: a high $\mathrm{RS}$  demonstrates that a class can be re-separated under the audit, but does not by itself establish recovery of memorized training information or forget-specific residual structure. Positive $\Delta\mathrm{RS}$ provides stronger evidence that the released unlearned model is more recoverable than a model that never observed the forget class.

\noindent \textbf{Limitations and future work.}
Our evaluation focuses on classifier-input representations, as auditing earlier layers requires depth-specific probes and updates. In the multi-class setting, assigning unlabeled boundary probes to several forget classes introduces additional ambiguity and can produce non-monotone behavior as the forget set changes. Moreover, failure to observe relearning does not establish complete erasure. Future work will investigate efficient layer-wise auditing, improved multi-class probe assignment, and stronger statistical tests of forgetting.

%\begin{comment}
{
    \small
    \bibliographystyle{ieeenat_fullname}
    \bibliography{main}
}
%\end{comment}
\clearpage
\appendix
\counterwithin{figure}{section}
\counterwithin{table}{section}
\counterwithin{equation}{section}
\counterwithin{algocf}{section}

\section*{Appendix}

\begin{center}
    \textbf{Table of Contents}
\end{center}

\vspace{-0.5em}
\hrule
\vspace{0.5em}

\small

\noindent
\textbf{A}\quad Proof of Proposition~1
\dotfill \pageref{app:proof}\\

\noindent
\textbf{B}\quad Hyperparameter Settings
\dotfill \pageref{app:hyperparameters}\\

\noindent
\textbf{C}\quad Computational Cost and Efficiency
\dotfill \pageref{app:computational_efficiency}\\

\noindent
\textbf{D}\quad Post-hoc Synthetic--Real Alignment
\dotfill \pageref{app:alignment_analysis}\\

\noindent
\textbf{E}\quad Empirical Assessment of the Margin Approximation
\dotfill \pageref{app:margin_approximation}\\

\noindent
\textbf{F}\quad Confidence of Forget Class Assignments
\dotfill \pageref{app:forget_assignment_confidence}\\

\noindent
\textbf{G}\quad SFRA Without the Released Forget Class Output Row
\dotfill \pageref{app:forget_row_ablation}\\

\noindent
\textbf{H}\quad Sensitivity to the Number of Synthetic Probes
\dotfill \pageref{app:sensitivity}\\

\noindent
\textbf{I}\quad Retain--Forget Accuracy Trade-off
\dotfill \pageref{app:retain_forget_tradeoff}\\

\noindent
\textbf{J}\quad Additional Results and Details for Single-Class and Multi-Class SFRA
\dotfill \pageref{app:additional_results}\\

\noindent
\textbf{K}\quad Geometric Interpretation of Synthetic Boundary Probes
\dotfill \pageref{app:geometric_interpretation}\\

\noindent
\textbf{L}\quad $\mathrm{RS}$ Distribution Across Forget Classes
\dotfill \pageref{app:rs_distribution}\\

\noindent
\textbf{M}\quad Per-Class $\mathrm{RS}$ Heatmaps
\dotfill \pageref{app:rs_heatmaps}\\

\noindent
\textbf{N}\quad Absolute and Excess Recoverability
\dotfill \pageref{app:recoverability_diagnostic}\\

\noindent
\textbf{O}\quad Sampling Distribution Ablation
\dotfill \pageref{app:distribution_ablation}\\

\noindent
\textbf{P}\quad Uncertainty-Score Ablation
\dotfill \pageref{app:uncertainty_metric}\\

\noindent
\textbf{Q}\quad Effect of Gaussian Support on SFRA
\dotfill \pageref{app:gaussian_support}\\

\noindent
\textbf{R}\quad Detailed Per-Class Results and Linear Separability
\dotfill \pageref{app:linear_separability}

\vspace{0.5em}
\hrule
\vspace{1em}

\normalsize
\vfill

\section{Proof of proposition 1}
\label{app:proof}

Consider a single gradient update on the classification head parameters \(W\) (the classifier bias is omitted for simplicity). Let \(w_c\) denote the \(c\)-th row of \(W\). The pairwise logit margin for a real forget class embedding \(z\sim \mathcal{E}_f\)
between class \(c_f\) and a retain class \(j\in\mathcal{Y}_r\) is defined in proposition~1 as $m_{c_f,j}(z)$. After one gradient step with learning rate \(\eta>0\) computed using the cross-entropy loss on synthetic probs in \(\mathcal{S}_f\) (which are treated as labeled with class \(c_f\)), the change in the margin at real embedding \(z\) is as follows: 
\begin{equation}
\Delta m_{c_f,j}(z)
= (\Delta w_{c_f} - \Delta w_j)^\top z,
\label{eq:change_margin}
\end{equation}
Using the gradient of the cross-entropy loss, we obtain:
\begin{align}
    \Delta w_c =& -\eta \,\mathbb{E}_{s\sim \mathcal{S}_f}[\nabla_{w_c} \ell(h(s), c_f)]\\
    =&-\eta \,\mathbb{E}_{s\sim \mathcal{S}_f}\left[\big(p_c(s) - \mathbbm{1}\{c=c_f\}\big)\, s\right],\nonumber
\end{align}
where \(p_c(s)\) denotes the softmax probability assigned to class \(c\)
for input \(s\). Hence,
\begin{align}\label{eq:eqdelta}
    \Delta w_{c_f} - \Delta w_j= & -\eta\,\mathbb{E}_{s\sim\mathcal{S}_f}\big[(p_{c_f}(s)-1) - p_j(s)\big]\, s \nonumber \\
    = & \eta\,\mathbb{E}_{s\sim\mathcal{S}_f}\big[1 - p_{c_f}(s) + p_j(s)\big]\, s.
\end{align}

\noindent By substituting \eqref{eq:eqdelta} into the margin change in equation~\eqref{eq:change_margin} and taking expectation over real forget class embeddings \(z\sim \mathcal{E}_f\) we can say:
\begin{equation}
    \mathbb{E}_{z\sim \mathcal{E}_f}\left[\Delta m_{c_f,j}(z)\right]
= \eta\,\mathbb{E}_{\substack{s\sim\mathcal{S}_f\\z\sim\mathcal E_f}}\left[\left(1 - p_{c_f}(s) + p_j(s)\right)s^\top z\right].
\end{equation}
Because $s\sim\mathcal{S}_f$ and $z\sim\mathcal{E}_f$ are sampled
independently, the expectation factorizes, and the expected margin change
can be written as
\begin{equation}
\mathbb{E}_{z\sim\mathcal{E}_f}
\left[\Delta m_{c_f,j}(z)\right]
=
\eta\,
\mathbb{E}_{s\sim\mathcal{S}_f}
\left[\alpha_j(s)s\right]^\top
\mu_{\mathcal{E}_f}.
\label{eq:exact_weighted_margin}
\end{equation}
Defining
$\alpha_j(s)=1-p_{c_f}(s)+p_j(s)$
and writing $\alpha_j(s)=1+\delta_j(s)$ yields
\begin{equation}
\mathbb{E}_{z\sim\mathcal{E}_f}
\left[\Delta m_{c_f,j}(z)\right]
=
\eta\left(
\mu_{\mathcal{S}_f}^{\top}\mu_{\mathcal{E}_f}
+
r_j
\right),
\label{eq:margin_with_residual}
\end{equation}
where the residual introduced replacing $\alpha_j(s)$ with one is
\begin{equation}
r_j =
\mathbb{E}_{s\sim\mathcal{S}_f}
\left[\delta_j(s)s\right]^\top
\mu_{\mathcal{E}_f}.
\label{eq:residual}
\end{equation}
For the evaluated setting, Appendix~\ref{app:margin_approximation}
shows that the unweighted approximation closely tracks the weighted expression in aggregate. This empirical result supports using the mean-alignment condition as an approximation, but the exact condition remains $\mu_{\mathcal{S}_f}^{\top}\mu_{\mathcal{E}_f} + r_j > 0$. Consequently,
when the residual is small relative to the alignment term,
\begin{equation}
\mu_{\mathcal{S}_f}^{\top}\mu_{\mathcal{E}_f} > |r_j|,
\label{eq:robust_alignment_condition}
\end{equation}
the expected pairwise margin increases:
\begin{equation}
\mathbb{E}_{z\sim\mathcal{E}_f}
\left[\Delta m_{c_f,j}(z)\right] > 0.
\end{equation}
In particular, when $r_j\approx0$, this condition reduces to the
approximate alignment criterion
$\mu_{\mathcal{S}_f}^{\top}\mu_{\mathcal{E}_f}>0$.
This result characterizes the contribution of the synthetic forget-set
loss. The synthetic retain-set loss in Alg.~1 may contribute
an additional term to the complete update. Although derived for the
classifier-input representation, the same analysis may in principle be
extended to intermediate representations using an appropriate probe head.

\section{Hyperparameter Settings}
\label{app:hyperparameters}
For our proposed SFRA, we sample a pool of $N$ synthetic probs per retain class from a standard Gaussian in feature space, then mine two subsets: (i) the $M$ most confident samples (highest predicted probability) to form the synthetic retain set, and (ii) the $M$ least confident samples as boundary-adjacent probes, which are relabeled as the forget class and used to drive relearning updates. Table~\ref{tab:Hyperparameters_settings} reports the values of $(N,M)$ used across datasets and settings. $N$ controls how thoroughly we explore the feature space, while $M$ controls the size of the mined retain and boundary-probe subsets (and thus the relearning compute). During relearning, we update only the classifier head while keeping the feature encoder frozen. The classifier is optimized with Adam using learning rate $10^{-2}$ and weight decay $10^{-4}$. Synthetic relearning samples are processed with batch size 256, while evaluation uses batch size 1024. We use a held-out validation set to select the best-performing relearning model and apply early stopping when the validation performance does not improve for 10 consecutive epochs (patience = 10).

\begin{table}[H]
\centering
\small
\caption{Hyperparameters used throughout our experiments. We report \(N\), the total number of generated embeddings per retain class, and \(M\), the number of selected embeddings.}
\label{tab:Hyperparameters_settings}

\resizebox{0.65\columnwidth}{!}{
\begin{tabular}{l|l|cc}
\toprule
Setting & Dataset & $N$ & $M$ \\
\midrule
\multirow{3}{*}{Single-class}
& CIFAR-10     & $500K$ & $500$ \\
& CIFAR-100    & $100K$ & $50$  \\
& TinyImageNet & $50K$  & $25$  \\
\midrule
\multirow{2}{*}{Multi-class}
& CIFAR-10     & $500K$ & $500$ \\
& CIFAR-100    & $50K$  & $100$ \\
\bottomrule
\end{tabular}
}
\end{table}

\section{Computational Cost and Efficiency}
\label{app:computational_efficiency}
To assess the computational overhead of our proposed SFRA, we measure the wall-clock time required for synthetic probe construction across representative datasets, architectures, and unlearned checkpoints. Probe construction is performed directly in the classifier-input space using the released classifier head and therefore requires neither image generation nor forward passes through the full encoder. The timing benchmark reproduces the two-pass construction used by SFRA: one pass selects high-confidence retain probes, while the other selects low-confidence boundary probes that are relabeled as the forget class. We conduct the benchmark on a single NVIDIA RTX A6000 GPU using a sampling batch size of $65{,}536$. The reported times correspond to one unlearned checkpoint with forget class~$0$ and are extrapolated over all corresponding retain classes.

\begin{table}[H]
\centering
\caption{Probe-generation runtime for our proposed SFRA using single-class unlearning checkpoints with forget class 0. Each entry reports the full extrapolated probe-construction time for constructing Gaussian feature-space probes. The probe settings are CIFAR-10: $N=500K$ and $M=500$, CIFAR-100: $N=100K$ and $M=100$, and TinyImageNet: $N=50K$ and $M=25$.}
\label{tab:probe_generation_timing_architectures}
\scriptsize
\setlength{\tabcolsep}{3pt}
\renewcommand{\arraystretch}{0.95}
\resizebox{\columnwidth}{!}{%
\begin{tabular}{l|l|ccc}
\toprule
Backbone & Unlearning Method & \multicolumn{1}{c}{CIFAR-10} & \multicolumn{1}{c}{CIFAR-100} & \multicolumn{1}{c}{TinyImageNet} \\
\midrule
\multirow{10}{*}{ResNet-18} & Finetune \cite{golatkar2020eternal} & 0.86s & 3.4m & 1.8m \\
 & Negative Gradient \cite{golatkar2020eternal} & 0.92s & 5.1m & 1.8m \\
 & Negative Gradient+ \cite{kurmanji2023towards} & 0.92s & 4.1m & 1.8m \\
 & Random Label \cite{hayase2020selective} & 1.0s & 5.0m & 1.8m \\
 & Boundary Shrink \cite{chen2023boundary} & 1.1s & 4.4m & 1.8m \\
 & Learn to Unlearn \cite{cha2024learning} & 0.91s & 4.6m & 1.8m \\
 & SCRUB \cite{kurmanji2023towards} & 0.97s & 3.7m & 1.8m \\
 & Bad Teacher \cite{chundawat2023can} & 0.93s & 5.0m & 1.7m \\
 & SalUn \cite{fan2023salun} & 0.93s & 9.1m & 1.8m \\
 & DELETE \cite{zhou2025decoupled} & 0.91s & 3.9m & 1.8m \\
\midrule
\multirow{9}{*}{Swin-T} & Finetune \cite{golatkar2020eternal} & 1.2s & 4.9m & 3.2m \\
 & Negative Gradient \cite{golatkar2020eternal} & 1.2s & 5.8m & 2.9m \\
 & Negative Gradient+ \cite{kurmanji2023towards} & 1.3s & 6.0m & 2.9m \\
 & Random Label \cite{hayase2020selective} & 1.4s & 5.5m & 2.9m \\
 & Learn to Unlearn \cite{cha2024learning} & 1.3s & 5.1m & 2.9m \\
 & SCRUB \cite{kurmanji2023towards} & 1.2s & 4.7m & 2.9m \\
 & Bad Teacher \cite{chundawat2023can} & 1.2s & 6.0m & 2.9m \\
 & SalUn \cite{fan2023salun} & 1.5s & 6.3m & 3.0m \\
 & DELETE \cite{zhou2025decoupled} & 1.3s & 4.9m & 2.9m \\
\midrule
\multirow{9}{*}{ViT-B/16} & Finetune \cite{golatkar2020eternal} & 1.5s & 3.4m & 2.7m \\
 & Negative Gradient \cite{golatkar2020eternal} & 1.2s & 3.5m & 2.3m \\
 & Negative Gradient+ \cite{kurmanji2023towards} & 1.2s & 13.3m & 2.3m \\
 & Random Label \cite{hayase2020selective} & 1.5s & 4.5m & 2.3m \\
 & Learn to Unlearn \cite{cha2024learning} & 1.2s & 4.1m & 2.3m \\
 & SCRUB \cite{kurmanji2023towards} & 1.7s & 4.1m & 2.3m \\
 & Bad Teacher \cite{chundawat2023can} & 1.2s & 4.6m & 2.3m \\
 & SalUn \cite{fan2023salun} & 1.3s & 3.8m & 2.3m \\
 & DELETE \cite{zhou2025decoupled} & 1.2s & 4.2m & 2.3m \\
\bottomrule
\end{tabular}
}
\end{table}

Although SFRA considers relatively large candidate pools, candidate generation and confidence evaluation consist primarily of batched matrix operations and are therefore highly parallelizable on a GPU. Moreover, only a small fraction of the generated candidates is retained: for each retain class, SFRA selects $M$ high-confidence retain probes and $M$ low-confidence boundary probes from candidate pools of size $N$. Consequently, the synthetic dataset used for classifier-head optimization is substantially smaller than the generated candidate pools. The subsequent relearning stage is also lightweight because the encoder remains frozen and only the existing classifier head is updated. As reported in Table~\ref{tab:probe_generation_timing_architectures}, full probe construction requires $0.86$--$1.7$ seconds for CIFAR-10, $3.4$--$13.3$ minutes for CIFAR-100, and $1.7$--$3.2$ minutes for TinyImageNet across the evaluated methods and backbones. These measurements cover probe construction only and exclude classifier-head relearning. Overall, SFRA avoids encoder-level optimization, maintains a comparatively small synthetic relearning set, and can be parallelized across target classes or checkpoints.

\section{Post-hoc Synthetic--Real Alignment}
\label{app:alignment_analysis}
Proposition~1 identifies the alignment between the mean synthetic forget probes and the mean real forget class representation as a principal geometric term governing the expected single-step change in the forget class margin. To empirically examine this term, we perform a post-hoc analysis using CIFAR-10 with ResNet-18. For each unlearned checkpoint and designated forget class $c_f$, we extract the classifier-input representations of all real test samples belonging to $c_f$ using the frozen encoder. We denote their mean by
$\mu_{\mathcal{E}_f} = \frac{1}{|\mathcal{E}_f|} \sum_{z\in \mathcal{E}_f} z.$  These real forget samples are used exclusively for this post-hoc diagnostic and are never used for synthetic-probe construction, relearning, or hyperparameter selection. We independently reproduce the synthetic forget-probe generation used by our proposed SFRA. Specifically, Gaussian candidates are sampled in the classifier-input space and filtered using the released classifier head, after which the selected low-confidence probes are combined to form $\mathcal{S}_f$. Their mean is
$ \mu_{\mathcal{S}_f} = \frac{1}{|\mathcal{S}_f|} \sum_{s\in \mathcal{S}_f} s.$ We evaluate the signed inner-product alignment and cosine-normalized counterpart
\begin{equation}
    A_{\mathrm{IP}}
    =
    \mu_{\mathcal{S}_f}^{\top}\mu_{\mathcal{E}_f},  ~~~~~~~
    A_{\cos}
    =
    \frac{\mu_{\mathcal{S}_f}^{\top}\mu_{\mathcal{E}_f}}
    {\|\mu_{\mathcal{S}_f}\|_2\|\mu_{\mathcal{E}_f}\|_2}.
\end{equation}

Table~\ref{tab:synthetic_real_alignment_per_class} reports the alignment quantities and $\mathrm{RS}$ separately for each of the ten CIFAR-10 forget classes. We report the per-class results rather than only an aggregate mean$\pm$std because the alignment term in Proposition~1 is inherently class-dependent and exhibits substantial variation across designated forget classes.

Several observations emerge from the per-class analysis. First, the retrained reference exhibits consistently negative synthetic--real alignment across all forget classes, whereas the unlearned models generally produce alignment values that are closer to zero or positive. This indicates a systematic difference between the geometry induced by the released unlearned models and that of the retrained reference.

Second, the magnitude of the measured alignment does not exhibit a monotonic relationship with final $\mathrm{RS}$. For example, Negative Gradient+ shows positive alignment for several forget classes, including comparatively strong alignment for some classes, while its final $\mathrm{RS}$ remains relatively low. Conversely, Negative Gradient obtains consistently higher $\mathrm{RS}$ despite alignment values concentrated close to zero. Thus, the per-class results reveal that alignment is method- and class-dependent rather than a standalone predictor of final recoverability.

\begin{table}[t]
\centering
\caption{Per-forget-class synthetic--real alignment and $\mathrm{RS}$ values on CIFAR-10 using ResNet-18. Each forget class column corresponds to a separate unlearned checkpoint in which that class is designated for forgetting. $A_{\mathrm{IP}}$ and $A_{\mathrm{cos}}$ denote inner-product and cosine alignment, respectively.}
\label{tab:synthetic_real_alignment_per_class}
\scriptsize
\setlength{\tabcolsep}{2pt}
\renewcommand{\arraystretch}{0.82}
\resizebox{\columnwidth}{!}{%
\begin{tabular}{l|l|cccccccccc}
\toprule
\multirow{2}{*}{Unlearning Method} & \multirow{2}{*}{Metric} & \multicolumn{10}{c}{Forget Class} \\
 & & 0 & 1 & 2 & 3 & 4 & 5 & 6 & 7 & 8 & 9 \\
\midrule
\multirow{3}{*}{Retrained} & $\mathrm{RS}$ & $0.55$ & $0.22$ & $0.63$ & $0.58$ & $0.69$ & $0.31$ & $0.52$ & $0.49$ & $0.43$ & $0.36$ \\
 & $A_{\mathrm{IP}}$ & $-1.75$ & $-1.06$ & $-1.63$ & $-1.54$ & $-1.66$ & $-1.74$ & $-1.90$ & $-1.72$ & $-1.51$ & $-1.37$ \\
 & $A_{\mathrm{cos}}$ & $-0.70$ & $-0.41$ & $-0.73$ & $-0.65$ & $-0.74$ & $-0.64$ & $-0.72$ & $-0.68$ & $-0.60$ & $-0.51$ \\
\midrule
\multirow{3}{*}{Finetune \cite{golatkar2020eternal}} & $\mathrm{RS}$ & $0.58$ & $0.23$ & $0.61$ & $0.46$ & $0.57$ & $0.27$ & $0.70$ & $0.56$ & $0.42$ & $0.27$ \\
 & $A_{\mathrm{IP}}$ & $-0.22$ & $+0.13$ & $-0.17$ & $-0.22$ & $-0.31$ & $-0.35$ & $-0.23$ & $+0.02$ & $-0.05$ & $-0.02$ \\
 & $A_{\mathrm{cos}}$ & $-0.09$ & $+0.04$ & $-0.08$ & $-0.09$ & $-0.13$ & $-0.14$ & $-0.10$ & $+0.01$ & $-0.02$ & $-0.01$ \\
\midrule
\multirow{3}{*}{Negative Gradient \cite{golatkar2020eternal}} & $\mathrm{RS}$ & $0.61$ & $0.76$ & $0.56$ & $0.59$ & $0.68$ & $0.65$ & $0.69$ & $0.71$ & $0.66$ & $0.65$ \\
 & $A_{\mathrm{IP}}$ & $-0.11$ & $+0.08$ & $+0.02$ & $+0.10$ & $+0.08$ & $+0.01$ & $+0.02$ & $-0.01$ & $+0.01$ & $+0.05$ \\
 & $A_{\mathrm{cos}}$ & $-0.07$ & $+0.05$ & $+0.01$ & $+0.06$ & $+0.05$ & $+0.01$ & $+0.01$ & $-0.01$ & $+0.01$ & $+0.03$ \\
\midrule
\multirow{3}{*}{Negative Gradient+ \cite{kurmanji2023towards}} & $\mathrm{RS}$ & $0.31$ & $0.32$ & $0.21$ & $0.46$ & $0.49$ & $0.33$ & $0.48$ & $0.20$ & $0.41$ & $0.46$ \\
 & $A_{\mathrm{IP}}$ & $+0.02$ & $+0.00$ & $+0.26$ & $+0.04$ & $+0.07$ & $+0.51$ & $-0.07$ & $+0.18$ & $+0.13$ & $+0.15$ \\
 & $A_{\mathrm{cos}}$ & $+0.01$ & $+0.00$ & $+0.14$ & $+0.02$ & $+0.04$ & $+0.22$ & $-0.05$ & $+0.09$ & $+0.08$ & $+0.10$ \\
\midrule
\multirow{3}{*}{Random Label \cite{hayase2020selective}} & $\mathrm{RS}$ & $0.73$ & $0.80$ & $0.66$ & $0.71$ & $0.76$ & $0.72$ & $0.74$ & $0.76$ & $0.72$ & $0.77$ \\
 & $A_{\mathrm{IP}}$ & $-0.16$ & $+0.06$ & $-0.02$ & $+0.06$ & $+0.07$ & $-0.06$ & $-0.02$ & $+0.02$ & $+0.02$ & $+0.06$ \\
 & $A_{\mathrm{cos}}$ & $-0.10$ & $+0.04$ & $-0.01$ & $+0.03$ & $+0.05$ & $-0.04$ & $-0.01$ & $+0.01$ & $+0.01$ & $+0.04$ \\
\midrule
\multirow{3}{*}{Boundary Shrink \cite{chen2023boundary}} & $\mathrm{RS}$ & $0.71$ & $0.80$ & $0.63$ & $0.67$ & $0.77$ & $0.70$ & $0.74$ & $0.75$ & $0.72$ & $0.77$ \\
 & $A_{\mathrm{IP}}$ & $-0.15$ & $+0.06$ & $-0.04$ & $+0.05$ & $+0.09$ & $-0.05$ & $-0.02$ & $+0.02$ & $+0.03$ & $+0.08$ \\
 & $A_{\mathrm{cos}}$ & $-0.09$ & $+0.03$ & $-0.02$ & $+0.03$ & $+0.06$ & $-0.03$ & $-0.01$ & $+0.01$ & $+0.02$ & $+0.05$ \\
\midrule
\multirow{3}{*}{Learn to Unlearn \cite{cha2024learning}} & $\mathrm{RS}$ & $0.67$ & $0.68$ & $0.59$ & $0.56$ & $0.70$ & $0.70$ & $0.68$ & $0.71$ & $0.68$ & $0.68$ \\
 & $A_{\mathrm{IP}}$ & $-0.15$ & $+0.09$ & $-0.02$ & $+0.08$ & $+0.08$ & $+0.02$ & $+0.03$ & $+0.01$ & $+0.01$ & $+0.05$ \\
 & $A_{\mathrm{cos}}$ & $-0.09$ & $+0.05$ & $-0.01$ & $+0.05$ & $+0.05$ & $+0.01$ & $+0.02$ & $+0.01$ & $+0.01$ & $+0.03$ \\
\midrule
\multirow{3}{*}{SCRUB \cite{kurmanji2023towards}} & $\mathrm{RS}$ & $0.64$ & $0.34$ & $0.61$ & $0.51$ & $0.61$ & $0.35$ & $0.57$ & $0.40$ & $0.41$ & $0.42$ \\
 & $A_{\mathrm{IP}}$ & $+0.36$ & $+1.44$ & $+1.25$ & $+0.33$ & $+0.47$ & $+1.39$ & $+0.22$ & $+1.02$ & $+0.34$ & $+0.48$ \\
 & $A_{\mathrm{cos}}$ & $+0.20$ & $+0.42$ & $+0.39$ & $+0.16$ & $+0.23$ & $+0.41$ & $+0.12$ & $+0.31$ & $+0.18$ & $+0.25$ \\
\midrule
\multirow{3}{*}{Bad Teacher \cite{chundawat2023can}} & $\mathrm{RS}$ & $0.97$ & $0.98$ & $0.79$ & $0.96$ & $0.94$ & $0.97$ & $0.98$ & $0.98$ & $0.98$ & $0.97$ \\
 & $A_{\mathrm{IP}}$ & $-0.06$ & $-0.03$ & $-0.05$ & $-0.04$ & $-0.04$ & $-0.02$ & $-0.02$ & $-0.08$ & $-0.02$ & $-0.04$ \\
 & $A_{\mathrm{cos}}$ & $-0.13$ & $-0.06$ & $-0.14$ & $-0.09$ & $-0.15$ & $-0.03$ & $-0.04$ & $-0.16$ & $-0.04$ & $-0.07$ \\
\midrule
\multirow{3}{*}{SalUn \cite{fan2023salun}} & $\mathrm{RS}$ & $0.83$ & $0.76$ & $0.85$ & $0.79$ & $0.84$ & $0.78$ & $0.85$ & $0.87$ & $0.82$ & $0.84$ \\
 & $A_{\mathrm{IP}}$ & $-0.07$ & $+0.09$ & $-0.14$ & $-0.03$ & $-0.12$ & $-0.08$ & $-0.10$ & $-0.02$ & $-0.09$ & $-0.06$ \\
 & $A_{\mathrm{cos}}$ & $-0.04$ & $+0.05$ & $-0.09$ & $-0.02$ & $-0.08$ & $-0.06$ & $-0.06$ & $-0.01$ & $-0.05$ & $-0.04$ \\
\midrule
\multirow{3}{*}{DELETE \cite{zhou2025decoupled}} & $\mathrm{RS}$ & $0.96$ & $0.97$ & $0.95$ & $0.93$ & $0.97$ & $0.93$ & $0.97$ & $0.96$ & $0.97$ & $0.96$ \\
 & $A_{\mathrm{IP}}$ & $-0.01$ & $+0.01$ & $-0.08$ & $-0.03$ & $+0.05$ & $-0.02$ & $+0.07$ & $+0.05$ & $+0.13$ & $-0.02$ \\
 & $A_{\mathrm{cos}}$ & $-0.01$ & $+0.01$ & $-0.06$ & $-0.02$ & $+0.04$ & $-0.01$ & $+0.05$ & $+0.04$ & $+0.11$ & $-0.02$ \\
\bottomrule
\end{tabular}%
}
\end{table}

These observations are consistent with Proposition~1. The proposition depends on the complete class-dependent quantity $\mu_{\mathcal{S}_f}^{\top}\mu_{\mathcal{E}_f}+r_j$, whereas the present experiment measures only its directly observable principal alignment term. Consequently, a non-positive value of $\mu_{\mathcal{S}_f}^{\top}\mu_{\mathcal{E}_f}$ does not imply that the sufficient condition fails, since the residual term $r_j$ may be positive. Likewise, positive alignment alone does not establish the complete sufficient condition without estimating $r_j$.

Moreover, Proposition~1 characterizes a sufficient condition for an expected single-step increase in the forget class margin, whereas $\mathrm{RS}$ is measured after iterative classifier-head relearning. We therefore interpret the alignment experiment as a post-hoc geometric diagnostic of the principal term appearing in Proposition~1, rather than as a claim that alignment alone determines the final $\mathrm{RS}$.

\section{Empirical Assessment of the Margin Approximation}
\label{app:margin_approximation}
\begin{figure}[h]
    \centering
    \includegraphics[width=\columnwidth]
    {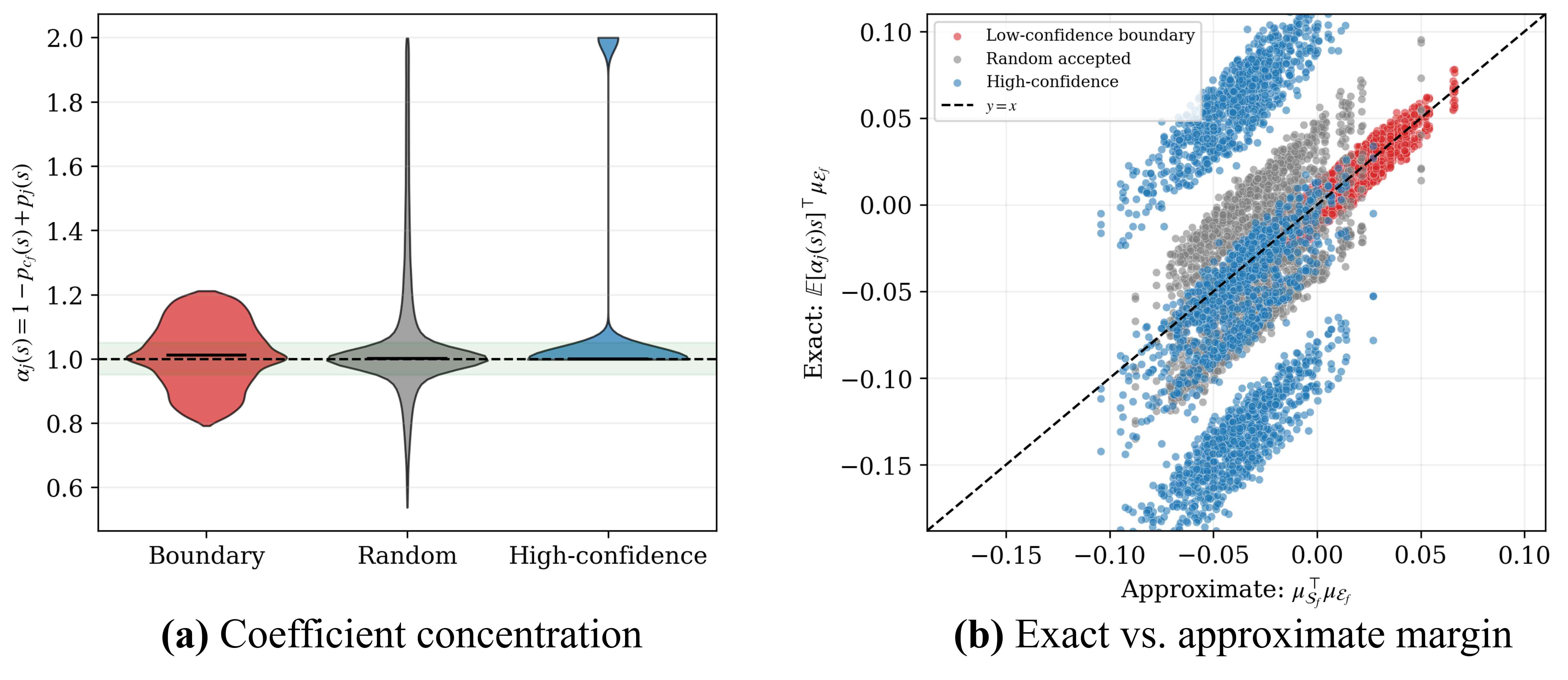}
    \caption{Empirical assessment of the margin approximation.
    Results for CIFAR-10 with ResNet-18, Bad Teacher, and class $7$ as the forget class. (a) Distribution of
    $\alpha_j(s)=1-p_{c_f}(s)+p_j(s)$ across selected probes and retain class competitors; the dashed line denotes one. (b) Exact probability-weighted margin expression versus the approximation based on the unweighted synthetic mean over bootstrap samples. The dashed diagonal denotes perfect agreement. The learning-rate factor is omitted because it scales both expressions
    equally.}
    \label{fig:coefficient_validation}
\end{figure}

We empirically assess the approximation used in Proposition~1 %\ref{prop:margin_increase}
 on CIFAR-10 with a ResNet-18 backbone, Bad Teacher unlearning, and class $7$ as the forget class. We generate $500{,}000$ accepted Gaussian embeddings per retain class and select the $500$ lowest-confidence embeddings from each retain class, yielding $4{,}500$ synthetic forget probes.
Figure~\ref{fig:coefficient_validation}~(a) shows the distribution of $\alpha_j(s)=1-p_{c_f}(s)+p_j(s)$. Although $\alpha_j(s)$ is not pointwise constant, its distribution is centered close to one, with mean $1.013$. Figure~\ref{fig:coefficient_validation}~(b) compares the exact weighted margin expression,
$\mathbb{E}_{s\sim\mathcal{S}_f}
[\alpha_j(s)s]^\top\mu_{\mathcal{E}_f}$,
with the approximation based on the unweighted synthetic mean,
$\mu_{\mathcal{S}_f}^{\top}\mu_{\mathcal{E}_f}.$
Across bootstrap samples and retain class competitors, the two expressions achieve Pearson correlation $r=0.939$, Spearman correlation $\rho=0.934$, regression slope $1.005$, and $88.9\%$ sign agreement. These results support the approximation in aggregate for this evaluated setting; the exact residual-based condition remains the formal statement.

\section{Confidence of Forget Class Assignments}
\label{app:forget_assignment_confidence}
To further examine whether real forget class samples are mapped to retain classes with low or high confidence, we analyze the prediction confidence of the unlearned model on the real test set. For each single-class unlearning checkpoint, we first evaluate all test samples using the unlearned model. Then, for each forget class, we compute two weighted average confidence values. The first value is the average confidence assigned to correctly classified retain samples. Specifically, for each retain class, we consider real retain samples whose ground-truth label and predicted label both match that retain class, and average the corresponding softmax confidence. The second value is the average confidence assigned to real forget class samples that are predicted as one of the retain classes. The averages are weighted by the number of samples assigned to each retain class. Table~\ref{tab:bad_teacher_cifar10_resnet18_weighted_confidence} shows the results for Bad Teacher on CIFAR-10 with a ResNet-18. Across forget classes, correctly classified retain samples receive high confidence, while forget class samples assigned to retain classes receive substantially lower confidence. This indicates that the unlearned model does not assign forget class samples to retain classes with the same confidence as genuine retain samples, suggesting that forget samples remain relatively uncertain under the unlearned classifier. This observation supports our probe-selection strategy: for each retain class, we treat low-confidence synthetic samples assigned to that retain class as candidate forget class probes, since real forget class samples assigned to retain classes also tend to receive lower confidence than genuine correctly classified retain samples.

\begin{table}[t]
\centering
\caption{Average confidence comparison for Bad Teacher on CIFAR-10 for ResNet-18 backbone. For each forget class, we report the weighted average confidence of correctly classified retain samples and forget-class samples assigned to retain classes.}
\label{tab:bad_teacher_cifar10_resnet18_weighted_confidence}
\footnotesize
\setlength{\tabcolsep}{3pt}
\renewcommand{\arraystretch}{1.0}
\begin{tabular}{c|ccc}
\toprule
\begin{tabular}[c]{@{}c@{}}Forget\\Class\end{tabular}
& \begin{tabular}[c]{@{}c@{}}Correct\\Retain\\Conf.\end{tabular}
& \begin{tabular}[c]{@{}c@{}}Forget\\Assigned\\Conf.\end{tabular}
& \begin{tabular}[c]{@{}c@{}}Gap\end{tabular} \\
\midrule
$0$ & $0.974$ & $0.247$ & $0.727$ \\
$1$ & $0.976$ & $0.230$ & $0.746$ \\
$2$ & $0.860$ & $0.362$ & $0.498$ \\
$3$ & $0.974$ & $0.294$ & $0.680$ \\
$4$ & $0.975$ & $0.230$ & $0.745$ \\
$5$ & $0.973$ & $0.284$ & $0.689$ \\
$6$ & $0.975$ & $0.238$ & $0.737$ \\
$7$ & $0.975$ & $0.244$ & $0.731$ \\
$8$ & $0.975$ & $0.230$ & $0.745$ \\
$9$ & $0.976$ & $0.248$ & $0.728$ \\
\midrule
\textbf{Average} & $0.963$ & $0.261$ & $0.703$ \\
\bottomrule
\end{tabular}
\end{table}

\section{SFRA Without the Released Forget Class Output Row}
\label{app:forget_row_ablation}

The standard SFRA formulation assumes that the released classifier retains an output corresponding to the known forget class. We examine whether this assumption can be relaxed when the corresponding output row has been removed. In this setting, we restore the missing output using a randomly initialized weight vector and bias and then apply the same audit procedure. Table~\ref{tab:forget_row_ablation_cifar10_resnet18_fg7} compares this setting with SFRA using the forget class output row provided by the unlearned checkpoint. The results show that SFRA does not require access to the learned parameters of the unlearned forget class output row, although the identity of the forget class must remain known.

\begin{table}[h]
\centering
\caption{Robustness of SFRA to removal of the forget class output row on CIFAR-10 with ResNet-18 and forget class~7. Here, $w_f$ denotes the forget class output-row parameters: Unlearned $w_f$ uses the row from the released unlearned checkpoint, whereas Random $w_f$ restores a missing row using random initialization before probe generation and relearning. All other audit settings are fixed.}
\label{tab:forget_row_ablation_cifar10_resnet18_fg7}
\resizebox{\columnwidth}{!}{%
\begin{tabular}{l|cc|cc|cc}
\toprule
\multirow{2}{*}{Unlearning Method} & \multicolumn{2}{c|}{$\mathcal{A}_r^t(\%)$} & \multicolumn{2}{c|}{$\mathcal{A}_f^t(\%)$} & \multicolumn{2}{c}{$\mathrm{RS}$} \\
 & Unlearned $w_f$ & Random $w_f$ & Unlearned $w_f$ & Random $w_f$ & Unlearned $w_f$ & Random $w_f$ \\
\midrule
Retrained & 92.22 & 89.96 & 37.70 & 48.80 & 0.54 & 0.65 \\
Finetune \cite{golatkar2020eternal} & 92.77 & 91.93 & 38.70 & 41.60 & 0.56 & 0.58 \\
Negative Gradient \cite{golatkar2020eternal} & 90.43 & 90.49 & 64.50 & 65.00 & 0.72 & 0.78 \\
Negative Gradient+ \cite{kurmanji2023towards} & 85.72 & 85.76 & 7.90 & 11.90 & 0.14 & 0.21 \\
Random Label \cite{hayase2020selective} & 92.04 & 91.79 & 75.10 & 75.90 & 0.77 & 0.86 \\
Boundary Shrink \cite{chen2023boundary} & 91.98 & 91.38 & 76.60 & 76.90 & 0.78 & 0.86 \\
Learn to Unlearn \cite{cha2024learning} & 90.29 & 90.04 & 67.50 & 68.40 & 0.74 & 0.81 \\
SCRUB \cite{kurmanji2023towards} & 84.62 & 91.90 & 35.90 & 17.30 & 0.52 & 0.29 \\
Bad Teacher \cite{chundawat2023can} & 92.77 & 92.88 & 98.70 & 98.80 & 0.98 & 0.98 \\
SalUn \cite{fan2023salun} & 87.81 & 87.31 & 90.80 & 89.70 & 0.88 & 0.92 \\
DELETE \cite{zhou2025decoupled} & 91.61 & 89.82 & 94.90 & 95.10 & 0.96 & 0.95 \\
\bottomrule
\end{tabular}%
}
\end{table}

\section{Sensitivity to the Number of Synthetic Probs}
\label{app:sensitivity}
\begin{figure}[h]
  \centering
  \includegraphics[width=0.95\columnwidth]{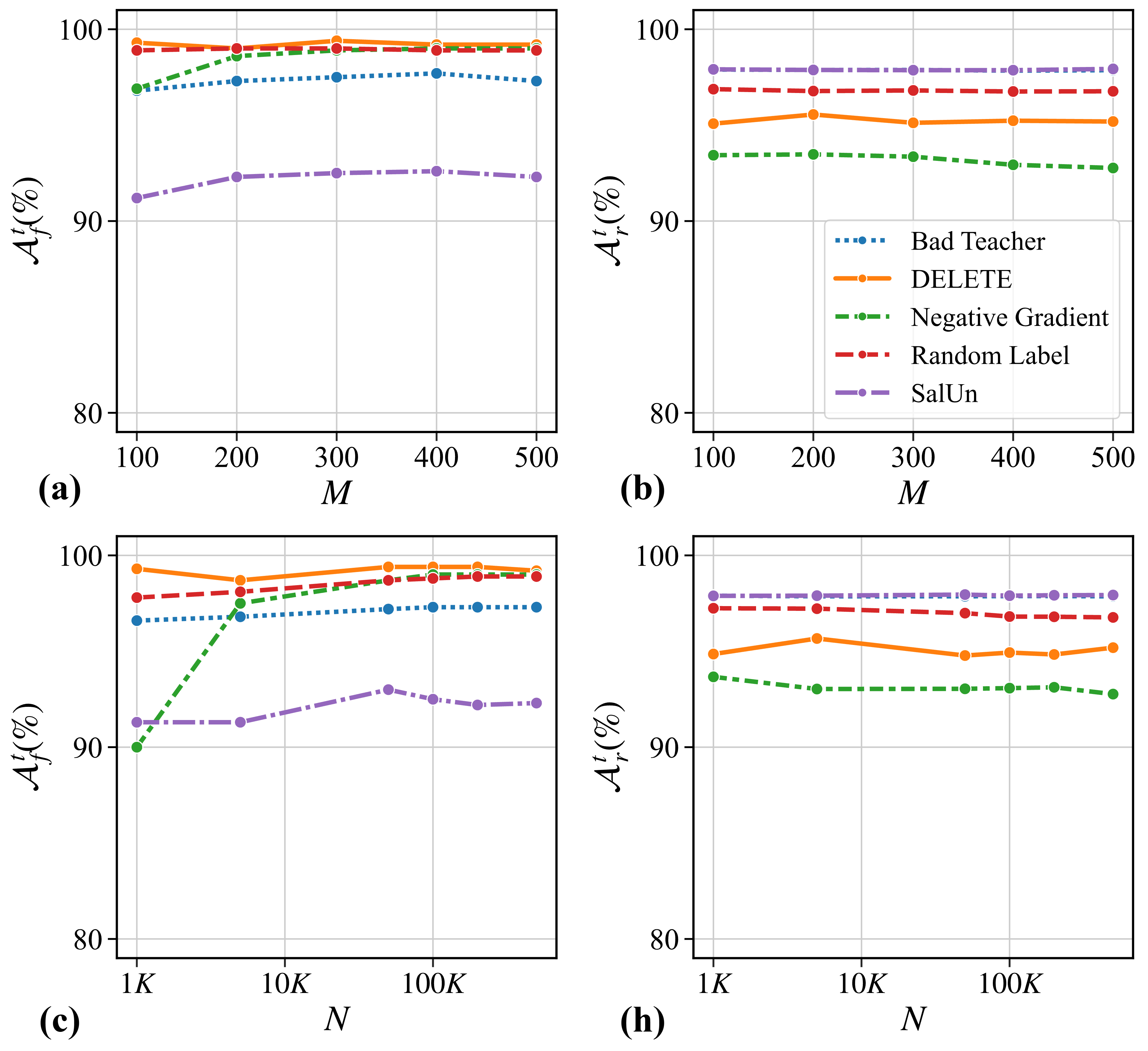}
  \caption{Impact of the number of embeddings on SFRA for CIFAR-10 with a ViT-B-16 under single-class unlearning of class~$9$. \textbf{(a,b)} Effect of varying \(M\) with \(N=500K\) on $\mathcal{A}_f^t$ and $\mathcal{A}_r^t$. \textbf{(c,d)} Effect of varying \(N\) with \(M=500\) on $\mathcal{A}_f^t$ and $\mathcal{A}_r^t$.}
  \label{fig:acc_mn_vit-b-16}
\end{figure}

ViT-B/16 exhibits lower sensitivity to both the number of selected embeddings \(M\) and the number of generated embeddings \(N\) compared with ResNet-18, as shown in Fig.~\ref{fig:acc_mn_vit-b-16}. When \(N\) is fixed, the retain accuracy ($\mathcal{A}_r^{t}$) remains stable within ($\pm 1\%$), while the forget accuracy ($\mathcal{A}_f^{t}$) increases and reaches saturation at smaller values of \(M\). When \(M\) is fixed, increasing \(N\) yields more consistent gains in ($\mathcal{A}_f^{t}$), suggesting that a larger candidate pool improves the chance of mining useful boundary probes. Overall, these results indicate that relearning on ViT-B/16 is stable and does not require aggressive tuning of \(M\) or \(N\).

\section{Retain--Forget Accuracy Trade-off}
\label{app:retain_forget_tradeoff}
\begin{figure}[h]
    \centering

    \begin{subfigure}{\columnwidth}
        \centering
        \includegraphics[width=\columnwidth]
        {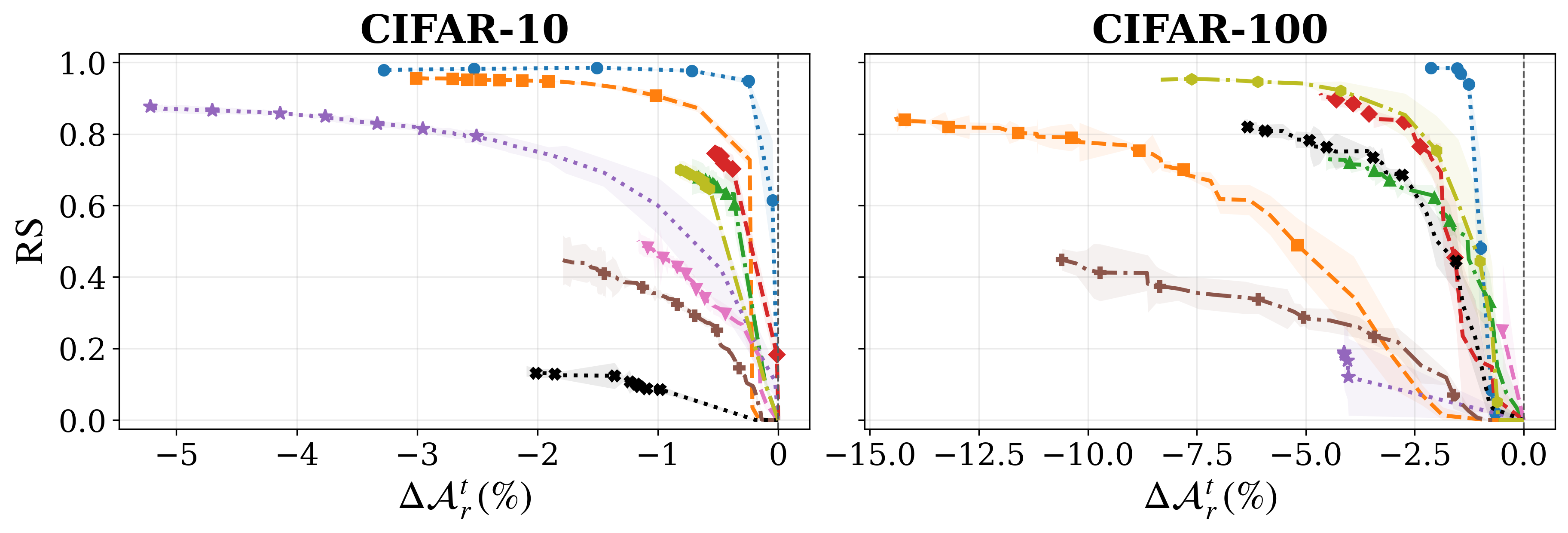}
        \caption{ResNet-18}
        \label{fig:retain_delta_RS_resnet18}
    \end{subfigure}

    \vspace{2mm}

    \begin{subfigure}{\columnwidth}
        \centering
        \includegraphics[width=\columnwidth]
        {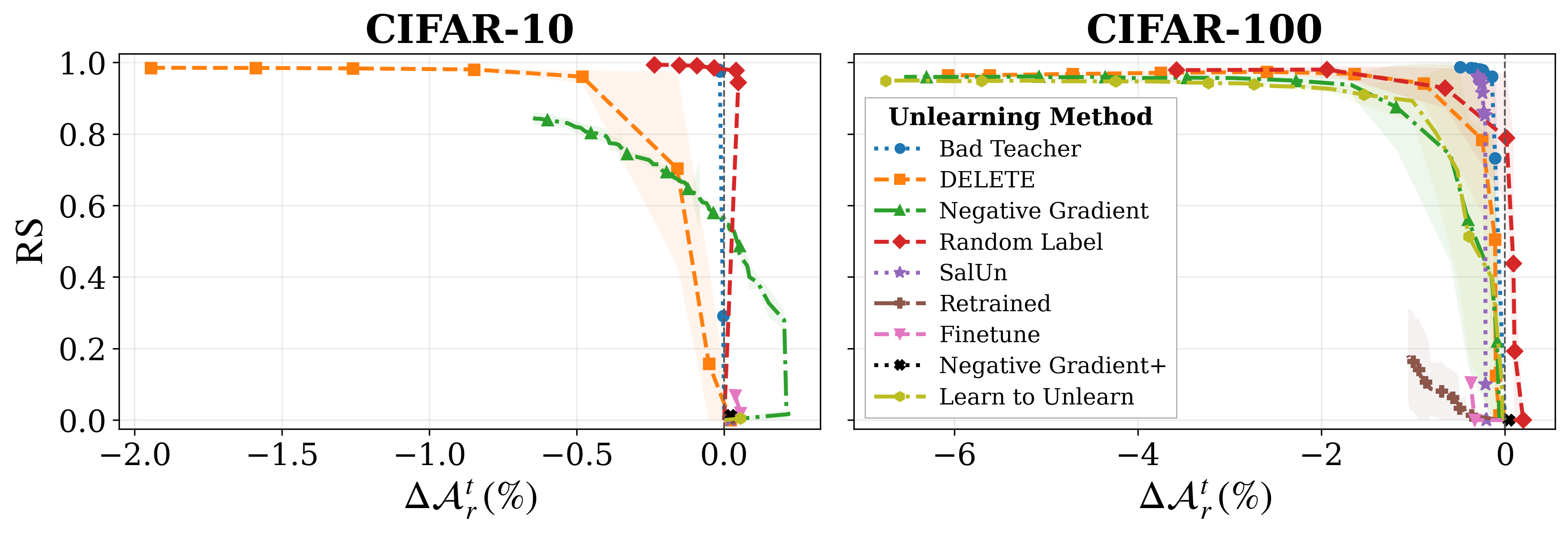}
        \caption{ViT-B/16}
        \label{fig:retain_delta_RS_vit}
    \end{subfigure}

    \begin{subfigure}{\columnwidth}
        \centering
        \includegraphics[width=\columnwidth]
        {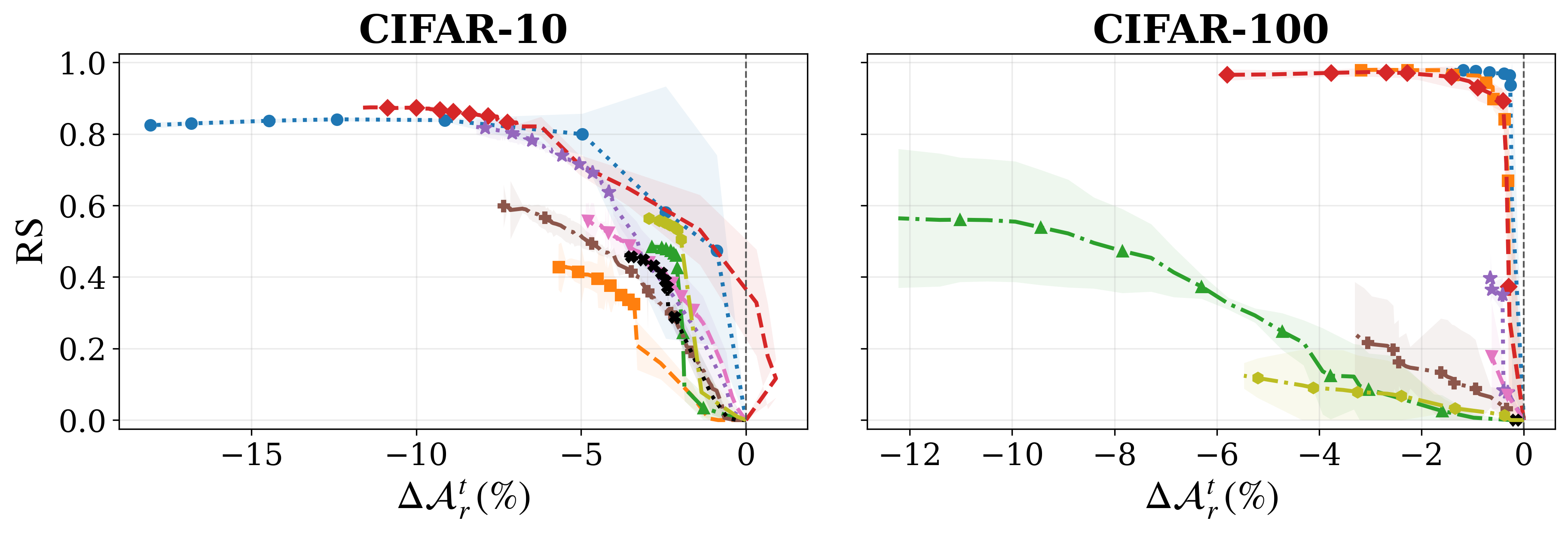}
        \caption{Swin-T}
        \label{fig:retain_delta_RS_swin-t}
    \end{subfigure}
    \caption{SFRA recoverability--utility trade-off on CIFAR-10 and CIFAR-100 using (a) ResNet-18, (b) ViT-B/16, and (c) Swin-T. We plot $\mathrm{RS}$  against the change in retain accuracy $\Delta\mathcal{A}_r=\mathcal{A}_r^{re}-\mathcal{A}_r^{un}$ along Pareto-efficient relearning checkpoints. Curves show the mean across three random seeds, with shaded regions indicating one standard deviation.}

    \label{fig:retain_delta_RS}
\end{figure}

To further examine the trade-off between forget class recovery and retain class preservation during SFRA, we analyze how the $\mathrm{RS}$  evolves as retain accuracy changes throughout classifier-head relearning. Figure~\ref{fig:retain_delta_RS} reports $\mathrm{RS}$  against $\Delta\mathcal{A}_r=\mathcal{A}_r^{re}-\mathcal{A}_r^{un}$ on CIFAR-10 and CIFAR-100 for ResNet-18, ViT-B/16, and Swin-T, where negative values of $\Delta\mathcal{A}_r^t$ indicate degradation relative to the unlearned checkpoint. The outlined initial point corresponds to the unlearned checkpoint, and the connected points trace the Pareto-efficient checkpoints obtained during SFRA.

Figure~\ref{fig:retain_delta_RS} reveals a clear but method-dependent trade-off between recoverability and retain class preservation. For several unlearning methods, $\mathrm{RS}$  increases substantially while $\Delta\mathcal{A}_r^t$ remains close to zero, indicating that considerable forget class recovery can be achieved with only limited loss of retain accuracy. In contrast, other methods exhibit appreciable increases in $\mathrm{RS}$  only after a larger decrease in retain accuracy, indicating a less favorable recovery--utility trade-off. The trajectories also tend to saturate: beyond a certain point, additional relearning provides limited improvement in $\mathrm{RS}$  while retain accuracy continues to decrease. This behavior motivates considering intermediate Pareto-efficient checkpoints rather than simply extending relearning for more epochs. Overall, the results show that high source-free recoverability is not necessarily a consequence of sacrificing retain performance, although the extent of this trade-off varies across unlearning methods, datasets, and backbone architectures.

The trade-off curves also enable evaluating recoverability under a user-specified retain-utility budget. In particular, one can define a maximum acceptable degradation $\epsilon$ in retain accuracy and restrict the analysis to checkpoints satisfying $\Delta\mathcal{A}_r^t \geq -\epsilon$. The largest $\mathrm{RS}$  attained within this region then quantifies how much relearning is achievable without exceeding the prescribed loss in retain performance. For example, setting $\epsilon=0.05$ evaluates the strongest recoverability attainable while allowing at most a five-percentage-point decrease in retain accuracy. This provides a utility-aware interpretation of SFRA: rather than considering recovery in isolation, one can assess how much forget class information can be recovered under a specified tolerance for retain class degradation.

\section{Additional Results and Details for Single-Class and Multi-Class SFRA}
\label{app:additional_results}
This section presents class unlearning methods and the additional single-class results for the remaining backbone architectures, as well as multi-class unlearning and relearning results for different numbers of forget classes.

\noindent \textbf{Class Unlearning Baselines:}
To clarify the class-unlearning methods evaluated in our relearning audit, we briefly describe the ten baselines used in our experiments. Finetune~\cite{golatkar2020eternal} fine-tunes the model using only retain data. Negative Gradient~\cite{golatkar2020eternal} performs gradient ascent on the forget-set loss to reduce performance on the designated forget samples. Negative Gradient+~\cite{kurmanji2023towards} combines gradient ascent on the forget-set loss with gradient descent on the retain-set loss to preserve utility. Random Label~\cite{hayase2020selective} replaces the labels of forget samples with randomly selected retain class labels. Boundary Shrink~\cite{chen2023boundary} relabels each forget sample as its nearest incorrect class, thereby shifting the corresponding decision boundary. Learn to Unlearn~\cite{cha2024learning} applies adversarial perturbations to forget samples to induce forgetting while maintaining retain class performance without requiring retain data. SCRUB~\cite{kurmanji2023towards} combines cross-entropy supervision and teacher--student distillation on retain data with negative distillation on forget data, encouraging the student to diverge from the teacher on the forget set. Bad Teacher~\cite{chundawat2023can} trains the student to match a competent teacher on retain samples and an incompetent teacher on forget samples. Saliency Unlearning (SalUn)~\cite{fan2023salun} combines weight saliency with random relabeling, updating only parameters identified as important for forgetting while freezing the remaining parameters. Finally, DELETE~\cite{zhou2025decoupled} masks the forget class logit and distills the remaining soft predictions from a frozen teacher model to preserve performance on the retain classes.

\noindent \textbf{Single-Class SFRA:}  While the main paper reports the ResNet-18 results, Tables~\ref{tab:ViT-B_16_single_class_all_datasets} and \ref{tab:Swin-T_single_class_all_datasets}, 
report the corresponding results for ViT-B/16 and Swin-T respectively.  All experiments follow the same protocol, datasets, evaluation metrics, and synthetic prob generation settings used in the main paper. For CIFAR-10, we evaluate all 10 forget classes across all three backbones (ResNet-18, ViT-B/16, and Swin-T). For CIFAR-100, we evaluate 10 designated forget classes $\{0,10,20,30,40,50,60,70,80,90\}$ with ResNet-18, ViT-B/16 and Swin-T. For TinyImageNet, we evaluate 10 designated forget classes $\{0,20,40,60,80,100,120,140,160,180\}$ with ResNet-18, ViT-B/16 and Swin-T. These results show that SFRA is not specific to a backbone. At the same time, the magnitude of $\mathrm{RS}$  varies across backbones and unlearning methods, indicating that different architectures preserve or disrupt forget class geometry to different degrees after unlearning.

\noindent \textbf{Multi-Class SFRA:} We extend our proposed SFRA from a single forget class to a set of forget classes $\mathcal{Y}_f$, with $\mathcal{Y}_r=\mathcal{Y}\setminus\mathcal{Y}_f$. The main challenge in this setting is that a low-confidence synthetic probe does not have a natural label indicating which forget class it should represent. We therefore use the unlearned classifier to partition the selected boundary-adjacent probes among the forget classes. For each retain class $c_r\in\mathcal{Y}_r$, we construct a candidate pool $\mathcal{P}_{c_r}$ of $N$ Gaussian embeddings predicted as $c_r$. The $M$ highest-confidence embeddings form the synthetic
retain set for $c_r$, while the $|\mathcal{Y}_f|M$ lowest-confidence embeddings form a shared boundary-probe pool $\mathcal{Q}_{c_r}$. To obtain cardinality-balanced synthetic forget sets, we greedily assign these probes to the forget classes. Specifically, the forget classes are processed in ascending class-index order, and each $c_f\in\mathcal{Y}_f$ receives the $M$ currently unassigned probes with the highest $p_{c_f}(s)$. Thus, each boundary probe is assigned to exactly one forget class and every forget class receives $|\mathcal{Y}_r|M$ synthetic probes. The resulting synthetic sets are $\mathcal{S}_r=\bigcup_{c_r\in\mathcal{Y}_r}\mathcal{S}_{c_r}$ and $\mathcal{S}_f=\bigcup_{c_f\in\mathcal{Y}_f}\mathcal{S}_{c_f}$. As in single-class SFRA, the encoder is frozen and only the classifier head is updated, without using any real data. The complete procedure is summarized in Alg.~\ref{alg:multiclass}.

\begin{table}[t]
\centering
\caption{Comparison of unlearning methods under our proposed SFRA and the source-dependent PRA baseline for 2-class unlearning on CIFAR-10 and CIFAR-100 with ResNet-18. Within each dataset, the highest and second-highest $\mathrm{RS}$ and $\Delta\mathrm{RS}$ values are shown in \textbf{bold} and \underline{underlined}, respectively.}
\label{tab:ResNet-18_multi_class_2}
\vspace{-3mm}
\fontsize{10.5}{10.5}\selectfont
\resizebox{\columnwidth}{!}{%
\begin{tabular}{c|c|cccc|cccc}
\toprule
\toprule
\multirow{2}{*}{Unlearning Method} & \multirow{2}{*}{Model Variant} & \multicolumn{4}{c}{\textbf{CIFAR-10}} & \multicolumn{4}{|c}{\textbf{CIFAR-100}} \\
 &  & $\mathcal{A}^{t}_{r}(\%)$ & $\mathcal{A}^{t}_{f}(\%)$ & $\mathrm{RS}$ & $\Delta\mathrm{RS}$ & $\mathcal{A}^{t}_{r}(\%)$ & $\mathcal{A}^{t}_{f}(\%)$ & $\mathrm{RS}$ & $\Delta\mathrm{RS}$\\
\midrule
\midrule
Original & Original & $94.04$ & $97.10$ & - & - & $79.92$ & $82.00$ & - & - \\
\midrule
\midrule
\multirow{3}{*}{Retrained} & Unlearned & $93.76$ & $0.00$ & - & - & $80.40$ & $0.00$ & - & - \\
 & PRA \cite{ha2025unlearning} & $93.88$ & $2.80$ & $0.05$ & - & $80.02$ & $13.50$ & $0.24$ & - \\
 & SFRA (ours) & $87.61$ & $28.00$ & $0.43$ & - & $80.40$ & $0.00$ & $0.00$ & - \\
\midrule
\multirow{3}{*}{Finetune \cite{golatkar2020eternal}} & Unlearned & $93.19$ & $0.00$ & - & - & $79.90$ & $0.50$ & - & - \\
 & PRA \cite{ha2025unlearning} & $93.05$ & $4.65$ & $0.09$ & $+0.03$ & $79.03$ & $24.00$ & $0.38$ & $+0.14$ \\
 & SFRA (ours) & $87.46$ & $40.60$ & $0.57$ & $+0.14$ & $74.38$ & $29.50$ & $0.44$ & $+0.44$ \\
\midrule
\multirow{3}{*}{Negative Gradient \cite{golatkar2020eternal}} & Unlearned & $89.64$ & $1.30$ & - & - & $69.35$ & $0.00$ & - & - \\
 & PRA \cite{ha2025unlearning} & $89.17$ & $26.40$ & $0.40$ & $+0.35$ & $68.59$ & $54.50$ & $0.70$ & $+0.47$ \\
 & SFRA (ours) & $86.26$ & $13.40$ & $0.22$ & $-0.22$ & $65.36$ & $36.00$ & $0.52$ & $+0.52$ \\
\midrule
\multirow{3}{*}{Negative Gradient+ \cite{kurmanji2023towards}} & Unlearned & $91.64$ & $0.35$ & - & - & $75.73$ & $0.00$ & - & - \\
 & PRA \cite{ha2025unlearning} & $91.12$ & $54.20$ & $0.70$ & $+0.64$ & $75.28$ & $47.50$ & $0.64$ & $+0.41$ \\
 & SFRA (ours) & $85.96$ & $4.10$ & $0.07$ & $-0.36$ & $71.34$ & $17.50$ & $0.30$ & $+0.30$ \\
\midrule
\multirow{3}{*}{Random Label \cite{hayase2020selective}} & Unlearned & $92.11$ & $17.10$ & - & - & $69.19$ & $2.50$ & - & - \\
 & PRA \cite{ha2025unlearning} & $91.47$ & $46.85$ & $0.46$ & $+0.40$ & $68.68$ & $64.00$ & $0.76$ & $+0.52$ \\
 & SFRA (ours) & $86.12$ & $35.65$ & $0.31$ & $-0.12$ & $62.77$ & $63.50$ & $0.74$ & $\underline{+0.74}$ \\
\midrule
\multirow{3}{*}{Learn to Unlearn \cite{cha2024learning}} & Unlearned & $90.67$ & $0.85$ & - & - & $70.75$ & $0.00$ & - & - \\
 & PRA \cite{ha2025unlearning} & $90.28$ & $41.10$ & $0.57$ & $+0.52$ & $69.90$ & $63.00$ & $\underline{0.77}$ & $+0.53$ \\
 & SFRA (ours) & $87.47$ & $8.40$ & $0.14$ & $-0.29$ & $65.41$ & $56.00$ & $0.70$ & $+0.70$ \\
\midrule
\multirow{3}{*}{SCRUB \cite{kurmanji2023towards}} & Unlearned & $93.69$ & $0.00$ & - & - & $79.71$ & $0.00$ & - & - \\
 & PRA \cite{ha2025unlearning} & $93.60$ & $31.25$ & $0.48$ & $+0.42$ & $79.39$ & $47.50$ & $0.64$ & $+0.41$ \\
 & SFRA (ours) & $93.69$ & $0.00$ & $0.00$ & $-0.43$ & $79.71$ & $0.00$ & $0.00$ & $+0.00$ \\
\midrule
\multirow{3}{*}{Bad Teacher \cite{chundawat2023can}} & Unlearned & $94.25$ & $0.00$ & - & - & $79.69$ & $0.00$ & - & - \\
 & PRA \cite{ha2025unlearning} & $93.26$ & $59.00$ & $\underline{0.74}$ & $\underline{+0.68}$ & $79.21$ & $73.00$ & \textbf{\boldmath $0.84$} & $+0.60$ \\
 & SFRA (ours) & $90.54$ & $52.80$ & $0.68$ & $+0.25$ & $75.67$ & $61.00$ & $0.75$ & $\mathbf{+0.75}$ \\
\midrule
\multirow{3}{*}{SalUn \cite{fan2023salun}} & Unlearned & $93.12$ & $0.70$ & - & - & $79.03$ & $4.50$ & - & - \\
 & PRA \cite{ha2025unlearning} & $92.41$ & $31.75$ & $0.47$ & $+0.42$ & $78.20$ & $49.50$ & $0.62$ & $+0.38$ \\
 & SFRA (ours) & $84.79$ & $43.30$ & $0.58$ & $+0.15$ & $77.48$ & $35.50$ & $0.47$ & $+0.47$ \\
\midrule
\multirow{3}{*}{DELETE \cite{zhou2025decoupled}} & Unlearned & $94.42$ & $0.00$ & - & - & $78.87$ & $0.00$ & - & - \\
 & PRA \cite{ha2025unlearning} & $93.79$ & $78.30$ & \textbf{\boldmath $0.88$} & $\mathbf{+0.82}$ & $78.62$ & $40.00$ & $0.57$ & $+0.33$ \\
 & SFRA (ours) & $85.12$ & $17.35$ & $0.29$ & $-0.14$ & $71.71$ & $11.00$ & $0.20$ & $+0.20$ \\
\midrule
\bottomrule
\end{tabular}
}
\end{table}

We provide additional results for the 2-class unlearning setting on CIFAR-10 and CIFAR-100 using a ResNet-18 backbone. We randomly select $\mathcal{Y}_f=\{1,6\}$ for CIFAR-10, while for CIFAR-100 we use $\mathcal{Y}_f=\{25,58\}$, $\{25,58,38,23,96\}$, and $\{25,58,38,23,96,54,51,49,98,66\}$ for the 2-, 5-, and 10-class settings, respectively, following the class-selection protocol of \cite{zhou2025decoupled}. These results further evaluate the effectiveness of class relearning when multiple classes are forgotten simultaneously. Table~\ref{tab:ResNet-18_multi_class_2} reports the 2-class unlearning and relearning results on CIFAR-10 and CIFAR-100 with ResNet-18. Overall, the results show that our proposed SFRA extends to the multi-class setting: several unlearning methods exhibit substantial recovery of the jointly forget classes while largely preserving retain class performance. At the same time, the degree of recoverability remains method-dependent, with some methods showing substantially greater resistance to relearning than others. These results suggest that residual recoverable structure is not limited to the single-class setting and can persist when multiple classes are unlearned simultaneously. However, the multi-class audit should not be interpreted as monotone in $|\mathcal{Y}_f|$. Changing the forget set changes both the residual class geometry and the greedy assignment of unlabeled boundary probes. Consequently, a method may be unrecoverable for one forget-set composition yet partially recoverable for a larger one; the DELETE 5-class/10-class behavior in the main paper is an example. This assignment sensitivity is a limitation of the present multi-class extension rather than evidence that increasing the number of forget classes necessarily makes unlearning weaker.

\begin{table*}[t]
\centering
\caption{Comparison of unlearning methods using our proposed SFRA and the source-dependent PRA baseline on ViT-B/16 models under single-class unlearning across three datasets. For all model variants, retain accuracy $\mathcal{A}^t_r$ is reported as the mean $\pm$ standard deviation across forget classes, while forget accuracy $\mathcal{A}^t_f$ is reported as $(\min,\mathrm{mean},\max)$. $\mathrm{RS}$ and $\Delta\mathrm{RS}$ are independently reported as maxima across forget classes. Within each dataset, the highest and second-highest $\mathrm{RS}$ and $\Delta\mathrm{RS}$ values are shown in \textbf{bold} and \underline{underlined}, respectively.}
\label{tab:ViT-B_16_single_class_all_datasets}
\vspace{-0.3cm}
\resizebox{\textwidth}{!}{%
\begin{tabular}{c|c|cccc|cccc|cccc}
\toprule
\toprule
\multirow{2}{*}{Unlearning Method} & \multirow{2}{*}{Model Variant} & \multicolumn{4}{c}{\textbf{CIFAR-10}} & \multicolumn{4}{|c|}{\textbf{CIFAR-100}} & \multicolumn{4}{c}{\textbf{TinyImageNet}} \\
 &  & $\mathcal{A}^{t}_{r}(\%)$ & $\mathcal{A}^{t}_{f}(\%)$ & $\mathrm{RS}$ & $\Delta\mathrm{RS}$ & $\mathcal{A}^{t}_{r}(\%)$ & $\mathcal{A}^{t}_{f}(\%)$ & $\mathrm{RS}$ & $\Delta\mathrm{RS}$ & $\mathcal{A}^{t}_{r}(\%)$ & $\mathcal{A}^{t}_{f}(\%)$ & $\mathrm{RS}$ & $\Delta\mathrm{RS}$\\
\midrule
\midrule
Original & Original & $97.80{\scriptstyle\,\pm\,0.12}$ & $(95.90,97.80,98.80)$ & - & - & $87.78{\scriptstyle\,\pm\,0.08}$ & $(74.00,85.40,98.00)$ & - & - & $89.06{\scriptstyle\,\pm\,0.03}$ & $(78.00,88.00,100.00)$ & - & - \\
\midrule
\midrule
\multirow{3}{*}{Retrained} & Unlearned & $98.41{\scriptstyle\,\pm\,0.24}$ & $(0.00,0.00,0.00)$ &  &  & $87.30{\scriptstyle\,\pm\,0.41}$ & $(0.00,0.00,0.00)$ &  &  & $88.15{\scriptstyle\,\pm\,0.16}$ & $(0.00,0.00,0.00)$ &  &  \\
 & PRA \cite{ha2025unlearning} & $97.77{\scriptstyle\,\pm\,0.55}$ & $(19.60,65.86,89.40)$ & $0.94$ & - & $86.74{\scriptstyle\,\pm\,0.43}$ & $(39.00,69.90,86.00)$ & $0.92$ & - & $89.08{\scriptstyle\,\pm\,0.17}$ & $(72.00,84.60,100.00)$ & \textbf{\boldmath $1.00$} & - \\
 & SFRA (ours) & $98.25{\scriptstyle\,\pm\,0.21}$ & $(1.00,12.35,39.80)$ & $0.57$ & - & $86.53{\scriptstyle\,\pm\,0.52}$ & $(4.00,23.30,54.00)$ & $0.70$ & - & $82.12{\scriptstyle\,\pm\,1.42}$ & $(58.00,84.60,100.00)$ & $0.97$ & - \\
\midrule
\multirow{3}{*}{Finetune \cite{golatkar2020eternal}} & Unlearned & $95.64{\scriptstyle\,\pm\,3.82}$ & $(0.00,1.40,3.80)$ &  &  & $86.85{\scriptstyle\,\pm\,1.21}$ & $(0.00,2.90,14.00)$ &  &  & $76.38{\scriptstyle\,\pm\,1.62}$ & $(0.00,0.00,0.00)$ &  &  \\
 & PRA \cite{ha2025unlearning} & $95.17{\scriptstyle\,\pm\,3.70}$ & $(0.10,70.16,95.00)$ & $0.96$ & $+0.50$ & $86.70{\scriptstyle\,\pm\,1.34}$ & $(33.00,59.40,92.00)$ & $0.95$ & $+0.16$ & $76.38{\scriptstyle\,\pm\,1.62}$ & $(0.00,5.00,18.00)$ & $0.31$ & $-0.67$ \\
 & SFRA (ours) & $95.55{\scriptstyle\,\pm\,3.89}$ & $(0.30,5.71,16.90)$ & $0.24$ & $+0.07$ & $86.58{\scriptstyle\,\pm\,1.28}$ & $(2.00,15.10,39.00)$ & $0.42$ & $+0.07$ & $69.21{\scriptstyle\,\pm\,1.49}$ & $(46.00,64.00,88.00)$ & $0.90$ & $-0.04$ \\
\midrule
\multirow{3}{*}{Negative Gradient \cite{golatkar2020eternal}} & Unlearned & $94.70{\scriptstyle\,\pm\,2.42}$ & $(0.00,1.07,5.30)$ &  &  & $84.38{\scriptstyle\,\pm\,1.55}$ & $(0.00,0.90,7.00)$ &  &  & $87.15{\scriptstyle\,\pm\,0.68}$ & $(0.00,0.20,2.00)$ &  &  \\
 & PRA \cite{ha2025unlearning} & $94.05{\scriptstyle\,\pm\,1.96}$ & $(0.00,64.50,99.20)$ & $\underline{0.99}$ & $+0.63$ & $83.67{\scriptstyle\,\pm\,1.71}$ & $(0.00,60.90,100.00)$ & \textbf{\boldmath $0.99$} & $+0.38$ & $85.94{\scriptstyle\,\pm\,0.69}$ & $(58.00,83.20,94.00)$ & $0.96$ & $+0.05$ \\
 & SFRA (ours) & $92.35{\scriptstyle\,\pm\,4.73}$ & $(1.70,70.22,99.50)$ & $\underline{0.99}$ & $+0.92$ & $78.10{\scriptstyle\,\pm\,3.67}$ & $(3.00,58.70,99.00)$ & $0.96$ & $+0.63$ & $79.48{\scriptstyle\,\pm\,1.01}$ & $(68.00,86.00,96.00)$ & $0.95$ & $+0.17$ \\
\midrule
\multirow{3}{*}{Negative Gradient+ \cite{kurmanji2023towards}} & Unlearned & $97.53{\scriptstyle\,\pm\,0.45}$ & $(0.00,0.01,0.10)$ &  &  & $86.33{\scriptstyle\,\pm\,0.61}$ & $(0.00,0.00,0.00)$ &  &  & $87.59{\scriptstyle\,\pm\,0.65}$ & $(0.00,0.00,0.00)$ &  &  \\
 & PRA \cite{ha2025unlearning} & $96.27{\scriptstyle\,\pm\,0.59}$ & $(0.00,77.74,99.90)$ & $\underline{0.99}$ & $+0.65$ & $85.97{\scriptstyle\,\pm\,0.70}$ & $(0.00,39.20,97.00)$ & $\underline{0.98}$ & $+0.42$ & $86.71{\scriptstyle\,\pm\,0.80}$ & $(2.00,73.20,100.00)$ & \textbf{\boldmath $1.00$} & $+0.12$ \\
 & SFRA (ours) & $97.58{\scriptstyle\,\pm\,0.37}$ & $(0.00,21.08,94.60)$ & $0.97$ & $+0.71$ & $85.97{\scriptstyle\,\pm\,0.62}$ & $(0.00,21.80,91.00)$ & $0.95$ & $+0.60$ & $83.76{\scriptstyle\,\pm\,2.52}$ & $(12.00,53.20,100.00)$ & $\underline{0.99}$ & $+0.11$ \\
\midrule
\multirow{3}{*}{Random Label \cite{hayase2020selective}} & Unlearned & $97.67{\scriptstyle\,\pm\,0.35}$ & $(0.00,0.03,0.20)$ &  &  & $82.28{\scriptstyle\,\pm\,4.24}$ & $(0.00,0.00,0.00)$ &  &  & $87.88{\scriptstyle\,\pm\,0.62}$ & $(0.00,0.00,0.00)$ &  &  \\
 & PRA \cite{ha2025unlearning} & $96.48{\scriptstyle\,\pm\,0.61}$ & $(95.20,98.55,99.80)$ & \textbf{\boldmath $1.00$} & $+0.66$ & $81.31{\scriptstyle\,\pm\,3.91}$ & $(54.00,89.70,100.00)$ & \textbf{\boldmath $0.99$} & $+0.42$ & $86.50{\scriptstyle\,\pm\,0.53}$ & $(72.00,93.40,100.00)$ & $\underline{0.99}$ & $+0.13$ \\
 & SFRA (ours) & $96.14{\scriptstyle\,\pm\,1.73}$ & $(97.00,98.69,99.80)$ & \textbf{\boldmath $1.00$} & $\mathbf{+0.98}$ & $76.90{\scriptstyle\,\pm\,4.43}$ & $(19.00,87.30,100.00)$ & $\underline{0.98}$ & $\underline{+0.86}$ & $84.20{\scriptstyle\,\pm\,2.06}$ & $(92.00,97.20,100.00)$ & $\underline{0.99}$ & $\mathbf{+0.24}$ \\
\midrule
\multirow{3}{*}{Learn to Unlearn \cite{cha2024learning}} & Unlearned & $89.13{\scriptstyle\,\pm\,9.40}$ & $(0.00,0.30,2.80)$ &  &  & $84.77{\scriptstyle\,\pm\,1.96}$ & $(0.00,0.10,1.00)$ &  &  & $87.24{\scriptstyle\,\pm\,0.53}$ & $(0.00,0.00,0.00)$ &  &  \\
 & PRA \cite{ha2025unlearning} & $88.71{\scriptstyle\,\pm\,9.11}$ & $(0.00,34.01,98.40)$ & $0.98$ & $+0.36$ & $84.16{\scriptstyle\,\pm\,2.02}$ & $(1.00,63.40,100.00)$ & \textbf{\boldmath $0.99$} & $+0.37$ & $85.96{\scriptstyle\,\pm\,0.56}$ & $(72.00,86.40,94.00)$ & $0.97$ & $+0.08$ \\
 & SFRA (ours) & $88.15{\scriptstyle\,\pm\,8.86}$ & $(0.40,28.53,97.70)$ & $0.98$ & $+0.86$ & $79.68{\scriptstyle\,\pm\,3.67}$ & $(2.00,53.90,98.00)$ & $0.96$ & $+0.62$ & $81.23{\scriptstyle\,\pm\,2.07}$ & $(50.00,84.00,96.00)$ & $0.95$ & $+0.07$ \\
\midrule
\multirow{3}{*}{SCRUB \cite{kurmanji2023towards}} & Unlearned & $97.37{\scriptstyle\,\pm\,0.92}$ & $(0.00,1.12,8.10)$ &  &  & $83.39{\scriptstyle\,\pm\,1.51}$ & $(0.00,1.10,8.00)$ &  &  & $86.44{\scriptstyle\,\pm\,1.16}$ & $(0.00,0.60,4.00)$ &  &  \\
 & PRA \cite{ha2025unlearning} & $97.10{\scriptstyle\,\pm\,0.94}$ & $(1.40,23.73,65.30)$ & $0.73$ & $+0.09$ & $83.31{\scriptstyle\,\pm\,2.22}$ & $(0.00,63.10,96.00)$ & $0.97$ & $+0.38$ & $85.24{\scriptstyle\,\pm\,1.01}$ & $(42.00,77.00,94.00)$ & $0.94$ & $+0.05$ \\
 & SFRA (ours) & $96.57{\scriptstyle\,\pm\,2.72}$ & $(0.00,7.10,58.50)$ & $0.65$ & $+0.63$ & $78.95{\scriptstyle\,\pm\,3.93}$ & $(0.00,41.20,97.00)$ & $0.95$ & $+0.62$ & $79.23{\scriptstyle\,\pm\,1.24}$ & $(32.00,68.80,96.00)$ & $0.94$ & $+0.05$ \\
\midrule
\multirow{3}{*}{Bad Teacher \cite{chundawat2023can}} & Unlearned & $96.16{\scriptstyle\,\pm\,4.00}$ & $(0.00,3.23,14.80)$ &  &  & $87.71{\scriptstyle\,\pm\,0.11}$ & $(0.00,0.00,0.00)$ &  &  & $89.20{\scriptstyle\,\pm\,0.14}$ & $(0.00,0.40,4.00)$ &  &  \\
 & PRA \cite{ha2025unlearning} & $95.68{\scriptstyle\,\pm\,4.20}$ & $(91.10,97.34,99.60)$ & \textbf{\boldmath $1.00$} & $+0.63$ & $87.55{\scriptstyle\,\pm\,0.13}$ & $(81.00,90.20,98.00)$ & \textbf{\boldmath $0.99$} & $+0.39$ & $89.16{\scriptstyle\,\pm\,0.15}$ & $(82.00,90.40,100.00)$ & \textbf{\boldmath $1.00$} & $+0.11$ \\
 & SFRA (ours) & $95.31{\scriptstyle\,\pm\,5.55}$ & $(95.50,97.21,98.40)$ & $\underline{0.99}$ & $\underline{+0.96}$ & $86.74{\scriptstyle\,\pm\,0.20}$ & $(84.00,90.60,99.00)$ & \textbf{\boldmath $0.99$} & $+0.85$ & $85.36{\scriptstyle\,\pm\,2.90}$ & $(96.00,98.80,100.00)$ & \textbf{\boldmath $1.00$} & $\mathbf{+0.24}$ \\
\midrule
\multirow{3}{*}{SalUn \cite{fan2023salun}} & Unlearned & $98.08{\scriptstyle\,\pm\,0.23}$ & $(0.00,0.01,0.10)$ &  &  & $87.50{\scriptstyle\,\pm\,0.22}$ & $(0.00,0.00,0.00)$ &  &  & $88.53{\scriptstyle\,\pm\,0.11}$ & $(0.00,0.00,0.00)$ &  &  \\
 & PRA \cite{ha2025unlearning} & $97.08{\scriptstyle\,\pm\,0.76}$ & $(96.10,98.54,99.50)$ & $\underline{0.99}$ & $+0.66$ & $87.19{\scriptstyle\,\pm\,0.41}$ & $(74.00,87.70,97.00)$ & $\underline{0.98}$ & $+0.39$ & $88.51{\scriptstyle\,\pm\,0.11}$ & $(64.00,79.20,100.00)$ & \textbf{\boldmath $1.00$} & $+0.06$ \\
 & SFRA (ours) & $98.07{\scriptstyle\,\pm\,0.23}$ & $(5.90,77.29,97.30)$ & $\underline{0.99}$ & $+0.90$ & $87.36{\scriptstyle\,\pm\,0.26}$ & $(27.00,69.40,94.00)$ & $0.97$ & $+0.69$ & $87.53{\scriptstyle\,\pm\,0.47}$ & $(80.00,88.40,100.00)$ & \textbf{\boldmath $1.00$} & $\underline{+0.18}$ \\
\midrule
\multirow{3}{*}{DELETE \cite{zhou2025decoupled}} & Unlearned & $97.16{\scriptstyle\,\pm\,0.67}$ & $(0.00,0.00,0.00)$ &  &  & $83.78{\scriptstyle\,\pm\,1.96}$ & $(0.00,0.30,1.00)$ &  &  & $88.56{\scriptstyle\,\pm\,0.19}$ & $(0.00,0.00,0.00)$ &  &  \\
 & PRA \cite{ha2025unlearning} & $96.51{\scriptstyle\,\pm\,0.63}$ & $(0.00,80.23,99.70)$ & $\underline{0.99}$ & $+0.65$ & $83.02{\scriptstyle\,\pm\,1.57}$ & $(38.00,78.40,100.00)$ & \textbf{\boldmath $0.99$} & $+0.41$ & $87.32{\scriptstyle\,\pm\,0.69}$ & $(78.00,91.80,100.00)$ & $\underline{0.99}$ & $+0.11$ \\
 & SFRA (ours) & $94.74{\scriptstyle\,\pm\,1.80}$ & $(39.30,90.86,99.90)$ & $\underline{0.99}$ & $\underline{+0.96}$ & $78.45{\scriptstyle\,\pm\,3.35}$ & $(70.00,92.70,100.00)$ & $0.97$ & $\mathbf{+0.89}$ & $83.38{\scriptstyle\,\pm\,1.43}$ & $(92.00,97.00,100.00)$ & $0.98$ & $\mathbf{+0.24}$ \\
\midrule
\bottomrule
\end{tabular}
}
\end{table*}

\begin{table*}[t]
\centering
\caption{Comparison of unlearning methods using our proposed SFRA and the source-dependent PRA baseline on Swin-T models under single-class unlearning across three datasets. For all model variants, retain accuracy $\mathcal{A}^t_r$ is reported as the mean $\pm$ standard deviation across forget classes, while forget accuracy $\mathcal{A}^t_f$ is reported as $(\min,\mathrm{mean},\max)$. $\mathrm{RS}$ and $\Delta\mathrm{RS}$ are independently reported as maxima across forget classes. Within each dataset, the highest and second-highest $\mathrm{RS}$ and $\Delta\mathrm{RS}$ values are shown in \textbf{bold} and \underline{underlined}, respectively.}
\label{tab:Swin-T_single_class_all_datasets}
\vspace{-0.3cm}
\resizebox{\textwidth}{!}{%
\begin{tabular}{c|c|cccc|cccc|cccc}
\toprule
\toprule
\multirow{2}{*}{Unlearning Method} & \multirow{2}{*}{Model Variant} & \multicolumn{4}{c}{\textbf{CIFAR-10}} & \multicolumn{4}{|c|}{\textbf{CIFAR-100}} & \multicolumn{4}{c}{\textbf{TinyImageNet}} \\
 &  & $\mathcal{A}^{t}_{r}(\%)$ & $\mathcal{A}^{t}_{f}(\%)$ & $\mathrm{RS}$ & $\Delta\mathrm{RS}$ & $\mathcal{A}^{t}_{r}(\%)$ & $\mathcal{A}^{t}_{f}(\%)$ & $\mathrm{RS}$ & $\Delta\mathrm{RS}$ & $\mathcal{A}^{t}_{r}(\%)$ & $\mathcal{A}^{t}_{f}(\%)$ & $\mathrm{RS}$ & $\Delta\mathrm{RS}$\\
\midrule
\midrule
Original & Original & $82.69{\scriptstyle\,\pm\,0.94}$ & $(64.00,82.69,91.00)$ & - & - & $88.35{\scriptstyle\,\pm\,0.07}$ & $(74.00,87.70,95.00)$ & - & - & $86.32{\scriptstyle\,\pm\,0.03}$ & $(80.00,87.80,100.00)$ & - & - \\
\midrule
\midrule
\multirow{3}{*}{Retrained} & Unlearned & $82.61{\scriptstyle\,\pm\,1.47}$ & $(0.00,0.00,0.00)$ &  &  & $88.44{\scriptstyle\,\pm\,0.12}$ & $(0.00,0.00,0.00)$ &  &  & $86.18{\scriptstyle\,\pm\,0.12}$ & $(0.00,0.00,0.00)$ &  &  \\
 & PRA \cite{ha2025unlearning} & $82.70{\scriptstyle\,\pm\,1.46}$ & $(0.30,2.19,7.30)$ & $0.14$ & - & $87.74{\scriptstyle\,\pm\,0.29}$ & $(47.00,72.00,95.00)$ & $0.97$ & - & $86.25{\scriptstyle\,\pm\,0.22}$ & $(72.00,87.00,100.00)$ & \textbf{\boldmath $1.00$} & - \\
 & SFRA (ours) & $78.69{\scriptstyle\,\pm\,1.40}$ & $(13.60,22.35,34.50)$ & $0.51$ & - & $85.70{\scriptstyle\,\pm\,0.65}$ & $(4.00,28.80,73.00)$ & $0.83$ & - & $79.35{\scriptstyle\,\pm\,1.77}$ & $(50.00,78.20,92.00)$ & $0.92$ & - \\
\midrule
\multirow{3}{*}{Finetune \cite{golatkar2020eternal}} & Unlearned & $90.17{\scriptstyle\,\pm\,1.23}$ & $(0.00,0.00,0.00)$ &  &  & $86.98{\scriptstyle\,\pm\,0.67}$ & $(0.00,2.40,12.00)$ &  &  & $77.01{\scriptstyle\,\pm\,0.54}$ & $(0.00,0.00,0.00)$ &  &  \\
 & PRA \cite{ha2025unlearning} & $90.17{\scriptstyle\,\pm\,1.23}$ & $(0.00,0.05,0.20)$ & $0.00$ & $-0.00$ & $86.71{\scriptstyle\,\pm\,0.89}$ & $(46.00,67.80,87.00)$ & $0.88$ & $+0.20$ & $76.96{\scriptstyle\,\pm\,0.54}$ & $(0.00,12.40,32.00)$ & $0.48$ & $-0.42$ \\
 & SFRA (ours) & $83.61{\scriptstyle\,\pm\,2.37}$ & $(26.10,42.07,56.70)$ & $0.70$ & $+0.40$ & $86.15{\scriptstyle\,\pm\,0.83}$ & $(0.00,14.60,41.00)$ & $0.45$ & $+0.31$ & $70.18{\scriptstyle\,\pm\,0.90}$ & $(32.00,55.60,68.00)$ & $0.79$ & $+0.06$ \\
\midrule
\multirow{3}{*}{Negative Gradient \cite{golatkar2020eternal}} & Unlearned & $79.41{\scriptstyle\,\pm\,2.56}$ & $(0.40,1.41,2.00)$ &  &  & $85.79{\scriptstyle\,\pm\,1.61}$ & $(0.00,0.70,2.00)$ &  &  & $84.54{\scriptstyle\,\pm\,0.90}$ & $(0.00,1.60,8.00)$ &  &  \\
 & PRA \cite{ha2025unlearning} & $79.34{\scriptstyle\,\pm\,2.53}$ & $(0.10,1.68,5.50)$ & $0.09$ & $+0.08$ & $85.22{\scriptstyle\,\pm\,1.69}$ & $(16.00,57.90,96.00)$ & $0.97$ & $+0.06$ & $83.86{\scriptstyle\,\pm\,1.08}$ & $(52.00,75.80,94.00)$ & $0.97$ & $+0.01$ \\
 & SFRA (ours) & $74.35{\scriptstyle\,\pm\,2.27}$ & $(18.90,37.99,69.50)$ & $0.78$ & $+0.44$ & $78.04{\scriptstyle\,\pm\,1.35}$ & $(7.00,58.50,92.00)$ & $0.91$ & $+0.45$ & $77.36{\scriptstyle\,\pm\,0.97}$ & $(76.00,86.40,96.00)$ & $0.95$ & $+0.24$ \\
\midrule
\multirow{3}{*}{Negative Gradient+ \cite{kurmanji2023towards}} & Unlearned & $82.27{\scriptstyle\,\pm\,1.78}$ & $(0.20,0.37,0.50)$ &  &  & $86.38{\scriptstyle\,\pm\,0.92}$ & $(0.00,0.10,1.00)$ &  &  & $84.73{\scriptstyle\,\pm\,0.65}$ & $(0.00,0.20,2.00)$ &  &  \\
 & PRA \cite{ha2025unlearning} & $82.06{\scriptstyle\,\pm\,1.81}$ & $(0.10,2.83,5.10)$ & $0.09$ & $+0.08$ & $86.09{\scriptstyle\,\pm\,0.93}$ & $(6.00,45.10,87.00)$ & $0.93$ & $+0.13$ & $83.65{\scriptstyle\,\pm\,0.81}$ & $(0.00,73.20,96.00)$ & $0.98$ & $+0.04$ \\
 & SFRA (ours) & $76.72{\scriptstyle\,\pm\,2.54}$ & $(14.50,33.14,60.50)$ & $0.73$ & $+0.38$ & $86.20{\scriptstyle\,\pm\,0.84}$ & $(0.00,2.10,5.00)$ & $0.10$ & $-0.07$ & $82.33{\scriptstyle\,\pm\,1.30}$ & $(0.00,40.00,88.00)$ & $0.91$ & $-0.01$ \\
\midrule
\multirow{3}{*}{Random Label \cite{hayase2020selective}} & Unlearned & $73.93{\scriptstyle\,\pm\,1.83}$ & $(2.20,9.63,15.10)$ &  &  & $87.19{\scriptstyle\,\pm\,0.40}$ & $(0.00,0.10,1.00)$ &  &  & $84.75{\scriptstyle\,\pm\,0.75}$ & $(0.00,0.00,0.00)$ &  &  \\
 & PRA \cite{ha2025unlearning} & $73.35{\scriptstyle\,\pm\,1.65}$ & $(15.40,31.34,60.10)$ & $0.73$ & $+0.59$ & $86.16{\scriptstyle\,\pm\,0.69}$ & $(84.00,93.10,99.00)$ & \textbf{\boldmath $0.99$} & $+0.33$ & $83.66{\scriptstyle\,\pm\,1.07}$ & $(70.00,89.80,100.00)$ & \textbf{\boldmath $1.00$} & $+0.07$ \\
 & SFRA (ours) & $67.34{\scriptstyle\,\pm\,1.75}$ & $(53.80,76.98,96.00)$ & $\underline{0.88}$ & $+0.61$ & $81.58{\scriptstyle\,\pm\,2.02}$ & $(89.00,96.40,100.00)$ & $\underline{0.98}$ & $\underline{+0.90}$ & $78.70{\scriptstyle\,\pm\,2.04}$ & $(58.00,87.20,100.00)$ & $0.97$ & $\mathbf{+0.32}$ \\
\midrule
\multirow{3}{*}{Learn to Unlearn \cite{cha2024learning}} & Unlearned & $79.14{\scriptstyle\,\pm\,2.36}$ & $(0.30,1.43,2.70)$ &  &  & $86.25{\scriptstyle\,\pm\,0.94}$ & $(0.00,0.00,0.00)$ &  &  & $84.01{\scriptstyle\,\pm\,0.79}$ & $(0.00,0.00,0.00)$ &  &  \\
 & PRA \cite{ha2025unlearning} & $78.98{\scriptstyle\,\pm\,2.37}$ & $(0.20,2.68,8.20)$ & $0.14$ & $+0.13$ & $85.74{\scriptstyle\,\pm\,1.03}$ & $(16.00,56.60,82.00)$ & $0.90$ & $+0.23$ & $82.75{\scriptstyle\,\pm\,0.87}$ & $(68.00,81.60,98.00)$ & $\underline{0.99}$ & $+0.01$ \\
 & SFRA (ours) & $74.28{\scriptstyle\,\pm\,2.87}$ & $(34.40,48.09,71.10)$ & $0.79$ & $+0.46$ & $80.40{\scriptstyle\,\pm\,2.77}$ & $(5.00,44.40,92.00)$ & $0.92$ & $+0.39$ & $83.84{\scriptstyle\,\pm\,3.58}$ & $(0.00,49.13,77.94)$ & $0.88$ & $+0.04$ \\
\midrule
\multirow{3}{*}{SCRUB \cite{kurmanji2023towards}} & Unlearned & $83.25{\scriptstyle\,\pm\,1.45}$ & $(0.00,0.18,0.80)$ &  &  & $85.91{\scriptstyle\,\pm\,1.06}$ & $(0.00,0.00,0.00)$ &  &  & $83.58{\scriptstyle\,\pm\,0.71}$ & $(0.00,0.20,2.00)$ &  &  \\
 & PRA \cite{ha2025unlearning} & $83.13{\scriptstyle\,\pm\,1.44}$ & $(0.10,2.21,9.70)$ & $0.18$ & $+0.13$ & $85.48{\scriptstyle\,\pm\,1.28}$ & $(17.00,49.30,80.00)$ & $0.89$ & $+0.08$ & $82.69{\scriptstyle\,\pm\,0.89}$ & $(38.00,69.40,94.00)$ & $0.95$ & $+0.02$ \\
 & SFRA (ours) & $79.16{\scriptstyle\,\pm\,1.37}$ & $(9.80,25.64,48.90)$ & $0.64$ & $+0.21$ & $81.01{\scriptstyle\,\pm\,3.21}$ & $(0.00,31.00,76.00)$ & $0.84$ & $+0.25$ & $76.34{\scriptstyle\,\pm\,1.36}$ & $(14.00,63.20,86.00)$ & $0.89$ & $+0.06$ \\
\midrule
\multirow{3}{*}{Bad Teacher \cite{chundawat2023can}} & Unlearned & $81.14{\scriptstyle\,\pm\,1.40}$ & $(0.20,6.56,16.00)$ &  &  & $88.57{\scriptstyle\,\pm\,0.14}$ & $(0.00,0.30,2.00)$ &  &  & $85.91{\scriptstyle\,\pm\,0.10}$ & $(0.00,0.40,4.00)$ &  &  \\
 & PRA \cite{ha2025unlearning} & $80.65{\scriptstyle\,\pm\,1.47}$ & $(21.20,55.79,77.90)$ & $0.86$ & $\mathbf{+0.82}$ & $87.98{\scriptstyle\,\pm\,0.25}$ & $(93.00,97.00,100.00)$ & \textbf{\boldmath $0.99$} & $+0.35$ & $85.74{\scriptstyle\,\pm\,0.13}$ & $(88.00,94.20,100.00)$ & \textbf{\boldmath $1.00$} & $+0.10$ \\
 & SFRA (ours) & $75.40{\scriptstyle\,\pm\,1.38}$ & $(71.20,90.68,97.50)$ & \textbf{\boldmath $0.96$} & $\underline{+0.68}$ & $86.75{\scriptstyle\,\pm\,0.80}$ & $(86.00,94.00,99.00)$ & \textbf{\boldmath $0.99$} & $\mathbf{+0.91}$ & $81.42{\scriptstyle\,\pm\,2.37}$ & $(96.00,98.40,100.00)$ & $\underline{0.99}$ & $\underline{+0.30}$ \\
\midrule
\multirow{3}{*}{SalUn \cite{fan2023salun}} & Unlearned & $83.39{\scriptstyle\,\pm\,1.64}$ & $(0.80,2.33,4.20)$ &  &  & $87.89{\scriptstyle\,\pm\,0.52}$ & $(0.00,0.00,0.00)$ &  &  & $84.90{\scriptstyle\,\pm\,1.42}$ & $(0.00,1.00,6.00)$ &  &  \\
 & PRA \cite{ha2025unlearning} & $83.06{\scriptstyle\,\pm\,1.58}$ & $(4.00,17.73,52.60)$ & $0.65$ & $+0.59$ & $87.84{\scriptstyle\,\pm\,0.57}$ & $(25.00,48.00,79.00)$ & $0.88$ & $+0.02$ & $83.90{\scriptstyle\,\pm\,2.26}$ & $(76.00,90.20,100.00)$ & \textbf{\boldmath $1.00$} & $+0.08$ \\
 & SFRA (ours) & $75.43{\scriptstyle\,\pm\,1.95}$ & $(61.20,73.70,81.20)$ & $0.86$ & $+0.53$ & $87.29{\scriptstyle\,\pm\,0.54}$ & $(8.00,39.70,69.00)$ & $0.81$ & $+0.69$ & $80.98{\scriptstyle\,\pm\,4.79}$ & $(54.00,82.80,94.00)$ & $0.97$ & $+0.20$ \\
\midrule
\multirow{3}{*}{DELETE \cite{zhou2025decoupled}} & Unlearned & $83.40{\scriptstyle\,\pm\,1.54}$ & $(0.00,0.00,0.00)$ &  &  & $87.64{\scriptstyle\,\pm\,0.41}$ & $(0.00,0.00,0.00)$ &  &  & $85.76{\scriptstyle\,\pm\,0.29}$ & $(0.00,5.60,28.00)$ &  &  \\
 & PRA \cite{ha2025unlearning} & $83.01{\scriptstyle\,\pm\,1.62}$ & $(2.90,7.57,13.80)$ & $0.24$ & $+0.23$ & $86.68{\scriptstyle\,\pm\,0.65}$ & $(78.00,91.40,100.00)$ & \textbf{\boldmath $0.99$} & $+0.33$ & $85.19{\scriptstyle\,\pm\,0.89}$ & $(80.00,90.80,100.00)$ & \textbf{\boldmath $1.00$} & $+0.05$ \\
 & SFRA (ours) & $76.09{\scriptstyle\,\pm\,2.13}$ & $(13.40,32.84,53.20)$ & $0.67$ & $+0.25$ & $81.97{\scriptstyle\,\pm\,2.46}$ & $(89.00,96.90,100.00)$ & $\underline{0.98}$ & $\mathbf{+0.91}$ & $79.53{\scriptstyle\,\pm\,2.13}$ & $(92.00,96.60,100.00)$ & $\underline{0.99}$ & $\underline{+0.30}$ \\
\midrule
\bottomrule
\end{tabular}
}
\end{table*}

\begin{algorithm}[h]
\caption{Multi-Class SFRA}
\label{alg:multiclass}

\KwIn{
Unlearned classifier $\Phi_{un}=h\circ e$ with class set $\mathcal{Y}$;
retain-set $\mathcal{Y}_r$;
forget-set $\mathcal{Y}_f$;
accepted probs per retain class $N$;
selected probs per subset $M$, where
$(1+|\mathcal{Y}_f|)M\leq N$;
relearning loss $\mathcal{L}_{\mathrm{re}}$;
number of relearning steps $T$;
learning rate $\eta$.
}
\KwOut{Relearned classifier $\Phi_{re}=h'\circ e$.}

\BlankLine
\StepOne{Synthetic Prob Generation}

Initialize
$\mathcal{S}_{c_r}\gets\emptyset$
for every $c_r\in\mathcal{Y}_r$\;

Initialize
$\mathcal{S}_{c_f}\gets\emptyset$
for every $c_f\in\mathcal{Y}_f$\;

Let
$c_f^{(1)}<\cdots<c_f^{(|\mathcal{Y}_f|)}$
denote the forget classes sorted by ascending class index\;

\For{each retain class $c_r\in\mathcal{Y}_r$}{
    Initialize the shared candidate pool
    $\mathcal{P}_{c_r}\gets\emptyset$\;

    \While{$|\mathcal{P}_{c_r}|<N$}{
        Sample an embedding
        $s\sim\mathcal{N}(0,I_d)$\;

        Compute the class probabilities
        $p(s)=\operatorname{softmax}(h(s))$\;

        \If{$\arg\max_{c\in\mathcal{Y}}p_c(s)=c_r$}{
            Append $(s,p(s))$ to $\mathcal{P}_{c_r}$\;
        }
    }

    Sort $\mathcal{P}_{c_r}$ in descending order of
    $p_{c_r}(s)$\;

    Let $\mathcal{R}_{c_r}$ be the $M$ probs in
    $\mathcal{P}_{c_r}$ with the highest values of
    $p_{c_r}(s)$\;

    Append the probs in $\mathcal{R}_{c_r}$ to
    $\mathcal{S}_{c_r}$ with label $c_r$\;

    From the same candidate pool $\mathcal{P}_{c_r}$,
    select the $|\mathcal{Y}_f|M$ probs with the
    lowest values of $p_{c_r}(s)$ to form the
    boundary-probe pool $\mathcal{Q}_{c_r}$\;

    Initialize the assigned-probe set
    $\mathcal{U}_{c_r}\gets\emptyset$\;

    \For{$j\gets1$ \KwTo $|\mathcal{Y}_f|$}{
        Set $c_f\gets c_f^{(j)}$\;

        Select
        $
        \mathcal{Q}_{c_r,c_f}
        \gets
        \underset{
            \substack{
                \mathcal{V}\subseteq
                \mathcal{Q}_{c_r}\setminus\mathcal{U}_{c_r}\\
                |\mathcal{V}|=M
            }
        }{\arg\max}
        \sum_{s\in\mathcal{V}}p_{c_f}(s),
        $

        Append the probs in
        $\mathcal{Q}_{c_r,c_f}$ to
        $\mathcal{S}_{c_f}$ with label $c_f$\;

        Update
        $\mathcal{U}_{c_r}
        \gets
        \mathcal{U}_{c_r}\cup\mathcal{Q}_{c_r,c_f}$\;
    }
}

Define the complete labeled synthetic datasets
\[
\begin{aligned}
\mathcal{S}_r=\bigcup_{c_r\in\mathcal{Y}_r}
\mathcal{S}_{c_r},~~~
\mathcal{S}_f=\bigcup_{c_f\in\mathcal{Y}_f}
\mathcal{S}_{c_f}.
\end{aligned}
\]
\BlankLine
\StepTwo{Multi-Class Relearning}

Combine the synthetic datasets
$\mathcal{S}\gets\mathcal{S}_r\cup\mathcal{S}_f$\;

Freeze the feature extractor $e$ and initialize
$h'\gets h$\;

\For{$t=1,\ldots,T$}{
    Sample a mini-batch
    $\mathcal{B}\subseteq\mathcal{S}$\;

    Compute
    $\mathcal{L}_{\mathrm{re}}
    =-\frac{1}{|\mathcal{B}|}
    \sum_{(s,y)\in\mathcal{B}}
    \log p_y(s;\theta)$\;
    
    %Update $h$ by minimizing    $\mathcal{L}_{\mathrm{re}}$ on $\mathcal{B}$;
    Update
    $\theta\gets\theta-\eta
    \nabla_\theta\mathcal{L}_{\mathrm{re}}$\;
}
\Return{relearned model $\Phi_{re}=h'\circ e$}\;

\end{algorithm}

\clearpage
\section{Geometric Interpretation of Synthetic Boundary Probes}
\label{app:geometric_interpretation}

\begin{figure*}[h]
\centering

% ================= ORIGINAL MODEL =================
\begin{minipage}{\textwidth}
    \centering
    \textbf{Original Model}
\end{minipage}

\vspace{1mm}

\begin{subfigure}[h]{0.24\textwidth}
    \centering
    \includegraphics[width=\linewidth]
    {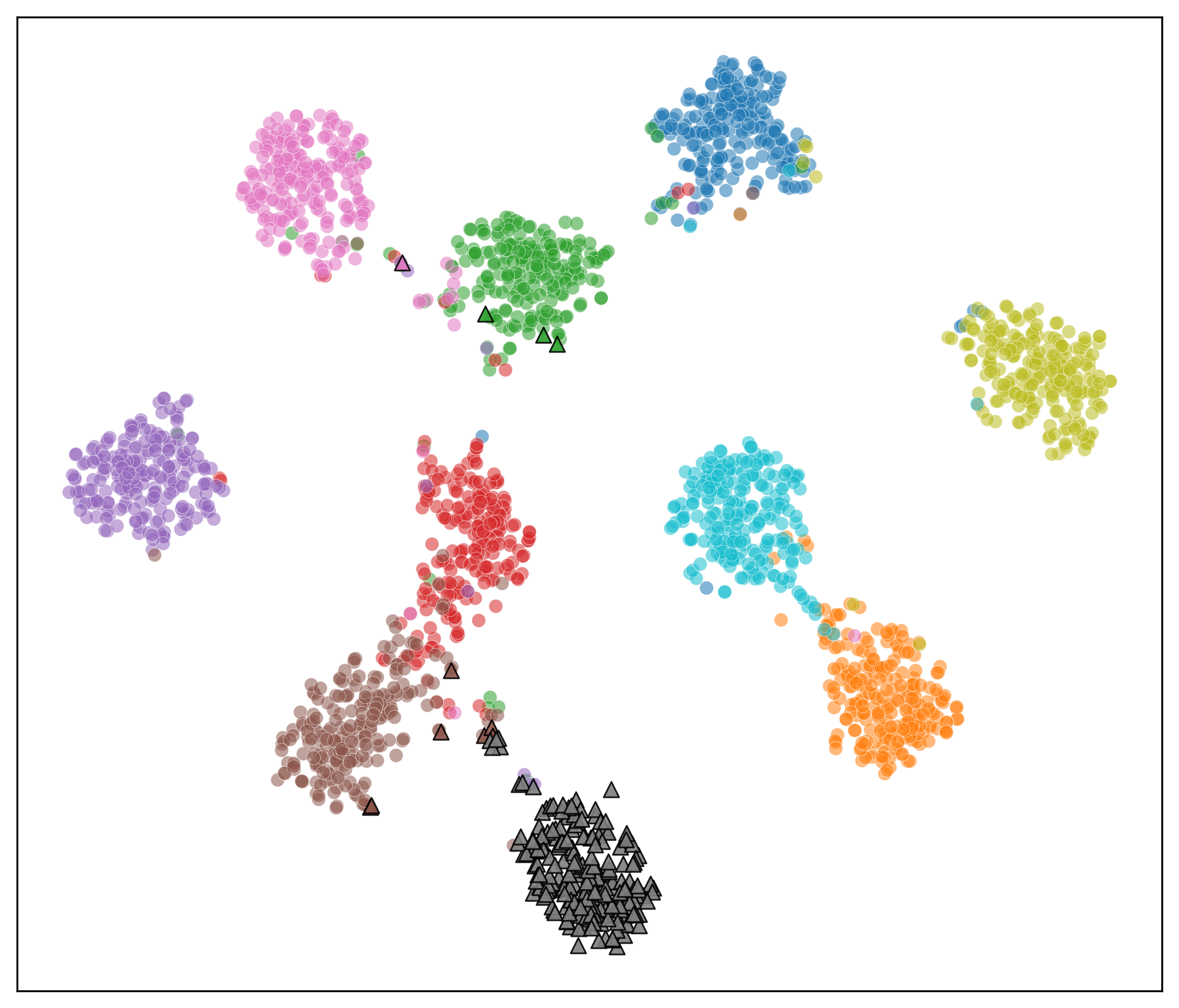}
    \caption{Original: feature space}
\end{subfigure}
\hspace{0.02\textwidth}
\begin{subfigure}[h]{0.24\textwidth}
    \centering
    \includegraphics[width=\linewidth]
    {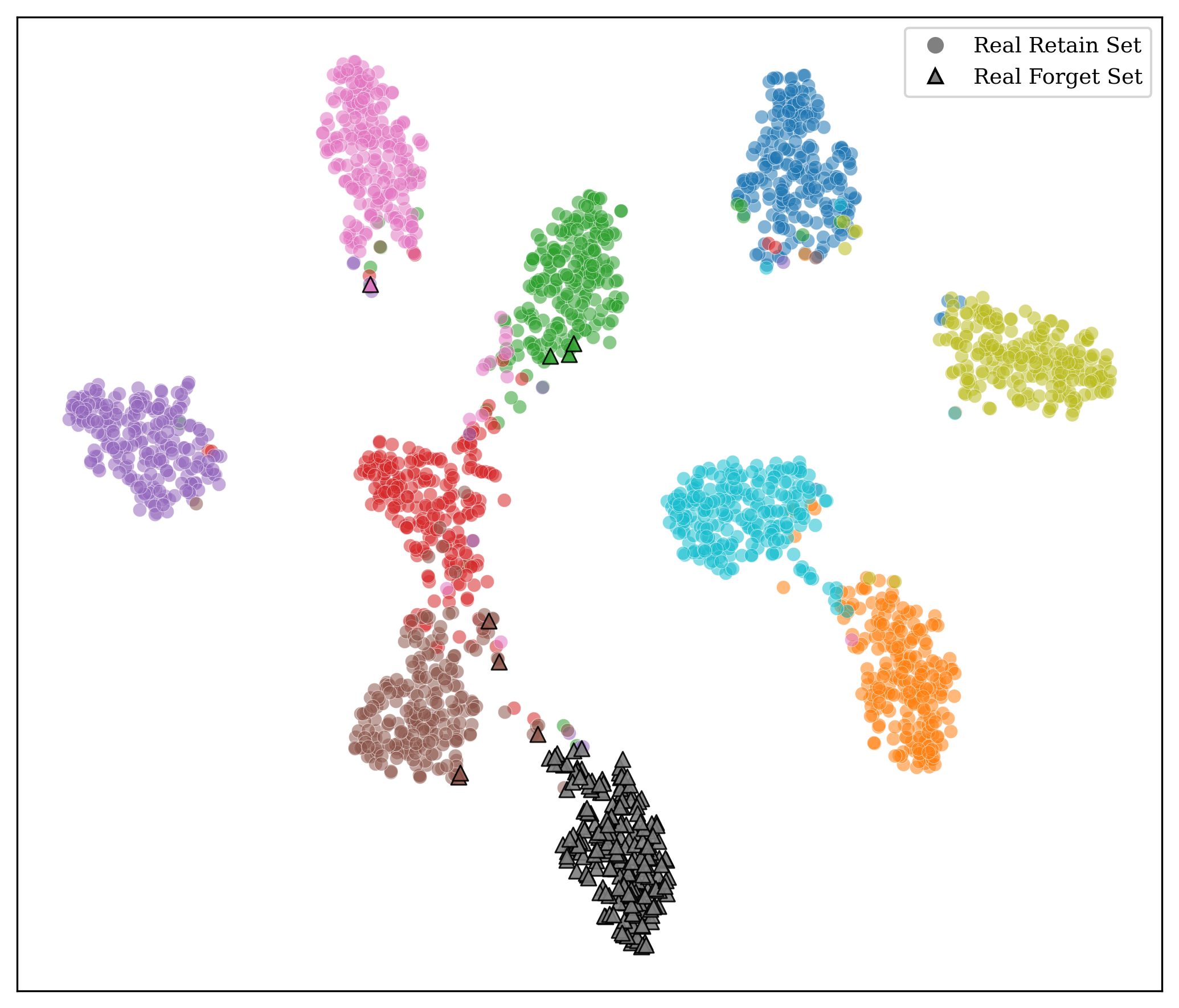}
    \caption{Original: logit space}
\end{subfigure}

\vspace{2mm}

% ================= GROUP HEADINGS =================
\begin{minipage}{0.49\textwidth}
    \centering
    \textbf{Higher Recoverability}
\end{minipage}
\hfill
\begin{minipage}{0.49\textwidth}
    \centering
    \textbf{Lower Recoverability}
\end{minipage}

\vspace{1mm}

% ================= ROW 2 =================

\begin{subfigure}[h]{0.24\textwidth}
    \centering
    \includegraphics[width=\linewidth]
    {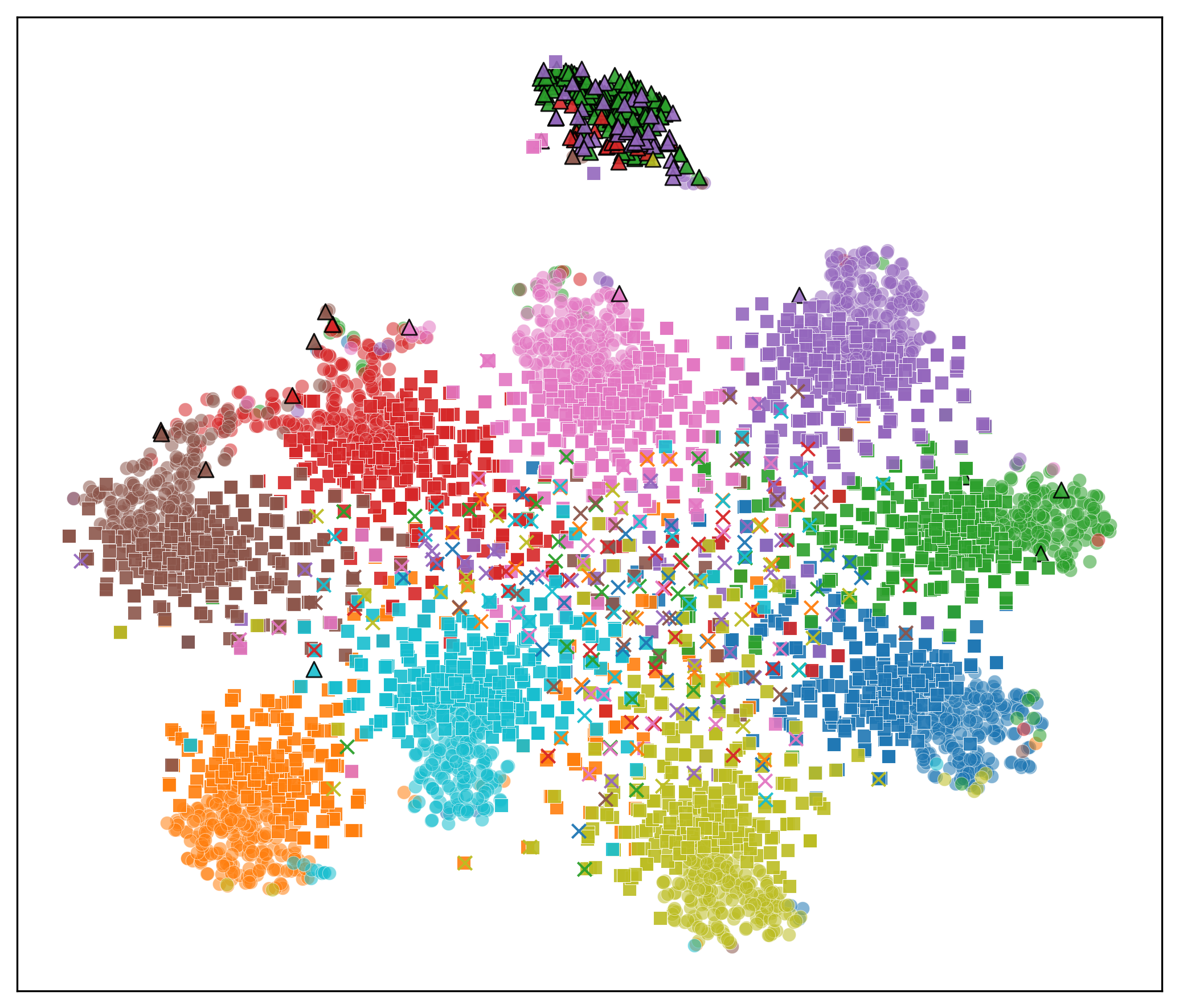}
    \caption{Bad Teacher: feature space}
\end{subfigure}
\hfill
\begin{subfigure}[h]{0.24\textwidth}
    \centering
    \includegraphics[width=\linewidth]
    {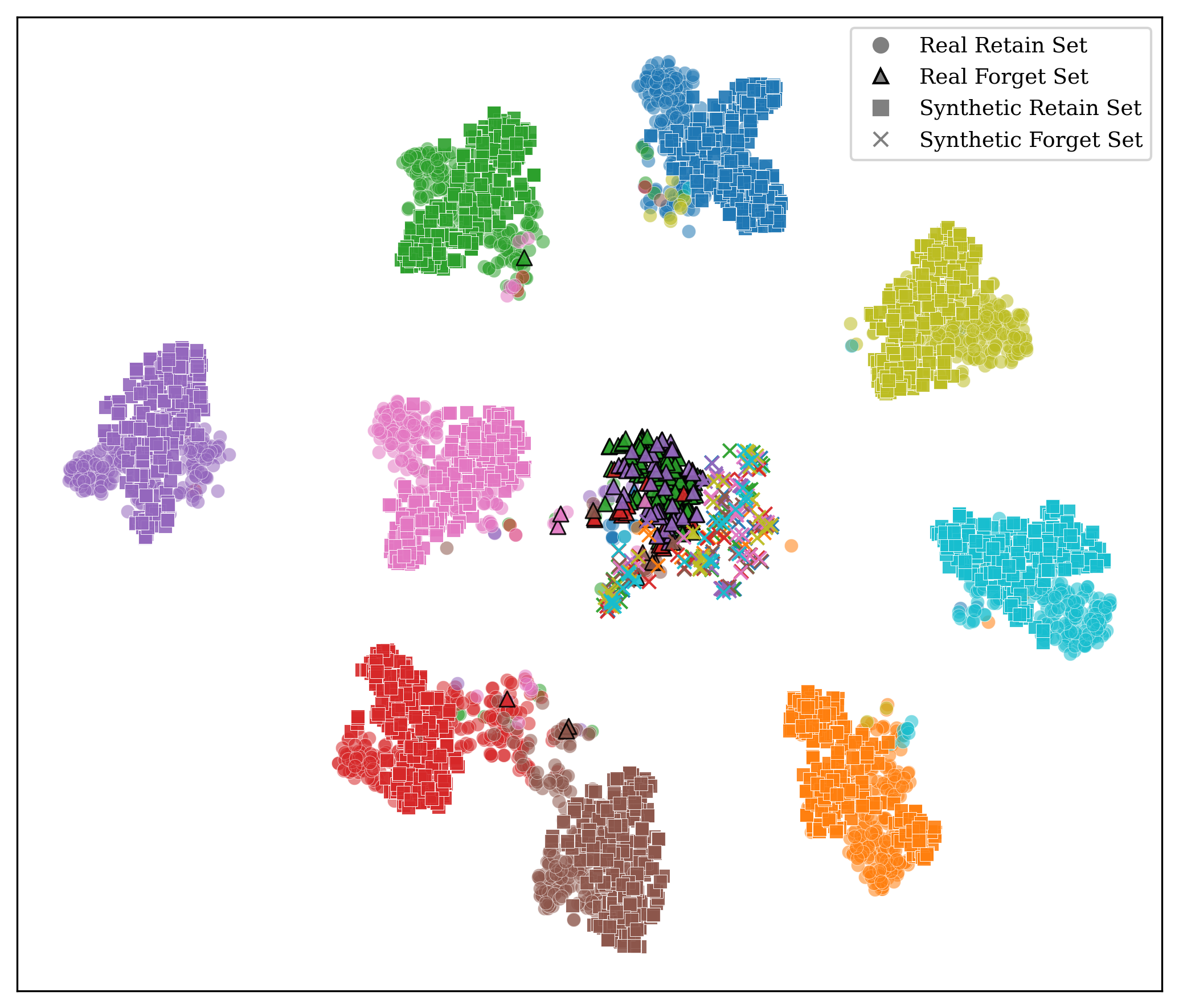}
    \caption{Bad Teacher: logit space}
\end{subfigure}
\hfill
\begin{subfigure}[h]{0.24\textwidth}
    \centering
    \includegraphics[width=\linewidth]
    {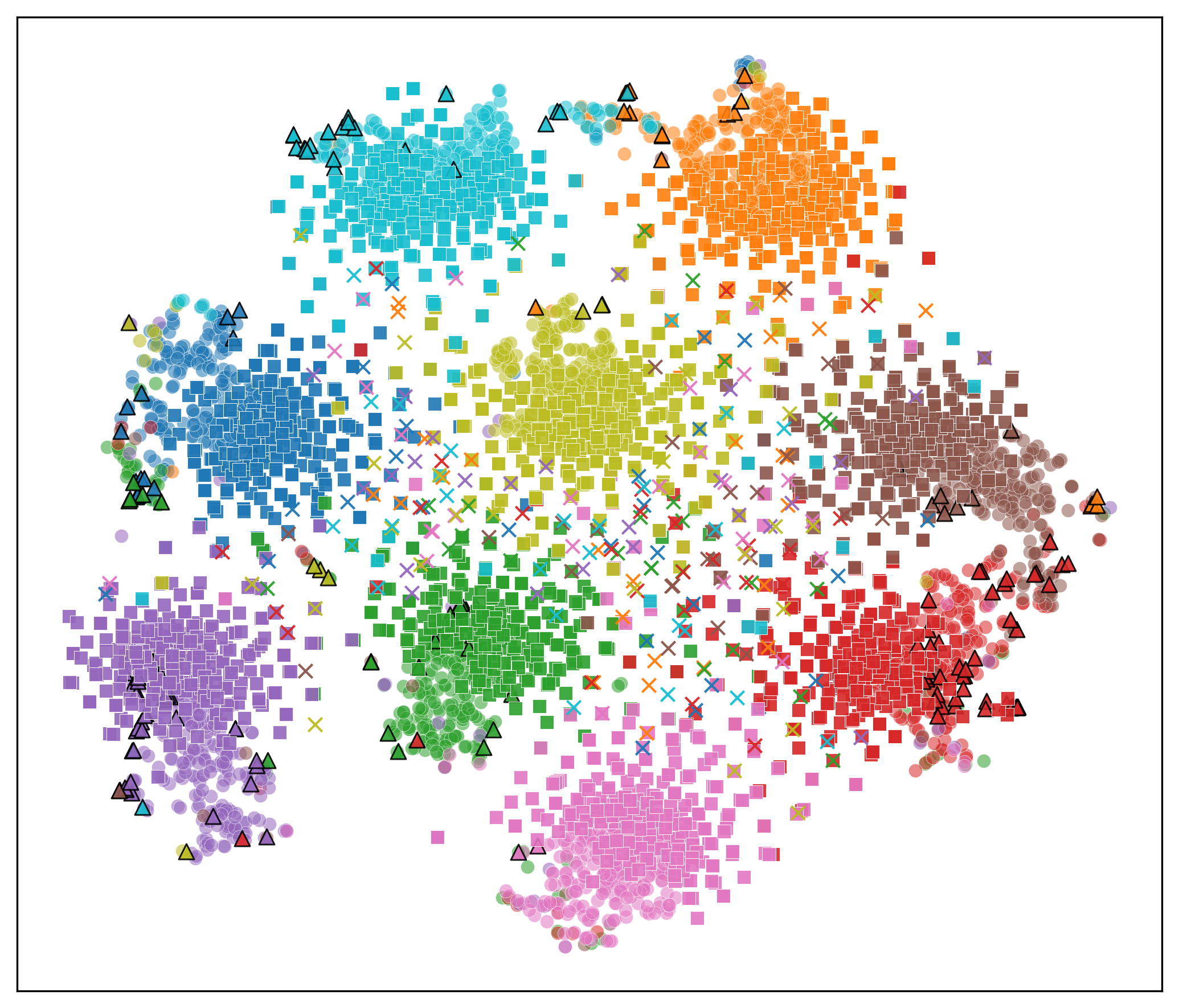}
    \caption{Negative Gradient+: feature space}
\end{subfigure}
\hfill
\begin{subfigure}[h]{0.24\textwidth}
    \centering
    \includegraphics[width=\linewidth]
    {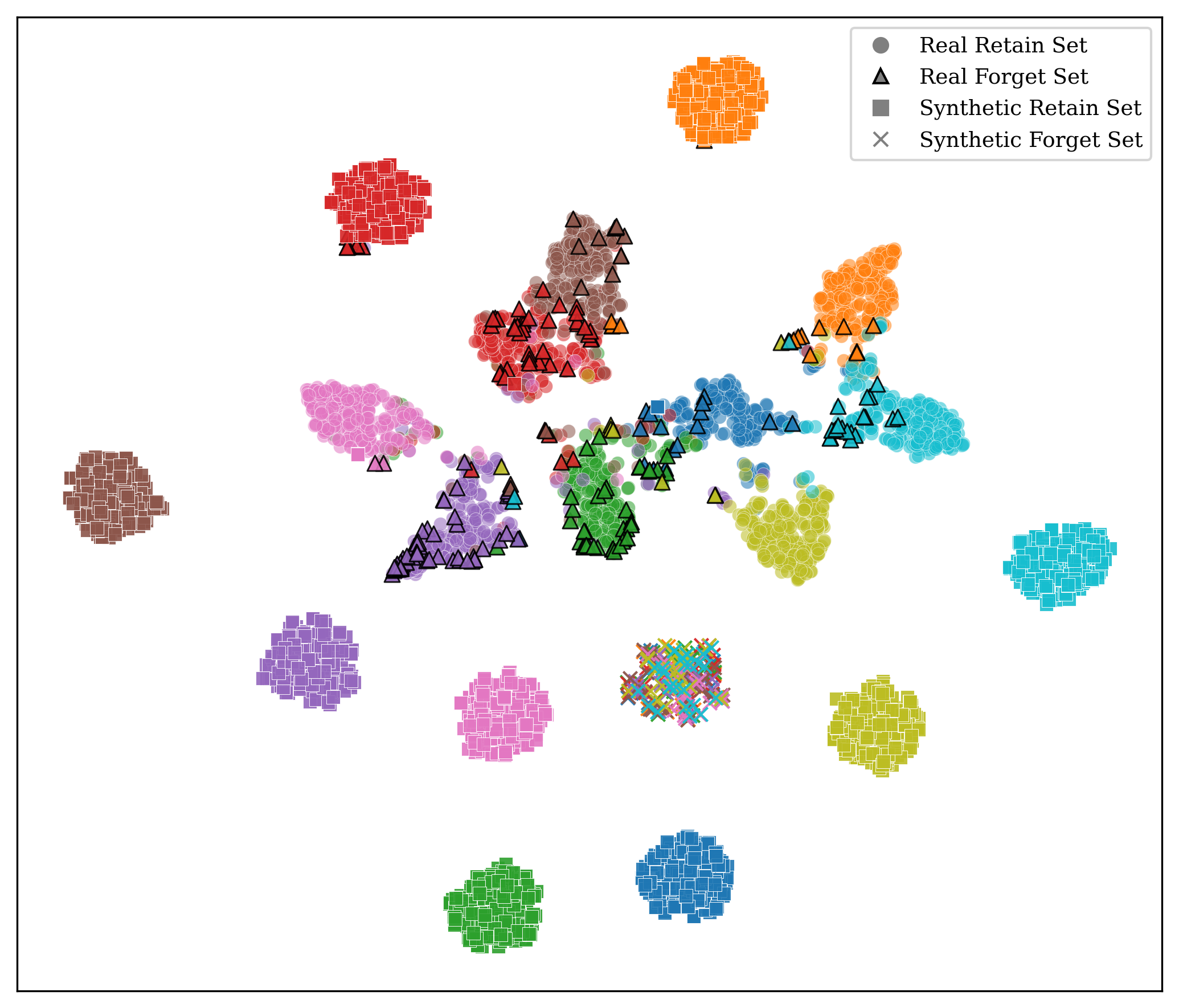}
    \caption{Negative Gradient+: logit space}
\end{subfigure}

\vspace{1mm}

% ================= ROW 3 =================

\begin{subfigure}[h]{0.24\textwidth}
    \centering
    \includegraphics[width=\linewidth]
    {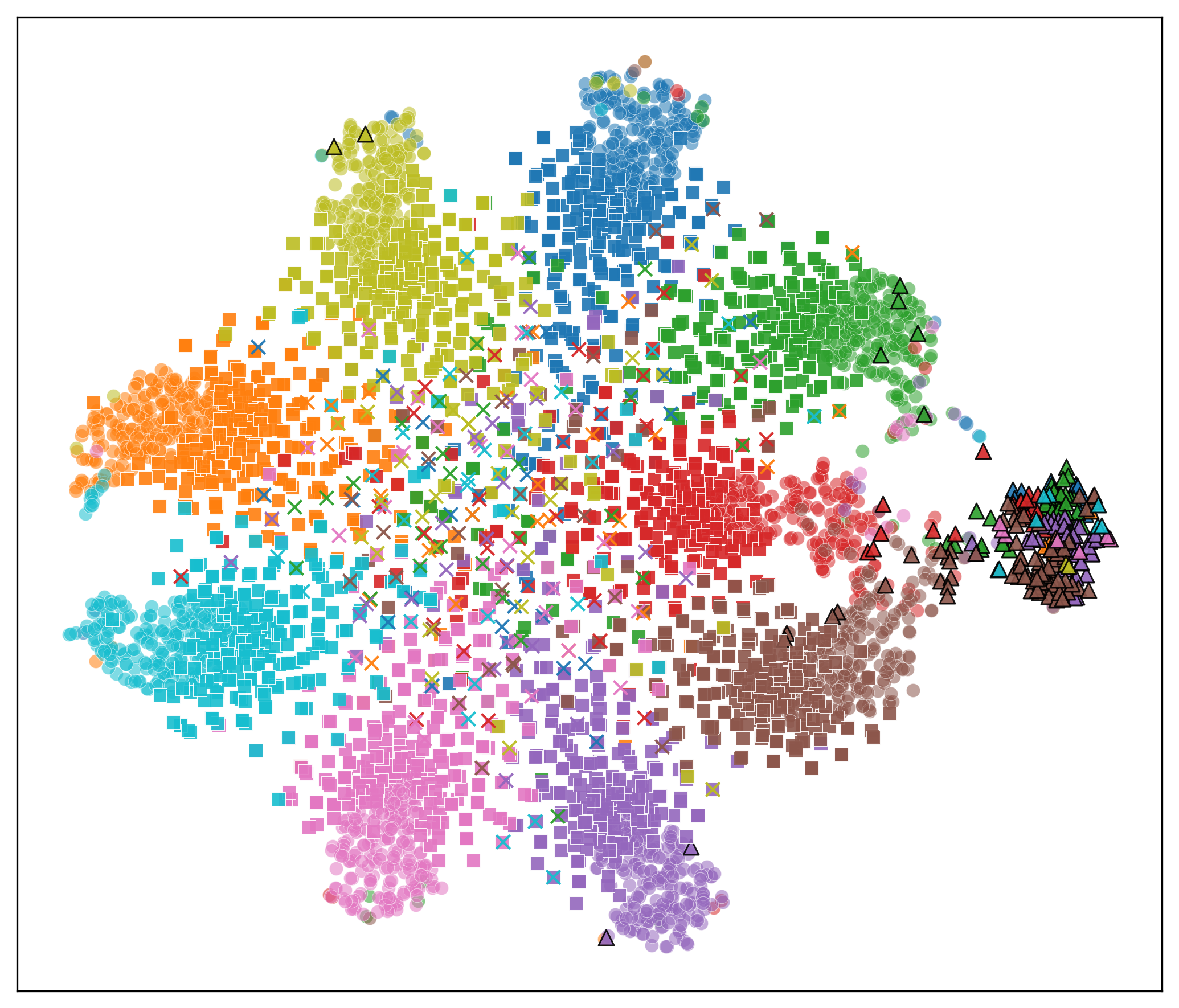}
    \caption{DELETE: feature space}
\end{subfigure}
\hfill
\begin{subfigure}[h]{0.24\textwidth}
    \centering
    \includegraphics[width=\linewidth]
    {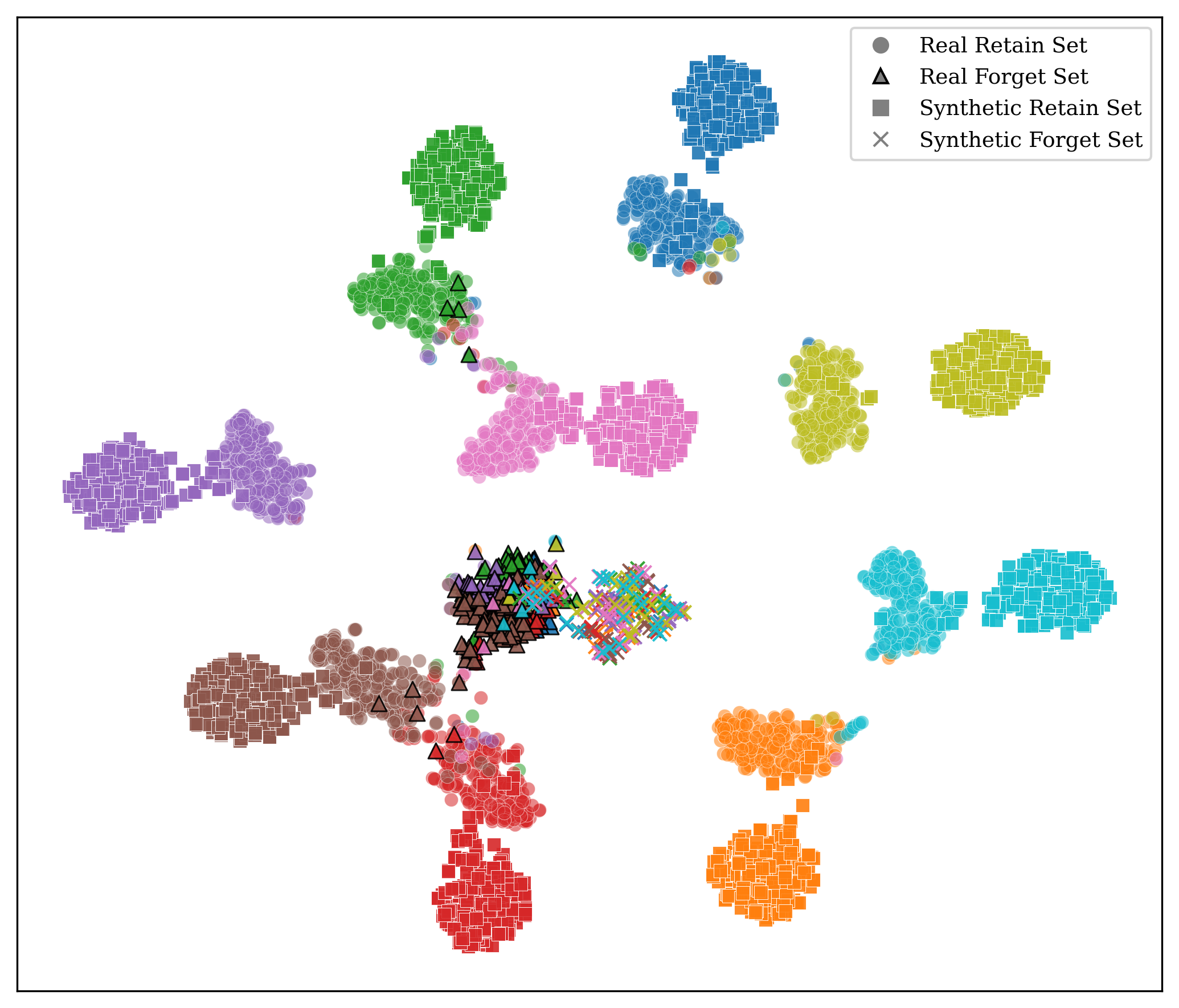}
    \caption{DELETE: logit space}
\end{subfigure}
\hfill
\begin{subfigure}[h]{0.24\textwidth}
    \centering
    \includegraphics[width=\linewidth]
    {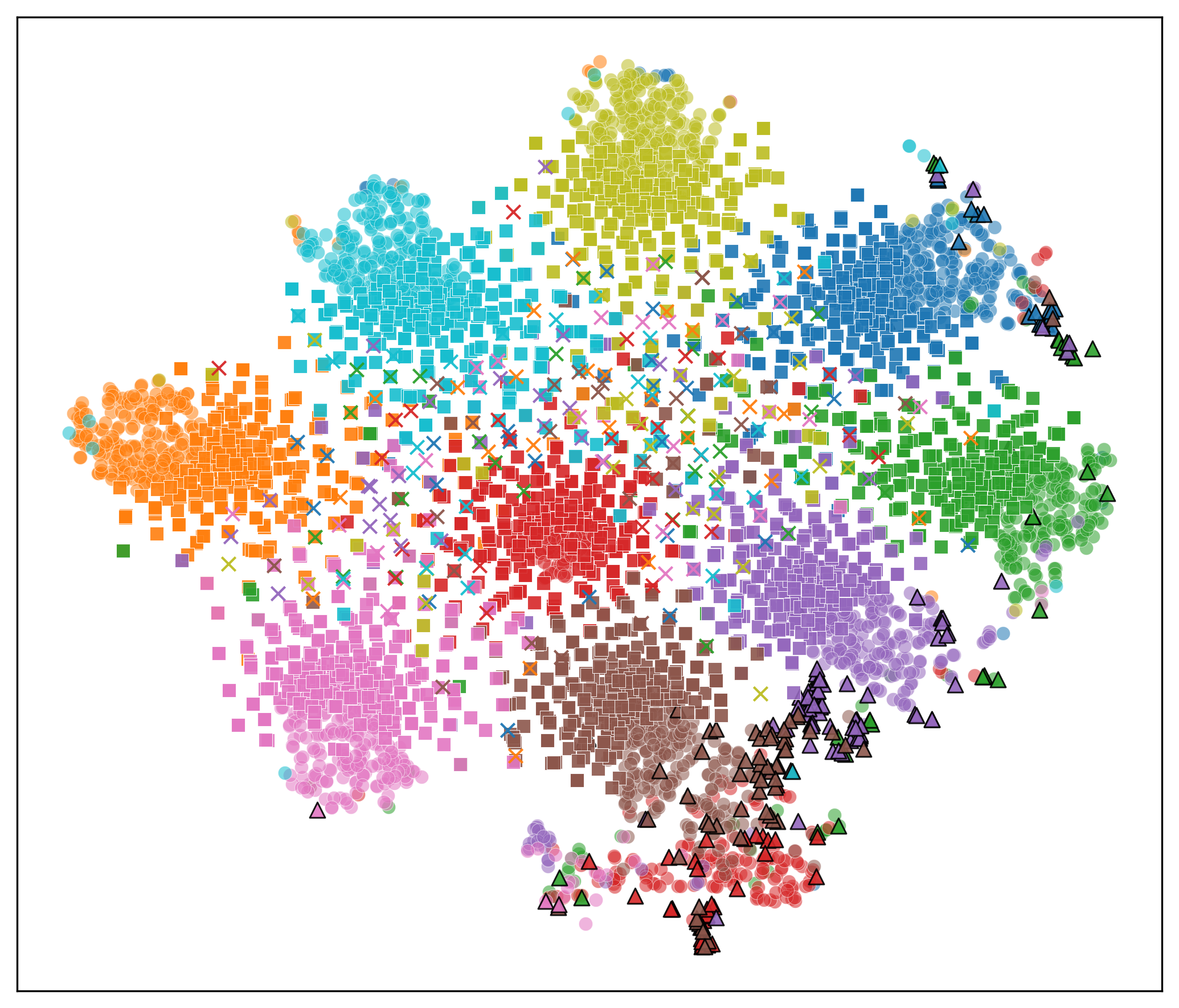}
    \caption{SCRUB: feature space}
\end{subfigure}
\hfill
\begin{subfigure}[h]{0.24\textwidth}
    \centering
    \includegraphics[width=\linewidth]
    {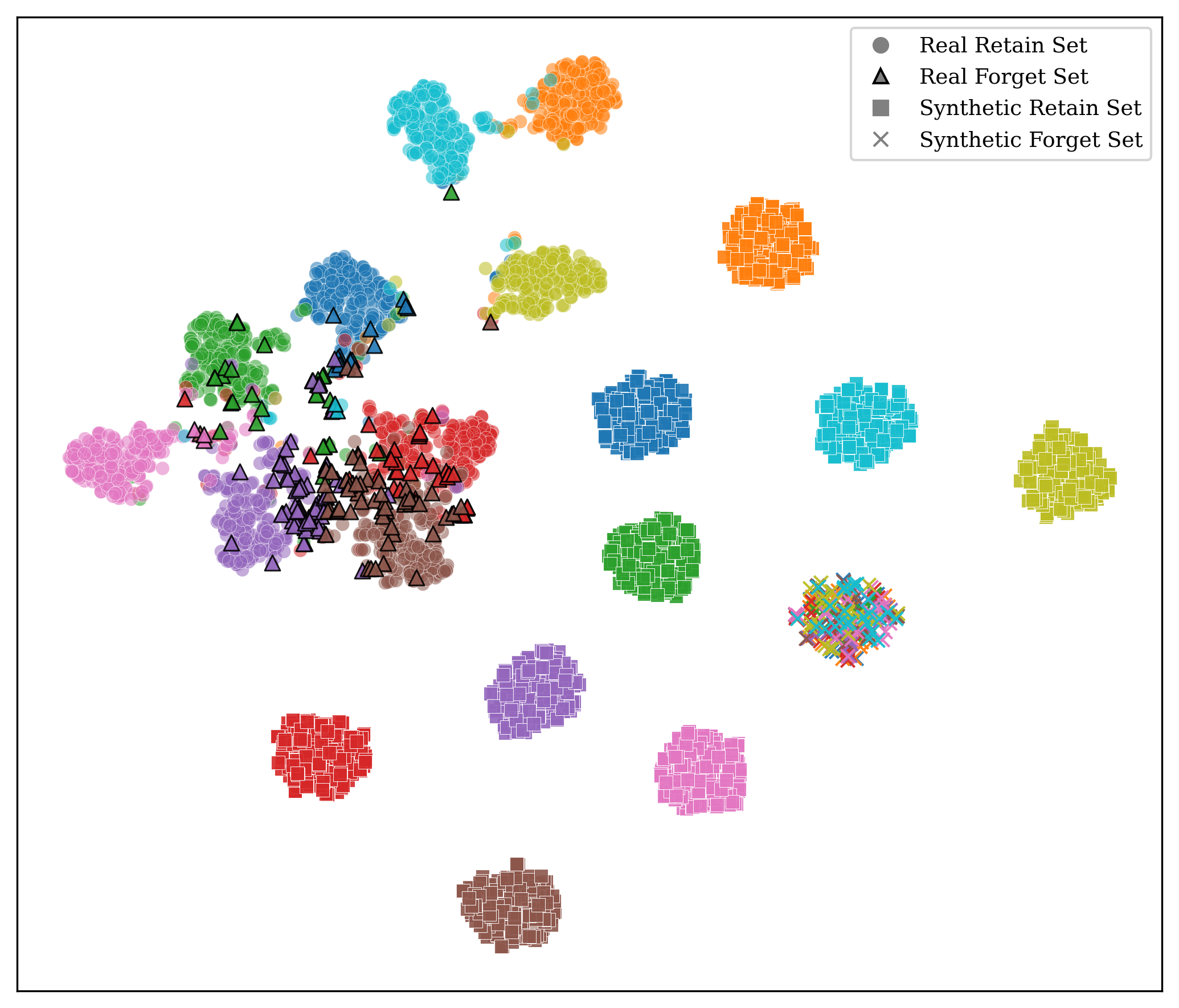}
    \caption{SCRUB: logit space}
\end{subfigure}

\vspace{1mm}

% ================= ROW 4 =================

\begin{subfigure}[h]{0.24\textwidth}
    \centering
    \includegraphics[width=\linewidth]
    {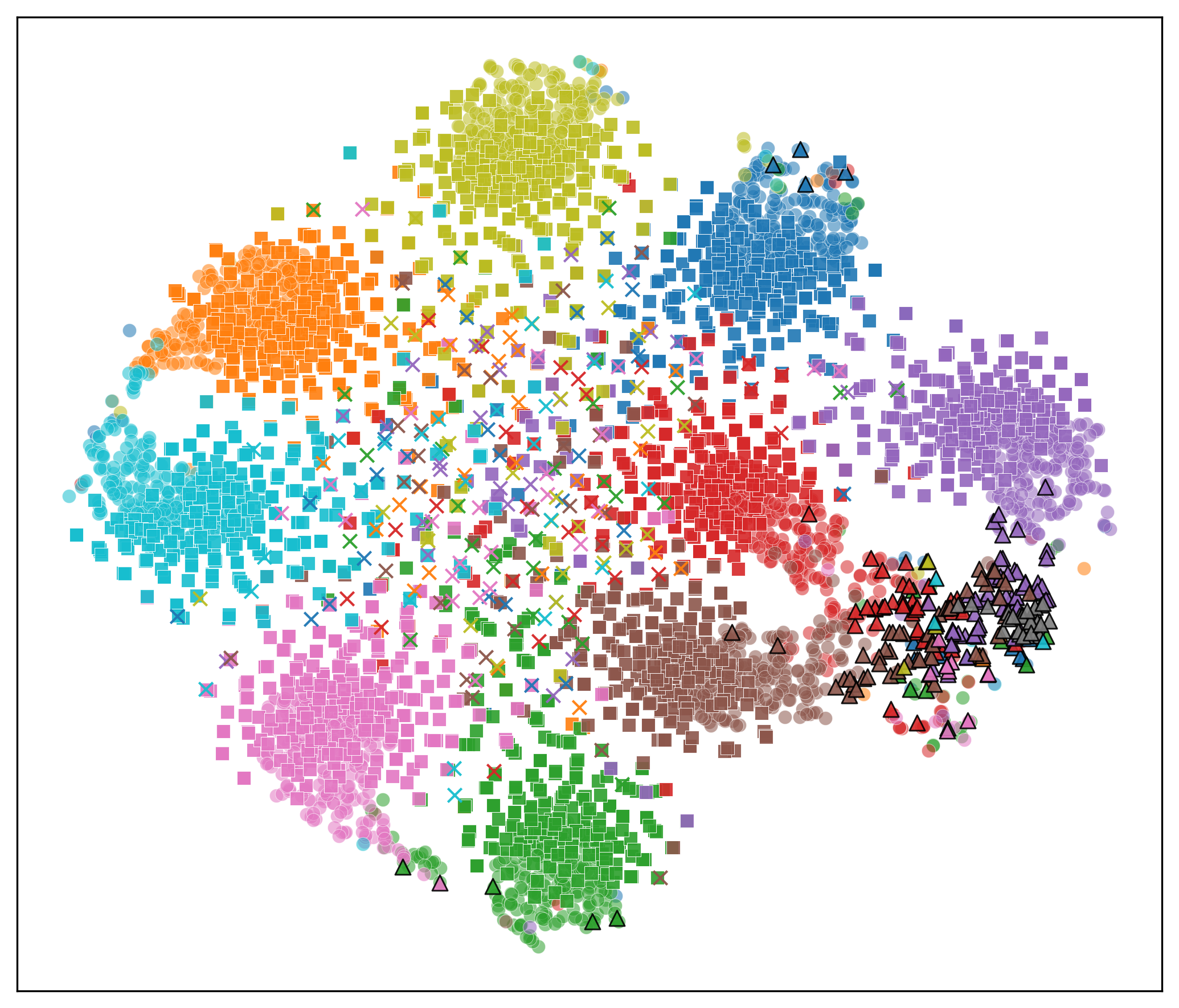}
    \caption{SalUn: feature space}
\end{subfigure}
\hfill
\begin{subfigure}[h]{0.24\textwidth}
    \centering
    \includegraphics[width=\linewidth]
    {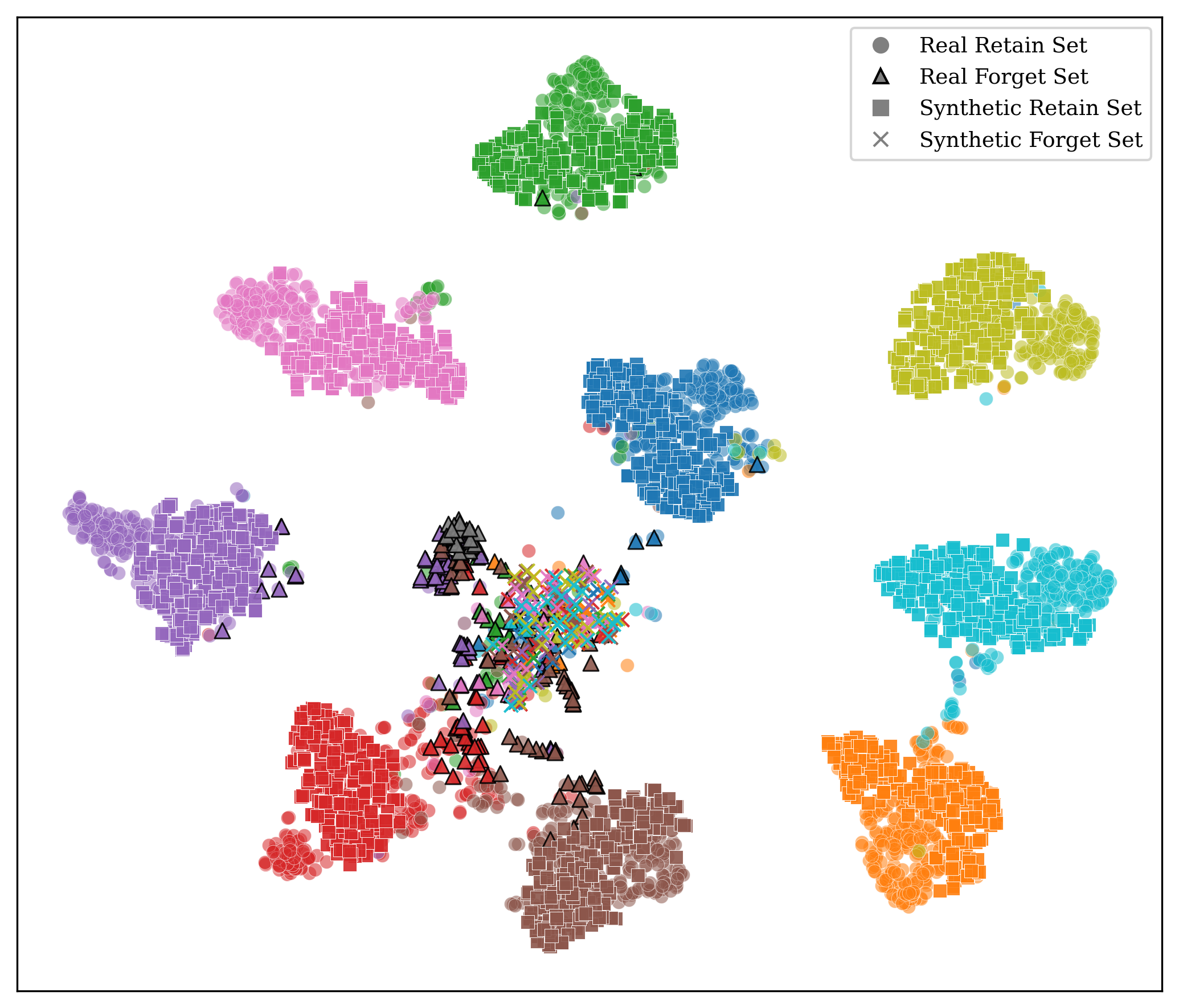}
    \caption{SalUn: logit space}
\end{subfigure}
\hfill
\begin{subfigure}[h]{0.24\textwidth}
    \centering
    \includegraphics[width=\linewidth]
    {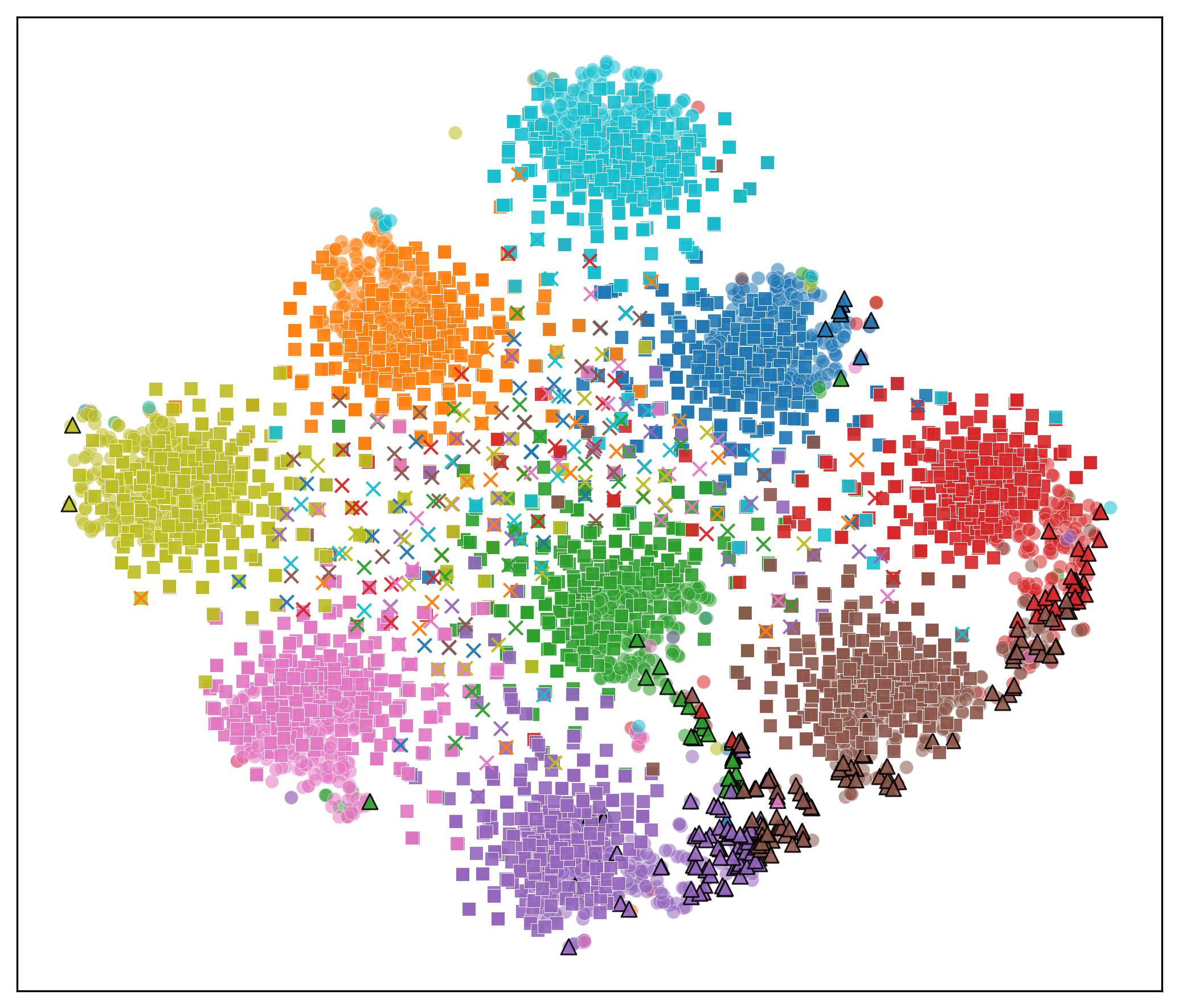}
    \caption{Retrained: feature space}
\end{subfigure}
\hfill
\begin{subfigure}[h]{0.24\textwidth}
    \centering
    \includegraphics[width=\linewidth]
    {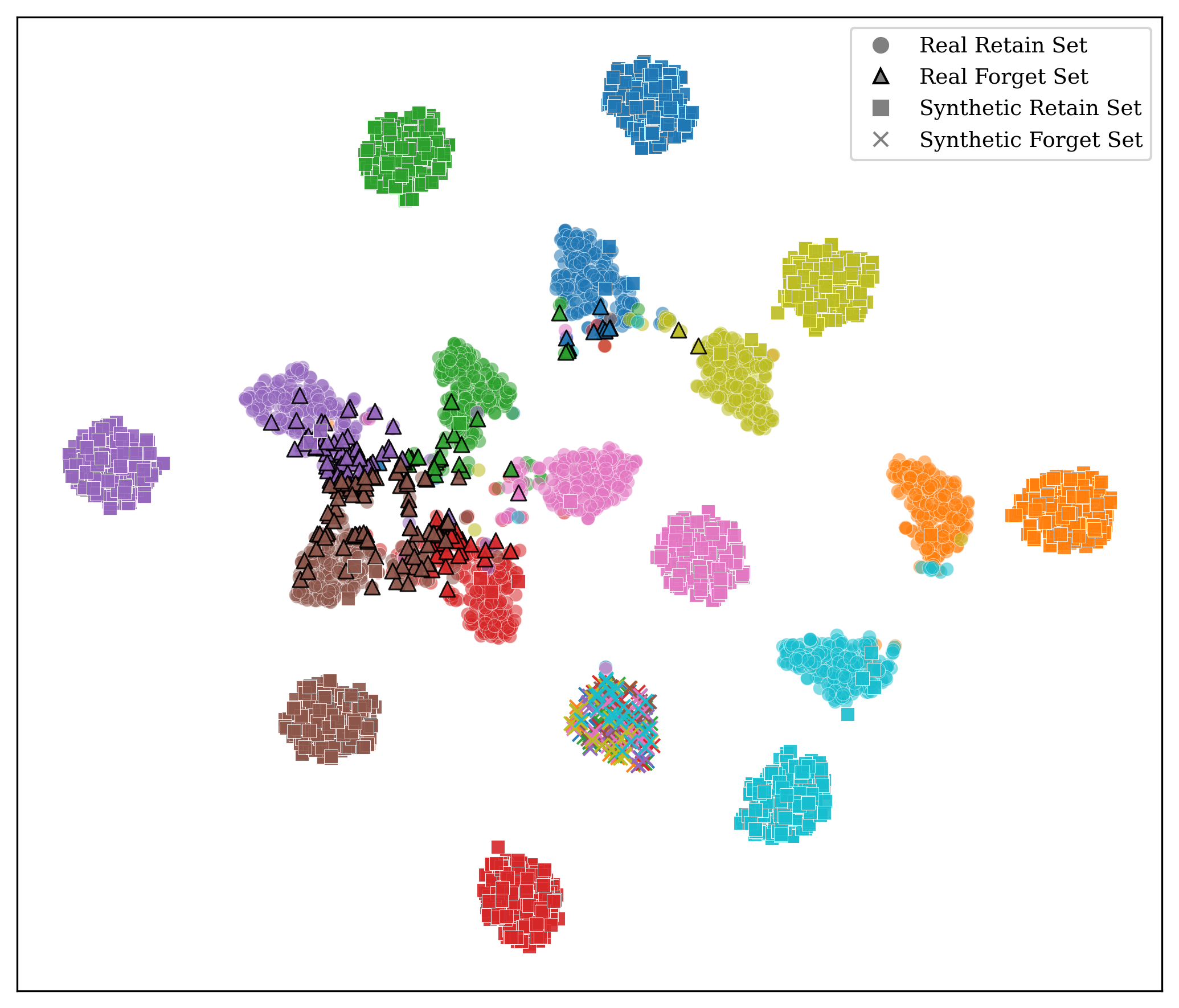}
    \caption{Retrained: logit space}
\end{subfigure}

\caption{
t-SNE visualization of representations on CIFAR-10 with ResNet-18 for class~$7$. The first row shows the original model as a common reference in the pre-classifier feature and classifier-head output spaces. The subsequent rows compare methods with higher recoverability (Bad Teacher, $\mathrm{RS}=0.98$; DELETE, $\mathrm{RS}=0.96$; SalUn, $\mathrm{RS}=0.87$) against methods with lower recoverability or the retrained reference (Negative Gradient+, $\mathrm{RS}=0.20$; SCRUB, $\mathrm{RS}=0.40$; Retrained, $\mathrm{RS}=0.52$). For each model, the left and right panels show the feature and logit spaces, respectively.}
\label{fig:tsne_recoverability_comparison}
\end{figure*}

Figure~\ref{fig:tsne_recoverability_comparison} provides qualitative geometric insight into the relationship between the selected synthetic probes and source-free recoverability. We compare methods exhibiting high recoverability (Bad Teacher, DELETE, and SalUn) with methods exhibiting lower recoverability (Negative Gradient+ and SCRUB), together with the retrained reference. For each method, we visualize both the pre-classifier feature space and the classifier-head output space. Importantly, the synthetic forget probes are not expected to reproduce the real forget class distribution. Indeed, in feature space, they generally do not coincide with the real forget embeddings, supporting their interpretation as boundary probes rather than synthetic reconstructions of the forget class.

A qualitative distinction emerges between the higher- and lower-recoverability cases. For highly recoverable methods, the real forget representations retain more coherent residual structure, while the model-selected synthetic forget probes exhibit a favorable relationship with the forget region, particularly after the classifier-head mapping. Consequently, a lightweight head update can exploit these probes to reform a decision region that recognizes the real forget samples. In contrast, for lower-recoverability methods, this relationship is substantially weaker and the real forget representations exhibit less recoverable organization, limiting the effectiveness of the same source-free update. The retrained model provides a matched reference for assessing the extent to which observed relearning can arise from generic representation-level separability rather than recoverability associated with the original forget class training.

These observations further clarify the role of Gaussian sampling in our method: the raw Gaussian distribution itself need not approximate the unknown forget class distribution. Rather, Gaussian sampling provides a broad candidate pool, while model-guided confidence filtering selects boundary-adjacent probes that can induce a useful update direction when recoverable forget class structure remains. This interpretation is consistent with Proposition~1, where successful relearning depends on sufficient alignment between the update induced by the selected synthetic probes and the residual real forget class representation, rather than on distributional matching.

\section{RS Distribution Across Forget Classes}
\label{app:rs_distribution}
The main paper includes the $\mathrm{RS}$ distribution analysis for the CIFAR-10 dataset with the ResNet-18 backbone as a representative case. In this section, we extend this analysis to all evaluated datasets and backbones, including CIFAR-10, CIFAR-100, and TinyImageNet with ResNet-18, ViT-B/16, and Swin-T. These plots provide a broader distributional view of $\mathrm{RS}$ across forget classes and show whether relearning is systematic or driven only by a few highly vulnerable classes. Each violin plot summarizes the $\mathrm{RS}$ values obtained by varying the designated forget class for a given dataset, backbone, and unlearning method. A higher median $\mathrm{RS}$ indicates that SFRA is common across forget classes, while a wider distribution indicates stronger class-specific variability.

\begin{figure*}[t]
    \centering

    \begin{subfigure}{\textwidth}
    \centering
    \includegraphics[width=\textwidth]{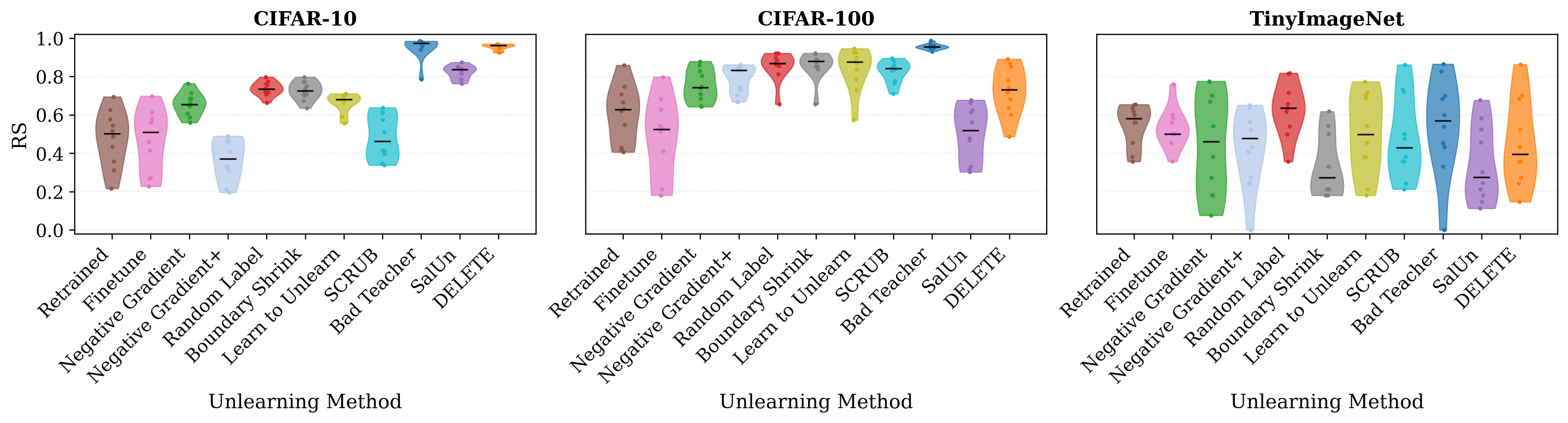}
    \caption{ResNet-18}
    \end{subfigure}

    \vspace{0.5em}

    \begin{subfigure}{\textwidth}
        \centering
        \includegraphics[width=\textwidth]{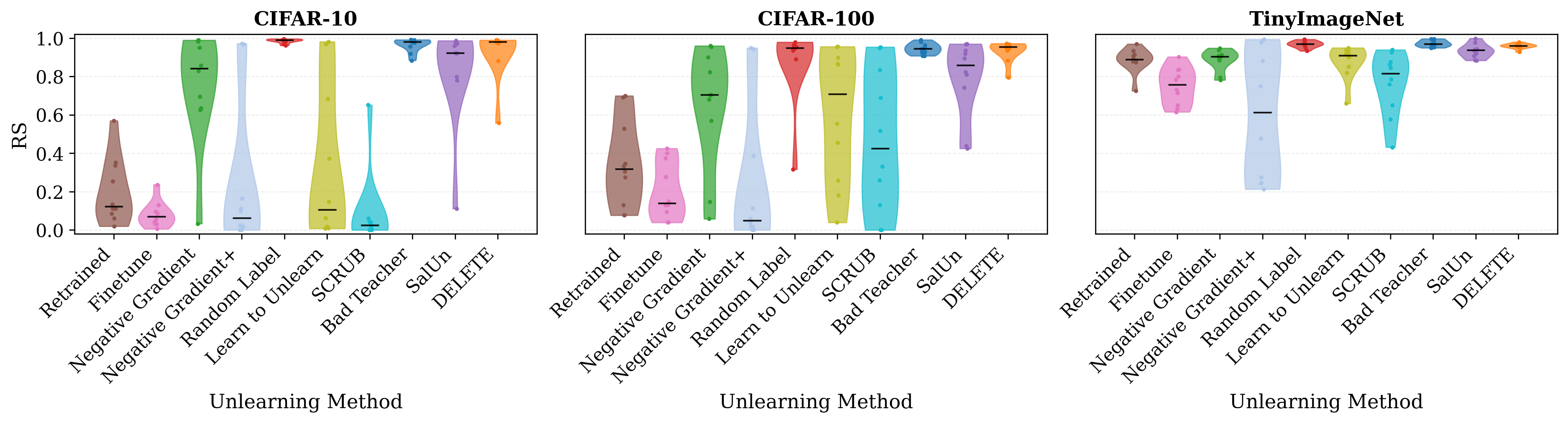}
        \caption{ViT-B/16}
    \end{subfigure}

    \vspace{0.5em}
    
    \begin{subfigure}{\textwidth}
        \centering
        \includegraphics[width=\textwidth]{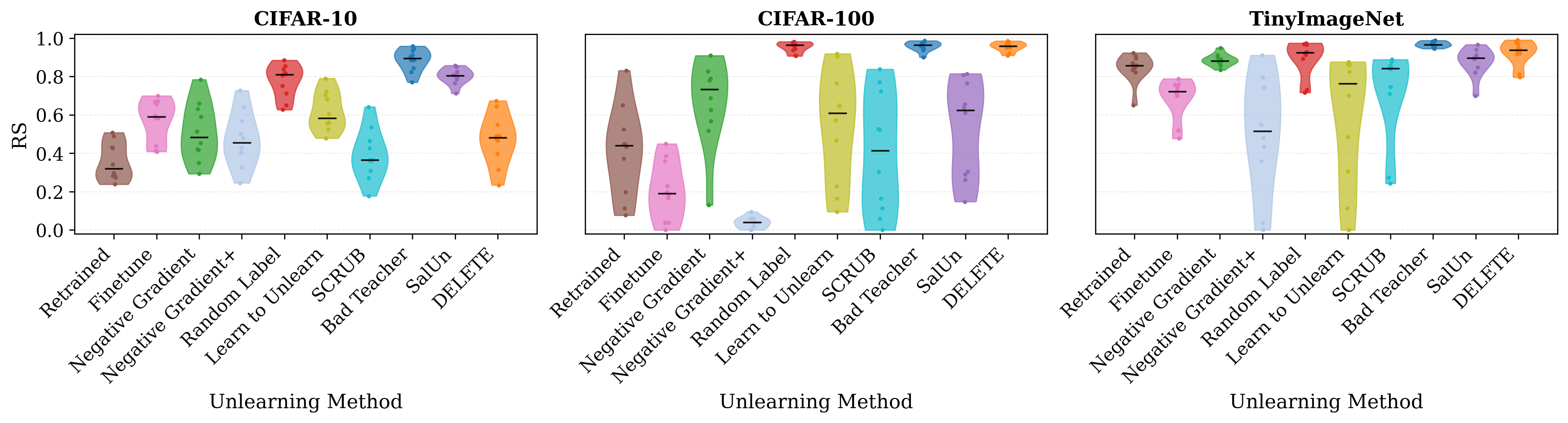}
        \caption{Swin-T}
    \end{subfigure}

    \vspace{0.5em}

    \caption{Distribution of $\mathrm{RS}$ across forget classes on three datasets (CIFAR-10, CIFAR-100, and TinyImageNet) and three backbones (ResNet-18, ViT-B/16, and Swin-T), enabling a comparison of unlearning methods. Rows correspond to backbones and columns to datasets; within each panel, each violin summarizes the per-class $\mathrm{RS}$ obtained by varying the designated forget class. The central marker denotes the median and the thick bar the interquartile range (IQR).}
    \label{fig:rs_violin_plot}
\end{figure*}

\section{Per-Class RS Heatmaps}
\label{app:rs_heatmaps}
The main paper reports the per-class $\mathrm{RS}$ heatmap for CIFAR-10 with the ResNet-18 backbone to illustrate class-specific relearning behavior in a compact setting. Here, we provide the full set of per-class $\mathrm{RS}$ heatmaps across all evaluated datasets and backbones, including CIFAR-10, CIFAR-100, and TinyImageNet with ResNet-18, ViT-B/16, and Swin-T. Each heatmap shows how vulnerable each forget class is to SFRA under different unlearning methods. Rows correspond to unlearning methods, columns correspond to forget classes, and each cell reports the $\mathrm{RS}$ obtained when the corresponding class is selected for forgetting. This extended analysis avoids hiding class-specific behavior behind averaged or worst-case summaries and helps identify both method-level and class-level relearning patterns.

\begin{figure*}[t]
    \centering

    \begin{subfigure}[t]{\textwidth}
        \centering
        \includegraphics[width=\linewidth]{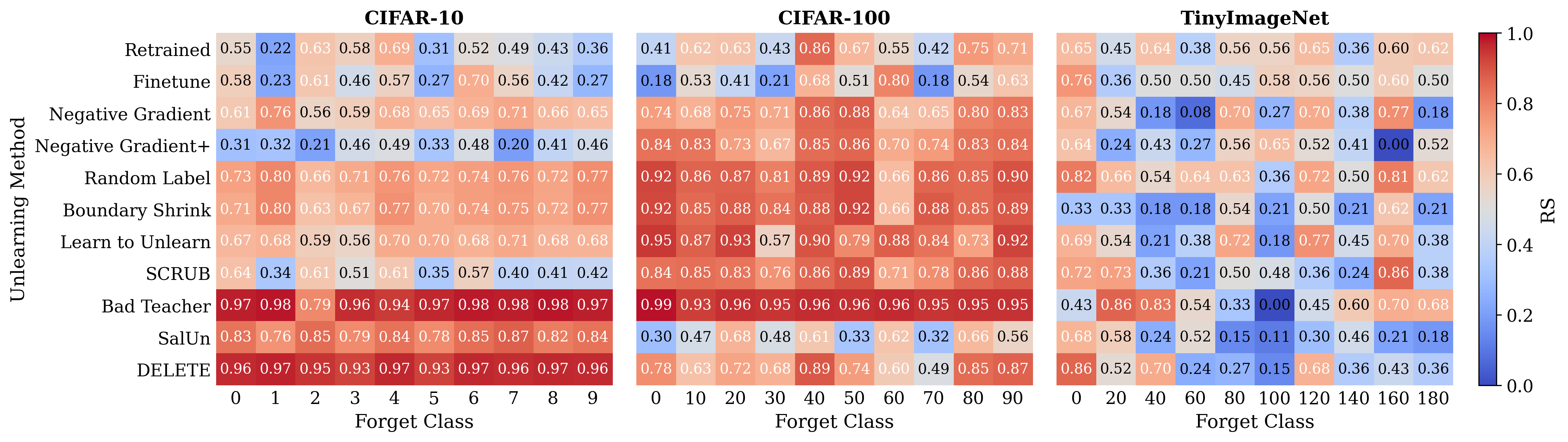}
        \caption{ResNet-18}
        \label{fig:rs_heatmap_resnet18}
    \end{subfigure}

    \vspace{0.5em}
  
    \begin{subfigure}[t]{\textwidth}
        \centering
        \includegraphics[width=\linewidth]{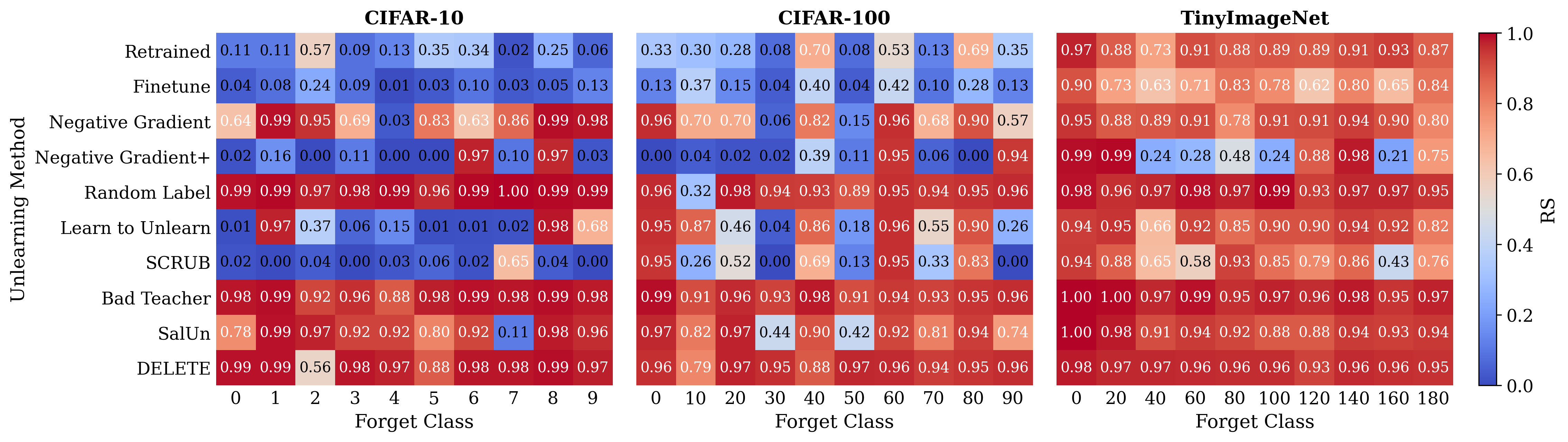}
        \caption{ViT-B/16}
        \label{fig:rs_heatmap_vitb16}
    \end{subfigure}
    
    \vspace{0.5em}

    \begin{subfigure}[t]{\textwidth}
        \centering
        \includegraphics[width=\linewidth]{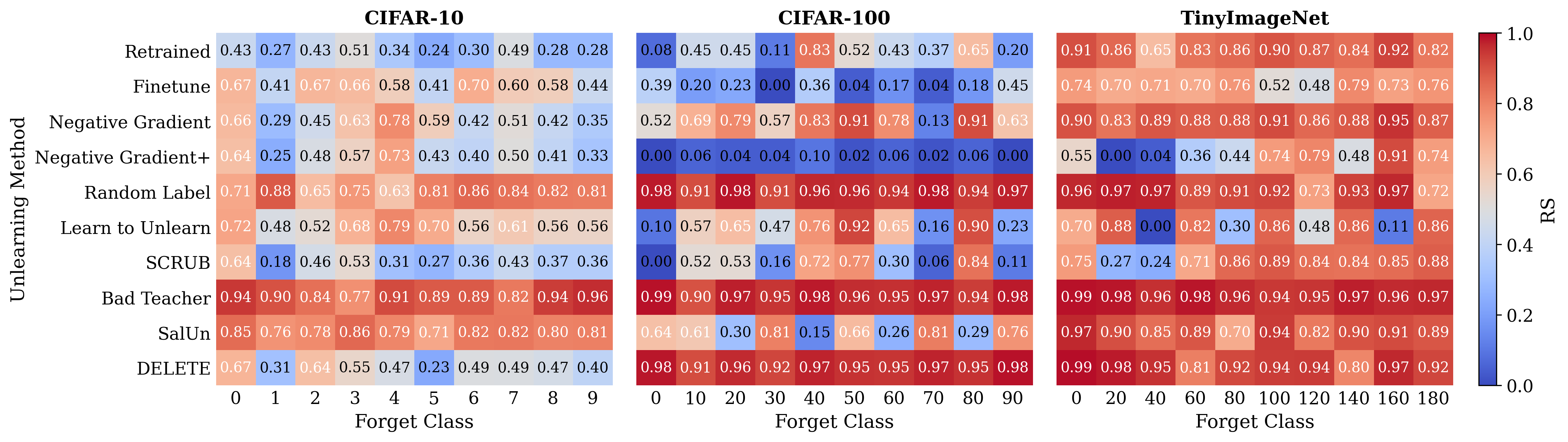}
        \caption{Swin-T}
        \label{fig:rs_heatmap_swint}
    \end{subfigure}
    
    \caption{Per-class $\mathrm{RS}$ heatmaps for three datasets (CIFAR-10, CIFAR-100, and TinyImageNet) and three backbones (ResNet-18, ViT-B/16, and Swin-T), comparing unlearning methods. Rows correspond to unlearning methods and columns to forget classes. Each cell reports the $\mathrm{RS}$ obtained when the corresponding class is designated for forgetting and color intensity encodes the $\mathrm{RS}$ magnitude.}
    \label{fig:rs_heatmap_plot}
\end{figure*}

\section{Absolute and Excess Recoverability}
\label{app:recoverability_diagnostic}

$\mathrm{RS}$ measures the absolute source-free recoverability of a forget class. However, a high $\mathrm{RS}$ does not necessarily indicate forget-specific residual structure, since a matched retrained model may also support relearning through generic representation transfer. We therefore report $\Delta\mathrm{RS}$, which measures excess recoverability relative to a model that never observed the forget class training data.

\begin{figure*}[t]
    \centering

    \begin{subfigure}{0.98\textwidth}
        \centering
        \includegraphics[width=\textwidth]
        {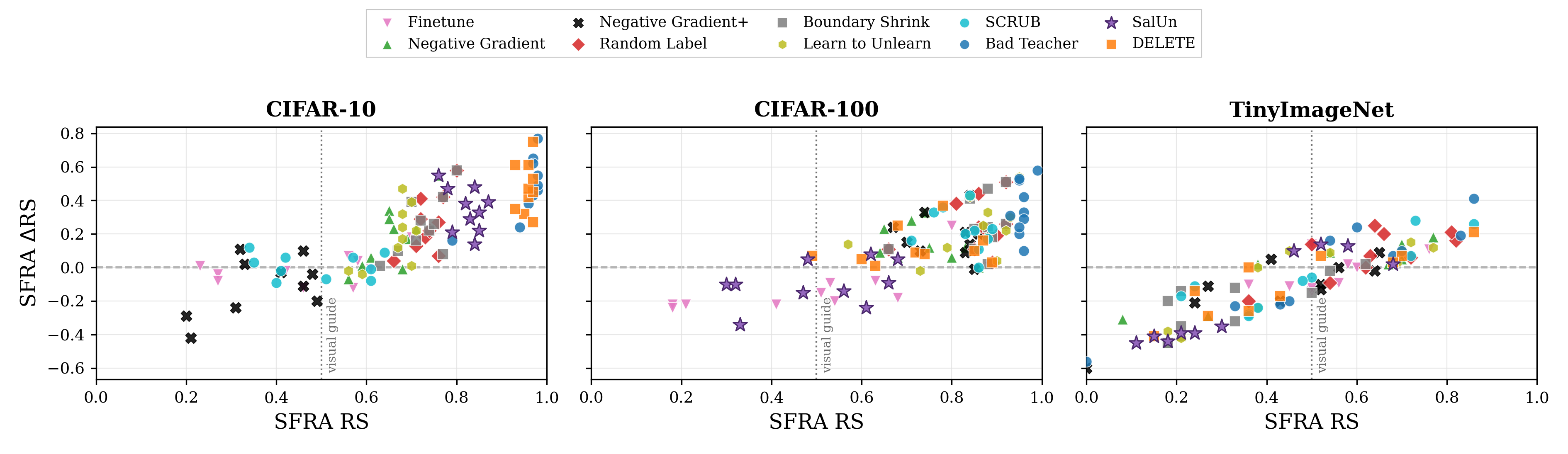}
        \caption{ResNet-18}
        \label{fig:rs_delta_resnet18}
    \end{subfigure}

    \vspace{2mm}

    \begin{subfigure}{0.98\textwidth}
        \centering
        \includegraphics[width=\textwidth]
        {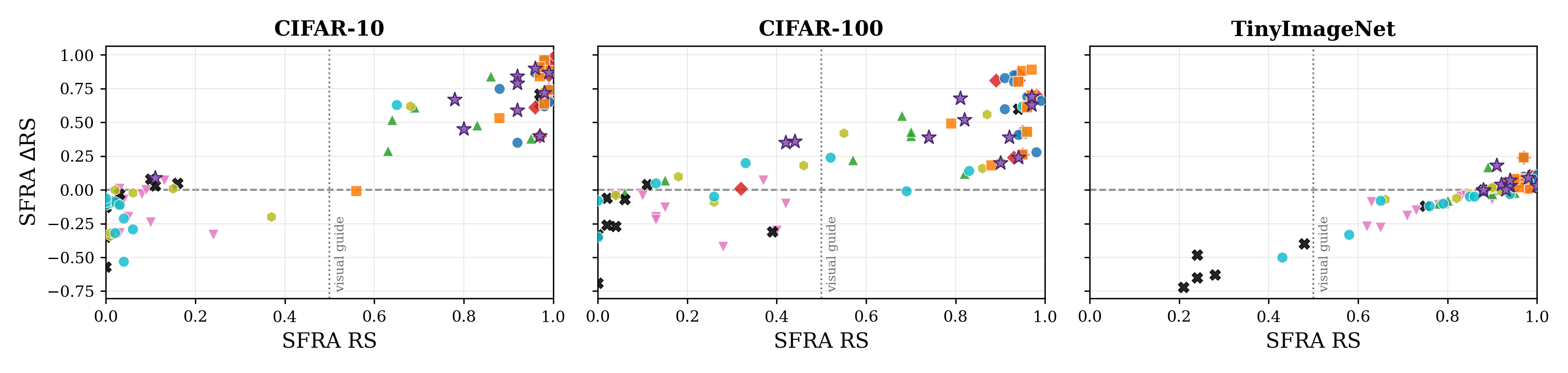}
        \caption{ViT-B/16}
        \label{fig:rs_delta_vit}
    \end{subfigure}

    \vspace{2mm}

    \begin{subfigure}{0.98\textwidth}
        \centering
        \includegraphics[width=\textwidth]
        {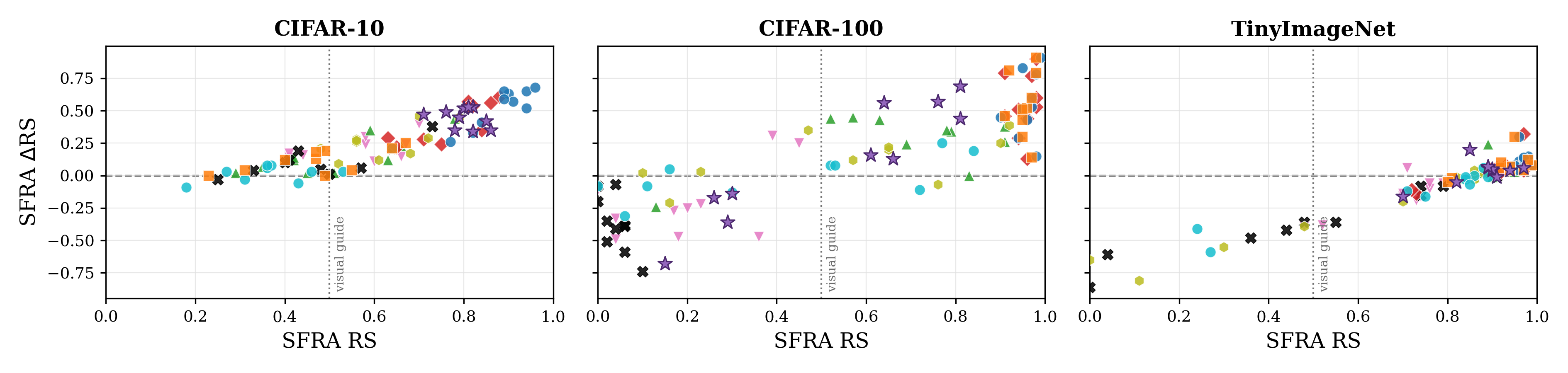}
        \caption{Swin-T}
        \label{fig:rs_delta_swin}
    \end{subfigure}

    \caption{
    Absolute and excess recoverability under SFRA.
    Each point represents one unlearning-method and forget class pair on CIFAR-10, CIFAR-100, or TinyImageNet. The horizontal dashed line marks $\Delta\mathrm{RS}=0$. Points above this line are more recoverable than their corresponding retrained references. The vertical dotted line at $\mathrm{RS}=0.5$ is a visual guide only and does not define a theoretical threshold. The upper-right region therefore identifies checkpoints with both high absolute recoverability and positive excess recoverability.
    }
    \label{fig:rs_delta_diagnostic}
\end{figure*}

Figure~\ref{fig:rs_delta_diagnostic} jointly visualizes $\mathrm{RS}$ and $\Delta\mathrm{RS}$ for ResNet-18, ViT-B/16, and Swin-T, with each point representing an (unlearning method--forget class) pair. The horizontal line at $\Delta\mathrm{RS}=0$ distinguishes checkpoints that are more recoverable than their matched retrained references, while the vertical line at $\mathrm{RS}=0.5$ serves only as a visual guide. High $\mathrm{RS}$ with positive $\Delta\mathrm{RS}$ provides the strongest evidence of recovery beyond generic relearnability. Many CIFAR-10 and CIFAR-100 pairs exhibit this behavior, whereas TinyImageNet contains more cases with lower $\mathrm{RS}$ or nonpositive $\Delta\mathrm{RS}$. Because $\Delta\mathrm{RS}$ is defined using $\mathrm{RS}$, we interpret the figure as a diagnostic decomposition rather than a correlation analysis.

\section{Sampling Distribution Ablation}
\label{app:distribution_ablation}
Our default implementation generates candidate embeddings from a standard Gaussian distribution, $s \sim \mathcal{N}(0,I_d)$. These samples are not assumed to lie on the natural feature manifold. In particular, for classifier inputs constrained by a final ReLU, a standard Gaussian produces negative coordinates that cannot occur for real post-ReLU features. This is intentional: the proposal distribution supplies candidate directions rather than synthetic reconstructions of real features, and model-guided confidence filtering determines which candidates become probes. To evaluate whether our proposed SFRA depends on this particular proposal distribution, we replace Gaussian sampling with Uniform and Laplace proposals while keeping all other components unchanged. Table~\ref{tab:cifar10_resnet18_sampling_distribution_rs} reports the per-class $\mathrm{RS}$ on CIFAR-10 with a ResNet-18 backbone. Although Uniform and Laplace sampling yield some differences in absolute $\mathrm{RS}$ values across classes, both generally preserve the relative recoverability patterns observed with Gaussian sampling. In particular, methods that are highly vulnerable or relatively resistant under Gaussian sampling tend to remain so under the alternative proposal distributions. Instead, the model-guided partitioning and confidence-based selection of boundary-adjacent forget probes and high-confidence retain probes play a more important role in determining the relearning outcome.

\begin{table}[t]
\centering
\caption{$\mathrm{RS}$ for Gaussian, Uniform, and Laplace synthesis distributions on CIFAR-10 using ResNet-18. Each forget class column corresponds to a separate unlearned checkpoint in which that class is designated for forgetting.}
\label{tab:cifar10_resnet18_sampling_distribution_rs}
\setlength{\tabcolsep}{2pt}
\renewcommand{\arraystretch}{0.88}
\scriptsize
\resizebox{\columnwidth}{!}{%
\begin{tabular}{l|l|cccccccccc}
\toprule
\multirow{2}{*}{Unlearning Method} & \multirow{2}{*}{\shortstack{Embedding\\Distribution}} & \multicolumn{10}{c}{Forget Class} \\
 & & 0 & 1 & 2 & 3 & 4 & 5 & 6 & 7 & 8 & 9 \\
\midrule
\multirow{3}{*}{Retrained} & Gaussian & $0.55$ & $0.22$ & $0.63$ & $0.58$ & $0.69$ & $0.31$ & $0.52$ & $0.49$ & $0.43$ & $0.36$ \\
 & Uniform & $0.49$ & $0.14$ & $0.48$ & $0.41$ & $0.54$ & $0.27$ & $0.37$ & $0.38$ & $0.30$ & $0.25$ \\
 & Laplace & $0.38$ & $0.13$ & $0.48$ & $0.42$ & $0.54$ & $0.19$ & $0.39$ & $0.39$ & $0.39$ & $0.23$ \\
\midrule
\multirow{3}{*}{Finetune \cite{golatkar2020eternal}} & Gaussian & $0.58$ & $0.23$ & $0.61$ & $0.46$ & $0.57$ & $0.27$ & $0.70$ & $0.56$ & $0.42$ & $0.27$ \\
 & Uniform & $0.59$ & $0.16$ & $0.56$ & $0.48$ & $0.56$ & $0.28$ & $0.67$ & $0.53$ & $0.39$ & $0.22$ \\
 & Laplace & $0.53$ & $0.13$ & $0.58$ & $0.45$ & $0.48$ & $0.25$ & $0.60$ & $0.44$ & $0.38$ & $0.19$ \\
\midrule
\multirow{3}{*}{Negative Gradient \cite{golatkar2020eternal}} & Gaussian & $0.61$ & $0.76$ & $0.56$ & $0.59$ & $0.68$ & $0.65$ & $0.69$ & $0.71$ & $0.66$ & $0.65$ \\
 & Uniform & $0.55$ & $0.69$ & $0.49$ & $0.54$ & $0.66$ & $0.65$ & $0.68$ & $0.70$ & $0.57$ & $0.58$ \\
 & Laplace & $0.53$ & $0.71$ & $0.50$ & $0.56$ & $0.65$ & $0.63$ & $0.65$ & $0.67$ & $0.62$ & $0.56$ \\
\midrule
\multirow{3}{*}{Negative Gradient+ \cite{kurmanji2023towards}} & Gaussian & $0.31$ & $0.32$ & $0.21$ & $0.46$ & $0.49$ & $0.33$ & $0.48$ & $0.20$ & $0.41$ & $0.46$ \\
 & Uniform & $0.25$ & $0.31$ & $0.18$ & $0.39$ & $0.40$ & $0.37$ & $0.42$ & $0.17$ & $0.32$ & $0.37$ \\
 & Laplace & $0.31$ & $0.26$ & $0.21$ & $0.39$ & $0.39$ & $0.25$ & $0.45$ & $0.11$ & $0.33$ & $0.37$ \\
\midrule
\multirow{3}{*}{Random Label \cite{hayase2020selective}} & Gaussian & $0.73$ & $0.80$ & $0.66$ & $0.71$ & $0.76$ & $0.72$ & $0.74$ & $0.76$ & $0.72$ & $0.77$ \\
 & Uniform & $0.72$ & $0.74$ & $0.64$ & $0.62$ & $0.76$ & $0.66$ & $0.69$ & $0.72$ & $0.66$ & $0.75$ \\
 & Laplace & $0.69$ & $0.74$ & $0.64$ & $0.66$ & $0.75$ & $0.68$ & $0.66$ & $0.70$ & $0.70$ & $0.70$ \\
\midrule
\multirow{3}{*}{Boundary Shrink \cite{chen2023boundary}} & Gaussian & $0.71$ & $0.80$ & $0.63$ & $0.67$ & $0.77$ & $0.70$ & $0.74$ & $0.75$ & $0.72$ & $0.77$ \\
 & Uniform & $0.64$ & $0.78$ & $0.59$ & $0.61$ & $0.70$ & $0.65$ & $0.69$ & $0.72$ & $0.71$ & $0.71$ \\
 & Laplace & $0.69$ & $0.74$ & $0.62$ & $0.60$ & $0.73$ & $0.69$ & $0.66$ & $0.70$ & $0.69$ & $0.73$ \\
\midrule
\multirow{3}{*}{Learn to Unlearn \cite{cha2024learning}} & Gaussian & $0.67$ & $0.68$ & $0.59$ & $0.56$ & $0.70$ & $0.70$ & $0.68$ & $0.71$ & $0.68$ & $0.68$ \\
 & Uniform & $0.67$ & $0.60$ & $0.50$ & $0.48$ & $0.63$ & $0.69$ & $0.67$ & $0.66$ & $0.67$ & $0.59$ \\
 & Laplace & $0.65$ & $0.63$ & $0.54$ & $0.49$ & $0.69$ & $0.61$ & $0.59$ & $0.65$ & $0.64$ & $0.62$ \\
\midrule
\multirow{3}{*}{SCRUB \cite{kurmanji2023towards}} & Gaussian & $0.64$ & $0.34$ & $0.61$ & $0.51$ & $0.61$ & $0.35$ & $0.57$ & $0.40$ & $0.41$ & $0.42$ \\
 & Uniform & $0.66$ & $0.40$ & $0.70$ & $0.50$ & $0.62$ & $0.51$ & $0.50$ & $0.51$ & $0.50$ & $0.34$ \\
 & Laplace & $0.61$ & $0.41$ & $0.71$ & $0.46$ & $0.52$ & $0.50$ & $0.50$ & $0.47$ & $0.40$ & $0.33$ \\
\midrule
\multirow{3}{*}{Bad Teacher \cite{chundawat2023can}} & Gaussian & $0.97$ & $0.98$ & $0.79$ & $0.96$ & $0.94$ & $0.97$ & $0.98$ & $0.98$ & $0.98$ & $0.97$ \\
 & Uniform & $0.98$ & $0.99$ & $0.79$ & $0.96$ & $0.94$ & $0.97$ & $0.98$ & $0.98$ & $0.99$ & $0.98$ \\
 & Laplace & $0.98$ & $0.99$ & $0.78$ & $0.96$ & $0.94$ & $0.97$ & $0.98$ & $0.99$ & $0.99$ & $0.98$ \\
\midrule
\multirow{3}{*}{SalUn \cite{fan2023salun}} & Gaussian & $0.83$ & $0.76$ & $0.85$ & $0.79$ & $0.84$ & $0.78$ & $0.85$ & $0.87$ & $0.82$ & $0.84$ \\
 & Uniform & $0.80$ & $0.76$ & $0.83$ & $0.80$ & $0.85$ & $0.82$ & $0.88$ & $0.87$ & $0.78$ & $0.86$ \\
 & Laplace & $0.84$ & $0.80$ & $0.85$ & $0.81$ & $0.85$ & $0.80$ & $0.86$ & $0.85$ & $0.82$ & $0.84$ \\
\midrule
\multirow{3}{*}{DELETE \cite{zhou2025decoupled}} & Gaussian & $0.96$ & $0.97$ & $0.95$ & $0.93$ & $0.97$ & $0.93$ & $0.97$ & $0.96$ & $0.97$ & $0.96$ \\
 & Uniform & $0.96$ & $0.97$ & $0.94$ & $0.92$ & $0.96$ & $0.92$ & $0.96$ & $0.96$ & $0.96$ & $0.96$ \\
 & Laplace & $0.96$ & $0.97$ & $0.94$ & $0.93$ & $0.96$ & $0.92$ & $0.97$ & $0.95$ & $0.96$ & $0.96$ \\
\bottomrule
\end{tabular}%
}
\end{table}

\section{Uncertainty-Score Ablation}
\label{app:uncertainty_metric}

We examine whether our proposed SFRA depends specifically on Softmax confidence by replacing its probe-ranking score with predictive entropy and energy. All other components of the audit, including the unlearned checkpoint, Gaussian candidate pool, number of selected probes, classifier-head optimization, and evaluation protocol, remain fixed. For a candidate embedding $s$ with classifier logits $z(s)$ and predicted probabilities $p(s)$, we consider the following scores:
\begin{align}
    U_{\mathrm{Softmax}}(s)
    &= 1-\max_k p_k(s),\\
    U_{\mathrm{Entropy}}(s)
    &= -\sum_k p_k(s)\log p_k(s),\\
    U_{\mathrm{Energy}}(s)
    &= -\log\sum_k \exp(z_k(s)).
\end{align}
For each score, candidates with the highest uncertainty are selected as synthetic forget probes and relabeled as the designated forget class, while the lowest-uncertainty remaining candidates form the synthetic retain set. The two selected subsets are disjoint. Consequently, this experiment changes only the criterion used to rank candidates from the same synthetic pool.

Tables~\ref{tab:uncertainty_per_class_resnet18},
\ref{tab:uncertainty_per_class_vit_b_16}, and
\ref{tab:uncertainty_per_class_swin_t} report the resulting $\mathrm{RS}$ for every CIFAR-10 forget class using ResNet-18, ViT-B/16, and Swin-T, respectively. Each entry is the mean $\pm$ standard deviation over three independent audit seeds. Across seeds, we resample the Gaussian candidate pools, repeat probe selection, and reinitialize the classifier-head optimization, while keeping the unlearned checkpoint and all audit hyperparameters fixed. Thus, the reported variation measures the stochasticity of the source-free audit rather than variation across independently trained unlearned checkpoints.

The results show that our proposed SFRA is not tied to Softmax confidence. Entropy- and energy-based selection also produce substantial $\mathrm{RS}$ for numerous combinations of unlearning method, forget class, and architecture. However, no uncertainty score uniformly dominates. On ResNet-18, Softmax obtains the highest average $\mathrm{RS}$ across the evaluated method--class pairs, although energy produces the highest $\mathrm{RS}$ for many individual pairs. On ViT-B/16, Softmax and entropy obtain
similar average performance, whereas energy is less effective on average but remains competitive for selected checkpoints. On Swin-T, Softmax again achieves the highest average $\mathrm{RS}$, while entropy and energy outperform it for some method--class combinations. The preferred uncertainty score is therefore both checkpoint- and class-dependent.

The seed-level results further show that most measurements are stable under the stochastic components of the audit. For ResNet-18, $88.2\%$ of the reported entries have an $\mathrm{RS}$ standard deviation no greater than $0.02$, and $99.7\%$ have a standard deviation no greater than $0.05$. The corresponding fractions within $0.05$ are $85.7\%$ for ViT-B/16 and $95.9\%$ for Swin-T. Nevertheless, a small number of method--class--score combinations exhibit substantially greater variability, particularly for ViT-B/16. This indicates that probe sampling and classifier-head optimization can affect the measured recoverability of certain checkpoints and motivates reporting results over multiple audit seeds. Overall, these experiments demonstrate that our SFRA is observed across several uncertainty definitions, architectures, forget classes, and random audit seeds. Softmax confidence remains a simple and generally strong default, but the results do not support treating any single uncertainty score as universally optimal.

\begin{table*}[p]
\centering
\caption{Uncertainty-score ablation on CIFAR-10 using a ResNet-18 backbone. Each class column corresponds to a separate unlearned checkpoint with the indicated forget class and reports $\mathrm{RS}$ as the mean $\pm$ standard deviation across three independent audit seeds, while Avg. gives the mean $\mathrm{RS}$ across all forget classes and seeds.}
\label{tab:uncertainty_per_class_resnet18}
\vspace{-0.15cm}
\setlength{\tabcolsep}{2pt}
\renewcommand{\arraystretch}{0.88}
\scriptsize
\resizebox{\textwidth}{!}{%
\begin{tabular}{l|l|cccccccccc|c}
\toprule
\multirow{2}{*}{Unlearning Method} & \multirow{2}{*}{Uncertainty} & \multicolumn{10}{c|}{Forget Class} & \multirow{2}{*}{Avg.} \\
 & & 0 & 1 & 2 & 3 & 4 & 5 & 6 & 7 & 8 & 9 & \\
\midrule
\multirow{3}{*}{Retrained} & Softmax & $0.54${\tiny $\,\pm\,0.00$} & $0.19${\tiny $\,\pm\,0.01$} & $0.64${\tiny $\,\pm\,0.01$} & $0.57${\tiny $\,\pm\,0.01$} & $0.68${\tiny $\,\pm\,0.01$} & $0.32${\tiny $\,\pm\,0.01$} & $0.55${\tiny $\,\pm\,0.01$} & $0.52${\tiny $\,\pm\,0.00$} & $0.44${\tiny $\,\pm\,0.01$} & $0.35${\tiny $\,\pm\,0.01$} & $0.48$ \\
 & Entropy & $0.52${\tiny $\,\pm\,0.01$} & $0.18${\tiny $\,\pm\,0.01$} & $0.61${\tiny $\,\pm\,0.00$} & $0.57${\tiny $\,\pm\,0.01$} & $0.65${\tiny $\,\pm\,0.02$} & $0.29${\tiny $\,\pm\,0.03$} & $0.52${\tiny $\,\pm\,0.01$} & $0.49${\tiny $\,\pm\,0.01$} & $0.40${\tiny $\,\pm\,0.02$} & $0.33${\tiny $\,\pm\,0.02$} & $0.46$ \\
 & Energy & \textbf{$0.59${\tiny $\,\pm\,0.01$}} & \textbf{$0.22${\tiny $\,\pm\,0.02$}} & \textbf{$0.68${\tiny $\,\pm\,0.00$}} & \textbf{$0.62${\tiny $\,\pm\,0.02$}} & \textbf{$0.74${\tiny $\,\pm\,0.00$}} & \textbf{$0.36${\tiny $\,\pm\,0.00$}} & \textbf{$0.60${\tiny $\,\pm\,0.02$}} & \textbf{$0.59${\tiny $\,\pm\,0.02$}} & \textbf{$0.46${\tiny $\,\pm\,0.04$}} & \textbf{$0.41${\tiny $\,\pm\,0.01$}} & \textbf{$0.53$} \\
\midrule
\multirow{3}{*}{Finetune \cite{golatkar2020eternal}} & Softmax & $0.59${\tiny $\,\pm\,0.01$} & $0.23${\tiny $\,\pm\,0.02$} & $0.60${\tiny $\,\pm\,0.02$} & $0.47${\tiny $\,\pm\,0.03$} & $0.56${\tiny $\,\pm\,0.02$} & $0.31${\tiny $\,\pm\,0.01$} & \textbf{$0.70${\tiny $\,\pm\,0.04$}} & $0.55${\tiny $\,\pm\,0.02$} & \textbf{$0.46${\tiny $\,\pm\,0.03$}} & \textbf{$0.28${\tiny $\,\pm\,0.02$}} & $0.48$ \\
 & Entropy & $0.60${\tiny $\,\pm\,0.01$} & $0.21${\tiny $\,\pm\,0.01$} & $0.59${\tiny $\,\pm\,0.03$} & $0.47${\tiny $\,\pm\,0.00$} & $0.57${\tiny $\,\pm\,0.01$} & $0.30${\tiny $\,\pm\,0.02$} & $0.65${\tiny $\,\pm\,0.01$} & $0.53${\tiny $\,\pm\,0.02$} & $0.44${\tiny $\,\pm\,0.02$} & $0.25${\tiny $\,\pm\,0.02$} & $0.46$ \\
 & Energy & \textbf{$0.61${\tiny $\,\pm\,0.01$}} & \textbf{$0.25${\tiny $\,\pm\,0.01$}} & \textbf{$0.64${\tiny $\,\pm\,0.02$}} & \textbf{$0.50${\tiny $\,\pm\,0.01$}} & \textbf{$0.59${\tiny $\,\pm\,0.02$}} & \textbf{$0.32${\tiny $\,\pm\,0.01$}} & $0.70${\tiny $\,\pm\,0.02$} & \textbf{$0.59${\tiny $\,\pm\,0.02$}} & $0.45${\tiny $\,\pm\,0.03$} & $0.25${\tiny $\,\pm\,0.04$} & \textbf{$0.49$} \\
\midrule
\multirow{3}{*}{Negative Gradient \cite{golatkar2020eternal}} & Softmax & $0.62${\tiny $\,\pm\,0.01$} & $0.74${\tiny $\,\pm\,0.01$} & \textbf{$0.57${\tiny $\,\pm\,0.01$}} & $0.59${\tiny $\,\pm\,0.01$} & \textbf{$0.69${\tiny $\,\pm\,0.00$}} & \textbf{$0.66${\tiny $\,\pm\,0.02$}} & $0.69${\tiny $\,\pm\,0.01$} & \textbf{$0.72${\tiny $\,\pm\,0.02$}} & $0.66${\tiny $\,\pm\,0.01$} & \textbf{$0.66${\tiny $\,\pm\,0.01$}} & $0.66$ \\
 & Entropy & $0.60${\tiny $\,\pm\,0.01$} & $0.73${\tiny $\,\pm\,0.01$} & $0.54${\tiny $\,\pm\,0.01$} & $0.59${\tiny $\,\pm\,0.02$} & $0.65${\tiny $\,\pm\,0.01$} & $0.64${\tiny $\,\pm\,0.03$} & $0.68${\tiny $\,\pm\,0.01$} & $0.69${\tiny $\,\pm\,0.01$} & $0.65${\tiny $\,\pm\,0.01$} & $0.64${\tiny $\,\pm\,0.01$} & $0.64$ \\
 & Energy & \textbf{$0.62${\tiny $\,\pm\,0.00$}} & \textbf{$0.76${\tiny $\,\pm\,0.01$}} & $0.56${\tiny $\,\pm\,0.01$} & \textbf{$0.62${\tiny $\,\pm\,0.01$}} & $0.67${\tiny $\,\pm\,0.01$} & $0.65${\tiny $\,\pm\,0.02$} & \textbf{$0.70${\tiny $\,\pm\,0.01$}} & $0.71${\tiny $\,\pm\,0.01$} & \textbf{$0.67${\tiny $\,\pm\,0.01$}} & $0.66${\tiny $\,\pm\,0.01$} & \textbf{$0.66$} \\
\midrule
\multirow{3}{*}{Negative Gradient+ \cite{kurmanji2023towards}} & Softmax & \textbf{$0.30${\tiny $\,\pm\,0.01$}} & \textbf{$0.32${\tiny $\,\pm\,0.01$}} & \textbf{$0.23${\tiny $\,\pm\,0.01$}} & \textbf{$0.46${\tiny $\,\pm\,0.01$}} & \textbf{$0.47${\tiny $\,\pm\,0.02$}} & \textbf{$0.36${\tiny $\,\pm\,0.05$}} & \textbf{$0.50${\tiny $\,\pm\,0.01$}} & \textbf{$0.16${\tiny $\,\pm\,0.02$}} & \textbf{$0.39${\tiny $\,\pm\,0.01$}} & \textbf{$0.47${\tiny $\,\pm\,0.01$}} & \textbf{$0.37$} \\
 & Entropy & $0.27${\tiny $\,\pm\,0.01$} & $0.28${\tiny $\,\pm\,0.01$} & $0.15${\tiny $\,\pm\,0.01$} & $0.42${\tiny $\,\pm\,0.01$} & $0.44${\tiny $\,\pm\,0.00$} & $0.09${\tiny $\,\pm\,0.01$} & $0.48${\tiny $\,\pm\,0.01$} & $0.08${\tiny $\,\pm\,0.01$} & $0.37${\tiny $\,\pm\,0.01$} & $0.43${\tiny $\,\pm\,0.02$} & $0.30$ \\
 & Energy & $0.28${\tiny $\,\pm\,0.01$} & $0.27${\tiny $\,\pm\,0.01$} & $0.06${\tiny $\,\pm\,0.01$} & $0.40${\tiny $\,\pm\,0.01$} & $0.46${\tiny $\,\pm\,0.01$} & $0.00${\tiny $\,\pm\,0.00$} & $0.50${\tiny $\,\pm\,0.01$} & $0.02${\tiny $\,\pm\,0.00$} & $0.38${\tiny $\,\pm\,0.01$} & $0.44${\tiny $\,\pm\,0.01$} & $0.28$ \\
\midrule
\multirow{3}{*}{Random Label \cite{hayase2020selective}} & Softmax & $0.73${\tiny $\,\pm\,0.00$} & $0.79${\tiny $\,\pm\,0.01$} & $0.65${\tiny $\,\pm\,0.02$} & $0.69${\tiny $\,\pm\,0.02$} & $0.77${\tiny $\,\pm\,0.01$} & \textbf{$0.72${\tiny $\,\pm\,0.01$}} & $0.74${\tiny $\,\pm\,0.01$} & $0.76${\tiny $\,\pm\,0.01$} & $0.73${\tiny $\,\pm\,0.01$} & $0.77${\tiny $\,\pm\,0.01$} & $0.73$ \\
 & Entropy & $0.73${\tiny $\,\pm\,0.01$} & $0.78${\tiny $\,\pm\,0.01$} & $0.63${\tiny $\,\pm\,0.00$} & $0.69${\tiny $\,\pm\,0.02$} & $0.75${\tiny $\,\pm\,0.01$} & $0.71${\tiny $\,\pm\,0.01$} & $0.73${\tiny $\,\pm\,0.01$} & $0.75${\tiny $\,\pm\,0.00$} & $0.72${\tiny $\,\pm\,0.01$} & $0.75${\tiny $\,\pm\,0.01$} & $0.72$ \\
 & Energy & \textbf{$0.74${\tiny $\,\pm\,0.01$}} & \textbf{$0.80${\tiny $\,\pm\,0.01$}} & \textbf{$0.66${\tiny $\,\pm\,0.00$}} & \textbf{$0.71${\tiny $\,\pm\,0.01$}} & \textbf{$0.78${\tiny $\,\pm\,0.01$}} & $0.72${\tiny $\,\pm\,0.02$} & \textbf{$0.75${\tiny $\,\pm\,0.01$}} & \textbf{$0.78${\tiny $\,\pm\,0.00$}} & \textbf{$0.73${\tiny $\,\pm\,0.00$}} & \textbf{$0.77${\tiny $\,\pm\,0.01$}} & \textbf{$0.74$} \\
\midrule
\multirow{3}{*}{Boundary Shrink \cite{chen2023boundary}} & Softmax & $0.71${\tiny $\,\pm\,0.01$} & $0.79${\tiny $\,\pm\,0.01$} & $0.65${\tiny $\,\pm\,0.02$} & \textbf{$0.69${\tiny $\,\pm\,0.01$}} & $0.77${\tiny $\,\pm\,0.00$} & $0.71${\tiny $\,\pm\,0.01$} & $0.74${\tiny $\,\pm\,0.01$} & $0.76${\tiny $\,\pm\,0.02$} & $0.73${\tiny $\,\pm\,0.01$} & \textbf{$0.76${\tiny $\,\pm\,0.01$}} & $0.73$ \\
 & Entropy & $0.69${\tiny $\,\pm\,0.02$} & $0.78${\tiny $\,\pm\,0.00$} & $0.63${\tiny $\,\pm\,0.01$} & $0.68${\tiny $\,\pm\,0.01$} & $0.76${\tiny $\,\pm\,0.01$} & $0.70${\tiny $\,\pm\,0.01$} & $0.73${\tiny $\,\pm\,0.01$} & $0.76${\tiny $\,\pm\,0.02$} & $0.72${\tiny $\,\pm\,0.01$} & $0.75${\tiny $\,\pm\,0.01$} & $0.72$ \\
 & Energy & \textbf{$0.72${\tiny $\,\pm\,0.00$}} & \textbf{$0.80${\tiny $\,\pm\,0.01$}} & \textbf{$0.65${\tiny $\,\pm\,0.01$}} & $0.69${\tiny $\,\pm\,0.01$} & \textbf{$0.77${\tiny $\,\pm\,0.01$}} & \textbf{$0.72${\tiny $\,\pm\,0.01$}} & \textbf{$0.75${\tiny $\,\pm\,0.00$}} & \textbf{$0.77${\tiny $\,\pm\,0.01$}} & \textbf{$0.74${\tiny $\,\pm\,0.01$}} & $0.76${\tiny $\,\pm\,0.01$} & \textbf{$0.74$} \\
\midrule
\multirow{3}{*}{Learn to Unlearn \cite{cha2024learning}} & Softmax & \textbf{$0.69${\tiny $\,\pm\,0.01$}} & $0.68${\tiny $\,\pm\,0.01$} & \textbf{$0.58${\tiny $\,\pm\,0.01$}} & $0.56${\tiny $\,\pm\,0.02$} & $0.69${\tiny $\,\pm\,0.01$} & $0.70${\tiny $\,\pm\,0.01$} & $0.69${\tiny $\,\pm\,0.01$} & \textbf{$0.72${\tiny $\,\pm\,0.01$}} & \textbf{$0.69${\tiny $\,\pm\,0.01$}} & $0.68${\tiny $\,\pm\,0.01$} & $0.67$ \\
 & Entropy & $0.67${\tiny $\,\pm\,0.00$} & $0.67${\tiny $\,\pm\,0.01$} & $0.57${\tiny $\,\pm\,0.03$} & $0.55${\tiny $\,\pm\,0.01$} & $0.68${\tiny $\,\pm\,0.02$} & $0.68${\tiny $\,\pm\,0.01$} & $0.67${\tiny $\,\pm\,0.01$} & $0.71${\tiny $\,\pm\,0.00$} & $0.66${\tiny $\,\pm\,0.01$} & $0.67${\tiny $\,\pm\,0.01$} & $0.65$ \\
 & Energy & $0.68${\tiny $\,\pm\,0.00$} & \textbf{$0.70${\tiny $\,\pm\,0.00$}} & $0.58${\tiny $\,\pm\,0.00$} & \textbf{$0.58${\tiny $\,\pm\,0.01$}} & \textbf{$0.70${\tiny $\,\pm\,0.00$}} & \textbf{$0.70${\tiny $\,\pm\,0.02$}} & \textbf{$0.69${\tiny $\,\pm\,0.01$}} & $0.72${\tiny $\,\pm\,0.01$} & $0.69${\tiny $\,\pm\,0.01$} & \textbf{$0.68${\tiny $\,\pm\,0.01$}} & \textbf{$0.67$} \\
\midrule
\multirow{3}{*}{SCRUB \cite{kurmanji2023towards}} & Softmax & \textbf{$0.63${\tiny $\,\pm\,0.01$}} & \textbf{$0.42${\tiny $\,\pm\,0.01$}} & \textbf{$0.71${\tiny $\,\pm\,0.00$}} & \textbf{$0.53${\tiny $\,\pm\,0.02$}} & \textbf{$0.63${\tiny $\,\pm\,0.04$}} & \textbf{$0.52${\tiny $\,\pm\,0.04$}} & \textbf{$0.63${\tiny $\,\pm\,0.02$}} & \textbf{$0.51${\tiny $\,\pm\,0.01$}} & \textbf{$0.42${\tiny $\,\pm\,0.02$}} & \textbf{$0.41${\tiny $\,\pm\,0.00$}} & \textbf{$0.54$} \\
 & Entropy & $0.53${\tiny $\,\pm\,0.02$} & $0.17${\tiny $\,\pm\,0.04$} & $0.30${\tiny $\,\pm\,0.08$} & $0.37${\tiny $\,\pm\,0.01$} & $0.44${\tiny $\,\pm\,0.01$} & $0.10${\tiny $\,\pm\,0.04$} & $0.47${\tiny $\,\pm\,0.03$} & $0.15${\tiny $\,\pm\,0.02$} & $0.31${\tiny $\,\pm\,0.02$} & $0.30${\tiny $\,\pm\,0.01$} & $0.31$ \\
 & Energy & $0.28${\tiny $\,\pm\,0.03$} & $0.00${\tiny $\,\pm\,0.00$} & $0.00${\tiny $\,\pm\,0.00$} & $0.10${\tiny $\,\pm\,0.02$} & $0.15${\tiny $\,\pm\,0.02$} & $0.00${\tiny $\,\pm\,0.00$} & $0.22${\tiny $\,\pm\,0.01$} & $0.00${\tiny $\,\pm\,0.00$} & $0.09${\tiny $\,\pm\,0.01$} & $0.10${\tiny $\,\pm\,0.01$} & $0.09$ \\
\midrule
\multirow{3}{*}{Bad Teacher \cite{chundawat2023can}} & Softmax & $0.98${\tiny $\,\pm\,0.00$} & \textbf{$0.99${\tiny $\,\pm\,0.00$}} & \textbf{$0.80${\tiny $\,\pm\,0.04$}} & $0.96${\tiny $\,\pm\,0.00$} & \textbf{$0.94${\tiny $\,\pm\,0.00$}} & $0.97${\tiny $\,\pm\,0.00$} & $0.98${\tiny $\,\pm\,0.00$} & \textbf{$0.99${\tiny $\,\pm\,0.00$}} & $0.99${\tiny $\,\pm\,0.00$} & $0.98${\tiny $\,\pm\,0.00$} & \textbf{$0.96$} \\
 & Entropy & $0.98${\tiny $\,\pm\,0.00$} & $0.99${\tiny $\,\pm\,0.00$} & $0.68${\tiny $\,\pm\,0.01$} & \textbf{$0.96${\tiny $\,\pm\,0.00$}} & $0.93${\tiny $\,\pm\,0.00$} & $0.97${\tiny $\,\pm\,0.00$} & \textbf{$0.98${\tiny $\,\pm\,0.00$}} & $0.99${\tiny $\,\pm\,0.00$} & \textbf{$0.99${\tiny $\,\pm\,0.00$}} & \textbf{$0.98${\tiny $\,\pm\,0.00$}} & $0.94$ \\
 & Energy & \textbf{$0.98${\tiny $\,\pm\,0.00$}} & $0.99${\tiny $\,\pm\,0.00$} & $0.70${\tiny $\,\pm\,0.00$} & $0.96${\tiny $\,\pm\,0.00$} & $0.91${\tiny $\,\pm\,0.00$} & \textbf{$0.97${\tiny $\,\pm\,0.00$}} & $0.98${\tiny $\,\pm\,0.00$} & $0.98${\tiny $\,\pm\,0.00$} & $0.99${\tiny $\,\pm\,0.00$} & $0.98${\tiny $\,\pm\,0.00$} & $0.94$ \\
\midrule
\multirow{3}{*}{SalUn \cite{fan2023salun}} & Softmax & \textbf{$0.85${\tiny $\,\pm\,0.01$}} & \textbf{$0.82${\tiny $\,\pm\,0.01$}} & \textbf{$0.86${\tiny $\,\pm\,0.00$}} & $0.82${\tiny $\,\pm\,0.00$} & $0.86${\tiny $\,\pm\,0.00$} & $0.85${\tiny $\,\pm\,0.01$} & \textbf{$0.88${\tiny $\,\pm\,0.00$}} & $0.88${\tiny $\,\pm\,0.00$} & \textbf{$0.83${\tiny $\,\pm\,0.00$}} & $0.87${\tiny $\,\pm\,0.00$} & \textbf{$0.85$} \\
 & Entropy & $0.84${\tiny $\,\pm\,0.01$} & $0.82${\tiny $\,\pm\,0.01$} & $0.85${\tiny $\,\pm\,0.01$} & $0.82${\tiny $\,\pm\,0.01$} & $0.86${\tiny $\,\pm\,0.00$} & $0.85${\tiny $\,\pm\,0.00$} & $0.88${\tiny $\,\pm\,0.00$} & $0.88${\tiny $\,\pm\,0.00$} & $0.83${\tiny $\,\pm\,0.00$} & \textbf{$0.87${\tiny $\,\pm\,0.00$}} & $0.85$ \\
 & Energy & $0.79${\tiny $\,\pm\,0.01$} & $0.77${\tiny $\,\pm\,0.02$} & $0.85${\tiny $\,\pm\,0.00$} & \textbf{$0.82${\tiny $\,\pm\,0.00$}} & \textbf{$0.86${\tiny $\,\pm\,0.00$}} & \textbf{$0.86${\tiny $\,\pm\,0.00$}} & $0.88${\tiny $\,\pm\,0.00$} & \textbf{$0.89${\tiny $\,\pm\,0.00$}} & $0.80${\tiny $\,\pm\,0.00$} & $0.86${\tiny $\,\pm\,0.01$} & $0.84$ \\
\midrule
\multirow{3}{*}{DELETE \cite{zhou2025decoupled}} & Softmax & $0.96${\tiny $\,\pm\,0.00$} & $0.97${\tiny $\,\pm\,0.00$} & $0.95${\tiny $\,\pm\,0.00$} & $0.93${\tiny $\,\pm\,0.00$} & $0.96${\tiny $\,\pm\,0.00$} & $0.93${\tiny $\,\pm\,0.00$} & $0.97${\tiny $\,\pm\,0.00$} & $0.96${\tiny $\,\pm\,0.00$} & $0.96${\tiny $\,\pm\,0.00$} & \textbf{$0.97${\tiny $\,\pm\,0.00$}} & $0.96$ \\
 & Entropy & $0.96${\tiny $\,\pm\,0.00$} & \textbf{$0.97${\tiny $\,\pm\,0.00$}} & $0.95${\tiny $\,\pm\,0.00$} & $0.93${\tiny $\,\pm\,0.00$} & $0.97${\tiny $\,\pm\,0.00$} & $0.93${\tiny $\,\pm\,0.00$} & \textbf{$0.97${\tiny $\,\pm\,0.00$}} & \textbf{$0.96${\tiny $\,\pm\,0.00$}} & \textbf{$0.97${\tiny $\,\pm\,0.00$}} & $0.96${\tiny $\,\pm\,0.00$} & $0.96$ \\
 & Energy & \textbf{$0.96${\tiny $\,\pm\,0.00$}} & $0.97${\tiny $\,\pm\,0.00$} & \textbf{$0.95${\tiny $\,\pm\,0.00$}} & \textbf{$0.93${\tiny $\,\pm\,0.00$}} & \textbf{$0.97${\tiny $\,\pm\,0.00$}} & \textbf{$0.93${\tiny $\,\pm\,0.00$}} & $0.97${\tiny $\,\pm\,0.00$} & $0.96${\tiny $\,\pm\,0.00$} & $0.96${\tiny $\,\pm\,0.00$} & $0.97${\tiny $\,\pm\,0.00$} & \textbf{$0.96$} \\
\bottomrule
\end{tabular}%
}
\end{table*}

\begin{table*}[p]
\centering
\caption{Uncertainty-score ablation on CIFAR-10 using a ViT-B/16 backbone. Each class column corresponds to a separate unlearned checkpoint with the indicated forget class and reports $\mathrm{RS}$ as the mean $\pm$ standard deviation across three independent audit seeds, while Avg. gives the mean $\mathrm{RS}$ across all forget classes and seeds.}
\label{tab:uncertainty_per_class_vit_b_16}
\vspace{-0.15cm}
\setlength{\tabcolsep}{2pt}
\renewcommand{\arraystretch}{0.88}
\scriptsize
\resizebox{\textwidth}{!}{%
\begin{tabular}{l|l|cccccccccc|c}
\toprule
\multirow{2}{*}{Unlearning Method} & \multirow{2}{*}{Uncertainty} & \multicolumn{10}{c|}{Forget Class} & \multirow{2}{*}{Avg.} \\
 & & 0 & 1 & 2 & 3 & 4 & 5 & 6 & 7 & 8 & 9 & \\
\midrule
\multirow{3}{*}{Retrained} & Softmax & \textbf{$0.12${\tiny $\,\pm\,0.01$}} & \textbf{$0.10${\tiny $\,\pm\,0.03$}} & \textbf{$0.56${\tiny $\,\pm\,0.01$}} & \textbf{$0.08${\tiny $\,\pm\,0.00$}} & $0.13${\tiny $\,\pm\,0.01$} & \textbf{$0.35${\tiny $\,\pm\,0.00$}} & \textbf{$0.35${\tiny $\,\pm\,0.00$}} & \textbf{$0.02${\tiny $\,\pm\,0.01$}} & \textbf{$0.28${\tiny $\,\pm\,0.01$}} & \textbf{$0.06${\tiny $\,\pm\,0.00$}} & \textbf{$0.20$} \\
 & Entropy & $0.08${\tiny $\,\pm\,0.01$} & $0.08${\tiny $\,\pm\,0.00$} & $0.55${\tiny $\,\pm\,0.01$} & $0.07${\tiny $\,\pm\,0.00$} & \textbf{$0.14${\tiny $\,\pm\,0.01$}} & $0.34${\tiny $\,\pm\,0.01$} & $0.31${\tiny $\,\pm\,0.00$} & $0.01${\tiny $\,\pm\,0.00$} & $0.25${\tiny $\,\pm\,0.01$} & $0.04${\tiny $\,\pm\,0.01$} & $0.19$ \\
 & Energy & $0.01${\tiny $\,\pm\,0.00$} & $0.00${\tiny $\,\pm\,0.00$} & $0.16${\tiny $\,\pm\,0.08$} & $0.00${\tiny $\,\pm\,0.00$} & $0.00${\tiny $\,\pm\,0.00$} & $0.05${\tiny $\,\pm\,0.02$} & $0.01${\tiny $\,\pm\,0.01$} & $0.00${\tiny $\,\pm\,0.00$} & $0.05${\tiny $\,\pm\,0.02$} & $0.00${\tiny $\,\pm\,0.00$} & $0.03$ \\
\midrule
\multirow{3}{*}{Finetune \cite{golatkar2020eternal}} & Softmax & $0.06${\tiny $\,\pm\,0.02$} & \textbf{$0.08${\tiny $\,\pm\,0.04$}} & \textbf{$0.22${\tiny $\,\pm\,0.01$}} & \textbf{$0.08${\tiny $\,\pm\,0.01$}} & $0.00${\tiny $\,\pm\,0.00$} & \textbf{$0.07${\tiny $\,\pm\,0.03$}} & $0.15${\tiny $\,\pm\,0.01$} & $0.07${\tiny $\,\pm\,0.01$} & $0.07${\tiny $\,\pm\,0.00$} & $0.06${\tiny $\,\pm\,0.00$} & $0.09$ \\
 & Entropy & \textbf{$0.06${\tiny $\,\pm\,0.00$}} & $0.02${\tiny $\,\pm\,0.02$} & $0.22${\tiny $\,\pm\,0.03$} & $0.07${\tiny $\,\pm\,0.00$} & \textbf{$0.01${\tiny $\,\pm\,0.00$}} & $0.04${\tiny $\,\pm\,0.02$} & \textbf{$0.16${\tiny $\,\pm\,0.01$}} & \textbf{$0.13${\tiny $\,\pm\,0.02$}} & \textbf{$0.09${\tiny $\,\pm\,0.03$}} & \textbf{$0.06${\tiny $\,\pm\,0.01$}} & \textbf{$0.09$} \\
 & Energy & $0.00${\tiny $\,\pm\,0.00$} & $0.00${\tiny $\,\pm\,0.00$} & $0.08${\tiny $\,\pm\,0.03$} & $0.01${\tiny $\,\pm\,0.00$} & $0.00${\tiny $\,\pm\,0.00$} & $0.00${\tiny $\,\pm\,0.00$} & $0.00${\tiny $\,\pm\,0.00$} & $0.00${\tiny $\,\pm\,0.00$} & $0.00${\tiny $\,\pm\,0.00$} & $0.00${\tiny $\,\pm\,0.00$} & $0.01$ \\
\midrule
\multirow{3}{*}{Negative Gradient \cite{golatkar2020eternal}} & Softmax & \textbf{$0.62${\tiny $\,\pm\,0.02$}} & \textbf{$0.99${\tiny $\,\pm\,0.00$}} & \textbf{$0.95${\tiny $\,\pm\,0.00$}} & $0.63${\tiny $\,\pm\,0.02$} & \textbf{$0.12${\tiny $\,\pm\,0.07$}} & \textbf{$0.81${\tiny $\,\pm\,0.01$}} & $0.33${\tiny $\,\pm\,0.08$} & \textbf{$0.87${\tiny $\,\pm\,0.02$}} & $0.99${\tiny $\,\pm\,0.00$} & $0.99${\tiny $\,\pm\,0.00$} & $0.73$ \\
 & Entropy & $0.61${\tiny $\,\pm\,0.01$} & $0.98${\tiny $\,\pm\,0.00$} & $0.95${\tiny $\,\pm\,0.01$} & \textbf{$0.69${\tiny $\,\pm\,0.06$}} & $0.08${\tiny $\,\pm\,0.02$} & $0.79${\tiny $\,\pm\,0.02$} & \textbf{$0.61${\tiny $\,\pm\,0.13$}} & $0.85${\tiny $\,\pm\,0.00$} & \textbf{$0.99${\tiny $\,\pm\,0.00$}} & \textbf{$0.99${\tiny $\,\pm\,0.00$}} & \textbf{$0.75$} \\
 & Energy & $0.00${\tiny $\,\pm\,0.00$} & $0.98${\tiny $\,\pm\,0.01$} & $0.01${\tiny $\,\pm\,0.02$} & $0.15${\tiny $\,\pm\,0.11$} & $0.04${\tiny $\,\pm\,0.06$} & $0.00${\tiny $\,\pm\,0.00$} & $0.02${\tiny $\,\pm\,0.03$} & $0.00${\tiny $\,\pm\,0.00$} & $0.22${\tiny $\,\pm\,0.29$} & $0.20${\tiny $\,\pm\,0.03$} & $0.16$ \\
\midrule
\multirow{3}{*}{Negative Gradient+ \cite{kurmanji2023towards}} & Softmax & $0.06${\tiny $\,\pm\,0.08$} & $0.11${\tiny $\,\pm\,0.07$} & \textbf{$0.00${\tiny $\,\pm\,0.00$}} & $0.40${\tiny $\,\pm\,0.16$} & \textbf{$0.00${\tiny $\,\pm\,0.00$}} & $0.00${\tiny $\,\pm\,0.00$} & $0.39${\tiny $\,\pm\,0.21$} & $0.04${\tiny $\,\pm\,0.01$} & $0.77${\tiny $\,\pm\,0.27$} & $0.03${\tiny $\,\pm\,0.01$} & $0.18$ \\
 & Entropy & $0.05${\tiny $\,\pm\,0.01$} & $0.12${\tiny $\,\pm\,0.01$} & \textbf{$0.00${\tiny $\,\pm\,0.00$}} & \textbf{$0.87${\tiny $\,\pm\,0.03$}} & \textbf{$0.00${\tiny $\,\pm\,0.00$}} & $0.00${\tiny $\,\pm\,0.00$} & \textbf{$0.64${\tiny $\,\pm\,0.38$}} & \textbf{$0.13${\tiny $\,\pm\,0.06$}} & \textbf{$0.90${\tiny $\,\pm\,0.07$}} & $0.04${\tiny $\,\pm\,0.02$} & $0.28$ \\
 & Energy & \textbf{$0.83${\tiny $\,\pm\,0.16$}} & \textbf{$0.91${\tiny $\,\pm\,0.10$}} & \textbf{$0.00${\tiny $\,\pm\,0.00$}} & $0.83${\tiny $\,\pm\,0.04$} & \textbf{$0.00${\tiny $\,\pm\,0.00$}} & \textbf{$0.57${\tiny $\,\pm\,0.37$}} & $0.33${\tiny $\,\pm\,0.34$} & $0.01${\tiny $\,\pm\,0.00$} & $0.28${\tiny $\,\pm\,0.16$} & \textbf{$0.05${\tiny $\,\pm\,0.02$}} & \textbf{$0.38$} \\
\midrule
\multirow{3}{*}{Random Label \cite{hayase2020selective}} & Softmax & $0.99${\tiny $\,\pm\,0.00$} & \textbf{$0.99${\tiny $\,\pm\,0.00$}} & \textbf{$0.98${\tiny $\,\pm\,0.00$}} & $0.98${\tiny $\,\pm\,0.00$} & $0.99${\tiny $\,\pm\,0.00$} & \textbf{$0.97${\tiny $\,\pm\,0.00$}} & $0.99${\tiny $\,\pm\,0.00$} & \textbf{$1.00${\tiny $\,\pm\,0.00$}} & \textbf{$0.99${\tiny $\,\pm\,0.00$}} & $0.99${\tiny $\,\pm\,0.00$} & \textbf{$0.99$} \\
 & Entropy & \textbf{$0.99${\tiny $\,\pm\,0.00$}} & $0.99${\tiny $\,\pm\,0.00$} & $0.97${\tiny $\,\pm\,0.00$} & \textbf{$0.98${\tiny $\,\pm\,0.00$}} & \textbf{$0.99${\tiny $\,\pm\,0.00$}} & $0.97${\tiny $\,\pm\,0.00$} & \textbf{$0.99${\tiny $\,\pm\,0.00$}} & $1.00${\tiny $\,\pm\,0.00$} & $0.99${\tiny $\,\pm\,0.00$} & \textbf{$0.99${\tiny $\,\pm\,0.00$}} & $0.99$ \\
 & Energy & $0.96${\tiny $\,\pm\,0.00$} & $0.99${\tiny $\,\pm\,0.00$} & $0.96${\tiny $\,\pm\,0.02$} & $0.97${\tiny $\,\pm\,0.01$} & $0.91${\tiny $\,\pm\,0.06$} & $0.94${\tiny $\,\pm\,0.01$} & $0.37${\tiny $\,\pm\,0.44$} & $0.99${\tiny $\,\pm\,0.00$} & $0.90${\tiny $\,\pm\,0.01$} & $0.97${\tiny $\,\pm\,0.01$} & $0.89$ \\
\midrule
\multirow{3}{*}{Learn to Unlearn \cite{cha2024learning}} & Softmax & \textbf{$0.01${\tiny $\,\pm\,0.00$}} & $0.97${\tiny $\,\pm\,0.00$} & $0.63${\tiny $\,\pm\,0.54$} & $0.06${\tiny $\,\pm\,0.00$} & $0.02${\tiny $\,\pm\,0.01$} & \textbf{$0.01${\tiny $\,\pm\,0.00$}} & \textbf{$0.03${\tiny $\,\pm\,0.02$}} & $0.03${\tiny $\,\pm\,0.01$} & $0.97${\tiny $\,\pm\,0.02$} & \textbf{$0.76${\tiny $\,\pm\,0.03$}} & $0.35$ \\
 & Entropy & $0.01${\tiny $\,\pm\,0.00$} & \textbf{$0.97${\tiny $\,\pm\,0.00$}} & \textbf{$0.64${\tiny $\,\pm\,0.55$}} & \textbf{$0.06${\tiny $\,\pm\,0.00$}} & \textbf{$0.07${\tiny $\,\pm\,0.03$}} & $0.01${\tiny $\,\pm\,0.00$} & $0.02${\tiny $\,\pm\,0.00$} & $0.06${\tiny $\,\pm\,0.04$} & \textbf{$0.99${\tiny $\,\pm\,0.01$}} & $0.75${\tiny $\,\pm\,0.02$} & \textbf{$0.36$} \\
 & Energy & $0.00${\tiny $\,\pm\,0.00$} & $0.97${\tiny $\,\pm\,0.00$} & $0.01${\tiny $\,\pm\,0.01$} & $0.01${\tiny $\,\pm\,0.00$} & $0.02${\tiny $\,\pm\,0.02$} & $0.00${\tiny $\,\pm\,0.00$} & $0.00${\tiny $\,\pm\,0.00$} & \textbf{$0.21${\tiny $\,\pm\,0.10$}} & $0.41${\tiny $\,\pm\,0.39$} & $0.00${\tiny $\,\pm\,0.00$} & $0.16$ \\
\midrule
\multirow{3}{*}{SCRUB \cite{kurmanji2023towards}} & Softmax & \textbf{$0.00${\tiny $\,\pm\,0.00$}} & \textbf{$0.20${\tiny $\,\pm\,0.02$}} & $0.29${\tiny $\,\pm\,0.01$} & \textbf{$0.00${\tiny $\,\pm\,0.00$}} & $0.04${\tiny $\,\pm\,0.01$} & \textbf{$0.05${\tiny $\,\pm\,0.01$}} & \textbf{$0.01${\tiny $\,\pm\,0.00$}} & \textbf{$0.65${\tiny $\,\pm\,0.04$}} & \textbf{$0.00${\tiny $\,\pm\,0.00$}} & \textbf{$0.10${\tiny $\,\pm\,0.04$}} & \textbf{$0.13$} \\
 & Entropy & \textbf{$0.00${\tiny $\,\pm\,0.00$}} & $0.15${\tiny $\,\pm\,0.02$} & \textbf{$0.31${\tiny $\,\pm\,0.07$}} & \textbf{$0.00${\tiny $\,\pm\,0.00$}} & \textbf{$0.05${\tiny $\,\pm\,0.00$}} & $0.05${\tiny $\,\pm\,0.01$} & $0.01${\tiny $\,\pm\,0.00$} & $0.63${\tiny $\,\pm\,0.01$} & \textbf{$0.00${\tiny $\,\pm\,0.00$}} & $0.09${\tiny $\,\pm\,0.03$} & $0.13$ \\
 & Energy & \textbf{$0.00${\tiny $\,\pm\,0.00$}} & $0.00${\tiny $\,\pm\,0.00$} & $0.00${\tiny $\,\pm\,0.00$} & \textbf{$0.00${\tiny $\,\pm\,0.00$}} & $0.00${\tiny $\,\pm\,0.00$} & $0.00${\tiny $\,\pm\,0.00$} & $0.00${\tiny $\,\pm\,0.00$} & $0.00${\tiny $\,\pm\,0.00$} & \textbf{$0.00${\tiny $\,\pm\,0.00$}} & $0.01${\tiny $\,\pm\,0.01$} & $0.00$ \\
\midrule
\multirow{3}{*}{Bad Teacher \cite{chundawat2023can}} & Softmax & $0.94${\tiny $\,\pm\,0.00$} & $0.98${\tiny $\,\pm\,0.00$} & $0.92${\tiny $\,\pm\,0.00$} & \textbf{$0.95${\tiny $\,\pm\,0.00$}} & $0.88${\tiny $\,\pm\,0.00$} & $0.97${\tiny $\,\pm\,0.00$} & \textbf{$0.98${\tiny $\,\pm\,0.00$}} & \textbf{$0.99${\tiny $\,\pm\,0.00$}} & $0.91${\tiny $\,\pm\,0.00$} & $0.98${\tiny $\,\pm\,0.00$} & $0.95$ \\
 & Entropy & \textbf{$0.94${\tiny $\,\pm\,0.00$}} & \textbf{$0.98${\tiny $\,\pm\,0.00$}} & \textbf{$0.92${\tiny $\,\pm\,0.00$}} & $0.95${\tiny $\,\pm\,0.00$} & \textbf{$0.89${\tiny $\,\pm\,0.00$}} & \textbf{$0.97${\tiny $\,\pm\,0.00$}} & $0.98${\tiny $\,\pm\,0.00$} & $0.99${\tiny $\,\pm\,0.00$} & \textbf{$0.92${\tiny $\,\pm\,0.00$}} & \textbf{$0.99${\tiny $\,\pm\,0.00$}} & \textbf{$0.95$} \\
 & Energy & $0.41${\tiny $\,\pm\,0.08$} & $0.76${\tiny $\,\pm\,0.08$} & $0.79${\tiny $\,\pm\,0.01$} & $0.68${\tiny $\,\pm\,0.05$} & $0.46${\tiny $\,\pm\,0.10$} & $0.67${\tiny $\,\pm\,0.05$} & $0.90${\tiny $\,\pm\,0.02$} & $0.33${\tiny $\,\pm\,0.04$} & $0.74${\tiny $\,\pm\,0.00$} & $0.97${\tiny $\,\pm\,0.00$} & $0.67$ \\
\midrule
\multirow{3}{*}{SalUn \cite{fan2023salun}} & Softmax & $0.71${\tiny $\,\pm\,0.04$} & \textbf{$0.99${\tiny $\,\pm\,0.00$}} & $0.97${\tiny $\,\pm\,0.00$} & $0.92${\tiny $\,\pm\,0.00$} & \textbf{$0.92${\tiny $\,\pm\,0.00$}} & \textbf{$0.80${\tiny $\,\pm\,0.03$}} & $0.92${\tiny $\,\pm\,0.00$} & $0.01${\tiny $\,\pm\,0.01$} & \textbf{$0.98${\tiny $\,\pm\,0.00$}} & $0.96${\tiny $\,\pm\,0.00$} & $0.82$ \\
 & Entropy & \textbf{$0.74${\tiny $\,\pm\,0.06$}} & $0.99${\tiny $\,\pm\,0.00$} & \textbf{$0.97${\tiny $\,\pm\,0.00$}} & \textbf{$0.92${\tiny $\,\pm\,0.00$}} & $0.92${\tiny $\,\pm\,0.01$} & $0.79${\tiny $\,\pm\,0.02$} & \textbf{$0.92${\tiny $\,\pm\,0.00$}} & \textbf{$0.02${\tiny $\,\pm\,0.01$}} & $0.98${\tiny $\,\pm\,0.00$} & \textbf{$0.96${\tiny $\,\pm\,0.00$}} & \textbf{$0.82$} \\
 & Energy & $0.00${\tiny $\,\pm\,0.00$} & $0.98${\tiny $\,\pm\,0.00$} & $0.91${\tiny $\,\pm\,0.01$} & $0.79${\tiny $\,\pm\,0.10$} & $0.61${\tiny $\,\pm\,0.36$} & $0.38${\tiny $\,\pm\,0.20$} & $0.01${\tiny $\,\pm\,0.01$} & $0.00${\tiny $\,\pm\,0.00$} & $0.03${\tiny $\,\pm\,0.02$} & $0.96${\tiny $\,\pm\,0.00$} & $0.47$ \\
\midrule
\multirow{3}{*}{DELETE \cite{zhou2025decoupled}} & Softmax & $0.54${\tiny $\,\pm\,0.15$} & $0.99${\tiny $\,\pm\,0.00$} & \textbf{$0.72${\tiny $\,\pm\,0.15$}} & $0.98${\tiny $\,\pm\,0.00$} & $0.97${\tiny $\,\pm\,0.00$} & $0.93${\tiny $\,\pm\,0.00$} & $0.98${\tiny $\,\pm\,0.00$} & \textbf{$0.99${\tiny $\,\pm\,0.00$}} & $0.99${\tiny $\,\pm\,0.00$} & $0.99${\tiny $\,\pm\,0.00$} & \textbf{$0.91$} \\
 & Entropy & $0.01${\tiny $\,\pm\,0.01$} & \textbf{$0.99${\tiny $\,\pm\,0.00$}} & $0.16${\tiny $\,\pm\,0.23$} & \textbf{$0.98${\tiny $\,\pm\,0.00$}} & \textbf{$0.97${\tiny $\,\pm\,0.00$}} & $0.93${\tiny $\,\pm\,0.01$} & \textbf{$0.98${\tiny $\,\pm\,0.00$}} & $0.99${\tiny $\,\pm\,0.00$} & \textbf{$0.99${\tiny $\,\pm\,0.00$}} & \textbf{$0.99${\tiny $\,\pm\,0.00$}} & $0.80$ \\
 & Energy & \textbf{$0.99${\tiny $\,\pm\,0.01$}} & $0.96${\tiny $\,\pm\,0.02$} & $0.28${\tiny $\,\pm\,0.39$} & $0.61${\tiny $\,\pm\,0.12$} & $0.40${\tiny $\,\pm\,0.21$} & \textbf{$0.94${\tiny $\,\pm\,0.00$}} & $0.16${\tiny $\,\pm\,0.01$} & $0.97${\tiny $\,\pm\,0.01$} & $0.25${\tiny $\,\pm\,0.14$} & $0.97${\tiny $\,\pm\,0.03$} & $0.65$ \\
\bottomrule
\end{tabular}%
}
\end{table*}

\begin{table*}[p]
\centering
\caption{Uncertainty-score ablation on CIFAR-10 using a Swin-T backbone. Each class column corresponds to a separate unlearned checkpoint with the indicated forget class and reports $\mathrm{RS}$ as the mean $\pm$ standard deviation across three independent audit seeds, while Avg. gives the mean $\mathrm{RS}$ across all forget classes and seeds.}
\label{tab:uncertainty_per_class_swin_t}
\vspace{-0.15cm}
\setlength{\tabcolsep}{2pt}
\renewcommand{\arraystretch}{0.88}
\scriptsize
\resizebox{\textwidth}{!}{%
\begin{tabular}{l|l|cccccccccc|c}
\toprule
\multirow{2}{*}{Unlearning Method} & \multirow{2}{*}{Uncertainty} & \multicolumn{10}{c|}{Forget Class} & \multirow{2}{*}{Avg.} \\
 & & 0 & 1 & 2 & 3 & 4 & 5 & 6 & 7 & 8 & 9 & \\
\midrule
\multirow{3}{*}{Retrained} & Softmax & \textbf{$0.59${\tiny $\,\pm\,0.01$}} & \textbf{$0.38${\tiny $\,\pm\,0.03$}} & \textbf{$0.58${\tiny $\,\pm\,0.02$}} & \textbf{$0.62${\tiny $\,\pm\,0.02$}} & \textbf{$0.50${\tiny $\,\pm\,0.01$}} & \textbf{$0.35${\tiny $\,\pm\,0.01$}} & \textbf{$0.48${\tiny $\,\pm\,0.02$}} & \textbf{$0.63${\tiny $\,\pm\,0.02$}} & \textbf{$0.42${\tiny $\,\pm\,0.01$}} & \textbf{$0.43${\tiny $\,\pm\,0.01$}} & \textbf{$0.50$} \\
 & Entropy & $0.52${\tiny $\,\pm\,0.00$} & $0.31${\tiny $\,\pm\,0.00$} & $0.45${\tiny $\,\pm\,0.01$} & $0.53${\tiny $\,\pm\,0.02$} & $0.37${\tiny $\,\pm\,0.02$} & $0.23${\tiny $\,\pm\,0.02$} & $0.39${\tiny $\,\pm\,0.03$} & $0.51${\tiny $\,\pm\,0.01$} & $0.34${\tiny $\,\pm\,0.01$} & $0.38${\tiny $\,\pm\,0.01$} & $0.40$ \\
 & Energy & $0.51${\tiny $\,\pm\,0.01$} & $0.29${\tiny $\,\pm\,0.01$} & $0.39${\tiny $\,\pm\,0.02$} & $0.48${\tiny $\,\pm\,0.03$} & $0.28${\tiny $\,\pm\,0.03$} & $0.18${\tiny $\,\pm\,0.01$} & $0.33${\tiny $\,\pm\,0.02$} & $0.45${\tiny $\,\pm\,0.02$} & $0.30${\tiny $\,\pm\,0.01$} & $0.36${\tiny $\,\pm\,0.01$} & $0.36$ \\
\midrule
\multirow{3}{*}{Finetune \cite{golatkar2020eternal}} & Softmax & \textbf{$0.65${\tiny $\,\pm\,0.02$}} & $0.41${\tiny $\,\pm\,0.01$} & \textbf{$0.64${\tiny $\,\pm\,0.02$}} & \textbf{$0.62${\tiny $\,\pm\,0.01$}} & \textbf{$0.60${\tiny $\,\pm\,0.00$}} & \textbf{$0.42${\tiny $\,\pm\,0.00$}} & \textbf{$0.70${\tiny $\,\pm\,0.01$}} & \textbf{$0.59${\tiny $\,\pm\,0.02$}} & $0.60${\tiny $\,\pm\,0.01$} & $0.45${\tiny $\,\pm\,0.01$} & \textbf{$0.57$} \\
 & Entropy & $0.65${\tiny $\,\pm\,0.00$} & \textbf{$0.42${\tiny $\,\pm\,0.01$}} & $0.59${\tiny $\,\pm\,0.02$} & $0.61${\tiny $\,\pm\,0.02$} & $0.56${\tiny $\,\pm\,0.00$} & $0.37${\tiny $\,\pm\,0.01$} & $0.66${\tiny $\,\pm\,0.02$} & $0.57${\tiny $\,\pm\,0.02$} & $0.67${\tiny $\,\pm\,0.01$} & \textbf{$0.46${\tiny $\,\pm\,0.01$}} & $0.56$ \\
 & Energy & $0.64${\tiny $\,\pm\,0.01$} & $0.38${\tiny $\,\pm\,0.01$} & $0.55${\tiny $\,\pm\,0.01$} & $0.59${\tiny $\,\pm\,0.00$} & $0.49${\tiny $\,\pm\,0.02$} & $0.32${\tiny $\,\pm\,0.02$} & $0.65${\tiny $\,\pm\,0.02$} & $0.52${\tiny $\,\pm\,0.02$} & \textbf{$0.68${\tiny $\,\pm\,0.02$}} & $0.42${\tiny $\,\pm\,0.01$} & $0.52$ \\
\midrule
\multirow{3}{*}{Negative Gradient \cite{golatkar2020eternal}} & Softmax & \textbf{$0.68${\tiny $\,\pm\,0.01$}} & $0.26${\tiny $\,\pm\,0.03$} & \textbf{$0.47${\tiny $\,\pm\,0.03$}} & \textbf{$0.66${\tiny $\,\pm\,0.01$}} & \textbf{$0.78${\tiny $\,\pm\,0.00$}} & \textbf{$0.57${\tiny $\,\pm\,0.04$}} & \textbf{$0.41${\tiny $\,\pm\,0.03$}} & \textbf{$0.50${\tiny $\,\pm\,0.02$}} & \textbf{$0.41${\tiny $\,\pm\,0.01$}} & \textbf{$0.36${\tiny $\,\pm\,0.02$}} & \textbf{$0.51$} \\
 & Entropy & $0.64${\tiny $\,\pm\,0.02$} & \textbf{$0.29${\tiny $\,\pm\,0.02$}} & $0.42${\tiny $\,\pm\,0.01$} & $0.61${\tiny $\,\pm\,0.03$} & $0.74${\tiny $\,\pm\,0.01$} & $0.51${\tiny $\,\pm\,0.02$} & $0.33${\tiny $\,\pm\,0.01$} & $0.45${\tiny $\,\pm\,0.02$} & $0.37${\tiny $\,\pm\,0.02$} & $0.35${\tiny $\,\pm\,0.01$} & $0.47$ \\
 & Energy & $0.66${\tiny $\,\pm\,0.01$} & $0.26${\tiny $\,\pm\,0.01$} & $0.38${\tiny $\,\pm\,0.01$} & $0.59${\tiny $\,\pm\,0.01$} & $0.72${\tiny $\,\pm\,0.01$} & $0.53${\tiny $\,\pm\,0.02$} & $0.31${\tiny $\,\pm\,0.01$} & $0.44${\tiny $\,\pm\,0.02$} & $0.41${\tiny $\,\pm\,0.03$} & $0.36${\tiny $\,\pm\,0.01$} & $0.47$ \\
\midrule
\multirow{3}{*}{Negative Gradient+ \cite{kurmanji2023towards}} & Softmax & \textbf{$0.65${\tiny $\,\pm\,0.01$}} & $0.24${\tiny $\,\pm\,0.00$} & \textbf{$0.50${\tiny $\,\pm\,0.00$}} & $0.59${\tiny $\,\pm\,0.01$} & \textbf{$0.72${\tiny $\,\pm\,0.01$}} & \textbf{$0.45${\tiny $\,\pm\,0.03$}} & \textbf{$0.39${\tiny $\,\pm\,0.02$}} & \textbf{$0.48${\tiny $\,\pm\,0.01$}} & $0.38${\tiny $\,\pm\,0.02$} & $0.32${\tiny $\,\pm\,0.01$} & \textbf{$0.47$} \\
 & Entropy & $0.61${\tiny $\,\pm\,0.02$} & $0.23${\tiny $\,\pm\,0.01$} & $0.43${\tiny $\,\pm\,0.02$} & $0.56${\tiny $\,\pm\,0.01$} & $0.67${\tiny $\,\pm\,0.02$} & $0.36${\tiny $\,\pm\,0.03$} & $0.27${\tiny $\,\pm\,0.02$} & $0.42${\tiny $\,\pm\,0.02$} & $0.36${\tiny $\,\pm\,0.03$} & $0.27${\tiny $\,\pm\,0.01$} & $0.42$ \\
 & Energy & $0.64${\tiny $\,\pm\,0.02$} & \textbf{$0.29${\tiny $\,\pm\,0.02$}} & $0.45${\tiny $\,\pm\,0.02$} & \textbf{$0.60${\tiny $\,\pm\,0.01$}} & $0.65${\tiny $\,\pm\,0.02$} & $0.41${\tiny $\,\pm\,0.02$} & $0.31${\tiny $\,\pm\,0.01$} & $0.47${\tiny $\,\pm\,0.01$} & \textbf{$0.38${\tiny $\,\pm\,0.03$}} & \textbf{$0.38${\tiny $\,\pm\,0.01$}} & $0.46$ \\
\midrule
\multirow{3}{*}{Random Label \cite{hayase2020selective}} & Softmax & $0.81${\tiny $\,\pm\,0.10$} & $0.88${\tiny $\,\pm\,0.00$} & $0.66${\tiny $\,\pm\,0.05$} & \textbf{$0.76${\tiny $\,\pm\,0.00$}} & $0.70${\tiny $\,\pm\,0.05$} & $0.83${\tiny $\,\pm\,0.01$} & $0.85${\tiny $\,\pm\,0.00$} & $0.84${\tiny $\,\pm\,0.00$} & $0.78${\tiny $\,\pm\,0.04$} & $0.83${\tiny $\,\pm\,0.02$} & $0.79$ \\
 & Entropy & \textbf{$0.90${\tiny $\,\pm\,0.01$}} & \textbf{$0.89${\tiny $\,\pm\,0.00$}} & $0.75${\tiny $\,\pm\,0.10$} & $0.65${\tiny $\,\pm\,0.01$} & \textbf{$0.74${\tiny $\,\pm\,0.09$}} & \textbf{$0.83${\tiny $\,\pm\,0.00$}} & \textbf{$0.89${\tiny $\,\pm\,0.00$}} & \textbf{$0.87${\tiny $\,\pm\,0.01$}} & \textbf{$0.86${\tiny $\,\pm\,0.00$}} & \textbf{$0.85${\tiny $\,\pm\,0.00$}} & \textbf{$0.82$} \\
 & Energy & $0.84${\tiny $\,\pm\,0.01$} & $0.88${\tiny $\,\pm\,0.00$} & \textbf{$0.78${\tiny $\,\pm\,0.01$}} & $0.65${\tiny $\,\pm\,0.05$} & $0.70${\tiny $\,\pm\,0.09$} & $0.82${\tiny $\,\pm\,0.01$} & $0.88${\tiny $\,\pm\,0.00$} & $0.84${\tiny $\,\pm\,0.00$} & $0.85${\tiny $\,\pm\,0.01$} & $0.84${\tiny $\,\pm\,0.00$} & $0.81$ \\
\midrule
\multirow{3}{*}{Learn to Unlearn \cite{cha2024learning}} & Softmax & \textbf{$0.73${\tiny $\,\pm\,0.01$}} & $0.48${\tiny $\,\pm\,0.01$} & \textbf{$0.53${\tiny $\,\pm\,0.01$}} & \textbf{$0.71${\tiny $\,\pm\,0.00$}} & \textbf{$0.79${\tiny $\,\pm\,0.00$}} & \textbf{$0.68${\tiny $\,\pm\,0.01$}} & \textbf{$0.55${\tiny $\,\pm\,0.04$}} & \textbf{$0.58${\tiny $\,\pm\,0.02$}} & \textbf{$0.58${\tiny $\,\pm\,0.02$}} & $0.55${\tiny $\,\pm\,0.00$} & \textbf{$0.62$} \\
 & Entropy & $0.70${\tiny $\,\pm\,0.03$} & \textbf{$0.52${\tiny $\,\pm\,0.01$}} & $0.45${\tiny $\,\pm\,0.03$} & $0.68${\tiny $\,\pm\,0.01$} & $0.77${\tiny $\,\pm\,0.01$} & $0.65${\tiny $\,\pm\,0.02$} & $0.44${\tiny $\,\pm\,0.01$} & $0.56${\tiny $\,\pm\,0.03$} & $0.48${\tiny $\,\pm\,0.01$} & $0.53${\tiny $\,\pm\,0.01$} & $0.58$ \\
 & Energy & $0.69${\tiny $\,\pm\,0.01$} & $0.46${\tiny $\,\pm\,0.02$} & $0.44${\tiny $\,\pm\,0.02$} & $0.70${\tiny $\,\pm\,0.01$} & $0.76${\tiny $\,\pm\,0.01$} & $0.66${\tiny $\,\pm\,0.00$} & $0.46${\tiny $\,\pm\,0.02$} & $0.55${\tiny $\,\pm\,0.00$} & $0.55${\tiny $\,\pm\,0.01$} & \textbf{$0.57${\tiny $\,\pm\,0.02$}} & $0.59$ \\
\midrule
\multirow{3}{*}{SCRUB \cite{kurmanji2023towards}} & Softmax & $0.70${\tiny $\,\pm\,0.01$} & $0.23${\tiny $\,\pm\,0.01$} & \textbf{$0.53${\tiny $\,\pm\,0.02$}} & \textbf{$0.67${\tiny $\,\pm\,0.01$}} & \textbf{$0.48${\tiny $\,\pm\,0.03$}} & $0.35${\tiny $\,\pm\,0.03$} & \textbf{$0.41${\tiny $\,\pm\,0.03$}} & $0.49${\tiny $\,\pm\,0.06$} & \textbf{$0.48${\tiny $\,\pm\,0.02$}} & $0.38${\tiny $\,\pm\,0.02$} & $0.47$ \\
 & Entropy & $0.67${\tiny $\,\pm\,0.03$} & $0.16${\tiny $\,\pm\,0.02$} & $0.46${\tiny $\,\pm\,0.04$} & $0.64${\tiny $\,\pm\,0.02$} & $0.29${\tiny $\,\pm\,0.02$} & $0.29${\tiny $\,\pm\,0.03$} & $0.29${\tiny $\,\pm\,0.03$} & $0.37${\tiny $\,\pm\,0.10$} & $0.46${\tiny $\,\pm\,0.02$} & $0.38${\tiny $\,\pm\,0.03$} & $0.40$ \\
 & Energy & \textbf{$0.70${\tiny $\,\pm\,0.01$}} & \textbf{$0.28${\tiny $\,\pm\,0.03$}} & $0.52${\tiny $\,\pm\,0.01$} & $0.65${\tiny $\,\pm\,0.01$} & $0.42${\tiny $\,\pm\,0.03$} & \textbf{$0.44${\tiny $\,\pm\,0.03$}} & $0.35${\tiny $\,\pm\,0.01$} & \textbf{$0.53${\tiny $\,\pm\,0.05$}} & $0.47${\tiny $\,\pm\,0.01$} & \textbf{$0.43${\tiny $\,\pm\,0.01$}} & \textbf{$0.48$} \\
\midrule
\multirow{3}{*}{Bad Teacher \cite{chundawat2023can}} & Softmax & $0.94${\tiny $\,\pm\,0.00$} & $0.90${\tiny $\,\pm\,0.01$} & $0.83${\tiny $\,\pm\,0.04$} & \textbf{$0.83${\tiny $\,\pm\,0.04$}} & $0.90${\tiny $\,\pm\,0.02$} & \textbf{$0.89${\tiny $\,\pm\,0.00$}} & $0.88${\tiny $\,\pm\,0.00$} & $0.84${\tiny $\,\pm\,0.00$} & $0.94${\tiny $\,\pm\,0.00$} & \textbf{$0.96${\tiny $\,\pm\,0.00$}} & \textbf{$0.89$} \\
 & Entropy & \textbf{$0.95${\tiny $\,\pm\,0.00$}} & \textbf{$0.90${\tiny $\,\pm\,0.00$}} & \textbf{$0.83${\tiny $\,\pm\,0.01$}} & $0.53${\tiny $\,\pm\,0.46$} & \textbf{$0.92${\tiny $\,\pm\,0.00$}} & $0.58${\tiny $\,\pm\,0.51$} & $0.88${\tiny $\,\pm\,0.00$} & \textbf{$0.84${\tiny $\,\pm\,0.00$}} & $0.94${\tiny $\,\pm\,0.00$} & $0.96${\tiny $\,\pm\,0.00$} & $0.83$ \\
 & Energy & $0.94${\tiny $\,\pm\,0.00$} & $0.90${\tiny $\,\pm\,0.00$} & $0.82${\tiny $\,\pm\,0.01$} & $0.78${\tiny $\,\pm\,0.02$} & $0.91${\tiny $\,\pm\,0.00$} & $0.87${\tiny $\,\pm\,0.01$} & \textbf{$0.89${\tiny $\,\pm\,0.00$}} & $0.83${\tiny $\,\pm\,0.01$} & \textbf{$0.94${\tiny $\,\pm\,0.00$}} & $0.96${\tiny $\,\pm\,0.00$} & $0.88$ \\
\midrule
\multirow{3}{*}{SalUn \cite{fan2023salun}} & Softmax & $0.85${\tiny $\,\pm\,0.01$} & $0.76${\tiny $\,\pm\,0.01$} & $0.82${\tiny $\,\pm\,0.00$} & \textbf{$0.86${\tiny $\,\pm\,0.00$}} & $0.78${\tiny $\,\pm\,0.00$} & $0.69${\tiny $\,\pm\,0.01$} & \textbf{$0.82${\tiny $\,\pm\,0.00$}} & \textbf{$0.82${\tiny $\,\pm\,0.01$}} & \textbf{$0.81${\tiny $\,\pm\,0.01$}} & $0.81${\tiny $\,\pm\,0.01$} & $0.80$ \\
 & Entropy & \textbf{$0.86${\tiny $\,\pm\,0.01$}} & $0.77${\tiny $\,\pm\,0.02$} & $0.84${\tiny $\,\pm\,0.00$} & $0.85${\tiny $\,\pm\,0.00$} & \textbf{$0.82${\tiny $\,\pm\,0.01$}} & \textbf{$0.72${\tiny $\,\pm\,0.01$}} & $0.81${\tiny $\,\pm\,0.01$} & $0.81${\tiny $\,\pm\,0.01$} & $0.81${\tiny $\,\pm\,0.01$} & \textbf{$0.83${\tiny $\,\pm\,0.00$}} & \textbf{$0.81$} \\
 & Energy & $0.85${\tiny $\,\pm\,0.01$} & \textbf{$0.78${\tiny $\,\pm\,0.01$}} & \textbf{$0.85${\tiny $\,\pm\,0.01$}} & $0.83${\tiny $\,\pm\,0.01$} & $0.78${\tiny $\,\pm\,0.00$} & $0.65${\tiny $\,\pm\,0.00$} & $0.81${\tiny $\,\pm\,0.01$} & $0.81${\tiny $\,\pm\,0.01$} & $0.81${\tiny $\,\pm\,0.01$} & $0.80${\tiny $\,\pm\,0.01$} & $0.80$ \\
\midrule
\multirow{3}{*}{DELETE \cite{zhou2025decoupled}} & Softmax & $0.69${\tiny $\,\pm\,0.02$} & $0.32${\tiny $\,\pm\,0.02$} & \textbf{$0.66${\tiny $\,\pm\,0.02$}} & $0.57${\tiny $\,\pm\,0.02$} & \textbf{$0.51${\tiny $\,\pm\,0.02$}} & $0.20${\tiny $\,\pm\,0.01$} & \textbf{$0.52${\tiny $\,\pm\,0.03$}} & $0.47${\tiny $\,\pm\,0.03$} & \textbf{$0.52${\tiny $\,\pm\,0.02$}} & $0.42${\tiny $\,\pm\,0.01$} & $0.49$ \\
 & Entropy & $0.67${\tiny $\,\pm\,0.03$} & $0.30${\tiny $\,\pm\,0.02$} & $0.61${\tiny $\,\pm\,0.01$} & $0.59${\tiny $\,\pm\,0.02$} & $0.42${\tiny $\,\pm\,0.03$} & $0.19${\tiny $\,\pm\,0.03$} & $0.36${\tiny $\,\pm\,0.05$} & $0.38${\tiny $\,\pm\,0.03$} & $0.37${\tiny $\,\pm\,0.04$} & $0.36${\tiny $\,\pm\,0.01$} & $0.42$ \\
 & Energy & \textbf{$0.74${\tiny $\,\pm\,0.00$}} & \textbf{$0.40${\tiny $\,\pm\,0.03$}} & $0.64${\tiny $\,\pm\,0.01$} & \textbf{$0.63${\tiny $\,\pm\,0.02$}} & $0.42${\tiny $\,\pm\,0.04$} & \textbf{$0.30${\tiny $\,\pm\,0.02$}} & $0.44${\tiny $\,\pm\,0.01$} & \textbf{$0.50${\tiny $\,\pm\,0.02$}} & $0.51${\tiny $\,\pm\,0.02$} & \textbf{$0.49${\tiny $\,\pm\,0.03$}} & \textbf{$0.51$} \\
\bottomrule
\end{tabular}%
}
\end{table*}

\section{Effect of Gaussian Support on SFRA}
\label{app:gaussian_support}

SFRA uses Gaussian embeddings as queries to the released classifier head, not as an estimate of the empirical feature distribution. Nevertheless, when the encoder representation is produced after a ReLU activation, its natural support is non-negative, whereas the standard Gaussian proposal contains both positive and negative coordinates. We therefore examine whether signed coordinates are necessary for successful relearning. We compare three proposals:
\begin{equation}
    s_{\mathrm{signed}}=z,\\
    s_{\mathrm{ReLU}}=\max(0,z),\\
    s_{\mathrm{abs}}=|z|,~z\sim\mathcal{N}(0,I).
\end{equation}
The ReLU-Gaussian proposal constrains embeddings to the non-negative orthant but introduces zeros and reduces their expected squared norm. The absolute Gaussian proposal is also non-negative but preserves the norm of every corresponding signed sample, since $\lVert |z|\rVert_2=\lVert z\rVert_2$. It therefore provides a control for separating coordinate support from probe norm. We use the CIFAR-10 ResNet-18 checkpoint produced by Bad Teacher with forget class~7. For every proposal, the checkpoint, underlying Gaussian random streams, candidate and probe counts, low-confidence forget selection, high-confidence retain selection, classifier-head optimization, and evaluation protocol are fixed. Specifically, for each of the nine retain classes, we collect $N=500{,}000$ accepted candidate embeddings predicted as that class by the unlearned classifier. From each class-specific candidate pool, we select the $M=500$ highest-confidence embeddings as retain probes and the $M=500$ lowest-confidence embeddings as forget probes. The latter are relabeled as forget class~7. Consequently, each run uses $9M=4{,}500$ synthetic retain probes and $9M=4{,}500$ synthetic forget probes. Here, $N$ denotes the number of accepted candidates rather than the number of raw proposal draws. Because the classifier acceptance rate depends on the proposal distribution, the number of raw draws required to obtain the same fixed $N$ varies across the three conditions and is reported separately in
Table~\ref{tab:gaussian_support_performance}. 

\begin{table}[H]
    \centering
    \caption{Effect of Gaussian support on SFRA for Bad Teacher on CIFAR-10 using ResNet-18 and forget class~7. Draws denote the total number of proposals required to construct the accepted class-specific pools, and time denotes their total generation time.}
    \label{tab:gaussian_support_performance}
    \setlength{\tabcolsep}{2.5pt}
    \renewcommand{\arraystretch}{1.1}
    \scriptsize
    \begin{tabular}{lccccc}
        \toprule
        Proposal & $\mathcal{A}_r^t$(\%) & $\mathcal{A}_f^t$(\%) &
        $\mathrm{RS}$ & Draws (M) & Time (s) \\
        \midrule
        Signed Gaussian
        & $93.02$ & $98.60$ & $0.99$ & $44.95$ & $7.54$ \\
        ReLU-Gaussian
        & $94.38$ & $88.20$ & $0.94$ & $54.67$ & $7.87$ \\
        Absolute Gaussian
        & $94.56$ & $0.10$ & $0.00$ & $138.15$ & $12.53$ \\
        \bottomrule
    \end{tabular}
\end{table}

Table~\ref{tab:gaussian_support_performance} shows that signed Gaussian probes provide the strongest and most efficient recovery. Constraining the probes to the non-negative orthant with ReLU still yields substantial recovery, with $\mathrm{RS}=0.94$, showing that negative coordinates are not required for SFRA to succeed. The absolute Gaussian condition, however, produces almost no recovery despite preserving the signed Gaussian norms. The difference between the signed and ReLU conditions therefore cannot be explained by norm alone.

\begin{table}[H]
    \centering
    \caption{Coordinate-support diagnostics for the same Gaussian-support ablation, computed over the selected synthetic forget set. The accepted-pool measurements exhibit the same sign pattern.}
    \label{tab:gaussian_support_diagnostics}
    \setlength{\tabcolsep}{3pt}
    \renewcommand{\arraystretch}{1.1}
    \scriptsize
    \begin{tabular}{lcccc}
        \toprule
        Proposal & Neg. & Zero & Non-neg. & Mean \\
        & coord. (\%) & coord. (\%) & vectors (\%) & $\ell_2$ norm \\
        \midrule
        Signed Gaussian   & $49.98$ & $0.00$ & $0.00$ & $22.49$ \\
        ReLU-Gaussian     & $0.00$  & $50.33$ & $100.00$ & $15.74$ \\
        Absolute Gaussian & $0.00$  & $0.00$ & $100.00$ & $22.50$ \\
        \bottomrule
    \end{tabular}
\end{table}

The sign diagnostics in Table~\ref{tab:gaussian_support_diagnostics} also show that classifier-based rejection sampling does not implicitly remove negative coordinates. The raw signed proposals contain $49.99\%$ negative coordinates, the accepted candidate pools contain $49.96\%$, and the selected retain and forget sets contain $49.93\%$ and $49.98\%$, respectively. Thus, the original SFRA procedure genuinely uses signed, potentially off-manifold queries. The comparison further distinguishes non-negativity from the structure induced by ReLU. ReLU-Gaussian probes contain approximately $50\%$ exact zeros and remain effective, whereas absolute-Gaussian probes are dense, strictly non-negative almost surely, and fail despite matching the signed-probe norms. This result suggests that restricting the proposal to non-negative support is compatible with SFRA, but that the particular geometry of the proposal, including its sparsity pattern, materially affects recovery. Overall, the audit does not depend on negative coordinates, although signed Gaussian probes provide substantially stronger and faster recovery in this setting.

\section{Detailed Per-Class Results and Linear Separability}
\label{app:linear_separability}

This section provides detailed per-class results for the single-class unlearning experiments, complementing the aggregate and worst-case results in the main paper. For each forget class, we report the unlearned performance, source-dependent PRA and our proposed SFRA results, and frozen-encoder linear-probe accuracy.

We further examine whether forget class information remains linearly accessible in the unlearned representation independently of the released classifier head. For each unlearned checkpoint, we freeze the encoder and train a linear classifier on its representations using real labeled training samples, reporting the forget class test accuracy as $\mathcal{A}_f^{\mathrm{LP}}$. A high $\mathcal{A}_f^{\mathrm{LP}}$
indicates that the forget class remains linearly separable in the post-unlearning representation. This is a post-hoc, source-dependent diagnostic: the real samples used for linear probing are never used for synthetic-probe construction or our proposed SFRA. Thus,
$\mathcal{A}_f^{\mathrm{LP}}$ measures supervised representation-level accessibility, whereas $\mathrm{RS}$ measures source-free recoverability using only the released model and synthetic probes. For ResNet-18, the per-class results on CIFAR-10, CIFAR-100, and TinyImageNet are reported in Tables~\ref{tab:cifar10_resnet18_variant_by_forget},
\ref{tab:cifar100_resnet18_variant_by_forget}, and
\ref{tab:tiny_imagenet_resnet18_variant_by_forget}, respectively. The corresponding ViT-B/16 results are reported in Tables~\ref{tab:cifar10_vit-b-16_variant_by_forget}, \ref{tab:cifar100_vit-b-16_variant_by_forget}, and
\ref{tab:tiny_imagenet_vit-b-16_variant_by_forget}.
For Swin-T, the results are reported in Tables~\ref{tab:cifar10_swin-t_variant_by_forget},
\ref{tab:cifar100_swin-t_variant_by_forget}, and
\ref{tab:tiny_imagenet_swin-t_variant_by_forget}.
These results reveal class-specific differences in recoverability and representation-level linear separability.

\begin{table*}[t]
\centering
\caption{Per-forget-class single-class unlearning and relearning results on CIFAR-10 using ResNet-18. We report the unlearned checkpoint, the source-dependent PRA baseline, our proposed SFRA, and frozen-encoder linear probing. Each forget class column corresponds to a separate unlearned checkpoint in which that class is designated for forgetting.}
\label{tab:cifar10_resnet18_variant_by_forget}
\begin{minipage}[t]{0.49\textwidth}
\centering
\scriptsize
\setlength{\tabcolsep}{2pt}
\renewcommand{\arraystretch}{0.80}
\resizebox{\columnwidth}{!}{%
\begin{tabular}{l|l|l|cccccccccc}
\toprule
\multirow{2}{*}{Unlearning Method} & \multirow{2}{*}{Metric} & \multirow{2}{*}{Variant} & \multicolumn{10}{c}{Forget Class} \\
 & & & 0 & 1 & 2 & 3 & 4 & 5 & 6 & 7 & 8 & 9 \\
\midrule
\multirow{2}{*}{Original} & \multirow{1}{*}{$\mathcal{A}^{t}_{r}(\%)$} & Original & $94.60$ & $94.29$ & $94.89$ & $95.28$ & $94.54$ & $95.01$ & $94.47$ & $94.54$ & $94.36$ & $94.52$ \\
\cmidrule(lr){2-13}
 & \multirow{1}{*}{$\mathcal{A}^{t}_{f}(\%)$} & Original & $95.10$ & $97.90$ & $92.50$ & $89.00$ & $95.60$ & $91.40$ & $96.30$ & $95.60$ & $97.30$ & $95.80$ \\
\midrule
\multirow{9}{*}{Retrained} & \multirow{3}{*}{$\mathcal{A}^{t}_{r}(\%)$} & Unlearned & $95.19$ & $94.81$ & $95.21$ & $96.44$ & $94.83$ & $95.94$ & $94.64$ & $94.72$ & $94.87$ & $95.21$ \\
 &  & PRA \cite{ha2025unlearning} & $94.99$ & $94.88$ & $95.04$ & $96.37$ & $94.52$ & $95.70$ & $94.40$ & $94.52$ & $94.88$ & $95.17$ \\
 &  & SFRA (ours) & $92.67$ & $91.48$ & $92.79$ & $94.11$ & $91.60$ & $93.43$ & $92.06$ & $92.59$ & $91.44$ & $92.07$ \\
\cmidrule(lr){2-13}
 & \multirow{3}{*}{$\mathcal{A}^{t}_{f}(\%)$} & Unlearned & $0.00$ & $0.00$ & $0.00$ & $0.00$ & $0.00$ & $0.00$ & $0.00$ & $0.00$ & $0.00$ & $0.00$ \\
 &  & PRA \cite{ha2025unlearning} & $14.40$ & $2.90$ & $17.80$ & $9.00$ & $14.70$ & $6.00$ & $14.30$ & $14.50$ & $3.70$ & $7.80$ \\
 &  & SFRA (ours) & $37.90$ & $12.20$ & $46.10$ & $40.90$ & $54.10$ & $18.60$ & $35.10$ & $32.40$ & $28.00$ & $21.90$ \\
\cmidrule(lr){2-13}
 & \multirow{1}{*}{$\mathcal{A}^{LP}_{f}(\%)$} & Linear Probe & $80.10$ & $76.60$ & $79.10$ & $77.40$ & $86.40$ & $69.40$ & $83.60$ & $83.40$ & $78.20$ & $84.40$ \\
\cmidrule(lr){2-13}
 & \multirow{2}{*}{$\mathrm{RS}$} & PRA \cite{ha2025unlearning} & $0.25$ & $0.06$ & $0.30$ & $0.17$ & $0.26$ & $0.11$ & $0.25$ & $0.25$ & $0.07$ & $0.14$ \\
 &  & SFRA (ours) & $0.55$ & $0.22$ & $0.63$ & $0.58$ & $0.69$ & $0.31$ & $0.52$ & $0.49$ & $0.43$ & $0.36$ \\
\midrule
\multirow{11}{*}{Finetune \cite{golatkar2020eternal}} & \multirow{3}{*}{$\mathcal{A}^{t}_{r}(\%)$} & Unlearned & $94.71$ & $94.03$ & $94.97$ & $96.01$ & $94.62$ & $95.30$ & $94.27$ & $94.26$ & $94.26$ & $94.89$ \\
 &  & PRA \cite{ha2025unlearning} & $94.63$ & $94.02$ & $94.87$ & $96.00$ & $94.54$ & $95.18$ & $94.24$ & $94.22$ & $94.21$ & $94.88$ \\
 &  & SFRA (ours) & $93.27$ & $91.98$ & $93.66$ & $94.54$ & $93.58$ & $94.26$ & $92.22$ & $92.73$ & $92.87$ & $93.46$ \\
\cmidrule(lr){2-13}
 & \multirow{3}{*}{$\mathcal{A}^{t}_{f}(\%)$} & Unlearned & $0.00$ & $0.00$ & $0.00$ & $0.00$ & $0.00$ & $0.00$ & $0.00$ & $0.00$ & $0.00$ & $0.00$ \\
 &  & PRA \cite{ha2025unlearning} & $8.10$ & $0.70$ & $6.90$ & $5.30$ & $4.60$ & $2.80$ & $7.20$ & $6.30$ & $3.20$ & $2.50$ \\
 &  & SFRA (ours) & $41.30$ & $12.90$ & $44.50$ & $30.00$ & $40.00$ & $15.50$ & $54.20$ & $39.10$ & $26.40$ & $15.90$ \\
\cmidrule(lr){2-13}
 & \multirow{1}{*}{$\mathcal{A}^{LP}_{f}(\%)$} & Linear Probe & $93.70$ & $95.10$ & $91.30$ & $89.60$ & $93.70$ & $87.30$ & $93.30$ & $94.30$ & $92.40$ & $95.30$ \\
\cmidrule(lr){2-13}
 & \multirow{2}{*}{$\mathrm{RS}$} & PRA \cite{ha2025unlearning} & $0.15$ & $0.01$ & $0.13$ & $0.10$ & $0.09$ & $0.05$ & $0.13$ & $0.12$ & $0.06$ & $0.05$ \\
 &  & SFRA (ours) & $0.58$ & $0.23$ & $0.61$ & $0.46$ & $0.57$ & $0.27$ & $0.70$ & $0.56$ & $0.42$ & $0.27$ \\
\cmidrule(lr){2-13}
 & \multirow{2}{*}{$\Delta\mathrm{RS}$} & PRA \cite{ha2025unlearning} & $-0.10$ & $-0.04$ & $-0.17$ & $-0.06$ & $-0.17$ & $-0.06$ & $-0.12$ & $-0.13$ & $-0.01$ & $-0.10$ \\
 &  & SFRA (ours) & $+0.04$ & $+0.01$ & $-0.01$ & $-0.12$ & $-0.12$ & $-0.04$ & $+0.18$ & $+0.07$ & $-0.02$ & $-0.08$ \\
\midrule
\multirow{11}{*}{Negative Gradient \cite{golatkar2020eternal}} & \multirow{3}{*}{$\mathcal{A}^{t}_{r}(\%)$} & Unlearned & $90.24$ & $90.66$ & $89.98$ & $93.46$ & $88.97$ & $88.58$ & $89.28$ & $91.61$ & $88.71$ & $89.69$ \\
 &  & PRA \cite{ha2025unlearning} & $89.64$ & $89.88$ & $89.78$ & $93.04$ & $87.82$ & $87.60$ & $88.78$ & $91.16$ & $88.17$ & $89.67$ \\
 &  & SFRA (ours) & $89.63$ & $90.62$ & $88.68$ & $92.88$ & $87.32$ & $86.88$ & $87.47$ & $90.61$ & $88.59$ & $89.32$ \\
\cmidrule(lr){2-13}
 & \multirow{3}{*}{$\mathcal{A}^{t}_{f}(\%)$} & Unlearned & $8.90$ & $6.70$ & $11.40$ & $5.50$ & $5.30$ & $7.30$ & $7.10$ & $7.70$ & $8.00$ & $7.30$ \\
 &  & PRA \cite{ha2025unlearning} & $20.40$ & $35.50$ & $22.30$ & $24.20$ & $22.60$ & $24.10$ & $28.20$ & $30.50$ & $25.30$ & $16.40$ \\
 &  & SFRA (ours) & $52.50$ & $68.40$ & $50.50$ & $47.10$ & $57.80$ & $55.60$ & $59.80$ & $63.50$ & $57.40$ & $55.00$ \\
\cmidrule(lr){2-13}
 & \multirow{1}{*}{$\mathcal{A}^{LP}_{f}(\%)$} & Linear Probe & $86.50$ & $92.00$ & $82.40$ & $78.40$ & $86.20$ & $81.50$ & $85.90$ & $90.20$ & $90.10$ & $87.00$ \\
\cmidrule(lr){2-13}
 & \multirow{2}{*}{$\mathrm{RS}$} & PRA \cite{ha2025unlearning} & $0.21$ & $0.45$ & $0.20$ & $0.31$ & $0.29$ & $0.29$ & $0.35$ & $0.37$ & $0.29$ & $0.17$ \\
 &  & SFRA (ours) & $0.61$ & $0.76$ & $0.56$ & $0.59$ & $0.68$ & $0.65$ & $0.69$ & $0.71$ & $0.66$ & $0.65$ \\
\cmidrule(lr){2-13}
 & \multirow{2}{*}{$\Delta\mathrm{RS}$} & PRA \cite{ha2025unlearning} & $-0.05$ & $+0.39$ & $-0.11$ & $+0.15$ & $+0.04$ & $+0.17$ & $+0.10$ & $+0.12$ & $+0.22$ & $+0.02$ \\
 &  & SFRA (ours) & $+0.06$ & $+0.55$ & $-0.07$ & $+0.01$ & $-0.01$ & $+0.34$ & $+0.17$ & $+0.23$ & $+0.23$ & $+0.29$ \\
\midrule
\multirow{11}{*}{Negative Gradient+ \cite{kurmanji2023towards}} & \multirow{3}{*}{$\mathcal{A}^{t}_{r}(\%)$} & Unlearned & $86.03$ & $89.28$ & $89.67$ & $91.48$ & $88.14$ & $86.76$ & $87.71$ & $87.91$ & $88.60$ & $90.04$ \\
 &  & PRA \cite{ha2025unlearning} & $86.00$ & $89.24$ & $89.67$ & $91.49$ & $88.08$ & $86.74$ & $87.66$ & $87.90$ & $88.56$ & $90.04$ \\
 &  & SFRA (ours) & $84.13$ & $87.90$ & $87.77$ & $89.86$ & $86.03$ & $81.99$ & $85.38$ & $84.44$ & $86.69$ & $88.97$ \\
\cmidrule(lr){2-13}
 & \multirow{3}{*}{$\mathcal{A}^{t}_{f}(\%)$} & Unlearned & $0.00$ & $0.00$ & $0.00$ & $0.00$ & $0.00$ & $0.00$ & $0.00$ & $0.10$ & $0.00$ & $0.00$ \\
 &  & PRA \cite{ha2025unlearning} & $1.50$ & $1.10$ & $0.50$ & $2.10$ & $2.80$ & $0.20$ & $5.00$ & $0.20$ & $2.10$ & $2.50$ \\
 &  & SFRA (ours) & $18.40$ & $19.30$ & $11.80$ & $30.20$ & $32.60$ & $20.10$ & $31.70$ & $11.00$ & $25.80$ & $29.80$ \\
\cmidrule(lr){2-13}
 & \multirow{1}{*}{$\mathcal{A}^{LP}_{f}(\%)$} & Linear Probe & $83.30$ & $87.70$ & $72.40$ & $71.20$ & $85.80$ & $76.50$ & $86.50$ & $77.70$ & $88.90$ & $88.40$ \\
\cmidrule(lr){2-13}
 & \multirow{2}{*}{$\mathrm{RS}$} & PRA \cite{ha2025unlearning} & $0.03$ & $0.02$ & $0.01$ & $0.04$ & $0.05$ & $0.00$ & $0.10$ & $0.00$ & $0.04$ & $0.05$ \\
 &  & SFRA (ours) & $0.31$ & $0.32$ & $0.21$ & $0.46$ & $0.49$ & $0.33$ & $0.48$ & $0.20$ & $0.41$ & $0.46$ \\
\cmidrule(lr){2-13}
 & \multirow{2}{*}{$\Delta\mathrm{RS}$} & PRA \cite{ha2025unlearning} & $-0.22$ & $-0.03$ & $-0.29$ & $-0.12$ & $-0.20$ & $-0.11$ & $-0.15$ & $-0.25$ & $-0.03$ & $-0.10$ \\
 &  & SFRA (ours) & $-0.24$ & $+0.11$ & $-0.42$ & $-0.11$ & $-0.20$ & $+0.02$ & $-0.04$ & $-0.29$ & $-0.03$ & $+0.10$ \\
\midrule
\multirow{11}{*}{Random Label \cite{hayase2020selective}} & \multirow{3}{*}{$\mathcal{A}^{t}_{r}(\%)$} & Unlearned & $92.31$ & $91.51$ & $92.24$ & $94.46$ & $91.41$ & $91.38$ & $90.59$ & $92.88$ & $91.10$ & $92.29$ \\
 &  & PRA \cite{ha2025unlearning} & $91.77$ & $90.59$ & $92.07$ & $94.18$ & $90.39$ & $90.34$ & $89.84$ & $92.28$ & $90.56$ & $92.29$ \\
 &  & SFRA (ours) & $91.10$ & $91.08$ & $91.08$ & $93.46$ & $90.40$ & $90.16$ & $88.89$ & $92.11$ & $90.70$ & $91.86$ \\
\cmidrule(lr){2-13}
 & \multirow{3}{*}{$\mathcal{A}^{t}_{f}(\%)$} & Unlearned & $12.70$ & $14.00$ & $16.70$ & $8.70$ & $9.00$ & $12.80$ & $11.10$ & $12.50$ & $15.20$ & $15.70$ \\
 &  & PRA \cite{ha2025unlearning} & $39.80$ & $54.70$ & $36.90$ & $33.80$ & $40.00$ & $39.20$ & $41.30$ & $46.30$ & $41.80$ & $41.60$ \\
 &  & SFRA (ours) & $70.60$ & $80.40$ & $66.60$ & $63.60$ & $70.70$ & $69.20$ & $70.10$ & $73.60$ & $71.80$ & $79.10$ \\
\cmidrule(lr){2-13}
 & \multirow{1}{*}{$\mathcal{A}^{LP}_{f}(\%)$} & Linear Probe & $89.70$ & $94.00$ & $86.80$ & $83.40$ & $90.20$ & $85.00$ & $90.10$ & $92.60$ & $93.20$ & $91.70$ \\
\cmidrule(lr){2-13}
 & \multirow{2}{*}{$\mathrm{RS}$} & PRA \cite{ha2025unlearning} & $0.43$ & $0.58$ & $0.34$ & $0.40$ & $0.47$ & $0.42$ & $0.46$ & $0.50$ & $0.42$ & $0.41$ \\
 &  & SFRA (ours) & $0.73$ & $0.80$ & $0.66$ & $0.71$ & $0.76$ & $0.72$ & $0.74$ & $0.76$ & $0.72$ & $0.77$ \\
\cmidrule(lr){2-13}
 & \multirow{2}{*}{$\Delta\mathrm{RS}$} & PRA \cite{ha2025unlearning} & $+0.17$ & $+0.52$ & $+0.03$ & $+0.24$ & $+0.22$ & $+0.30$ & $+0.21$ & $+0.25$ & $+0.35$ & $+0.27$ \\
 &  & SFRA (ours) & $+0.18$ & $+0.58$ & $+0.04$ & $+0.13$ & $+0.07$ & $+0.41$ & $+0.22$ & $+0.27$ & $+0.29$ & $+0.42$ \\
\bottomrule
\end{tabular}
}
\end{minipage}%
\hfill
\begin{minipage}[t]{0.49\textwidth}
\centering
\scriptsize
\setlength{\tabcolsep}{2pt}
\renewcommand{\arraystretch}{0.80}
\resizebox{\columnwidth}{!}{%
\begin{tabular}{l|l|l|cccccccccc}
\toprule
\multirow{2}{*}{Unlearning Method} & \multirow{2}{*}{Metric} & \multirow{2}{*}{Variant} & \multicolumn{10}{c}{Forget Class} \\
 & & & 0 & 1 & 2 & 3 & 4 & 5 & 6 & 7 & 8 & 9 \\
\midrule
\multirow{11}{*}{Boundary Shrink \cite{chen2023boundary}} & \multirow{3}{*}{$\mathcal{A}^{t}_{r}(\%)$} & Unlearned & $92.33$ & $91.80$ & $92.43$ & $94.68$ & $91.00$ & $91.17$ & $90.69$ & $92.83$ & $91.12$ & $92.23$ \\
 &  & PRA \cite{ha2025unlearning} & $91.77$ & $90.97$ & $92.29$ & $94.41$ & $89.90$ & $90.21$ & $89.96$ & $92.26$ & $90.56$ & $92.20$ \\
 &  & SFRA (ours) & $91.43$ & $91.06$ & $91.97$ & $93.87$ & $89.61$ & $89.68$ & $88.57$ & $92.48$ & $90.42$ & $91.94$ \\
\cmidrule(lr){2-13}
 & \multirow{3}{*}{$\mathcal{A}^{t}_{f}(\%)$} & Unlearned & $13.00$ & $14.40$ & $16.90$ & $9.30$ & $8.50$ & $12.70$ & $11.70$ & $12.60$ & $15.20$ & $15.50$ \\
 &  & PRA \cite{ha2025unlearning} & $38.40$ & $55.10$ & $36.60$ & $33.10$ & $38.60$ & $39.30$ & $41.10$ & $44.90$ & $41.00$ & $40.40$ \\
 &  & SFRA (ours) & $67.90$ & $80.90$ & $63.50$ & $60.30$ & $71.90$ & $67.10$ & $70.60$ & $72.80$ & $71.20$ & $78.60$ \\
\cmidrule(lr){2-13}
 & \multirow{1}{*}{$\mathcal{A}^{LP}_{f}(\%)$} & Linear Probe & $89.10$ & $94.40$ & $86.90$ & $82.30$ & $89.60$ & $85.00$ & $89.30$ & $92.50$ & $92.90$ & $92.20$ \\
\cmidrule(lr){2-13}
 & \multirow{2}{*}{$\mathrm{RS}$} & PRA \cite{ha2025unlearning} & $0.40$ & $0.58$ & $0.33$ & $0.38$ & $0.46$ & $0.42$ & $0.45$ & $0.49$ & $0.41$ & $0.40$ \\
 &  & SFRA (ours) & $0.71$ & $0.80$ & $0.63$ & $0.67$ & $0.77$ & $0.70$ & $0.74$ & $0.75$ & $0.72$ & $0.77$ \\
\cmidrule(lr){2-13}
 & \multirow{2}{*}{$\Delta\mathrm{RS}$} & PRA \cite{ha2025unlearning} & $+0.15$ & $+0.52$ & $+0.03$ & $+0.22$ & $+0.21$ & $+0.31$ & $+0.20$ & $+0.23$ & $+0.34$ & $+0.25$ \\
 &  & SFRA (ours) & $+0.16$ & $+0.58$ & $+0.01$ & $+0.10$ & $+0.08$ & $+0.39$ & $+0.22$ & $+0.26$ & $+0.28$ & $+0.42$ \\
\midrule
\multirow{11}{*}{Learn to Unlearn \cite{cha2024learning}} & \multirow{3}{*}{$\mathcal{A}^{t}_{r}(\%)$} & Unlearned & $91.64$ & $89.40$ & $91.02$ & $93.21$ & $89.71$ & $89.68$ & $88.89$ & $91.61$ & $89.17$ & $89.11$ \\
 &  & PRA \cite{ha2025unlearning} & $91.02$ & $88.88$ & $90.91$ & $93.02$ & $88.88$ & $88.77$ & $88.50$ & $91.28$ & $88.66$ & $89.10$ \\
 &  & SFRA (ours) & $90.74$ & $89.00$ & $89.81$ & $92.36$ & $88.26$ & $87.92$ & $87.28$ & $90.88$ & $89.16$ & $88.46$ \\
\cmidrule(lr){2-13}
 & \multirow{3}{*}{$\mathcal{A}^{t}_{f}(\%)$} & Unlearned & $10.70$ & $7.20$ & $13.60$ & $5.70$ & $6.70$ & $9.00$ & $7.90$ & $8.40$ & $11.30$ & $6.50$ \\
 &  & PRA \cite{ha2025unlearning} & $31.60$ & $22.90$ & $25.00$ & $16.10$ & $25.80$ & $29.60$ & $25.70$ & $27.60$ & $27.70$ & $15.80$ \\
 &  & SFRA (ours) & $60.90$ & $59.10$ & $55.60$ & $44.40$ & $60.90$ & $63.40$ & $60.20$ & $63.70$ & $62.40$ & $58.10$ \\
\cmidrule(lr){2-13}
 & \multirow{1}{*}{$\mathcal{A}^{LP}_{f}(\%)$} & Linear Probe & $88.60$ & $90.10$ & $82.40$ & $75.20$ & $86.50$ & $82.00$ & $85.80$ & $89.50$ & $90.90$ & $88.00$ \\
\cmidrule(lr){2-13}
 & \multirow{2}{*}{$\mathrm{RS}$} & PRA \cite{ha2025unlearning} & $0.35$ & $0.27$ & $0.20$ & $0.19$ & $0.32$ & $0.34$ & $0.30$ & $0.32$ & $0.28$ & $0.17$ \\
 &  & SFRA (ours) & $0.67$ & $0.68$ & $0.59$ & $0.56$ & $0.70$ & $0.70$ & $0.68$ & $0.71$ & $0.68$ & $0.68$ \\
\cmidrule(lr){2-13}
 & \multirow{2}{*}{$\Delta\mathrm{RS}$} & PRA \cite{ha2025unlearning} & $+0.09$ & $+0.21$ & $-0.10$ & $+0.02$ & $+0.06$ & $+0.23$ & $+0.05$ & $+0.07$ & $+0.21$ & $+0.03$ \\
 &  & SFRA (ours) & $+0.12$ & $+0.47$ & $-0.04$ & $-0.02$ & $+0.01$ & $+0.39$ & $+0.17$ & $+0.22$ & $+0.24$ & $+0.32$ \\
\midrule
\multirow{11}{*}{SCRUB \cite{kurmanji2023towards}} & \multirow{3}{*}{$\mathcal{A}^{t}_{r}(\%)$} & Unlearned & $94.48$ & $93.57$ & $94.30$ & $95.62$ & $94.38$ & $95.11$ & $93.89$ & $93.43$ & $94.23$ & $93.99$ \\
 &  & PRA \cite{ha2025unlearning} & $94.48$ & $93.57$ & $94.30$ & $95.61$ & $94.38$ & $95.11$ & $93.89$ & $93.43$ & $94.22$ & $93.99$ \\
 &  & SFRA (ours) & $92.87$ & $89.77$ & $89.68$ & $91.92$ & $90.74$ & $90.77$ & $92.20$ & $89.18$ & $92.70$ & $90.72$ \\
\cmidrule(lr){2-13}
 & \multirow{3}{*}{$\mathcal{A}^{t}_{f}(\%)$} & Unlearned & $0.00$ & $0.00$ & $0.00$ & $0.00$ & $0.00$ & $0.00$ & $0.00$ & $0.00$ & $0.00$ & $0.00$ \\
 &  & PRA \cite{ha2025unlearning} & $6.60$ & $0.00$ & $0.80$ & $1.10$ & $2.40$ & $0.00$ & $2.20$ & $0.00$ & $0.20$ & $0.60$ \\
 &  & SFRA (ours) & $47.10$ & $20.50$ & $45.30$ & $34.70$ & $44.80$ & $21.20$ & $40.60$ & $25.00$ & $26.10$ & $26.50$ \\
\cmidrule(lr){2-13}
 & \multirow{1}{*}{$\mathcal{A}^{LP}_{f}(\%)$} & Linear Probe & $92.70$ & $96.10$ & $85.20$ & $85.10$ & $92.50$ & $90.40$ & $95.00$ & $91.50$ & $93.00$ & $94.30$ \\
\cmidrule(lr){2-13}
 & \multirow{2}{*}{$\mathrm{RS}$} & PRA \cite{ha2025unlearning} & $0.12$ & $0.00$ & $0.02$ & $0.02$ & $0.05$ & $0.00$ & $0.04$ & $0.00$ & $0.00$ & $0.01$ \\
 &  & SFRA (ours) & $0.64$ & $0.34$ & $0.61$ & $0.51$ & $0.61$ & $0.35$ & $0.57$ & $0.40$ & $0.41$ & $0.42$ \\
\cmidrule(lr){2-13}
 & \multirow{2}{*}{$\Delta\mathrm{RS}$} & PRA \cite{ha2025unlearning} & $-0.13$ & $-0.06$ & $-0.29$ & $-0.14$ & $-0.21$ & $-0.11$ & $-0.21$ & $-0.25$ & $-0.07$ & $-0.13$ \\
 &  & SFRA (ours) & $+0.09$ & $+0.12$ & $-0.01$ & $-0.07$ & $-0.08$ & $+0.03$ & $+0.06$ & $-0.09$ & $-0.02$ & $+0.06$ \\
\midrule
\multirow{11}{*}{Bad Teacher \cite{chundawat2023can}} & \multirow{3}{*}{$\mathcal{A}^{t}_{r}(\%)$} & Unlearned & $94.96$ & $94.28$ & $80.30$ & $95.87$ & $84.33$ & $95.64$ & $94.50$ & $94.62$ & $94.37$ & $94.74$ \\
 &  & PRA \cite{ha2025unlearning} & $94.80$ & $94.16$ & $80.22$ & $95.76$ & $84.33$ & $95.37$ & $94.40$ & $94.56$ & $94.29$ & $94.66$ \\
 &  & SFRA (ours) & $90.24$ & $91.37$ & $76.10$ & $91.32$ & $78.94$ & $91.18$ & $90.81$ & $90.02$ & $91.50$ & $90.03$ \\
\cmidrule(lr){2-13}
 & \multirow{3}{*}{$\mathcal{A}^{t}_{f}(\%)$} & Unlearned & $0.00$ & $0.00$ & $10.30$ & $0.10$ & $0.80$ & $0.00$ & $0.10$ & $0.00$ & $0.00$ & $0.00$ \\
 &  & PRA \cite{ha2025unlearning} & $89.90$ & $95.80$ & $9.10$ & $71.40$ & $3.50$ & $80.30$ & $85.90$ & $87.20$ & $92.20$ & $93.30$ \\
 &  & SFRA (ours) & $99.70$ & $99.40$ & $76.90$ & $96.40$ & $93.90$ & $97.90$ & $99.30$ & $99.80$ & $99.60$ & $99.50$ \\
\cmidrule(lr){2-13}
 & \multirow{1}{*}{$\mathcal{A}^{LP}_{f}(\%)$} & Linear Probe & $96.30$ & $98.10$ & $73.30$ & $88.70$ & $87.30$ & $92.20$ & $97.50$ & $96.10$ & $98.10$ & $96.00$ \\
\cmidrule(lr){2-13}
 & \multirow{2}{*}{$\mathrm{RS}$} & PRA \cite{ha2025unlearning} & $0.95$ & $0.98$ & $0.00$ & $0.83$ & $0.05$ & $0.89$ & $0.92$ & $0.93$ & $0.96$ & $0.96$ \\
 &  & SFRA (ours) & $0.97$ & $0.98$ & $0.79$ & $0.96$ & $0.94$ & $0.97$ & $0.98$ & $0.98$ & $0.98$ & $0.97$ \\
\cmidrule(lr){2-13}
 & \multirow{2}{*}{$\Delta\mathrm{RS}$} & PRA \cite{ha2025unlearning} & $+0.69$ & $+0.92$ & $-0.30$ & $+0.67$ & $-0.20$ & $+0.78$ & $+0.67$ & $+0.68$ & $+0.89$ & $+0.82$ \\
 &  & SFRA (ours) & $+0.43$ & $+0.77$ & $+0.16$ & $+0.38$ & $+0.24$ & $+0.65$ & $+0.46$ & $+0.49$ & $+0.55$ & $+0.62$ \\
\midrule
\multirow{11}{*}{SalUn \cite{fan2023salun}} & \multirow{3}{*}{$\mathcal{A}^{t}_{r}(\%)$} & Unlearned & $94.11$ & $93.63$ & $94.78$ & $95.69$ & $93.78$ & $94.91$ & $93.81$ & $93.67$ & $93.66$ & $94.10$ \\
 &  & PRA \cite{ha2025unlearning} & $93.52$ & $93.38$ & $94.32$ & $94.82$ & $92.52$ & $93.98$ & $93.09$ & $92.86$ & $93.31$ & $93.53$ \\
 &  & SFRA (ours) & $89.44$ & $88.97$ & $90.06$ & $91.10$ & $89.11$ & $90.18$ & $89.18$ & $89.12$ & $88.99$ & $89.49$ \\
\cmidrule(lr){2-13}
 & \multirow{3}{*}{$\mathcal{A}^{t}_{f}(\%)$} & Unlearned & $7.80$ & $5.50$ & $5.80$ & $12.30$ & $14.50$ & $8.50$ & $6.70$ & $8.20$ & $8.60$ & $6.20$ \\
 &  & PRA \cite{ha2025unlearning} & $37.20$ & $24.00$ & $41.10$ & $48.20$ & $66.20$ & $41.40$ & $47.50$ & $55.50$ & $35.40$ & $47.00$ \\
 &  & SFRA (ours) & $81.80$ & $69.10$ & $82.60$ & $79.00$ & $89.40$ & $74.90$ & $82.80$ & $88.70$ & $79.80$ & $81.10$ \\
\cmidrule(lr){2-13}
 & \multirow{1}{*}{$\mathcal{A}^{LP}_{f}(\%)$} & Linear Probe & $91.50$ & $94.00$ & $89.50$ & $86.10$ & $93.00$ & $84.80$ & $93.60$ & $94.80$ & $91.60$ & $94.00$ \\
\cmidrule(lr){2-13}
 & \multirow{2}{*}{$\mathrm{RS}$} & PRA \cite{ha2025unlearning} & $0.45$ & $0.31$ & $0.52$ & $0.53$ & $0.68$ & $0.49$ & $0.58$ & $0.64$ & $0.42$ & $0.58$ \\
 &  & SFRA (ours) & $0.83$ & $0.76$ & $0.85$ & $0.79$ & $0.84$ & $0.78$ & $0.85$ & $0.87$ & $0.82$ & $0.84$ \\
\cmidrule(lr){2-13}
 & \multirow{2}{*}{$\Delta\mathrm{RS}$} & PRA \cite{ha2025unlearning} & $+0.20$ & $+0.26$ & $+0.22$ & $+0.36$ & $+0.42$ & $+0.38$ & $+0.33$ & $+0.39$ & $+0.35$ & $+0.43$ \\
 &  & SFRA (ours) & $+0.29$ & $+0.55$ & $+0.22$ & $+0.21$ & $+0.14$ & $+0.47$ & $+0.33$ & $+0.39$ & $+0.38$ & $+0.48$ \\
\midrule
\multirow{11}{*}{DELETE \cite{zhou2025decoupled}} & \multirow{3}{*}{$\mathcal{A}^{t}_{r}(\%)$} & Unlearned & $95.07$ & $94.49$ & $95.29$ & $96.29$ & $94.67$ & $95.60$ & $94.66$ & $94.59$ & $94.48$ & $94.61$ \\
 &  & PRA \cite{ha2025unlearning} & $93.99$ & $91.10$ & $94.24$ & $94.88$ & $93.29$ & $90.17$ & $91.82$ & $93.16$ & $91.39$ & $92.00$ \\
 &  & SFRA (ours) & $91.63$ & $90.63$ & $90.01$ & $92.64$ & $90.31$ & $89.90$ & $90.24$ & $91.29$ & $90.02$ & $91.56$ \\
\cmidrule(lr){2-13}
 & \multirow{3}{*}{$\mathcal{A}^{t}_{f}(\%)$} & Unlearned & $0.00$ & $0.00$ & $0.00$ & $0.00$ & $0.00$ & $0.00$ & $0.00$ & $0.00$ & $0.00$ & $0.00$ \\
 &  & PRA \cite{ha2025unlearning} & $75.20$ & $85.00$ & $67.40$ & $59.10$ & $71.70$ & $75.20$ & $86.60$ & $78.00$ & $81.40$ & $79.80$ \\
 &  & SFRA (ours) & $95.90$ & $97.70$ & $94.60$ & $89.00$ & $97.50$ & $91.10$ & $97.70$ & $95.30$ & $97.50$ & $95.60$ \\
\cmidrule(lr){2-13}
 & \multirow{1}{*}{$\mathcal{A}^{LP}_{f}(\%)$} & Linear Probe & $95.70$ & $96.90$ & $94.30$ & $93.10$ & $97.80$ & $91.60$ & $96.90$ & $96.30$ & $96.20$ & $97.40$ \\
\cmidrule(lr){2-13}
 & \multirow{2}{*}{$\mathrm{RS}$} & PRA \cite{ha2025unlearning} & $0.85$ & $0.90$ & $0.80$ & $0.74$ & $0.83$ & $0.84$ & $0.92$ & $0.87$ & $0.88$ & $0.88$ \\
 &  & SFRA (ours) & $0.96$ & $0.97$ & $0.95$ & $0.93$ & $0.97$ & $0.93$ & $0.97$ & $0.96$ & $0.97$ & $0.96$ \\
\cmidrule(lr){2-13}
 & \multirow{2}{*}{$\Delta\mathrm{RS}$} & PRA \cite{ha2025unlearning} & $+0.60$ & $+0.85$ & $+0.50$ & $+0.57$ & $+0.57$ & $+0.72$ & $+0.67$ & $+0.62$ & $+0.81$ & $+0.73$ \\
 &  & SFRA (ours) & $+0.42$ & $+0.75$ & $+0.32$ & $+0.35$ & $+0.27$ & $+0.61$ & $+0.45$ & $+0.47$ & $+0.53$ & $+0.61$ \\
\bottomrule
\end{tabular}
}
\end{minipage}%

\end{table*}

\begin{table*}[t]
\centering
\caption{Per-forget-class single-class unlearning and relearning results on CIFAR-100 using ResNet-18. We report the unlearned checkpoint, the source-dependent PRA baseline, our proposed SFRA, and frozen-encoder linear probing. Each forget class column corresponds to a separate unlearned checkpoint in which that class is designated for forgetting.}
\label{tab:cifar100_resnet18_variant_by_forget}
\begin{minipage}[t]{0.49\textwidth}
\centering
\scriptsize
\setlength{\tabcolsep}{2pt}
\renewcommand{\arraystretch}{0.80}
\resizebox{\columnwidth}{!}{%
\begin{tabular}{l|l|l|cccccccccc}
\toprule
\multirow{2}{*}{Unlearning Method} & \multirow{2}{*}{Metric} & \multirow{2}{*}{Variant} & \multicolumn{10}{c}{Forget Class} \\
 & & & 0 & 10 & 20 & 30 & 40 & 50 & 60 & 70 & 80 & 90 \\
\midrule
\multirow{2}{*}{Original} & \multirow{1}{*}{$\mathcal{A}^{t}_{r}(\%)$} & Original & $79.84$ & $80.13$ & $79.85$ & $80.05$ & $79.98$ & $80.04$ & $79.89$ & $79.98$ & $79.98$ & $79.90$ \\
\cmidrule(lr){2-13}
 & \multirow{1}{*}{$\mathcal{A}^{t}_{f}(\%)$} & Original & $92.00$ & $63.00$ & $91.00$ & $71.00$ & $78.00$ & $72.00$ & $87.00$ & $78.00$ & $78.00$ & $86.00$ \\
\midrule
\multirow{9}{*}{Retrained} & \multirow{3}{*}{$\mathcal{A}^{t}_{r}(\%)$} & Unlearned & $80.32$ & $80.44$ & $80.24$ & $80.17$ & $80.11$ & $80.18$ & $80.29$ & $79.58$ & $80.39$ & $80.13$ \\
 &  & PRA \cite{ha2025unlearning} & $79.56$ & $78.18$ & $78.51$ & $79.64$ & $77.39$ & $77.71$ & $79.64$ & $77.87$ & $77.82$ & $78.04$ \\
 &  & SFRA (ours) & $73.22$ & $72.56$ & $72.43$ & $72.29$ & $72.71$ & $72.57$ & $72.87$ & $71.72$ & $72.53$ & $72.82$ \\
\cmidrule(lr){2-13}
 & \multirow{3}{*}{$\mathcal{A}^{t}_{f}(\%)$} & Unlearned & $0.00$ & $0.00$ & $0.00$ & $0.00$ & $0.00$ & $0.00$ & $0.00$ & $0.00$ & $0.00$ & $0.00$ \\
 &  & PRA \cite{ha2025unlearning} & $27.00$ & $42.00$ & $37.00$ & $23.00$ & $69.00$ & $53.00$ & $23.00$ & $21.00$ & $50.00$ & $51.00$ \\
 &  & SFRA (ours) & $26.00$ & $47.00$ & $48.00$ & $28.00$ & $80.00$ & $52.00$ & $39.00$ & $27.00$ & $63.00$ & $57.00$ \\
\cmidrule(lr){2-13}
 & \multirow{1}{*}{$\mathcal{A}^{LP}_{f}(\%)$} & Linear Probe & $92.00$ & $71.00$ & $83.00$ & $76.00$ & $84.00$ & $82.00$ & $85.00$ & $79.00$ & $74.00$ & $85.00$ \\
\cmidrule(lr){2-13}
 & \multirow{2}{*}{$\mathrm{RS}$} & PRA \cite{ha2025unlearning} & $0.42$ & $0.59$ & $0.54$ & $0.37$ & $0.81$ & $0.69$ & $0.37$ & $0.35$ & $0.66$ & $0.67$ \\
 &  & SFRA (ours) & $0.41$ & $0.62$ & $0.63$ & $0.43$ & $0.86$ & $0.67$ & $0.55$ & $0.42$ & $0.75$ & $0.71$ \\
\midrule
\multirow{11}{*}{Finetune \cite{golatkar2020eternal}} & \multirow{3}{*}{$\mathcal{A}^{t}_{r}(\%)$} & Unlearned & $79.42$ & $77.97$ & $79.76$ & $79.31$ & $79.33$ & $80.13$ & $78.69$ & $79.53$ & $80.26$ & $76.91$ \\
 &  & PRA \cite{ha2025unlearning} & $79.32$ & $76.97$ & $78.94$ & $79.11$ & $78.31$ & $78.98$ & $78.26$ & $79.01$ & $78.89$ & $75.77$ \\
 &  & SFRA (ours) & $78.98$ & $73.53$ & $79.01$ & $76.57$ & $75.36$ & $76.68$ & $77.10$ & $78.78$ & $75.77$ & $75.29$ \\
\cmidrule(lr){2-13}
 & \multirow{3}{*}{$\mathcal{A}^{t}_{f}(\%)$} & Unlearned & $0.00$ & $0.00$ & $1.00$ & $0.00$ & $0.00$ & $0.00$ & $0.00$ & $0.00$ & $0.00$ & $0.00$ \\
 &  & PRA \cite{ha2025unlearning} & $11.00$ & $27.00$ & $23.00$ & $10.00$ & $49.00$ & $31.00$ & $29.00$ & $10.00$ & $37.00$ & $36.00$ \\
 &  & SFRA (ours) & $10.00$ & $37.00$ & $27.00$ & $12.00$ & $53.00$ & $35.00$ & $67.00$ & $10.00$ & $38.00$ & $46.00$ \\
\cmidrule(lr){2-13}
 & \multirow{1}{*}{$\mathcal{A}^{LP}_{f}(\%)$} & Linear Probe & $90.00$ & $75.00$ & $89.00$ & $77.00$ & $82.00$ & $81.00$ & $87.00$ & $81.00$ & $85.00$ & $87.00$ \\
\cmidrule(lr){2-13}
 & \multirow{2}{*}{$\mathrm{RS}$} & PRA \cite{ha2025unlearning} & $0.20$ & $0.42$ & $0.36$ & $0.18$ & $0.66$ & $0.47$ & $0.45$ & $0.18$ & $0.54$ & $0.53$ \\
 &  & SFRA (ours) & $0.18$ & $0.53$ & $0.41$ & $0.21$ & $0.68$ & $0.51$ & $0.80$ & $0.18$ & $0.54$ & $0.63$ \\
\cmidrule(lr){2-13}
 & \multirow{2}{*}{$\Delta\mathrm{RS}$} & PRA \cite{ha2025unlearning} & $-0.23$ & $-0.16$ & $-0.18$ & $-0.19$ & $-0.15$ & $-0.21$ & $+0.08$ & $-0.16$ & $-0.12$ & $-0.14$ \\
 &  & SFRA (ours) & $-0.22$ & $-0.09$ & $-0.22$ & $-0.22$ & $-0.18$ & $-0.15$ & $+0.25$ & $-0.24$ & $-0.20$ & $-0.08$ \\
\midrule
\multirow{11}{*}{Negative Gradient \cite{golatkar2020eternal}} & \multirow{3}{*}{$\mathcal{A}^{t}_{r}(\%)$} & Unlearned & $72.81$ & $71.29$ & $67.66$ & $73.62$ & $73.62$ & $74.56$ & $70.32$ & $72.16$ & $75.91$ & $74.35$ \\
 &  & PRA \cite{ha2025unlearning} & $72.47$ & $70.71$ & $67.02$ & $73.47$ & $72.90$ & $72.59$ & $70.04$ & $71.48$ & $74.27$ & $72.98$ \\
 &  & SFRA (ours) & $68.72$ & $67.41$ & $61.93$ & $66.92$ & $67.73$ & $69.10$ & $63.59$ & $67.34$ & $68.69$ & $69.93$ \\
\cmidrule(lr){2-13}
 & \multirow{3}{*}{$\mathcal{A}^{t}_{f}(\%)$} & Unlearned & $0.00$ & $0.00$ & $8.00$ & $0.00$ & $1.00$ & $0.00$ & $3.00$ & $2.00$ & $0.00$ & $0.00$ \\
 &  & PRA \cite{ha2025unlearning} & $32.00$ & $33.00$ & $49.00$ & $22.00$ & $43.00$ & $65.00$ & $35.00$ & $27.00$ & $50.00$ & $57.00$ \\
 &  & SFRA (ours) & $60.00$ & $53.00$ & $70.00$ & $57.00$ & $80.00$ & $82.00$ & $52.00$ & $51.00$ & $71.00$ & $73.00$ \\
\cmidrule(lr){2-13}
 & \multirow{1}{*}{$\mathcal{A}^{LP}_{f}(\%)$} & Linear Probe & $88.00$ & $54.00$ & $85.00$ & $64.00$ & $73.00$ & $60.00$ & $82.00$ & $67.00$ & $68.00$ & $83.00$ \\
\cmidrule(lr){2-13}
 & \multirow{2}{*}{$\mathrm{RS}$} & PRA \cite{ha2025unlearning} & $0.48$ & $0.50$ & $0.58$ & $0.36$ & $0.59$ & $0.78$ & $0.48$ & $0.40$ & $0.66$ & $0.72$ \\
 &  & SFRA (ours) & $0.74$ & $0.68$ & $0.75$ & $0.71$ & $0.86$ & $0.88$ & $0.64$ & $0.65$ & $0.80$ & $0.83$ \\
\cmidrule(lr){2-13}
 & \multirow{2}{*}{$\Delta\mathrm{RS}$} & PRA \cite{ha2025unlearning} & $+0.06$ & $-0.09$ & $+0.04$ & $-0.01$ & $-0.22$ & $+0.09$ & $+0.11$ & $+0.05$ & $+0.00$ & $+0.05$ \\
 &  & SFRA (ours) & $+0.33$ & $+0.06$ & $+0.12$ & $+0.28$ & $+0.00$ & $+0.21$ & $+0.09$ & $+0.23$ & $+0.06$ & $+0.12$ \\
\midrule
\multirow{11}{*}{Negative Gradient+ \cite{kurmanji2023towards}} & \multirow{3}{*}{$\mathcal{A}^{t}_{r}(\%)$} & Unlearned & $73.77$ & $77.64$ & $68.43$ & $75.14$ & $75.69$ & $77.07$ & $68.99$ & $73.45$ & $77.11$ & $76.04$ \\
 &  & PRA \cite{ha2025unlearning} & $73.24$ & $76.64$ & $67.41$ & $74.82$ & $74.60$ & $74.96$ & $68.52$ & $72.13$ & $75.64$ & $74.63$ \\
 &  & SFRA (ours) & $67.29$ & $72.44$ & $61.87$ & $68.16$ & $68.67$ & $70.34$ & $63.21$ & $66.27$ & $69.52$ & $69.47$ \\
\cmidrule(lr){2-13}
 & \multirow{3}{*}{$\mathcal{A}^{t}_{f}(\%)$} & Unlearned & $0.00$ & $0.00$ & $0.00$ & $0.00$ & $0.00$ & $0.00$ & $0.00$ & $2.00$ & $0.00$ & $0.00$ \\
 &  & PRA \cite{ha2025unlearning} & $46.00$ & $53.00$ & $41.00$ & $44.00$ & $52.00$ & $72.00$ & $42.00$ & $46.00$ & $61.00$ & $64.00$ \\
 &  & SFRA (ours) & $76.00$ & $74.00$ & $60.00$ & $52.00$ & $78.00$ & $80.00$ & $56.00$ & $64.00$ & $76.00$ & $77.00$ \\
\cmidrule(lr){2-13}
 & \multirow{1}{*}{$\mathcal{A}^{LP}_{f}(\%)$} & Linear Probe & $91.00$ & $59.00$ & $84.00$ & $63.00$ & $75.00$ & $64.00$ & $86.00$ & $71.00$ & $70.00$ & $83.00$ \\
\cmidrule(lr){2-13}
 & \multirow{2}{*}{$\mathrm{RS}$} & PRA \cite{ha2025unlearning} & $0.63$ & $0.69$ & $0.58$ & $0.61$ & $0.68$ & $0.83$ & $0.59$ & $0.61$ & $0.75$ & $0.78$ \\
 &  & SFRA (ours) & $0.84$ & $0.83$ & $0.73$ & $0.67$ & $0.85$ & $0.86$ & $0.70$ & $0.74$ & $0.83$ & $0.84$ \\
\cmidrule(lr){2-13}
 & \multirow{2}{*}{$\Delta\mathrm{RS}$} & PRA \cite{ha2025unlearning} & $+0.20$ & $+0.10$ & $+0.04$ & $+0.24$ & $-0.13$ & $+0.14$ & $+0.22$ & $+0.26$ & $+0.09$ & $+0.11$ \\
 &  & SFRA (ours) & $+0.43$ & $+0.21$ & $+0.10$ & $+0.24$ & $-0.01$ & $+0.20$ & $+0.15$ & $+0.33$ & $+0.09$ & $+0.14$ \\
\midrule
\multirow{11}{*}{Random Label \cite{hayase2020selective}} & \multirow{3}{*}{$\mathcal{A}^{t}_{r}(\%)$} & Unlearned & $66.87$ & $77.12$ & $63.51$ & $71.07$ & $71.94$ & $74.52$ & $57.80$ & $67.73$ & $72.58$ & $71.40$ \\
 &  & PRA \cite{ha2025unlearning} & $66.27$ & $75.44$ & $63.28$ & $70.33$ & $71.01$ & $72.79$ & $57.17$ & $67.39$ & $70.80$ & $69.94$ \\
 &  & SFRA (ours) & $60.99$ & $72.43$ & $57.51$ & $64.04$ & $65.94$ & $68.72$ & $53.18$ & $61.14$ & $65.58$ & $64.33$ \\
\cmidrule(lr){2-13}
 & \multirow{3}{*}{$\mathcal{A}^{t}_{f}(\%)$} & Unlearned & $2.00$ & $2.00$ & $5.00$ & $2.00$ & $4.00$ & $0.00$ & $4.00$ & $4.00$ & $0.00$ & $4.00$ \\
 &  & PRA \cite{ha2025unlearning} & $62.00$ & $63.00$ & $47.00$ & $59.00$ & $48.00$ & $73.00$ & $53.00$ & $39.00$ & $61.00$ & $77.00$ \\
 &  & SFRA (ours) & $92.00$ & $81.00$ & $86.00$ & $74.00$ & $88.00$ & $90.00$ & $54.00$ & $84.00$ & $79.00$ & $91.00$ \\
\cmidrule(lr){2-13}
 & \multirow{1}{*}{$\mathcal{A}^{LP}_{f}(\%)$} & Linear Probe & $82.00$ & $61.00$ & $80.00$ & $64.00$ & $73.00$ & $68.00$ & $81.00$ & $63.00$ & $74.00$ & $85.00$ \\
\cmidrule(lr){2-13}
 & \multirow{2}{*}{$\mathrm{RS}$} & PRA \cite{ha2025unlearning} & $0.75$ & $0.75$ & $0.59$ & $0.72$ & $0.61$ & $0.84$ & $0.66$ & $0.52$ & $0.75$ & $0.84$ \\
 &  & SFRA (ours) & $0.92$ & $0.86$ & $0.87$ & $0.81$ & $0.89$ & $0.92$ & $0.66$ & $0.86$ & $0.85$ & $0.90$ \\
\cmidrule(lr){2-13}
 & \multirow{2}{*}{$\Delta\mathrm{RS}$} & PRA \cite{ha2025unlearning} & $+0.32$ & $+0.17$ & $+0.05$ & $+0.35$ & $-0.20$ & $+0.15$ & $+0.28$ & $+0.17$ & $+0.09$ & $+0.17$ \\
 &  & SFRA (ours) & $+0.51$ & $+0.24$ & $+0.24$ & $+0.38$ & $+0.03$ & $+0.26$ & $+0.11$ & $+0.44$ & $+0.11$ & $+0.19$ \\
\bottomrule
\end{tabular}
}
\end{minipage}%
\hfill
\begin{minipage}[t]{0.49\textwidth}
\centering
\scriptsize
\setlength{\tabcolsep}{2pt}
\renewcommand{\arraystretch}{0.80}
\resizebox{\columnwidth}{!}{%
\begin{tabular}{l|l|l|cccccccccc}
\toprule
\multirow{2}{*}{Unlearning Method} & \multirow{2}{*}{Metric} & \multirow{2}{*}{Variant} & \multicolumn{10}{c}{Forget Class} \\
 & & & 0 & 10 & 20 & 30 & 40 & 50 & 60 & 70 & 80 & 90 \\
\midrule
\multirow{11}{*}{Boundary Shrink \cite{chen2023boundary}} & \multirow{3}{*}{$\mathcal{A}^{t}_{r}(\%)$} & Unlearned & $66.99$ & $77.12$ & $63.59$ & $69.59$ & $71.86$ & $74.37$ & $57.89$ & $67.71$ & $72.62$ & $71.36$ \\
 &  & PRA \cite{ha2025unlearning} & $66.34$ & $75.49$ & $63.37$ & $68.72$ & $70.89$ & $72.63$ & $57.25$ & $67.38$ & $70.81$ & $69.90$ \\
 &  & SFRA (ours) & $61.45$ & $72.63$ & $58.75$ & $63.09$ & $65.94$ & $68.68$ & $53.22$ & $61.01$ & $65.43$ & $64.63$ \\
\cmidrule(lr){2-13}
 & \multirow{3}{*}{$\mathcal{A}^{t}_{f}(\%)$} & Unlearned & $2.00$ & $2.00$ & $5.00$ & $0.00$ & $4.00$ & $0.00$ & $5.00$ & $4.00$ & $0.00$ & $4.00$ \\
 &  & PRA \cite{ha2025unlearning} & $62.00$ & $64.00$ & $47.00$ & $59.00$ & $48.00$ & $73.00$ & $53.00$ & $39.00$ & $61.00$ & $76.00$ \\
 &  & SFRA (ours) & $91.00$ & $79.00$ & $86.00$ & $76.00$ & $87.00$ & $90.00$ & $55.00$ & $88.00$ & $79.00$ & $89.00$ \\
\cmidrule(lr){2-13}
 & \multirow{1}{*}{$\mathcal{A}^{LP}_{f}(\%)$} & Linear Probe & $82.00$ & $61.00$ & $81.00$ & $64.00$ & $71.00$ & $66.00$ & $81.00$ & $62.00$ & $74.00$ & $86.00$ \\
\cmidrule(lr){2-13}
 & \multirow{2}{*}{$\mathrm{RS}$} & PRA \cite{ha2025unlearning} & $0.75$ & $0.76$ & $0.59$ & $0.74$ & $0.61$ & $0.84$ & $0.65$ & $0.52$ & $0.75$ & $0.83$ \\
 &  & SFRA (ours) & $0.92$ & $0.85$ & $0.88$ & $0.84$ & $0.88$ & $0.92$ & $0.66$ & $0.88$ & $0.85$ & $0.89$ \\
\cmidrule(lr){2-13}
 & \multirow{2}{*}{$\Delta\mathrm{RS}$} & PRA \cite{ha2025unlearning} & $+0.32$ & $+0.17$ & $+0.05$ & $+0.37$ & $-0.20$ & $+0.15$ & $+0.27$ & $+0.17$ & $+0.09$ & $+0.16$ \\
 &  & SFRA (ours) & $+0.51$ & $+0.23$ & $+0.24$ & $+0.41$ & $+0.02$ & $+0.26$ & $+0.11$ & $+0.47$ & $+0.11$ & $+0.18$ \\
\midrule
\multirow{11}{*}{Learn to Unlearn \cite{cha2024learning}} & \multirow{3}{*}{$\mathcal{A}^{t}_{r}(\%)$} & Unlearned & $73.37$ & $75.50$ & $72.76$ & $72.32$ & $75.38$ & $74.27$ & $71.21$ & $72.35$ & $75.00$ & $72.36$ \\
 &  & PRA \cite{ha2025unlearning} & $71.84$ & $74.18$ & $71.26$ & $71.14$ & $72.73$ & $73.18$ & $70.00$ & $71.54$ & $74.19$ & $71.35$ \\
 &  & SFRA (ours) & $68.40$ & $68.58$ & $66.12$ & $67.14$ & $68.10$ & $67.15$ & $64.26$ & $65.91$ & $67.78$ & $66.23$ \\
\cmidrule(lr){2-13}
 & \multirow{3}{*}{$\mathcal{A}^{t}_{f}(\%)$} & Unlearned & $0.00$ & $0.00$ & $0.00$ & $0.00$ & $3.00$ & $0.00$ & $0.00$ & $0.00$ & $0.00$ & $0.00$ \\
 &  & PRA \cite{ha2025unlearning} & $73.00$ & $61.00$ & $73.00$ & $59.00$ & $73.00$ & $65.00$ & $63.00$ & $56.00$ & $52.00$ & $65.00$ \\
 &  & SFRA (ours) & $94.00$ & $82.00$ & $92.00$ & $41.00$ & $90.00$ & $68.00$ & $83.00$ & $76.00$ & $60.00$ & $91.00$ \\
\cmidrule(lr){2-13}
 & \multirow{1}{*}{$\mathcal{A}^{LP}_{f}(\%)$} & Linear Probe & $85.00$ & $63.00$ & $82.00$ & $68.00$ & $73.00$ & $65.00$ & $79.00$ & $66.00$ & $63.00$ & $81.00$ \\
\cmidrule(lr){2-13}
 & \multirow{2}{*}{$\mathrm{RS}$} & PRA \cite{ha2025unlearning} & $0.84$ & $0.75$ & $0.84$ & $0.74$ & $0.81$ & $0.78$ & $0.77$ & $0.72$ & $0.68$ & $0.78$ \\
 &  & SFRA (ours) & $0.95$ & $0.87$ & $0.93$ & $0.57$ & $0.90$ & $0.79$ & $0.88$ & $0.84$ & $0.73$ & $0.92$ \\
\cmidrule(lr){2-13}
 & \multirow{2}{*}{$\Delta\mathrm{RS}$} & PRA \cite{ha2025unlearning} & $+0.41$ & $+0.17$ & $+0.30$ & $+0.37$ & $+0.01$ & $+0.10$ & $+0.40$ & $+0.37$ & $+0.02$ & $+0.11$ \\
 &  & SFRA (ours) & $+0.54$ & $+0.25$ & $+0.30$ & $+0.14$ & $+0.04$ & $+0.12$ & $+0.33$ & $+0.42$ & $-0.02$ & $+0.22$ \\
\midrule
\multirow{11}{*}{SCRUB \cite{kurmanji2023towards}} & \multirow{3}{*}{$\mathcal{A}^{t}_{r}(\%)$} & Unlearned & $69.26$ & $76.78$ & $62.87$ & $74.27$ & $72.37$ & $76.57$ & $62.29$ & $69.55$ & $74.36$ & $73.01$ \\
 &  & PRA \cite{ha2025unlearning} & $68.27$ & $75.60$ & $62.56$ & $73.80$ & $71.11$ & $74.00$ & $61.74$ & $69.08$ & $73.57$ & $72.34$ \\
 &  & SFRA (ours) & $64.22$ & $72.31$ & $57.67$ & $66.91$ & $65.95$ & $69.80$ & $56.32$ & $63.43$ & $68.17$ & $66.44$ \\
\cmidrule(lr){2-13}
 & \multirow{3}{*}{$\mathcal{A}^{t}_{f}(\%)$} & Unlearned & $0.00$ & $0.00$ & $2.00$ & $3.00$ & $2.00$ & $1.00$ & $0.00$ & $1.00$ & $0.00$ & $1.00$ \\
 &  & PRA \cite{ha2025unlearning} & $53.00$ & $52.00$ & $32.00$ & $50.00$ & $52.00$ & $73.00$ & $35.00$ & $31.00$ & $40.00$ & $54.00$ \\
 &  & SFRA (ours) & $75.00$ & $76.00$ & $76.00$ & $68.00$ & $81.00$ & $87.00$ & $57.00$ & $67.00$ & $79.00$ & $84.00$ \\
\cmidrule(lr){2-13}
 & \multirow{1}{*}{$\mathcal{A}^{LP}_{f}(\%)$} & Linear Probe & $85.00$ & $64.00$ & $82.00$ & $66.00$ & $71.00$ & $67.00$ & $80.00$ & $61.00$ & $73.00$ & $85.00$ \\
\cmidrule(lr){2-13}
 & \multirow{2}{*}{$\mathrm{RS}$} & PRA \cite{ha2025unlearning} & $0.69$ & $0.68$ & $0.46$ & $0.64$ & $0.66$ & $0.83$ & $0.52$ & $0.46$ & $0.57$ & $0.69$ \\
 &  & SFRA (ours) & $0.84$ & $0.85$ & $0.83$ & $0.76$ & $0.86$ & $0.89$ & $0.71$ & $0.78$ & $0.86$ & $0.88$ \\
\cmidrule(lr){2-13}
 & \multirow{2}{*}{$\Delta\mathrm{RS}$} & PRA \cite{ha2025unlearning} & $+0.27$ & $+0.09$ & $-0.08$ & $+0.26$ & $-0.14$ & $+0.14$ & $+0.14$ & $+0.11$ & $-0.09$ & $+0.02$ \\
 &  & SFRA (ours) & $+0.43$ & $+0.22$ & $+0.20$ & $+0.33$ & $-0.00$ & $+0.23$ & $+0.16$ & $+0.36$ & $+0.11$ & $+0.17$ \\
\midrule
\multirow{11}{*}{Bad Teacher \cite{chundawat2023can}} & \multirow{3}{*}{$\mathcal{A}^{t}_{r}(\%)$} & Unlearned & $79.80$ & $79.93$ & $79.84$ & $79.95$ & $79.90$ & $79.73$ & $79.92$ & $79.59$ & $79.71$ & $79.67$ \\
 &  & PRA \cite{ha2025unlearning} & $79.57$ & $76.41$ & $76.23$ & $77.57$ & $79.56$ & $79.11$ & $79.33$ & $75.87$ & $77.26$ & $76.41$ \\
 &  & SFRA (ours) & $76.98$ & $73.03$ & $74.14$ & $75.32$ & $72.20$ & $73.09$ & $74.91$ & $71.90$ & $71.99$ & $73.80$ \\
\cmidrule(lr){2-13}
 & \multirow{3}{*}{$\mathcal{A}^{t}_{f}(\%)$} & Unlearned & $0.00$ & $0.00$ & $0.00$ & $0.00$ & $0.00$ & $0.00$ & $0.00$ & $0.00$ & $0.00$ & $0.00$ \\
 &  & PRA \cite{ha2025unlearning} & $100.00$ & $75.00$ & $95.00$ & $88.00$ & $89.00$ & $81.00$ & $93.00$ & $85.00$ & $83.00$ & $94.00$ \\
 &  & SFRA (ours) & $100.00$ & $93.00$ & $97.00$ & $94.00$ & $100.00$ & $98.00$ & $98.00$ & $97.00$ & $98.00$ & $96.00$ \\
\cmidrule(lr){2-13}
 & \multirow{1}{*}{$\mathcal{A}^{LP}_{f}(\%)$} & Linear Probe & $95.00$ & $58.00$ & $87.00$ & $73.00$ & $72.00$ & $68.00$ & $94.00$ & $69.00$ & $61.00$ & $75.00$ \\
\cmidrule(lr){2-13}
 & \multirow{2}{*}{$\mathrm{RS}$} & PRA \cite{ha2025unlearning} & $1.00$ & $0.84$ & $0.96$ & $0.93$ & $0.94$ & $0.89$ & $0.96$ & $0.90$ & $0.90$ & $0.95$ \\
 &  & SFRA (ours) & $0.99$ & $0.93$ & $0.96$ & $0.95$ & $0.96$ & $0.96$ & $0.96$ & $0.95$ & $0.95$ & $0.95$ \\
\cmidrule(lr){2-13}
 & \multirow{2}{*}{$\Delta\mathrm{RS}$} & PRA \cite{ha2025unlearning} & $+0.57$ & $+0.26$ & $+0.42$ & $+0.55$ & $+0.13$ & $+0.21$ & $+0.59$ & $+0.56$ & $+0.24$ & $+0.28$ \\
 &  & SFRA (ours) & $+0.58$ & $+0.31$ & $+0.33$ & $+0.52$ & $+0.10$ & $+0.29$ & $+0.42$ & $+0.53$ & $+0.20$ & $+0.24$ \\
\midrule
\multirow{11}{*}{SalUn \cite{fan2023salun}} & \multirow{3}{*}{$\mathcal{A}^{t}_{r}(\%)$} & Unlearned & $76.37$ & $77.75$ & $77.48$ & $77.47$ & $77.56$ & $77.73$ & $76.50$ & $77.35$ & $76.50$ & $76.96$ \\
 &  & PRA \cite{ha2025unlearning} & $73.93$ & $74.18$ & $74.13$ & $71.77$ & $71.95$ & $74.20$ & $74.72$ & $72.24$ & $73.98$ & $73.17$ \\
 &  & SFRA (ours) & $71.43$ & $73.75$ & $71.00$ & $71.84$ & $70.01$ & $72.62$ & $69.55$ & $72.82$ & $71.19$ & $71.06$ \\
\cmidrule(lr){2-13}
 & \multirow{3}{*}{$\mathcal{A}^{t}_{f}(\%)$} & Unlearned & $0.00$ & $0.00$ & $0.00$ & $0.00$ & $0.00$ & $0.00$ & $0.00$ & $0.00$ & $0.00$ & $0.00$ \\
 &  & PRA \cite{ha2025unlearning} & $15.00$ & $37.00$ & $40.00$ & $30.00$ & $72.00$ & $53.00$ & $19.00$ & $30.00$ & $59.00$ & $56.00$ \\
 &  & SFRA (ours) & $18.00$ & $31.00$ & $53.00$ & $32.00$ & $46.00$ & $20.00$ & $47.00$ & $19.00$ & $51.00$ & $40.00$ \\
\cmidrule(lr){2-13}
 & \multirow{1}{*}{$\mathcal{A}^{LP}_{f}(\%)$} & Linear Probe & $94.00$ & $69.00$ & $89.00$ & $70.00$ & $83.00$ & $80.00$ & $89.00$ & $82.00$ & $74.00$ & $93.00$ \\
\cmidrule(lr){2-13}
 & \multirow{2}{*}{$\mathrm{RS}$} & PRA \cite{ha2025unlearning} & $0.26$ & $0.53$ & $0.57$ & $0.46$ & $0.82$ & $0.68$ & $0.32$ & $0.46$ & $0.74$ & $0.71$ \\
 &  & SFRA (ours) & $0.30$ & $0.47$ & $0.68$ & $0.48$ & $0.61$ & $0.33$ & $0.62$ & $0.32$ & $0.66$ & $0.56$ \\
\cmidrule(lr){2-13}
 & \multirow{2}{*}{$\Delta\mathrm{RS}$} & PRA \cite{ha2025unlearning} & $-0.16$ & $-0.05$ & $+0.03$ & $+0.08$ & $+0.01$ & $-0.00$ & $-0.06$ & $+0.11$ & $+0.07$ & $+0.04$ \\
 &  & SFRA (ours) & $-0.10$ & $-0.15$ & $+0.05$ & $+0.05$ & $-0.24$ & $-0.34$ & $+0.08$ & $-0.10$ & $-0.09$ & $-0.14$ \\
\midrule
\multirow{11}{*}{DELETE \cite{zhou2025decoupled}} & \multirow{3}{*}{$\mathcal{A}^{t}_{r}(\%)$} & Unlearned & $75.82$ & $78.35$ & $76.50$ & $77.25$ & $79.02$ & $78.86$ & $73.14$ & $77.04$ & $78.80$ & $78.02$ \\
 &  & PRA \cite{ha2025unlearning} & $72.67$ & $75.82$ & $74.80$ & $74.01$ & $70.68$ & $75.15$ & $73.08$ & $73.90$ & $71.38$ & $73.81$ \\
 &  & SFRA (ours) & $68.33$ & $71.91$ & $69.03$ & $70.17$ & $71.18$ & $71.21$ & $67.80$ & $69.48$ & $71.09$ & $70.33$ \\
\cmidrule(lr){2-13}
 & \multirow{3}{*}{$\mathcal{A}^{t}_{f}(\%)$} & Unlearned & $0.00$ & $0.00$ & $0.00$ & $0.00$ & $0.00$ & $0.00$ & $0.00$ & $0.00$ & $0.00$ & $0.00$ \\
 &  & PRA \cite{ha2025unlearning} & $66.00$ & $37.00$ & $81.00$ & $34.00$ & $85.00$ & $48.00$ & $19.00$ & $50.00$ & $78.00$ & $52.00$ \\
 &  & SFRA (ours) & $67.00$ & $48.00$ & $59.00$ & $54.00$ & $86.00$ & $62.00$ & $44.00$ & $33.00$ & $79.00$ & $82.00$ \\
\cmidrule(lr){2-13}
 & \multirow{1}{*}{$\mathcal{A}^{LP}_{f}(\%)$} & Linear Probe & $95.00$ & $65.00$ & $92.00$ & $75.00$ & $80.00$ & $70.00$ & $89.00$ & $76.00$ & $81.00$ & $87.00$ \\
\cmidrule(lr){2-13}
 & \multirow{2}{*}{$\mathrm{RS}$} & PRA \cite{ha2025unlearning} & $0.79$ & $0.54$ & $0.89$ & $0.50$ & $0.88$ & $0.64$ & $0.32$ & $0.66$ & $0.85$ & $0.67$ \\
 &  & SFRA (ours) & $0.78$ & $0.63$ & $0.72$ & $0.68$ & $0.89$ & $0.74$ & $0.60$ & $0.49$ & $0.85$ & $0.87$ \\
\cmidrule(lr){2-13}
 & \multirow{2}{*}{$\Delta\mathrm{RS}$} & PRA \cite{ha2025unlearning} & $+0.36$ & $-0.05$ & $+0.35$ & $+0.13$ & $+0.07$ & $-0.05$ & $-0.05$ & $+0.31$ & $+0.19$ & $+0.00$ \\
 &  & SFRA (ours) & $+0.37$ & $+0.01$ & $+0.09$ & $+0.25$ & $+0.03$ & $+0.08$ & $+0.05$ & $+0.07$ & $+0.10$ & $+0.16$ \\
\bottomrule
\end{tabular}
}
\end{minipage}%

\end{table*}

\begin{table*}[t]
\centering
\caption{Per-forget-class single-class unlearning and relearning results on TinyImageNet using ResNet-18. We report the unlearned checkpoint, the source-dependent PRA baseline, our proposed SFRA, and frozen-encoder linear probing. Each forget class column corresponds to a separate unlearned checkpoint in which that class is designated for forgetting.}
\label{tab:tiny_imagenet_resnet18_variant_by_forget}
\begin{minipage}[t]{0.49\textwidth}
\centering
\scriptsize
\setlength{\tabcolsep}{2pt}
\renewcommand{\arraystretch}{0.80}
\resizebox{\columnwidth}{!}{%
\begin{tabular}{l|l|l|cccccccccc}
\toprule
\multirow{2}{*}{Unlearning Method} & \multirow{2}{*}{Metric} & \multirow{2}{*}{Variant} & \multicolumn{10}{c}{Forget Class} \\
 & & & 0 & 20 & 40 & 60 & 80 & 100 & 120 & 140 & 160 & 180 \\
\midrule
\multirow{2}{*}{Original} & \multirow{1}{*}{$\mathcal{A}^{t}_{r}(\%)$} & Original & $71.31$ & $71.31$ & $71.47$ & $71.36$ & $71.47$ & $71.48$ & $71.46$ & $71.37$ & $71.35$ & $71.41$ \\
\cmidrule(lr){2-13}
 & \multirow{1}{*}{$\mathcal{A}^{t}_{f}(\%)$} & Original & $88.00$ & $88.00$ & $56.00$ & $78.00$ & $56.00$ & $54.00$ & $58.00$ & $76.00$ & $80.00$ & $68.00$ \\
\midrule
\multirow{9}{*}{Retrained} & \multirow{3}{*}{$\mathcal{A}^{t}_{r}(\%)$} & Unlearned & $70.59$ & $71.28$ & $71.16$ & $70.73$ & $69.93$ & $71.50$ & $71.26$ & $70.17$ & $70.43$ & $71.21$ \\
 &  & PRA \cite{ha2025unlearning} & $71.35$ & $71.90$ & $71.61$ & $71.32$ & $70.65$ & $72.05$ & $70.00$ & $70.71$ & $71.03$ & $71.98$ \\
 &  & SFRA (ours) & $64.88$ & $65.26$ & $65.02$ & $65.06$ & $63.66$ & $64.89$ & $64.76$ & $64.78$ & $64.58$ & $65.37$ \\
\cmidrule(lr){2-13}
 & \multirow{3}{*}{$\mathcal{A}^{t}_{f}(\%)$} & Unlearned & $0.00$ & $0.00$ & $0.00$ & $0.00$ & $0.00$ & $0.00$ & $0.00$ & $0.00$ & $0.00$ & $0.00$ \\
 &  & PRA \cite{ha2025unlearning} & $0.00$ & $0.00$ & $0.00$ & $0.00$ & $0.00$ & $0.00$ & $42.00$ & $0.00$ & $0.00$ & $0.00$ \\
 &  & SFRA (ours) & $50.00$ & $30.00$ & $48.00$ & $24.00$ & $40.00$ & $40.00$ & $50.00$ & $22.00$ & $44.00$ & $46.00$ \\
\cmidrule(lr){2-13}
 & \multirow{1}{*}{$\mathcal{A}^{LP}_{f}(\%)$} & Linear Probe & $88.00$ & $78.00$ & $52.00$ & $72.00$ & $34.00$ & $46.00$ & $50.00$ & $70.00$ & $72.00$ & $56.00$ \\
\cmidrule(lr){2-13}
 & \multirow{2}{*}{$\mathrm{RS}$} & PRA \cite{ha2025unlearning} & $0.00$ & $0.00$ & $0.00$ & $0.00$ & $0.00$ & $0.00$ & $0.59$ & $0.00$ & $0.00$ & $0.00$ \\
 &  & SFRA (ours) & $0.65$ & $0.45$ & $0.64$ & $0.38$ & $0.56$ & $0.56$ & $0.65$ & $0.36$ & $0.60$ & $0.62$ \\
\midrule
\multirow{11}{*}{Finetune \cite{golatkar2020eternal}} & \multirow{3}{*}{$\mathcal{A}^{t}_{r}(\%)$} & Unlearned & $67.16$ & $65.42$ & $67.38$ & $65.44$ & $67.93$ & $66.89$ & $66.21$ & $66.43$ & $67.00$ & $66.62$ \\
 &  & PRA \cite{ha2025unlearning} & $62.12$ & $64.78$ & $64.33$ & $65.00$ & $63.54$ & $62.36$ & $64.65$ & $63.80$ & $63.79$ & $65.44$ \\
 &  & SFRA (ours) & $60.54$ & $60.12$ & $61.32$ & $59.28$ & $61.73$ & $61.55$ & $59.62$ & $60.84$ & $61.71$ & $61.71$ \\
\cmidrule(lr){2-13}
 & \multirow{3}{*}{$\mathcal{A}^{t}_{f}(\%)$} & Unlearned & $0.00$ & $0.00$ & $0.00$ & $0.00$ & $0.00$ & $0.00$ & $0.00$ & $0.00$ & $0.00$ & $0.00$ \\
 &  & PRA \cite{ha2025unlearning} & $92.00$ & $16.00$ & $44.00$ & $12.00$ & $34.00$ & $64.00$ & $62.00$ & $56.00$ & $48.00$ & $48.00$ \\
 &  & SFRA (ours) & $64.00$ & $22.00$ & $34.00$ & $34.00$ & $30.00$ & $42.00$ & $40.00$ & $34.00$ & $44.00$ & $34.00$ \\
\cmidrule(lr){2-13}
 & \multirow{1}{*}{$\mathcal{A}^{LP}_{f}(\%)$} & Linear Probe & $92.00$ & $78.00$ & $68.00$ & $78.00$ & $58.00$ & $56.00$ & $56.00$ & $72.00$ & $76.00$ & $76.00$ \\
\cmidrule(lr){2-13}
 & \multirow{2}{*}{$\mathrm{RS}$} & PRA \cite{ha2025unlearning} & $0.93$ & $0.28$ & $0.61$ & $0.21$ & $0.50$ & $0.77$ & $0.76$ & $0.71$ & $0.64$ & $0.65$ \\
 &  & SFRA (ours) & $0.76$ & $0.36$ & $0.50$ & $0.50$ & $0.45$ & $0.58$ & $0.56$ & $0.50$ & $0.60$ & $0.50$ \\
\cmidrule(lr){2-13}
 & \multirow{2}{*}{$\Delta\mathrm{RS}$} & PRA \cite{ha2025unlearning} & $+0.93$ & $+0.28$ & $+0.61$ & $+0.21$ & $+0.50$ & $+0.77$ & $+0.17$ & $+0.71$ & $+0.64$ & $+0.65$ \\
 &  & SFRA (ours) & $+0.11$ & $-0.10$ & $-0.14$ & $+0.12$ & $-0.11$ & $+0.02$ & $-0.09$ & $+0.14$ & $+0.00$ & $-0.12$ \\
\midrule
\multirow{11}{*}{Negative Gradient \cite{golatkar2020eternal}} & \multirow{3}{*}{$\mathcal{A}^{t}_{r}(\%)$} & Unlearned & $66.89$ & $64.84$ & $65.12$ & $63.11$ & $67.03$ & $68.75$ & $68.77$ & $66.60$ & $67.15$ & $65.19$ \\
 &  & PRA \cite{ha2025unlearning} & $66.19$ & $64.45$ & $64.75$ & $62.79$ & $66.68$ & $68.29$ & $68.14$ & $66.38$ & $66.98$ & $64.96$ \\
 &  & SFRA (ours) & $60.69$ & $58.47$ & $58.80$ & $57.48$ & $60.52$ & $61.95$ & $62.39$ & $60.13$ & $60.98$ & $59.47$ \\
\cmidrule(lr){2-13}
 & \multirow{3}{*}{$\mathcal{A}^{t}_{f}(\%)$} & Unlearned & $2.00$ & $0.00$ & $0.00$ & $0.00$ & $0.00$ & $0.00$ & $0.00$ & $2.00$ & $0.00$ & $0.00$ \\
 &  & PRA \cite{ha2025unlearning} & $72.00$ & $60.00$ & $36.00$ & $36.00$ & $44.00$ & $50.00$ & $72.00$ & $44.00$ & $56.00$ & $36.00$ \\
 &  & SFRA (ours) & $54.00$ & $38.00$ & $10.00$ & $4.00$ & $56.00$ & $16.00$ & $56.00$ & $26.00$ & $66.00$ & $10.00$ \\
\cmidrule(lr){2-13}
 & \multirow{1}{*}{$\mathcal{A}^{LP}_{f}(\%)$} & Linear Probe & $88.00$ & $68.00$ & $44.00$ & $62.00$ & $44.00$ & $48.00$ & $60.00$ & $52.00$ & $50.00$ & $50.00$ \\
\cmidrule(lr){2-13}
 & \multirow{2}{*}{$\mathrm{RS}$} & PRA \cite{ha2025unlearning} & $0.82$ & $0.75$ & $0.53$ & $0.53$ & $0.61$ & $0.67$ & $0.83$ & $0.59$ & $0.72$ & $0.53$ \\
 &  & SFRA (ours) & $0.67$ & $0.54$ & $0.18$ & $0.08$ & $0.70$ & $0.27$ & $0.70$ & $0.38$ & $0.77$ & $0.18$ \\
\cmidrule(lr){2-13}
 & \multirow{2}{*}{$\Delta\mathrm{RS}$} & PRA \cite{ha2025unlearning} & $+0.82$ & $+0.75$ & $+0.53$ & $+0.53$ & $+0.61$ & $+0.67$ & $+0.25$ & $+0.59$ & $+0.72$ & $+0.53$ \\
 &  & SFRA (ours) & $+0.02$ & $+0.09$ & $-0.45$ & $-0.31$ & $+0.14$ & $-0.29$ & $+0.05$ & $+0.02$ & $+0.18$ & $-0.44$ \\
\midrule
\multirow{11}{*}{Negative Gradient+ \cite{kurmanji2023towards}} & \multirow{3}{*}{$\mathcal{A}^{t}_{r}(\%)$} & Unlearned & $70.30$ & $68.94$ & $70.32$ & $69.59$ & $70.33$ & $70.25$ & $70.34$ & $68.97$ & $67.48$ & $70.17$ \\
 &  & PRA \cite{ha2025unlearning} & $69.99$ & $68.94$ & $70.18$ & $69.47$ & $70.12$ & $70.14$ & $70.15$ & $68.93$ & $67.48$ & $70.07$ \\
 &  & SFRA (ours) & $64.81$ & $63.41$ & $64.63$ & $65.32$ & $65.76$ & $63.28$ & $65.87$ & $62.97$ & $67.48$ & $64.66$ \\
\cmidrule(lr){2-13}
 & \multirow{3}{*}{$\mathcal{A}^{t}_{f}(\%)$} & Unlearned & $2.00$ & $0.00$ & $4.00$ & $6.00$ & $0.00$ & $0.00$ & $0.00$ & $0.00$ & $0.00$ & $0.00$ \\
 &  & PRA \cite{ha2025unlearning} & $78.00$ & $6.00$ & $36.00$ & $60.00$ & $34.00$ & $36.00$ & $56.00$ & $14.00$ & $0.00$ & $50.00$ \\
 &  & SFRA (ours) & $50.00$ & $14.00$ & $32.00$ & $22.00$ & $40.00$ & $50.00$ & $36.00$ & $26.00$ & $0.00$ & $36.00$ \\
\cmidrule(lr){2-13}
 & \multirow{1}{*}{$\mathcal{A}^{LP}_{f}(\%)$} & Linear Probe & $80.00$ & $64.00$ & $38.00$ & $68.00$ & $48.00$ & $56.00$ & $52.00$ & $56.00$ & $42.00$ & $54.00$ \\
\cmidrule(lr){2-13}
 & \multirow{2}{*}{$\mathrm{RS}$} & PRA \cite{ha2025unlearning} & $0.86$ & $0.11$ & $0.48$ & $0.70$ & $0.51$ & $0.53$ & $0.72$ & $0.25$ & $0.00$ & $0.67$ \\
 &  & SFRA (ours) & $0.64$ & $0.24$ & $0.43$ & $0.27$ & $0.56$ & $0.65$ & $0.52$ & $0.41$ & $0.00$ & $0.52$ \\
\cmidrule(lr){2-13}
 & \multirow{2}{*}{$\Delta\mathrm{RS}$} & PRA \cite{ha2025unlearning} & $+0.86$ & $+0.11$ & $+0.48$ & $+0.70$ & $+0.51$ & $+0.53$ & $+0.13$ & $+0.25$ & $+0.00$ & $+0.67$ \\
 &  & SFRA (ours) & $-0.02$ & $-0.21$ & $-0.20$ & $-0.11$ & $+0.00$ & $+0.09$ & $-0.13$ & $+0.05$ & $-0.60$ & $-0.10$ \\
\midrule
\multirow{11}{*}{Random Label \cite{hayase2020selective}} & \multirow{3}{*}{$\mathcal{A}^{t}_{r}(\%)$} & Unlearned & $66.18$ & $67.44$ & $63.84$ & $66.47$ & $69.30$ & $68.92$ & $68.95$ & $66.91$ & $67.50$ & $66.97$ \\
 &  & PRA \cite{ha2025unlearning} & $65.10$ & $66.92$ & $63.37$ & $66.16$ & $68.37$ & $68.49$ & $68.32$ & $66.64$ & $66.86$ & $66.70$ \\
 &  & SFRA (ours) & $60.66$ & $63.81$ & $58.07$ & $60.35$ & $62.84$ & $63.57$ & $62.60$ & $60.69$ & $61.20$ & $60.65$ \\
\cmidrule(lr){2-13}
 & \multirow{3}{*}{$\mathcal{A}^{t}_{f}(\%)$} & Unlearned & $2.00$ & $2.00$ & $0.00$ & $0.00$ & $2.00$ & $0.00$ & $0.00$ & $2.00$ & $0.00$ & $0.00$ \\
 &  & PRA \cite{ha2025unlearning} & $90.00$ & $68.00$ & $46.00$ & $54.00$ & $58.00$ & $56.00$ & $66.00$ & $38.00$ & $70.00$ & $46.00$ \\
 &  & SFRA (ours) & $74.00$ & $52.00$ & $38.00$ & $48.00$ & $50.00$ & $22.00$ & $58.00$ & $36.00$ & $72.00$ & $46.00$ \\
\cmidrule(lr){2-13}
 & \multirow{1}{*}{$\mathcal{A}^{LP}_{f}(\%)$} & Linear Probe & $86.00$ & $76.00$ & $56.00$ & $68.00$ & $42.00$ & $52.00$ & $56.00$ & $64.00$ & $64.00$ & $56.00$ \\
\cmidrule(lr){2-13}
 & \multirow{2}{*}{$\mathrm{RS}$} & PRA \cite{ha2025unlearning} & $0.93$ & $0.79$ & $0.63$ & $0.70$ & $0.72$ & $0.72$ & $0.79$ & $0.53$ & $0.82$ & $0.63$ \\
 &  & SFRA (ours) & $0.82$ & $0.66$ & $0.54$ & $0.64$ & $0.63$ & $0.36$ & $0.72$ & $0.50$ & $0.81$ & $0.62$ \\
\cmidrule(lr){2-13}
 & \multirow{2}{*}{$\Delta\mathrm{RS}$} & PRA \cite{ha2025unlearning} & $+0.93$ & $+0.79$ & $+0.63$ & $+0.70$ & $+0.72$ & $+0.72$ & $+0.20$ & $+0.53$ & $+0.82$ & $+0.63$ \\
 &  & SFRA (ours) & $+0.16$ & $+0.20$ & $-0.09$ & $+0.25$ & $+0.07$ & $-0.20$ & $+0.06$ & $+0.14$ & $+0.21$ & $-0.00$ \\
\bottomrule
\end{tabular}
}
\end{minipage}%
\hfill
\begin{minipage}[t]{0.49\textwidth}
\centering
\scriptsize
\setlength{\tabcolsep}{2pt}
\renewcommand{\arraystretch}{0.80}
\resizebox{\columnwidth}{!}{%
\begin{tabular}{l|l|l|cccccccccc}
\toprule
\multirow{2}{*}{Unlearning Method} & \multirow{2}{*}{Metric} & \multirow{2}{*}{Variant} & \multicolumn{10}{c}{Forget Class} \\
 & & & 0 & 20 & 40 & 60 & 80 & 100 & 120 & 140 & 160 & 180 \\
\midrule
\multirow{11}{*}{Boundary Shrink \cite{chen2023boundary}} & \multirow{3}{*}{$\mathcal{A}^{t}_{r}(\%)$} & Unlearned & $61.82$ & $58.02$ & $64.44$ & $63.62$ & $65.61$ & $65.13$ & $65.95$ & $59.88$ & $64.42$ & $64.41$ \\
 &  & PRA \cite{ha2025unlearning} & $61.82$ & $56.97$ & $63.70$ & $62.94$ & $64.71$ & $64.60$ & $65.21$ & $59.50$ & $64.04$ & $63.84$ \\
 &  & SFRA (ours) & $57.75$ & $55.65$ & $59.11$ & $57.55$ & $60.30$ & $60.35$ & $60.38$ & $54.33$ & $58.27$ & $58.19$ \\
\cmidrule(lr){2-13}
 & \multirow{3}{*}{$\mathcal{A}^{t}_{f}(\%)$} & Unlearned & $2.00$ & $10.00$ & $4.00$ & $4.00$ & $2.00$ & $0.00$ & $0.00$ & $2.00$ & $8.00$ & $2.00$ \\
 &  & PRA \cite{ha2025unlearning} & $18.00$ & $76.00$ & $42.00$ & $40.00$ & $60.00$ & $54.00$ & $54.00$ & $28.00$ & $64.00$ & $46.00$ \\
 &  & SFRA (ours) & $22.00$ & $30.00$ & $14.00$ & $14.00$ & $40.00$ & $12.00$ & $34.00$ & $14.00$ & $54.00$ & $14.00$ \\
\cmidrule(lr){2-13}
 & \multirow{1}{*}{$\mathcal{A}^{LP}_{f}(\%)$} & Linear Probe & $78.00$ & $62.00$ & $40.00$ & $62.00$ & $42.00$ & $36.00$ & $52.00$ & $56.00$ & $62.00$ & $54.00$ \\
\cmidrule(lr){2-13}
 & \multirow{2}{*}{$\mathrm{RS}$} & PRA \cite{ha2025unlearning} & $0.28$ & $0.79$ & $0.55$ & $0.53$ & $0.73$ & $0.70$ & $0.70$ & $0.41$ & $0.72$ & $0.61$ \\
 &  & SFRA (ours) & $0.33$ & $0.33$ & $0.18$ & $0.18$ & $0.54$ & $0.21$ & $0.50$ & $0.21$ & $0.62$ & $0.21$ \\
\cmidrule(lr){2-13}
 & \multirow{2}{*}{$\Delta\mathrm{RS}$} & PRA \cite{ha2025unlearning} & $+0.28$ & $+0.79$ & $+0.55$ & $+0.53$ & $+0.73$ & $+0.70$ & $+0.11$ & $+0.41$ & $+0.72$ & $+0.61$ \\
 &  & SFRA (ours) & $-0.32$ & $-0.12$ & $-0.45$ & $-0.20$ & $-0.02$ & $-0.35$ & $-0.15$ & $-0.14$ & $+0.02$ & $-0.41$ \\
\midrule
\multirow{11}{*}{Learn to Unlearn \cite{cha2024learning}} & \multirow{3}{*}{$\mathcal{A}^{t}_{r}(\%)$} & Unlearned & $66.76$ & $66.21$ & $65.19$ & $64.35$ & $67.00$ & $68.78$ & $68.71$ & $67.52$ & $67.32$ & $66.47$ \\
 &  & PRA \cite{ha2025unlearning} & $66.09$ & $65.81$ & $64.80$ & $64.00$ & $66.70$ & $68.34$ & $68.12$ & $67.28$ & $67.15$ & $66.20$ \\
 &  & SFRA (ours) & $60.53$ & $60.04$ & $58.94$ & $57.99$ & $60.42$ & $62.85$ & $61.91$ & $61.01$ & $61.78$ & $59.98$ \\
\cmidrule(lr){2-13}
 & \multirow{3}{*}{$\mathcal{A}^{t}_{f}(\%)$} & Unlearned & $2.00$ & $2.00$ & $0.00$ & $2.00$ & $0.00$ & $0.00$ & $0.00$ & $2.00$ & $0.00$ & $0.00$ \\
 &  & PRA \cite{ha2025unlearning} & $70.00$ & $66.00$ & $36.00$ & $48.00$ & $44.00$ & $52.00$ & $72.00$ & $54.00$ & $58.00$ & $42.00$ \\
 &  & SFRA (ours) & $56.00$ & $40.00$ & $12.00$ & $26.00$ & $58.00$ & $10.00$ & $66.00$ & $32.00$ & $56.00$ & $24.00$ \\
\cmidrule(lr){2-13}
 & \multirow{1}{*}{$\mathcal{A}^{LP}_{f}(\%)$} & Linear Probe & $86.00$ & $68.00$ & $56.00$ & $66.00$ & $42.00$ & $48.00$ & $60.00$ & $58.00$ & $58.00$ & $50.00$ \\
\cmidrule(lr){2-13}
 & \multirow{2}{*}{$\mathrm{RS}$} & PRA \cite{ha2025unlearning} & $0.81$ & $0.78$ & $0.53$ & $0.63$ & $0.61$ & $0.68$ & $0.84$ & $0.68$ & $0.73$ & $0.59$ \\
 &  & SFRA (ours) & $0.69$ & $0.54$ & $0.21$ & $0.38$ & $0.72$ & $0.18$ & $0.77$ & $0.45$ & $0.70$ & $0.38$ \\
\cmidrule(lr){2-13}
 & \multirow{2}{*}{$\Delta\mathrm{RS}$} & PRA \cite{ha2025unlearning} & $+0.81$ & $+0.78$ & $+0.53$ & $+0.63$ & $+0.61$ & $+0.68$ & $+0.25$ & $+0.68$ & $+0.73$ & $+0.59$ \\
 &  & SFRA (ours) & $+0.03$ & $+0.09$ & $-0.42$ & $-0.00$ & $+0.15$ & $-0.38$ & $+0.12$ & $+0.10$ & $+0.10$ & $-0.24$ \\
\midrule
\multirow{11}{*}{SCRUB \cite{kurmanji2023towards}} & \multirow{3}{*}{$\mathcal{A}^{t}_{r}(\%)$} & Unlearned & $66.61$ & $66.00$ & $67.85$ & $66.73$ & $69.42$ & $69.31$ & $68.64$ & $66.09$ & $68.66$ & $68.01$ \\
 &  & PRA \cite{ha2025unlearning} & $66.59$ & $65.82$ & $67.77$ & $66.71$ & $69.38$ & $69.27$ & $67.32$ & $65.98$ & $68.65$ & $67.96$ \\
 &  & SFRA (ours) & $61.71$ & $60.39$ & $61.71$ & $60.42$ & $63.97$ & $62.56$ & $63.79$ & $59.94$ & $61.80$ & $62.11$ \\
\cmidrule(lr){2-13}
 & \multirow{3}{*}{$\mathcal{A}^{t}_{f}(\%)$} & Unlearned & $0.00$ & $0.00$ & $0.00$ & $0.00$ & $0.00$ & $0.00$ & $0.00$ & $0.00$ & $0.00$ & $0.00$ \\
 &  & PRA \cite{ha2025unlearning} & $64.00$ & $80.00$ & $46.00$ & $60.00$ & $40.00$ & $38.00$ & $88.00$ & $62.00$ & $54.00$ & $60.00$ \\
 &  & SFRA (ours) & $58.00$ & $60.00$ & $22.00$ & $12.00$ & $34.00$ & $32.00$ & $22.00$ & $14.00$ & $80.00$ & $24.00$ \\
\cmidrule(lr){2-13}
 & \multirow{1}{*}{$\mathcal{A}^{LP}_{f}(\%)$} & Linear Probe & $84.00$ & $72.00$ & $46.00$ & $62.00$ & $44.00$ & $46.00$ & $56.00$ & $70.00$ & $66.00$ & $58.00$ \\
\cmidrule(lr){2-13}
 & \multirow{2}{*}{$\mathrm{RS}$} & PRA \cite{ha2025unlearning} & $0.78$ & $0.89$ & $0.63$ & $0.75$ & $0.57$ & $0.55$ & $0.93$ & $0.77$ & $0.70$ & $0.75$ \\
 &  & SFRA (ours) & $0.72$ & $0.73$ & $0.36$ & $0.21$ & $0.50$ & $0.48$ & $0.36$ & $0.24$ & $0.86$ & $0.38$ \\
\cmidrule(lr){2-13}
 & \multirow{2}{*}{$\Delta\mathrm{RS}$} & PRA \cite{ha2025unlearning} & $+0.78$ & $+0.89$ & $+0.63$ & $+0.75$ & $+0.57$ & $+0.55$ & $+0.34$ & $+0.77$ & $+0.70$ & $+0.75$ \\
 &  & SFRA (ours) & $+0.07$ & $+0.28$ & $-0.28$ & $-0.17$ & $-0.06$ & $-0.08$ & $-0.29$ & $-0.11$ & $+0.26$ & $-0.24$ \\
\midrule
\multirow{11}{*}{Bad Teacher \cite{chundawat2023can}} & \multirow{3}{*}{$\mathcal{A}^{t}_{r}(\%)$} & Unlearned & $71.14$ & $71.00$ & $71.05$ & $70.94$ & $71.22$ & $71.19$ & $71.37$ & $71.03$ & $71.36$ & $71.12$ \\
 &  & PRA \cite{ha2025unlearning} & $70.40$ & $70.50$ & $69.77$ & $69.82$ & $69.76$ & $70.74$ & $70.47$ & $70.35$ & $70.59$ & $70.65$ \\
 &  & SFRA (ours) & $66.24$ & $65.00$ & $64.69$ & $63.87$ & $66.36$ & $69.74$ & $64.36$ & $64.68$ & $64.36$ & $64.43$ \\
\cmidrule(lr){2-13}
 & \multirow{3}{*}{$\mathcal{A}^{t}_{f}(\%)$} & Unlearned & $0.00$ & $0.00$ & $0.00$ & $0.00$ & $0.00$ & $16.00$ & $0.00$ & $0.00$ & $0.00$ & $18.00$ \\
 &  & PRA \cite{ha2025unlearning} & $96.00$ & $98.00$ & $90.00$ & $96.00$ & $72.00$ & $70.00$ & $86.00$ & $90.00$ & $94.00$ & $70.00$ \\
 &  & SFRA (ours) & $28.00$ & $80.00$ & $74.00$ & $38.00$ & $20.00$ & $0.00$ & $30.00$ & $44.00$ & $56.00$ & $72.00$ \\
\cmidrule(lr){2-13}
 & \multirow{1}{*}{$\mathcal{A}^{LP}_{f}(\%)$} & Linear Probe & $88.00$ & $88.00$ & $56.00$ & $76.00$ & $54.00$ & $50.00$ & $54.00$ & $78.00$ & $74.00$ & $46.00$ \\
\cmidrule(lr){2-13}
 & \multirow{2}{*}{$\mathrm{RS}$} & PRA \cite{ha2025unlearning} & $0.98$ & $0.99$ & $0.94$ & $0.97$ & $0.83$ & $0.70$ & $0.92$ & $0.94$ & $0.97$ & $0.68$ \\
 &  & SFRA (ours) & $0.43$ & $0.86$ & $0.83$ & $0.54$ & $0.33$ & $0.00$ & $0.45$ & $0.60$ & $0.70$ & $0.68$ \\
\cmidrule(lr){2-13}
 & \multirow{2}{*}{$\Delta\mathrm{RS}$} & PRA \cite{ha2025unlearning} & $+0.98$ & $+0.99$ & $+0.94$ & $+0.97$ & $+0.83$ & $+0.70$ & $+0.33$ & $+0.94$ & $+0.97$ & $+0.68$ \\
 &  & SFRA (ours) & $-0.22$ & $+0.41$ & $+0.19$ & $+0.16$ & $-0.23$ & $-0.56$ & $-0.20$ & $+0.24$ & $+0.10$ & $+0.07$ \\
\midrule
\multirow{11}{*}{SalUn \cite{fan2023salun}} & \multirow{3}{*}{$\mathcal{A}^{t}_{r}(\%)$} & Unlearned & $70.03$ & $69.31$ & $69.62$ & $69.59$ & $70.09$ & $70.47$ & $70.43$ & $69.33$ & $69.80$ & $69.98$ \\
 &  & PRA \cite{ha2025unlearning} & $69.94$ & $69.16$ & $69.24$ & $69.26$ & $69.32$ & $70.42$ & $70.34$ & $69.17$ & $69.74$ & $69.93$ \\
 &  & SFRA (ours) & $66.46$ & $64.62$ & $66.61$ & $66.30$ & $67.02$ & $67.62$ & $67.75$ & $65.46$ & $66.36$ & $65.26$ \\
\cmidrule(lr){2-13}
 & \multirow{3}{*}{$\mathcal{A}^{t}_{f}(\%)$} & Unlearned & $0.00$ & $0.00$ & $0.00$ & $0.00$ & $0.00$ & $0.00$ & $0.00$ & $0.00$ & $0.00$ & $0.00$ \\
 &  & PRA \cite{ha2025unlearning} & $66.00$ & $82.00$ & $62.00$ & $74.00$ & $56.00$ & $42.00$ & $48.00$ & $64.00$ & $40.00$ & $44.00$ \\
 &  & SFRA (ours) & $52.00$ & $42.00$ & $14.00$ & $36.00$ & $8.00$ & $6.00$ & $18.00$ & $30.00$ & $12.00$ & $10.00$ \\
\cmidrule(lr){2-13}
 & \multirow{1}{*}{$\mathcal{A}^{LP}_{f}(\%)$} & Linear Probe & $90.00$ & $82.00$ & $48.00$ & $76.00$ & $48.00$ & $38.00$ & $50.00$ & $74.00$ & $72.00$ & $60.00$ \\
\cmidrule(lr){2-13}
 & \multirow{2}{*}{$\mathrm{RS}$} & PRA \cite{ha2025unlearning} & $0.79$ & $0.90$ & $0.76$ & $0.85$ & $0.72$ & $0.59$ & $0.65$ & $0.78$ & $0.57$ & $0.61$ \\
 &  & SFRA (ours) & $0.68$ & $0.58$ & $0.24$ & $0.52$ & $0.15$ & $0.11$ & $0.30$ & $0.46$ & $0.21$ & $0.18$ \\
\cmidrule(lr){2-13}
 & \multirow{2}{*}{$\Delta\mathrm{RS}$} & PRA \cite{ha2025unlearning} & $+0.79$ & $+0.90$ & $+0.76$ & $+0.85$ & $+0.72$ & $+0.59$ & $+0.06$ & $+0.78$ & $+0.57$ & $+0.61$ \\
 &  & SFRA (ours) & $+0.02$ & $+0.13$ & $-0.39$ & $+0.14$ & $-0.41$ & $-0.45$ & $-0.35$ & $+0.10$ & $-0.39$ & $-0.44$ \\
\midrule
\multirow{11}{*}{DELETE \cite{zhou2025decoupled}} & \multirow{3}{*}{$\mathcal{A}^{t}_{r}(\%)$} & Unlearned & $69.63$ & $66.39$ & $70.65$ & $65.77$ & $70.84$ & $70.45$ & $69.88$ & $68.43$ & $69.87$ & $70.92$ \\
 &  & PRA \cite{ha2025unlearning} & $68.93$ & $66.12$ & $70.64$ & $65.54$ & $69.44$ & $70.20$ & $69.15$ & $68.12$ & $68.45$ & $70.83$ \\
 &  & SFRA (ours) & $63.12$ & $63.30$ & $64.24$ & $59.42$ & $63.85$ & $64.74$ & $63.05$ & $62.67$ & $64.39$ & $64.97$ \\
\cmidrule(lr){2-13}
 & \multirow{3}{*}{$\mathcal{A}^{t}_{f}(\%)$} & Unlearned & $0.00$ & $0.00$ & $0.00$ & $0.00$ & $0.00$ & $0.00$ & $0.00$ & $2.00$ & $0.00$ & $0.00$ \\
 &  & PRA \cite{ha2025unlearning} & $68.00$ & $52.00$ & $24.00$ & $24.00$ & $54.00$ & $32.00$ & $76.00$ & $54.00$ & $64.00$ & $36.00$ \\
 &  & SFRA (ours) & $80.00$ & $36.00$ & $56.00$ & $14.00$ & $16.00$ & $8.00$ & $54.00$ & $24.00$ & $28.00$ & $22.00$ \\
\cmidrule(lr){2-13}
 & \multirow{1}{*}{$\mathcal{A}^{LP}_{f}(\%)$} & Linear Probe & $84.00$ & $64.00$ & $52.00$ & $62.00$ & $54.00$ & $50.00$ & $56.00$ & $66.00$ & $70.00$ & $56.00$ \\
\cmidrule(lr){2-13}
 & \multirow{2}{*}{$\mathrm{RS}$} & PRA \cite{ha2025unlearning} & $0.81$ & $0.68$ & $0.39$ & $0.39$ & $0.70$ & $0.48$ & $0.86$ & $0.68$ & $0.78$ & $0.53$ \\
 &  & SFRA (ours) & $0.86$ & $0.52$ & $0.70$ & $0.24$ & $0.27$ & $0.15$ & $0.68$ & $0.36$ & $0.43$ & $0.36$ \\
\cmidrule(lr){2-13}
 & \multirow{2}{*}{$\Delta\mathrm{RS}$} & PRA \cite{ha2025unlearning} & $+0.81$ & $+0.68$ & $+0.39$ & $+0.39$ & $+0.70$ & $+0.48$ & $+0.27$ & $+0.68$ & $+0.78$ & $+0.53$ \\
 &  & SFRA (ours) & $+0.21$ & $+0.07$ & $+0.07$ & $-0.14$ & $-0.29$ & $-0.41$ & $+0.03$ & $-0.00$ & $-0.17$ & $-0.26$ \\
\bottomrule
\end{tabular}
}
\end{minipage}%

\end{table*}

\begin{table*}[t]
\centering
\caption{Per-forget-class single-class unlearning and relearning results on CIFAR-10 using ViT-B/16. We report the unlearned checkpoint, the source-dependent PRA baseline, our proposed SFRA, and frozen-encoder linear probing. Each forget class column corresponds to a separate unlearned checkpoint in which that class is designated for forgetting.}
\label{tab:cifar10_vit-b-16_variant_by_forget}
\begin{minipage}[t]{0.49\textwidth}
\centering
\scriptsize
\setlength{\tabcolsep}{2pt}
\renewcommand{\arraystretch}{0.80}
\resizebox{\columnwidth}{!}{%
\begin{tabular}{l|l|l|cccccccccc}
\toprule
\multirow{2}{*}{Unlearning Method} & \multirow{2}{*}{Metric} & \multirow{2}{*}{Variant} & \multicolumn{10}{c}{Forget Class} \\
 & & & 0 & 1 & 2 & 3 & 4 & 5 & 6 & 7 & 8 & 9 \\
\midrule
\multirow{2}{*}{Original} & \multirow{1}{*}{$\mathcal{A}^{t}_{r}(\%)$} & Original & $97.72$ & $97.69$ & $97.87$ & $98.01$ & $97.84$ & $97.98$ & $97.70$ & $97.71$ & $97.70$ & $97.78$ \\
\cmidrule(lr){2-13}
 & \multirow{1}{*}{$\mathcal{A}^{t}_{f}(\%)$} & Original & $98.50$ & $98.80$ & $97.20$ & $95.90$ & $97.40$ & $96.20$ & $98.70$ & $98.60$ & $98.70$ & $98.00$ \\
\midrule
\multirow{9}{*}{Retrained} & \multirow{3}{*}{$\mathcal{A}^{t}_{r}(\%)$} & Unlearned & $98.39$ & $98.38$ & $98.21$ & $98.86$ & $98.67$ & $98.67$ & $98.17$ & $98.28$ & $98.20$ & $98.31$ \\
 &  & PRA \cite{ha2025unlearning} & $97.77$ & $98.26$ & $96.98$ & $98.72$ & $97.29$ & $98.26$ & $97.49$ & $97.19$ & $97.72$ & $98.04$ \\
 &  & SFRA (ours) & $98.28$ & $98.27$ & $98.17$ & $98.72$ & $98.21$ & $98.44$ & $97.99$ & $98.29$ & $98.07$ & $98.10$ \\
\cmidrule(lr){2-13}
 & \multirow{3}{*}{$\mathcal{A}^{t}_{f}(\%)$} & Unlearned & $0.00$ & $0.00$ & $0.00$ & $0.00$ & $0.00$ & $0.00$ & $0.00$ & $0.00$ & $0.00$ & $0.00$ \\
 &  & PRA \cite{ha2025unlearning} & $75.90$ & $44.60$ & $79.70$ & $19.60$ & $83.20$ & $71.40$ & $78.20$ & $55.70$ & $89.40$ & $60.90$ \\
 &  & SFRA (ours) & $5.90$ & $5.90$ & $39.80$ & $4.50$ & $7.20$ & $21.40$ & $20.20$ & $1.00$ & $14.50$ & $3.10$ \\
\cmidrule(lr){2-13}
 & \multirow{1}{*}{$\mathcal{A}^{LP}_{f}(\%)$} & Linear Probe & $93.00$ & $97.40$ & $93.70$ & $91.50$ & $92.40$ & $91.50$ & $93.60$ & $93.00$ & $95.80$ & $95.00$ \\
\cmidrule(lr){2-13}
 & \multirow{2}{*}{$\mathrm{RS}$} & PRA \cite{ha2025unlearning} & $0.86$ & $0.62$ & $0.88$ & $0.33$ & $0.90$ & $0.83$ & $0.88$ & $0.71$ & $0.94$ & $0.76$ \\
 &  & SFRA (ours) & $0.11$ & $0.11$ & $0.57$ & $0.09$ & $0.13$ & $0.35$ & $0.34$ & $0.02$ & $0.25$ & $0.06$ \\
\midrule
\multirow{11}{*}{Finetune \cite{golatkar2020eternal}} & \multirow{3}{*}{$\mathcal{A}^{t}_{r}(\%)$} & Unlearned & $97.81$ & $96.72$ & $96.16$ & $97.38$ & $84.93$ & $97.16$ & $95.57$ & $96.41$ & $96.96$ & $97.28$ \\
 &  & PRA \cite{ha2025unlearning} & $97.74$ & $96.09$ & $95.89$ & $96.62$ & $84.92$ & $96.13$ & $94.41$ & $96.34$ & $96.42$ & $97.11$ \\
 &  & SFRA (ours) & $97.76$ & $96.74$ & $96.01$ & $97.34$ & $84.66$ & $97.13$ & $95.34$ & $96.39$ & $96.99$ & $97.16$ \\
\cmidrule(lr){2-13}
 & \multirow{3}{*}{$\mathcal{A}^{t}_{f}(\%)$} & Unlearned & $0.10$ & $1.30$ & $3.50$ & $2.10$ & $0.00$ & $0.20$ & $3.80$ & $1.60$ & $1.20$ & $0.20$ \\
 &  & PRA \cite{ha2025unlearning} & $65.70$ & $95.00$ & $65.20$ & $73.00$ & $0.10$ & $84.90$ & $89.20$ & $62.00$ & $91.10$ & $75.40$ \\
 &  & SFRA (ours) & $2.20$ & $5.70$ & $16.90$ & $6.80$ & $0.30$ & $1.90$ & $8.90$ & $3.20$ & $4.00$ & $7.20$ \\
\cmidrule(lr){2-13}
 & \multirow{1}{*}{$\mathcal{A}^{LP}_{f}(\%)$} & Linear Probe & $97.00$ & $97.00$ & $94.90$ & $89.80$ & $84.70$ & $91.90$ & $97.40$ & $96.20$ & $97.00$ & $96.20$ \\
\cmidrule(lr){2-13}
 & \multirow{2}{*}{$\mathrm{RS}$} & PRA \cite{ha2025unlearning} & $0.79$ & $0.96$ & $0.76$ & $0.83$ & $0.00$ & $0.91$ & $0.92$ & $0.75$ & $0.94$ & $0.86$ \\
 &  & SFRA (ours) & $0.04$ & $0.08$ & $0.24$ & $0.09$ & $0.01$ & $0.03$ & $0.10$ & $0.03$ & $0.05$ & $0.13$ \\
\cmidrule(lr){2-13}
 & \multirow{2}{*}{$\Delta\mathrm{RS}$} & PRA \cite{ha2025unlearning} & $-0.07$ & $+0.35$ & $-0.12$ & $+0.50$ & $-0.90$ & $+0.08$ & $+0.04$ & $+0.04$ & $+0.00$ & $+0.10$ \\
 &  & SFRA (ours) & $-0.07$ & $-0.03$ & $-0.33$ & $+0.00$ & $-0.13$ & $-0.32$ & $-0.24$ & $+0.01$ & $-0.20$ & $+0.07$ \\
\midrule
\multirow{11}{*}{Negative Gradient \cite{golatkar2020eternal}} & \multirow{3}{*}{$\mathcal{A}^{t}_{r}(\%)$} & Unlearned & $94.03$ & $96.79$ & $89.90$ & $97.67$ & $95.13$ & $92.48$ & $93.21$ & $95.92$ & $97.39$ & $94.51$ \\
 &  & PRA \cite{ha2025unlearning} & $93.88$ & $95.26$ & $89.90$ & $96.33$ & $95.13$ & $92.30$ & $92.68$ & $94.91$ & $96.03$ & $94.10$ \\
 &  & SFRA (ours) & $91.99$ & $95.77$ & $81.23$ & $97.16$ & $94.92$ & $89.67$ & $90.01$ & $95.26$ & $96.10$ & $91.39$ \\
\cmidrule(lr){2-13}
 & \multirow{3}{*}{$\mathcal{A}^{t}_{f}(\%)$} & Unlearned & $2.10$ & $0.50$ & $0.10$ & $1.60$ & $0.00$ & $5.30$ & $0.20$ & $0.50$ & $0.40$ & $0.00$ \\
 &  & PRA \cite{ha2025unlearning} & $77.20$ & $99.20$ & $0.10$ & $93.90$ & $0.00$ & $88.60$ & $1.20$ & $96.50$ & $99.20$ & $89.10$ \\
 &  & SFRA (ours) & $49.10$ & $98.70$ & $99.00$ & $55.00$ & $1.70$ & $77.40$ & $46.40$ & $75.90$ & $99.50$ & $99.50$ \\
\cmidrule(lr){2-13}
 & \multirow{1}{*}{$\mathcal{A}^{LP}_{f}(\%)$} & Linear Probe & $96.10$ & $98.10$ & $96.70$ & $92.50$ & $94.00$ & $93.50$ & $98.00$ & $96.20$ & $98.70$ & $98.20$ \\
\cmidrule(lr){2-13}
 & \multirow{2}{*}{$\mathrm{RS}$} & PRA \cite{ha2025unlearning} & $0.86$ & $0.99$ & $0.00$ & $0.95$ & $0.00$ & $0.91$ & $0.02$ & $0.97$ & $0.99$ & $0.94$ \\
 &  & SFRA (ours) & $0.64$ & $0.99$ & $0.95$ & $0.69$ & $0.03$ & $0.83$ & $0.63$ & $0.86$ & $0.99$ & $0.98$ \\
\cmidrule(lr){2-13}
 & \multirow{2}{*}{$\Delta\mathrm{RS}$} & PRA \cite{ha2025unlearning} & $-0.00$ & $+0.37$ & $-0.88$ & $+0.63$ & $-0.90$ & $+0.08$ & $-0.86$ & $+0.26$ & $+0.04$ & $+0.18$ \\
 &  & SFRA (ours) & $+0.52$ & $+0.87$ & $+0.38$ & $+0.61$ & $-0.10$ & $+0.48$ & $+0.29$ & $+0.84$ & $+0.74$ & $+0.92$ \\
\midrule
\multirow{11}{*}{Negative Gradient+ \cite{kurmanji2023towards}} & \multirow{3}{*}{$\mathcal{A}^{t}_{r}(\%)$} & Unlearned & $97.62$ & $97.79$ & $96.90$ & $98.24$ & $96.97$ & $97.94$ & $97.61$ & $97.07$ & $97.34$ & $97.80$ \\
 &  & PRA \cite{ha2025unlearning} & $95.80$ & $96.42$ & $96.89$ & $96.40$ & $96.71$ & $97.39$ & $95.51$ & $95.92$ & $95.80$ & $95.88$ \\
 &  & SFRA (ours) & $97.57$ & $97.70$ & $97.40$ & $98.23$ & $97.23$ & $98.00$ & $97.47$ & $97.02$ & $97.33$ & $97.82$ \\
\cmidrule(lr){2-13}
 & \multirow{3}{*}{$\mathcal{A}^{t}_{f}(\%)$} & Unlearned & $0.00$ & $0.00$ & $0.00$ & $0.00$ & $0.00$ & $0.00$ & $0.00$ & $0.10$ & $0.00$ & $0.00$ \\
 &  & PRA \cite{ha2025unlearning} & $99.40$ & $98.70$ & $0.00$ & $97.40$ & $0.00$ & $88.00$ & $99.90$ & $96.40$ & $99.70$ & $97.90$ \\
 &  & SFRA (ours) & $0.90$ & $8.90$ & $0.00$ & $5.90$ & $0.00$ & $0.00$ & $94.60$ & $5.40$ & $93.80$ & $1.30$ \\
\cmidrule(lr){2-13}
 & \multirow{1}{*}{$\mathcal{A}^{LP}_{f}(\%)$} & Linear Probe & $98.20$ & $98.00$ & $97.80$ & $93.50$ & $96.60$ & $94.30$ & $98.40$ & $96.60$ & $99.00$ & $96.80$ \\
\cmidrule(lr){2-13}
 & \multirow{2}{*}{$\mathrm{RS}$} & PRA \cite{ha2025unlearning} & $0.99$ & $0.99$ & $0.00$ & $0.98$ & $0.00$ & $0.93$ & $0.99$ & $0.98$ & $0.99$ & $0.98$ \\
 &  & SFRA (ours) & $0.02$ & $0.16$ & $0.00$ & $0.11$ & $0.00$ & $0.00$ & $0.97$ & $0.10$ & $0.97$ & $0.03$ \\
\cmidrule(lr){2-13}
 & \multirow{2}{*}{$\Delta\mathrm{RS}$} & PRA \cite{ha2025unlearning} & $+0.13$ & $+0.37$ & $-0.88$ & $+0.65$ & $-0.90$ & $+0.10$ & $+0.11$ & $+0.26$ & $+0.05$ & $+0.22$ \\
 &  & SFRA (ours) & $-0.09$ & $+0.05$ & $-0.57$ & $+0.03$ & $-0.13$ & $-0.35$ & $+0.64$ & $+0.08$ & $+0.71$ & $-0.03$ \\
\midrule
\multirow{11}{*}{Random Label \cite{hayase2020selective}} & \multirow{3}{*}{$\mathcal{A}^{t}_{r}(\%)$} & Unlearned & $97.71$ & $97.52$ & $96.90$ & $98.26$ & $97.93$ & $97.80$ & $97.77$ & $97.48$ & $97.66$ & $97.64$ \\
 &  & PRA \cite{ha2025unlearning} & $95.54$ & $96.32$ & $96.58$ & $97.06$ & $96.61$ & $96.38$ & $97.63$ & $96.67$ & $95.64$ & $96.41$ \\
 &  & SFRA (ours) & $96.81$ & $97.24$ & $92.72$ & $96.53$ & $97.26$ & $93.07$ & $97.39$ & $96.90$ & $96.90$ & $96.56$ \\
\cmidrule(lr){2-13}
 & \multirow{3}{*}{$\mathcal{A}^{t}_{f}(\%)$} & Unlearned & $0.00$ & $0.00$ & $0.20$ & $0.00$ & $0.00$ & $0.00$ & $0.00$ & $0.00$ & $0.10$ & $0.00$ \\
 &  & PRA \cite{ha2025unlearning} & $99.60$ & $99.40$ & $95.30$ & $98.50$ & $99.60$ & $95.20$ & $99.30$ & $99.70$ & $99.80$ & $99.10$ \\
 &  & SFRA (ours) & $98.40$ & $99.20$ & $98.40$ & $98.40$ & $98.40$ & $97.00$ & $98.90$ & $99.80$ & $99.40$ & $99.00$ \\
\cmidrule(lr){2-13}
 & \multirow{1}{*}{$\mathcal{A}^{LP}_{f}(\%)$} & Linear Probe & $97.70$ & $98.60$ & $96.60$ & $97.10$ & $97.20$ & $94.50$ & $97.70$ & $99.10$ & $98.70$ & $97.70$ \\
\cmidrule(lr){2-13}
 & \multirow{2}{*}{$\mathrm{RS}$} & PRA \cite{ha2025unlearning} & $0.99$ & $0.99$ & $0.97$ & $0.99$ & $0.99$ & $0.97$ & $1.00$ & $0.99$ & $0.99$ & $0.99$ \\
 &  & SFRA (ours) & $0.99$ & $0.99$ & $0.97$ & $0.98$ & $0.99$ & $0.96$ & $0.99$ & $1.00$ & $0.99$ & $0.99$ \\
\cmidrule(lr){2-13}
 & \multirow{2}{*}{$\Delta\mathrm{RS}$} & PRA \cite{ha2025unlearning} & $+0.13$ & $+0.37$ & $+0.09$ & $+0.66$ & $+0.09$ & $+0.14$ & $+0.12$ & $+0.28$ & $+0.05$ & $+0.23$ \\
 &  & SFRA (ours) & $+0.88$ & $+0.88$ & $+0.40$ & $+0.90$ & $+0.85$ & $+0.61$ & $+0.66$ & $+0.98$ & $+0.74$ & $+0.93$ \\
\bottomrule
\end{tabular}
}
\end{minipage}%
\hfill
\begin{minipage}[t]{0.49\textwidth}
\centering
\scriptsize
\setlength{\tabcolsep}{2pt}
\renewcommand{\arraystretch}{0.80}
\resizebox{\columnwidth}{!}{%
\begin{tabular}{l|l|l|cccccccccc}
\toprule
\multirow{2}{*}{Unlearning Method} & \multirow{2}{*}{Metric} & \multirow{2}{*}{Variant} & \multicolumn{10}{c}{Forget Class} \\
 & & & 0 & 1 & 2 & 3 & 4 & 5 & 6 & 7 & 8 & 9 \\
\midrule
\multirow{11}{*}{Learn to Unlearn \cite{cha2024learning}} & \multirow{3}{*}{$\mathcal{A}^{t}_{r}(\%)$} & Unlearned & $77.00$ & $97.03$ & $83.23$ & $96.58$ & $94.68$ & $70.27$ & $90.42$ & $96.47$ & $96.38$ & $89.21$ \\
 &  & PRA \cite{ha2025unlearning} & $77.00$ & $95.93$ & $83.23$ & $95.81$ & $94.68$ & $70.27$ & $90.41$ & $95.26$ & $96.20$ & $88.28$ \\
 &  & SFRA (ours) & $78.88$ & $96.22$ & $81.38$ & $96.66$ & $93.67$ & $71.30$ & $86.74$ & $96.42$ & $94.87$ & $85.36$ \\
\cmidrule(lr){2-13}
 & \multirow{3}{*}{$\mathcal{A}^{t}_{f}(\%)$} & Unlearned & $0.00$ & $2.80$ & $0.10$ & $0.10$ & $0.00$ & $0.00$ & $0.00$ & $0.00$ & $0.00$ & $0.00$ \\
 &  & PRA \cite{ha2025unlearning} & $0.00$ & $98.40$ & $0.10$ & $45.50$ & $0.10$ & $1.80$ & $0.00$ & $89.00$ & $8.30$ & $96.90$ \\
 &  & SFRA (ours) & $0.40$ & $97.60$ & $23.10$ & $3.30$ & $8.00$ & $0.50$ & $0.70$ & $0.90$ & $97.70$ & $53.10$ \\
\cmidrule(lr){2-13}
 & \multirow{1}{*}{$\mathcal{A}^{LP}_{f}(\%)$} & Linear Probe & $96.10$ & $97.80$ & $96.50$ & $92.50$ & $94.90$ & $93.00$ & $98.00$ & $95.70$ & $99.10$ & $97.60$ \\
\cmidrule(lr){2-13}
 & \multirow{2}{*}{$\mathrm{RS}$} & PRA \cite{ha2025unlearning} & $0.00$ & $0.97$ & $0.00$ & $0.62$ & $0.00$ & $0.04$ & $0.00$ & $0.94$ & $0.15$ & $0.98$ \\
 &  & SFRA (ours) & $0.01$ & $0.97$ & $0.37$ & $0.06$ & $0.15$ & $0.01$ & $0.01$ & $0.02$ & $0.98$ & $0.68$ \\
\cmidrule(lr){2-13}
 & \multirow{2}{*}{$\Delta\mathrm{RS}$} & PRA \cite{ha2025unlearning} & $-0.86$ & $+0.36$ & $-0.88$ & $+0.30$ & $-0.90$ & $-0.80$ & $-0.88$ & $+0.22$ & $-0.79$ & $+0.22$ \\
 &  & SFRA (ours) & $-0.10$ & $+0.86$ & $-0.20$ & $-0.02$ & $+0.01$ & $-0.34$ & $-0.32$ & $-0.00$ & $+0.73$ & $+0.62$ \\
\midrule
\multirow{11}{*}{SCRUB \cite{kurmanji2023towards}} & \multirow{3}{*}{$\mathcal{A}^{t}_{r}(\%)$} & Unlearned & $97.59$ & $97.83$ & $95.18$ & $98.47$ & $97.78$ & $97.61$ & $97.71$ & $97.20$ & $96.47$ & $97.84$ \\
 &  & PRA \cite{ha2025unlearning} & $96.53$ & $97.79$ & $95.07$ & $98.20$ & $97.31$ & $97.51$ & $97.68$ & $96.79$ & $96.27$ & $97.84$ \\
 &  & SFRA (ours) & $97.58$ & $97.81$ & $95.17$ & $98.47$ & $97.78$ & $97.60$ & $97.69$ & $89.30$ & $96.47$ & $97.84$ \\
\cmidrule(lr){2-13}
 & \multirow{3}{*}{$\mathcal{A}^{t}_{f}(\%)$} & Unlearned & $0.70$ & $0.10$ & $0.20$ & $1.30$ & $0.10$ & $0.40$ & $0.00$ & $8.10$ & $0.10$ & $0.20$ \\
 &  & PRA \cite{ha2025unlearning} & $54.20$ & $5.30$ & $5.90$ & $28.10$ & $40.10$ & $12.20$ & $6.80$ & $65.30$ & $18.00$ & $1.40$ \\
 &  & SFRA (ours) & $1.80$ & $0.30$ & $2.30$ & $0.00$ & $1.50$ & $3.50$ & $0.80$ & $58.50$ & $2.20$ & $0.10$ \\
\cmidrule(lr){2-13}
 & \multirow{1}{*}{$\mathcal{A}^{LP}_{f}(\%)$} & Linear Probe & $93.80$ & $96.40$ & $82.50$ & $93.40$ & $93.20$ & $89.20$ & $95.40$ & $96.70$ & $93.60$ & $94.20$ \\
\cmidrule(lr){2-13}
 & \multirow{2}{*}{$\mathrm{RS}$} & PRA \cite{ha2025unlearning} & $0.69$ & $0.10$ & $0.11$ & $0.42$ & $0.57$ & $0.21$ & $0.13$ & $0.73$ & $0.30$ & $0.02$ \\
 &  & SFRA (ours) & $0.02$ & $0.00$ & $0.04$ & $0.00$ & $0.03$ & $0.06$ & $0.02$ & $0.65$ & $0.04$ & $0.00$ \\
\cmidrule(lr){2-13}
 & \multirow{2}{*}{$\Delta\mathrm{RS}$} & PRA \cite{ha2025unlearning} & $-0.17$ & $-0.52$ & $-0.77$ & $+0.09$ & $-0.33$ & $-0.62$ & $-0.75$ & $+0.01$ & $-0.64$ & $-0.73$ \\
 &  & SFRA (ours) & $-0.09$ & $-0.11$ & $-0.53$ & $-0.09$ & $-0.11$ & $-0.29$ & $-0.32$ & $+0.63$ & $-0.21$ & $-0.06$ \\
\midrule
\multirow{11}{*}{Bad Teacher \cite{chundawat2023can}} & \multirow{3}{*}{$\mathcal{A}^{t}_{r}(\%)$} & Unlearned & $97.73$ & $97.73$ & $91.89$ & $98.29$ & $86.17$ & $98.32$ & $97.74$ & $97.83$ & $97.97$ & $97.92$ \\
 &  & PRA \cite{ha2025unlearning} & $96.97$ & $97.48$ & $91.06$ & $97.98$ & $85.28$ & $97.71$ & $97.63$ & $97.44$ & $97.43$ & $97.87$ \\
 &  & SFRA (ours) & $97.61$ & $97.67$ & $88.60$ & $98.00$ & $81.87$ & $98.16$ & $97.69$ & $97.77$ & $97.88$ & $97.88$ \\
\cmidrule(lr){2-13}
 & \multirow{3}{*}{$\mathcal{A}^{t}_{f}(\%)$} & Unlearned & $0.40$ & $0.00$ & $10.60$ & $3.80$ & $14.80$ & $0.10$ & $0.80$ & $0.00$ & $0.10$ & $1.70$ \\
 &  & PRA \cite{ha2025unlearning} & $99.40$ & $98.90$ & $95.10$ & $96.00$ & $91.10$ & $97.20$ & $99.20$ & $99.60$ & $99.20$ & $97.70$ \\
 &  & SFRA (ours) & $97.40$ & $98.10$ & $98.10$ & $95.50$ & $96.90$ & $95.60$ & $98.10$ & $96.70$ & $98.40$ & $97.30$ \\
\cmidrule(lr){2-13}
 & \multirow{1}{*}{$\mathcal{A}^{LP}_{f}(\%)$} & Linear Probe & $98.10$ & $98.60$ & $94.10$ & $95.80$ & $91.50$ & $96.10$ & $98.00$ & $98.70$ & $98.70$ & $97.90$ \\
\cmidrule(lr){2-13}
 & \multirow{2}{*}{$\mathrm{RS}$} & PRA \cite{ha2025unlearning} & $0.99$ & $0.99$ & $0.91$ & $0.96$ & $0.86$ & $0.98$ & $0.99$ & $1.00$ & $0.99$ & $0.98$ \\
 &  & SFRA (ours) & $0.98$ & $0.99$ & $0.92$ & $0.96$ & $0.88$ & $0.98$ & $0.99$ & $0.98$ & $0.99$ & $0.98$ \\
\cmidrule(lr){2-13}
 & \multirow{2}{*}{$\Delta\mathrm{RS}$} & PRA \cite{ha2025unlearning} & $+0.13$ & $+0.38$ & $+0.03$ & $+0.63$ & $-0.04$ & $+0.15$ & $+0.12$ & $+0.28$ & $+0.05$ & $+0.22$ \\
 &  & SFRA (ours) & $+0.87$ & $+0.88$ & $+0.35$ & $+0.87$ & $+0.75$ & $+0.62$ & $+0.65$ & $+0.96$ & $+0.74$ & $+0.92$ \\
\midrule
\multirow{11}{*}{SalUn \cite{fan2023salun}} & \multirow{3}{*}{$\mathcal{A}^{t}_{r}(\%)$} & Unlearned & $98.06$ & $98.02$ & $97.90$ & $98.53$ & $98.07$ & $98.48$ & $97.82$ & $97.98$ & $98.00$ & $97.97$ \\
 &  & PRA \cite{ha2025unlearning} & $96.62$ & $97.73$ & $96.52$ & $97.98$ & $95.64$ & $96.42$ & $97.79$ & $97.23$ & $97.13$ & $97.73$ \\
 &  & SFRA (ours) & $98.08$ & $98.02$ & $97.90$ & $98.44$ & $98.09$ & $98.50$ & $97.78$ & $97.99$ & $97.94$ & $97.91$ \\
\cmidrule(lr){2-13}
 & \multirow{3}{*}{$\mathcal{A}^{t}_{f}(\%)$} & Unlearned & $0.00$ & $0.00$ & $0.00$ & $0.00$ & $0.00$ & $0.00$ & $0.00$ & $0.00$ & $0.10$ & $0.00$ \\
 &  & PRA \cite{ha2025unlearning} & $99.50$ & $99.00$ & $98.60$ & $97.50$ & $99.10$ & $98.20$ & $96.10$ & $99.40$ & $99.50$ & $98.50$ \\
 &  & SFRA (ours) & $64.00$ & $97.30$ & $94.30$ & $85.60$ & $85.60$ & $66.50$ & $85.40$ & $5.90$ & $95.80$ & $92.50$ \\
\cmidrule(lr){2-13}
 & \multirow{1}{*}{$\mathcal{A}^{LP}_{f}(\%)$} & Linear Probe & $98.70$ & $98.80$ & $98.10$ & $96.20$ & $96.80$ & $96.00$ & $98.60$ & $98.30$ & $98.50$ & $97.70$ \\
\cmidrule(lr){2-13}
 & \multirow{2}{*}{$\mathrm{RS}$} & PRA \cite{ha2025unlearning} & $0.99$ & $0.99$ & $0.99$ & $0.98$ & $0.98$ & $0.98$ & $0.98$ & $0.99$ & $0.99$ & $0.99$ \\
 &  & SFRA (ours) & $0.78$ & $0.99$ & $0.97$ & $0.92$ & $0.92$ & $0.80$ & $0.92$ & $0.11$ & $0.98$ & $0.96$ \\
\cmidrule(lr){2-13}
 & \multirow{2}{*}{$\Delta\mathrm{RS}$} & PRA \cite{ha2025unlearning} & $+0.13$ & $+0.38$ & $+0.10$ & $+0.66$ & $+0.08$ & $+0.15$ & $+0.10$ & $+0.28$ & $+0.05$ & $+0.23$ \\
 &  & SFRA (ours) & $+0.67$ & $+0.87$ & $+0.40$ & $+0.84$ & $+0.79$ & $+0.45$ & $+0.59$ & $+0.09$ & $+0.72$ & $+0.90$ \\
\midrule
\multirow{11}{*}{DELETE \cite{zhou2025decoupled}} & \multirow{3}{*}{$\mathcal{A}^{t}_{r}(\%)$} & Unlearned & $97.39$ & $97.13$ & $96.97$ & $98.10$ & $97.96$ & $96.01$ & $97.52$ & $97.38$ & $96.91$ & $96.22$ \\
 &  & PRA \cite{ha2025unlearning} & $97.30$ & $96.46$ & $96.97$ & $96.84$ & $96.00$ & $96.01$ & $96.72$ & $97.37$ & $95.92$ & $95.52$ \\
 &  & SFRA (ours) & $95.94$ & $95.22$ & $92.83$ & $96.52$ & $97.53$ & $93.52$ & $95.27$ & $94.21$ & $94.88$ & $91.43$ \\
\cmidrule(lr){2-13}
 & \multirow{3}{*}{$\mathcal{A}^{t}_{f}(\%)$} & Unlearned & $0.00$ & $0.00$ & $0.00$ & $0.00$ & $0.00$ & $0.00$ & $0.00$ & $0.00$ & $0.00$ & $0.00$ \\
 &  & PRA \cite{ha2025unlearning} & $93.40$ & $99.00$ & $44.60$ & $97.10$ & $99.10$ & $0.00$ & $97.50$ & $72.80$ & $99.70$ & $99.10$ \\
 &  & SFRA (ours) & $99.10$ & $99.30$ & $39.30$ & $97.80$ & $95.00$ & $80.30$ & $98.40$ & $99.70$ & $99.90$ & $99.80$ \\
\cmidrule(lr){2-13}
 & \multirow{1}{*}{$\mathcal{A}^{LP}_{f}(\%)$} & Linear Probe & $99.10$ & $98.40$ & $94.70$ & $95.00$ & $96.20$ & $88.90$ & $97.90$ & $98.70$ & $99.00$ & $98.10$ \\
\cmidrule(lr){2-13}
 & \multirow{2}{*}{$\mathrm{RS}$} & PRA \cite{ha2025unlearning} & $0.97$ & $0.99$ & $0.62$ & $0.98$ & $0.99$ & $0.00$ & $0.98$ & $0.84$ & $0.99$ & $0.99$ \\
 &  & SFRA (ours) & $0.99$ & $0.99$ & $0.56$ & $0.98$ & $0.97$ & $0.88$ & $0.98$ & $0.98$ & $0.99$ & $0.97$ \\
\cmidrule(lr){2-13}
 & \multirow{2}{*}{$\Delta\mathrm{RS}$} & PRA \cite{ha2025unlearning} & $+0.10$ & $+0.37$ & $-0.27$ & $+0.65$ & $+0.08$ & $-0.83$ & $+0.11$ & $+0.13$ & $+0.05$ & $+0.24$ \\
 &  & SFRA (ours) & $+0.88$ & $+0.88$ & $-0.01$ & $+0.89$ & $+0.84$ & $+0.53$ & $+0.64$ & $+0.96$ & $+0.74$ & $+0.91$ \\
\bottomrule
\end{tabular}
}
\end{minipage}%

\end{table*}

\begin{table*}[t]
\centering
\caption{Per-forget-class single-class unlearning and relearning results on CIFAR-100 using ViT-B/16. We report the unlearned checkpoint, the source-dependent PRA baseline, our proposed SFRA, and frozen-encoder linear probing. Each forget class column corresponds to a separate unlearned checkpoint in which that class is designated for forgetting.}
\label{tab:cifar100_vit-b-16_variant_by_forget}
\begin{minipage}[t]{0.49\textwidth}
\centering
\scriptsize
\setlength{\tabcolsep}{2pt}
\renewcommand{\arraystretch}{0.80}
\resizebox{\columnwidth}{!}{%
\begin{tabular}{l|l|l|cccccccccc}
\toprule
\multirow{2}{*}{Unlearning Method} & \multirow{2}{*}{Metric} & \multirow{2}{*}{Variant} & \multicolumn{10}{c}{Forget Class} \\
 & & & 0 & 10 & 20 & 30 & 40 & 50 & 60 & 70 & 80 & 90 \\
\midrule
\multirow{2}{*}{Original} & \multirow{1}{*}{$\mathcal{A}^{t}_{r}(\%)$} & Original & $87.66$ & $87.89$ & $87.74$ & $87.85$ & $87.75$ & $87.90$ & $87.77$ & $87.79$ & $87.79$ & $87.72$ \\
\cmidrule(lr){2-13}
 & \multirow{1}{*}{$\mathcal{A}^{t}_{f}(\%)$} & Original & $98.00$ & $75.00$ & $90.00$ & $79.00$ & $89.00$ & $74.00$ & $87.00$ & $85.00$ & $85.00$ & $92.00$ \\
\midrule
\multirow{9}{*}{Retrained} & \multirow{3}{*}{$\mathcal{A}^{t}_{r}(\%)$} & Unlearned & $86.81$ & $87.27$ & $87.00$ & $88.17$ & $87.00$ & $87.77$ & $87.47$ & $87.34$ & $87.07$ & $87.07$ \\
 &  & PRA \cite{ha2025unlearning} & $86.51$ & $86.99$ & $86.56$ & $87.38$ & $86.35$ & $87.14$ & $87.36$ & $86.29$ & $86.36$ & $86.43$ \\
 &  & SFRA (ours) & $86.00$ & $86.62$ & $86.40$ & $87.53$ & $86.12$ & $87.20$ & $86.89$ & $86.20$ & $86.35$ & $85.99$ \\
\cmidrule(lr){2-13}
 & \multirow{3}{*}{$\mathcal{A}^{t}_{f}(\%)$} & Unlearned & $0.00$ & $0.00$ & $0.00$ & $0.00$ & $0.00$ & $0.00$ & $0.00$ & $0.00$ & $0.00$ & $0.00$ \\
 &  & PRA \cite{ha2025unlearning} & $86.00$ & $56.00$ & $85.00$ & $59.00$ & $60.00$ & $69.00$ & $39.00$ & $80.00$ & $79.00$ & $86.00$ \\
 &  & SFRA (ours) & $20.00$ & $18.00$ & $16.00$ & $4.00$ & $54.00$ & $4.00$ & $36.00$ & $7.00$ & $53.00$ & $21.00$ \\
\cmidrule(lr){2-13}
 & \multirow{1}{*}{$\mathcal{A}^{LP}_{f}(\%)$} & Linear Probe & $97.00$ & $74.00$ & $85.00$ & $81.00$ & $83.00$ & $72.00$ & $86.00$ & $73.00$ & $86.00$ & $88.00$ \\
\cmidrule(lr){2-13}
 & \multirow{2}{*}{$\mathrm{RS}$} & PRA \cite{ha2025unlearning} & $0.92$ & $0.72$ & $0.91$ & $0.74$ & $0.75$ & $0.81$ & $0.56$ & $0.88$ & $0.88$ & $0.92$ \\
 &  & SFRA (ours) & $0.33$ & $0.30$ & $0.28$ & $0.08$ & $0.70$ & $0.08$ & $0.53$ & $0.13$ & $0.69$ & $0.35$ \\
\midrule
\multirow{11}{*}{Finetune \cite{golatkar2020eternal}} & \multirow{3}{*}{$\mathcal{A}^{t}_{r}(\%)$} & Unlearned & $87.20$ & $87.76$ & $85.79$ & $87.30$ & $85.92$ & $88.31$ & $86.92$ & $88.53$ & $86.10$ & $84.71$ \\
 &  & PRA \cite{ha2025unlearning} & $87.14$ & $87.74$ & $85.66$ & $86.89$ & $85.89$ & $88.19$ & $86.85$ & $88.52$ & $86.04$ & $84.05$ \\
 &  & SFRA (ours) & $86.74$ & $87.52$ & $85.58$ & $86.91$ & $85.82$ & $88.15$ & $86.52$ & $88.47$ & $85.94$ & $84.21$ \\
\cmidrule(lr){2-13}
 & \multirow{3}{*}{$\mathcal{A}^{t}_{f}(\%)$} & Unlearned & $0.00$ & $1.00$ & $0.00$ & $0.00$ & $14.00$ & $0.00$ & $5.00$ & $0.00$ & $8.00$ & $1.00$ \\
 &  & PRA \cite{ha2025unlearning} & $83.00$ & $41.00$ & $77.00$ & $73.00$ & $41.00$ & $34.00$ & $61.00$ & $33.00$ & $59.00$ & $92.00$ \\
 &  & SFRA (ours) & $7.00$ & $24.00$ & $8.00$ & $2.00$ & $39.00$ & $2.00$ & $32.00$ & $5.00$ & $24.00$ & $8.00$ \\
\cmidrule(lr){2-13}
 & \multirow{1}{*}{$\mathcal{A}^{LP}_{f}(\%)$} & Linear Probe & $94.00$ & $82.00$ & $93.00$ & $84.00$ & $93.00$ & $82.00$ & $89.00$ & $79.00$ & $94.00$ & $88.00$ \\
\cmidrule(lr){2-13}
 & \multirow{2}{*}{$\mathrm{RS}$} & PRA \cite{ha2025unlearning} & $0.91$ & $0.57$ & $0.87$ & $0.84$ & $0.43$ & $0.51$ & $0.72$ & $0.50$ & $0.68$ & $0.95$ \\
 &  & SFRA (ours) & $0.13$ & $0.37$ & $0.15$ & $0.04$ & $0.40$ & $0.04$ & $0.42$ & $0.10$ & $0.28$ & $0.13$ \\
\cmidrule(lr){2-13}
 & \multirow{2}{*}{$\Delta\mathrm{RS}$} & PRA \cite{ha2025unlearning} & $-0.01$ & $-0.14$ & $-0.05$ & $+0.10$ & $-0.32$ & $-0.31$ & $+0.16$ & $-0.39$ & $-0.20$ & $+0.03$ \\
 &  & SFRA (ours) & $-0.20$ & $+0.07$ & $-0.13$ & $-0.04$ & $-0.30$ & $-0.04$ & $-0.10$ & $-0.04$ & $-0.42$ & $-0.22$ \\
\midrule
\multirow{11}{*}{Negative Gradient \cite{golatkar2020eternal}} & \multirow{3}{*}{$\mathcal{A}^{t}_{r}(\%)$} & Unlearned & $85.52$ & $84.06$ & $82.44$ & $86.71$ & $82.39$ & $85.09$ & $84.17$ & $83.27$ & $83.62$ & $86.55$ \\
 &  & PRA \cite{ha2025unlearning} & $83.79$ & $82.94$ & $81.02$ & $86.71$ & $82.14$ & $85.09$ & $83.86$ & $82.99$ & $82.64$ & $85.55$ \\
 &  & SFRA (ours) & $78.56$ & $76.14$ & $74.37$ & $85.45$ & $74.92$ & $82.41$ & $78.25$ & $75.05$ & $75.59$ & $80.25$ \\
\cmidrule(lr){2-13}
 & \multirow{3}{*}{$\mathcal{A}^{t}_{f}(\%)$} & Unlearned & $0.00$ & $0.00$ & $0.00$ & $0.00$ & $7.00$ & $0.00$ & $0.00$ & $0.00$ & $1.00$ & $1.00$ \\
 &  & PRA \cite{ha2025unlearning} & $100.00$ & $55.00$ & $87.00$ & $0.00$ & $53.00$ & $5.00$ & $89.00$ & $48.00$ & $76.00$ & $96.00$ \\
 &  & SFRA (ours) & $99.00$ & $57.00$ & $57.00$ & $3.00$ & $81.00$ & $8.00$ & $97.00$ & $54.00$ & $89.00$ & $42.00$ \\
\cmidrule(lr){2-13}
 & \multirow{1}{*}{$\mathcal{A}^{LP}_{f}(\%)$} & Linear Probe & $97.00$ & $73.00$ & $87.00$ & $80.00$ & $89.00$ & $80.00$ & $86.00$ & $76.00$ & $83.00$ & $87.00$ \\
\cmidrule(lr){2-13}
 & \multirow{2}{*}{$\mathrm{RS}$} & PRA \cite{ha2025unlearning} & $0.99$ & $0.71$ & $0.92$ & $0.00$ & $0.63$ & $0.10$ & $0.94$ & $0.65$ & $0.85$ & $0.97$ \\
 &  & SFRA (ours) & $0.96$ & $0.70$ & $0.70$ & $0.06$ & $0.82$ & $0.15$ & $0.96$ & $0.68$ & $0.90$ & $0.57$ \\
\cmidrule(lr){2-13}
 & \multirow{2}{*}{$\Delta\mathrm{RS}$} & PRA \cite{ha2025unlearning} & $+0.07$ & $-0.01$ & $+0.01$ & $-0.74$ & $-0.12$ & $-0.72$ & $+0.38$ & $-0.24$ & $-0.03$ & $+0.05$ \\
 &  & SFRA (ours) & $+0.63$ & $+0.40$ & $+0.43$ & $-0.02$ & $+0.12$ & $+0.07$ & $+0.43$ & $+0.55$ & $+0.21$ & $+0.22$ \\
\midrule
\multirow{11}{*}{Negative Gradient+ \cite{kurmanji2023towards}} & \multirow{3}{*}{$\mathcal{A}^{t}_{r}(\%)$} & Unlearned & $86.25$ & $86.06$ & $86.52$ & $87.00$ & $85.02$ & $86.39$ & $86.95$ & $86.15$ & $85.92$ & $87.04$ \\
 &  & PRA \cite{ha2025unlearning} & $86.16$ & $86.05$ & $86.52$ & $87.00$ & $84.46$ & $86.25$ & $85.62$ & $85.37$ & $85.90$ & $86.35$ \\
 &  & SFRA (ours) & $86.30$ & $86.11$ & $86.52$ & $86.89$ & $84.59$ & $85.47$ & $85.92$ & $85.78$ & $86.17$ & $85.92$ \\
\cmidrule(lr){2-13}
 & \multirow{3}{*}{$\mathcal{A}^{t}_{f}(\%)$} & Unlearned & $0.00$ & $0.00$ & $0.00$ & $0.00$ & $0.00$ & $0.00$ & $0.00$ & $0.00$ & $0.00$ & $0.00$ \\
 &  & PRA \cite{ha2025unlearning} & $0.00$ & $0.00$ & $0.00$ & $0.00$ & $68.00$ & $69.00$ & $97.00$ & $67.00$ & $0.00$ & $91.00$ \\
 &  & SFRA (ours) & $0.00$ & $2.00$ & $1.00$ & $1.00$ & $24.00$ & $6.00$ & $91.00$ & $3.00$ & $0.00$ & $90.00$ \\
\cmidrule(lr){2-13}
 & \multirow{1}{*}{$\mathcal{A}^{LP}_{f}(\%)$} & Linear Probe & $89.00$ & $62.00$ & $93.00$ & $77.00$ & $86.00$ & $78.00$ & $91.00$ & $76.00$ & $67.00$ & $84.00$ \\
\cmidrule(lr){2-13}
 & \multirow{2}{*}{$\mathrm{RS}$} & PRA \cite{ha2025unlearning} & $0.00$ & $0.00$ & $0.00$ & $0.00$ & $0.81$ & $0.82$ & $0.98$ & $0.80$ & $0.00$ & $0.95$ \\
 &  & SFRA (ours) & $0.00$ & $0.04$ & $0.02$ & $0.02$ & $0.39$ & $0.11$ & $0.95$ & $0.06$ & $0.00$ & $0.94$ \\
\cmidrule(lr){2-13}
 & \multirow{2}{*}{$\Delta\mathrm{RS}$} & PRA \cite{ha2025unlearning} & $-0.92$ & $-0.72$ & $-0.91$ & $-0.74$ & $+0.06$ & $+0.00$ & $+0.42$ & $-0.08$ & $-0.88$ & $+0.03$ \\
 &  & SFRA (ours) & $-0.33$ & $-0.27$ & $-0.26$ & $-0.06$ & $-0.31$ & $+0.04$ & $+0.42$ & $-0.07$ & $-0.69$ & $+0.60$ \\
\midrule
\multirow{11}{*}{Random Label \cite{hayase2020selective}} & \multirow{3}{*}{$\mathcal{A}^{t}_{r}(\%)$} & Unlearned & $84.05$ & $80.07$ & $79.89$ & $84.95$ & $84.40$ & $87.05$ & $80.98$ & $72.59$ & $82.26$ & $86.52$ \\
 &  & PRA \cite{ha2025unlearning} & $82.67$ & $79.47$ & $79.42$ & $83.80$ & $83.76$ & $86.12$ & $79.87$ & $72.26$ & $81.27$ & $84.45$ \\
 &  & SFRA (ours) & $77.03$ & $74.08$ & $76.66$ & $77.23$ & $76.44$ & $84.27$ & $74.84$ & $68.50$ & $76.81$ & $83.15$ \\
\cmidrule(lr){2-13}
 & \multirow{3}{*}{$\mathcal{A}^{t}_{f}(\%)$} & Unlearned & $0.00$ & $0.00$ & $0.00$ & $0.00$ & $0.00$ & $0.00$ & $0.00$ & $0.00$ & $0.00$ & $0.00$ \\
 &  & PRA \cite{ha2025unlearning} & $100.00$ & $54.00$ & $98.00$ & $89.00$ & $86.00$ & $91.00$ & $97.00$ & $92.00$ & $92.00$ & $98.00$ \\
 &  & SFRA (ours) & $100.00$ & $19.00$ & $99.00$ & $96.00$ & $95.00$ & $82.00$ & $97.00$ & $93.00$ & $96.00$ & $96.00$ \\
\cmidrule(lr){2-13}
 & \multirow{1}{*}{$\mathcal{A}^{LP}_{f}(\%)$} & Linear Probe & $98.00$ & $71.00$ & $89.00$ & $83.00$ & $87.00$ & $74.00$ & $91.00$ & $84.00$ & $85.00$ & $93.00$ \\
\cmidrule(lr){2-13}
 & \multirow{2}{*}{$\mathrm{RS}$} & PRA \cite{ha2025unlearning} & $0.99$ & $0.70$ & $0.99$ & $0.94$ & $0.92$ & $0.95$ & $0.98$ & $0.96$ & $0.95$ & $0.98$ \\
 &  & SFRA (ours) & $0.96$ & $0.32$ & $0.98$ & $0.94$ & $0.93$ & $0.89$ & $0.95$ & $0.94$ & $0.95$ & $0.96$ \\
\cmidrule(lr){2-13}
 & \multirow{2}{*}{$\Delta\mathrm{RS}$} & PRA \cite{ha2025unlearning} & $+0.07$ & $-0.02$ & $+0.07$ & $+0.20$ & $+0.17$ & $+0.14$ & $+0.42$ & $+0.07$ & $+0.07$ & $+0.06$ \\
 &  & SFRA (ours) & $+0.63$ & $+0.01$ & $+0.70$ & $+0.86$ & $+0.24$ & $+0.81$ & $+0.43$ & $+0.81$ & $+0.26$ & $+0.62$ \\
\bottomrule
\end{tabular}
}
\end{minipage}%
\hfill
\begin{minipage}[t]{0.49\textwidth}
\centering
\scriptsize
\setlength{\tabcolsep}{2pt}
\renewcommand{\arraystretch}{0.80}
\resizebox{\columnwidth}{!}{%
\begin{tabular}{l|l|l|cccccccccc}
\toprule
\multirow{2}{*}{Unlearning Method} & \multirow{2}{*}{Metric} & \multirow{2}{*}{Variant} & \multicolumn{10}{c}{Forget Class} \\
 & & & 0 & 10 & 20 & 30 & 40 & 50 & 60 & 70 & 80 & 90 \\
\midrule
\multirow{11}{*}{Learn to Unlearn \cite{cha2024learning}} & \multirow{3}{*}{$\mathcal{A}^{t}_{r}(\%)$} & Unlearned & $86.09$ & $84.48$ & $84.30$ & $87.03$ & $80.06$ & $85.94$ & $85.29$ & $84.92$ & $83.44$ & $86.17$ \\
 &  & PRA \cite{ha2025unlearning} & $84.10$ & $83.61$ & $83.82$ & $87.03$ & $79.78$ & $85.94$ & $84.67$ & $84.20$ & $82.58$ & $85.86$ \\
 &  & SFRA (ours) & $79.15$ & $76.45$ & $79.22$ & $86.83$ & $74.77$ & $83.03$ & $81.41$ & $80.53$ & $75.18$ & $80.18$ \\
\cmidrule(lr){2-13}
 & \multirow{3}{*}{$\mathcal{A}^{t}_{f}(\%)$} & Unlearned & $0.00$ & $0.00$ & $0.00$ & $0.00$ & $0.00$ & $1.00$ & $0.00$ & $0.00$ & $0.00$ & $0.00$ \\
 &  & PRA \cite{ha2025unlearning} & $100.00$ & $52.00$ & $88.00$ & $1.00$ & $51.00$ & $13.00$ & $88.00$ & $75.00$ & $75.00$ & $91.00$ \\
 &  & SFRA (ours) & $98.00$ & $82.00$ & $30.00$ & $2.00$ & $79.00$ & $11.00$ & $95.00$ & $39.00$ & $88.00$ & $15.00$ \\
\cmidrule(lr){2-13}
 & \multirow{1}{*}{$\mathcal{A}^{LP}_{f}(\%)$} & Linear Probe & $96.00$ & $76.00$ & $87.00$ & $77.00$ & $87.00$ & $82.00$ & $85.00$ & $76.00$ & $80.00$ & $85.00$ \\
\cmidrule(lr){2-13}
 & \multirow{2}{*}{$\mathrm{RS}$} & PRA \cite{ha2025unlearning} & $0.99$ & $0.68$ & $0.93$ & $0.02$ & $0.67$ & $0.21$ & $0.93$ & $0.85$ & $0.85$ & $0.95$ \\
 &  & SFRA (ours) & $0.95$ & $0.87$ & $0.46$ & $0.04$ & $0.86$ & $0.18$ & $0.96$ & $0.55$ & $0.90$ & $0.26$ \\
\cmidrule(lr){2-13}
 & \multirow{2}{*}{$\Delta\mathrm{RS}$} & PRA \cite{ha2025unlearning} & $+0.07$ & $-0.03$ & $+0.02$ & $-0.72$ & $-0.07$ & $-0.60$ & $+0.37$ & $-0.03$ & $-0.03$ & $+0.03$ \\
 &  & SFRA (ours) & $+0.62$ & $+0.56$ & $+0.18$ & $-0.04$ & $+0.16$ & $+0.10$ & $+0.43$ & $+0.42$ & $+0.21$ & $-0.09$ \\
\midrule
\multirow{11}{*}{SCRUB \cite{kurmanji2023towards}} & \multirow{3}{*}{$\mathcal{A}^{t}_{r}(\%)$} & Unlearned & $84.72$ & $83.34$ & $81.93$ & $83.93$ & $81.03$ & $86.15$ & $82.56$ & $83.94$ & $84.16$ & $82.14$ \\
 &  & PRA \cite{ha2025unlearning} & $83.14$ & $83.22$ & $80.80$ & $83.93$ & $80.01$ & $86.15$ & $82.36$ & $83.03$ & $82.95$ & $87.52$ \\
 &  & SFRA (ours) & $77.42$ & $77.95$ & $73.76$ & $84.53$ & $73.33$ & $84.17$ & $77.20$ & $80.18$ & $78.27$ & $82.64$ \\
\cmidrule(lr){2-13}
 & \multirow{3}{*}{$\mathcal{A}^{t}_{f}(\%)$} & Unlearned & $0.00$ & $1.00$ & $0.00$ & $0.00$ & $8.00$ & $2.00$ & $0.00$ & $0.00$ & $0.00$ & $0.00$ \\
 &  & PRA \cite{ha2025unlearning} & $96.00$ & $30.00$ & $87.00$ & $0.00$ & $55.00$ & $24.00$ & $89.00$ & $79.00$ & $75.00$ & $96.00$ \\
 &  & SFRA (ours) & $97.00$ & $16.00$ & $36.00$ & $0.00$ & $63.00$ & $9.00$ & $96.00$ & $20.00$ & $75.00$ & $0.00$ \\
\cmidrule(lr){2-13}
 & \multirow{1}{*}{$\mathcal{A}^{LP}_{f}(\%)$} & Linear Probe & $96.00$ & $73.00$ & $86.00$ & $81.00$ & $90.00$ & $81.00$ & $88.00$ & $75.00$ & $80.00$ & $92.00$ \\
\cmidrule(lr){2-13}
 & \multirow{2}{*}{$\mathrm{RS}$} & PRA \cite{ha2025unlearning} & $0.97$ & $0.45$ & $0.93$ & $0.00$ & $0.64$ & $0.36$ & $0.94$ & $0.88$ & $0.85$ & $0.08$ \\
 &  & SFRA (ours) & $0.95$ & $0.26$ & $0.52$ & $0.00$ & $0.69$ & $0.13$ & $0.95$ & $0.33$ & $0.83$ & $0.00$ \\
\cmidrule(lr){2-13}
 & \multirow{2}{*}{$\Delta\mathrm{RS}$} & PRA \cite{ha2025unlearning} & $+0.05$ & $-0.27$ & $+0.01$ & $-0.74$ & $-0.11$ & $-0.45$ & $+0.38$ & $-0.01$ & $-0.03$ & $-0.84$ \\
 &  & SFRA (ours) & $+0.62$ & $-0.05$ & $+0.24$ & $-0.08$ & $-0.01$ & $+0.05$ & $+0.42$ & $+0.20$ & $+0.14$ & $-0.35$ \\
\midrule
\multirow{11}{*}{Bad Teacher \cite{chundawat2023can}} & \multirow{3}{*}{$\mathcal{A}^{t}_{r}(\%)$} & Unlearned & $87.67$ & $87.91$ & $87.68$ & $87.78$ & $87.69$ & $87.80$ & $87.50$ & $87.65$ & $87.70$ & $87.74$ \\
 &  & PRA \cite{ha2025unlearning} & $87.63$ & $87.76$ & $87.58$ & $87.57$ & $87.58$ & $87.29$ & $87.40$ & $87.62$ & $87.59$ & $87.47$ \\
 &  & SFRA (ours) & $86.83$ & $86.60$ & $86.68$ & $86.96$ & $86.68$ & $86.64$ & $86.44$ & $87.13$ & $86.85$ & $86.61$ \\
\cmidrule(lr){2-13}
 & \multirow{3}{*}{$\mathcal{A}^{t}_{f}(\%)$} & Unlearned & $0.00$ & $0.00$ & $0.00$ & $0.00$ & $0.00$ & $0.00$ & $0.00$ & $0.00$ & $0.00$ & $0.00$ \\
 &  & PRA \cite{ha2025unlearning} & $98.00$ & $81.00$ & $93.00$ & $87.00$ & $91.00$ & $90.00$ & $90.00$ & $87.00$ & $90.00$ & $95.00$ \\
 &  & SFRA (ours) & $99.00$ & $84.00$ & $94.00$ & $87.00$ & $97.00$ & $84.00$ & $90.00$ & $87.00$ & $91.00$ & $93.00$ \\
\cmidrule(lr){2-13}
 & \multirow{1}{*}{$\mathcal{A}^{LP}_{f}(\%)$} & Linear Probe & $98.00$ & $79.00$ & $92.00$ & $83.00$ & $91.00$ & $75.00$ & $87.00$ & $84.00$ & $87.00$ & $92.00$ \\
\cmidrule(lr){2-13}
 & \multirow{2}{*}{$\mathrm{RS}$} & PRA \cite{ha2025unlearning} & $0.99$ & $0.89$ & $0.96$ & $0.93$ & $0.95$ & $0.95$ & $0.95$ & $0.93$ & $0.95$ & $0.97$ \\
 &  & SFRA (ours) & $0.99$ & $0.91$ & $0.96$ & $0.93$ & $0.98$ & $0.91$ & $0.94$ & $0.93$ & $0.95$ & $0.96$ \\
\cmidrule(lr){2-13}
 & \multirow{2}{*}{$\Delta\mathrm{RS}$} & PRA \cite{ha2025unlearning} & $+0.07$ & $+0.18$ & $+0.05$ & $+0.19$ & $+0.20$ & $+0.13$ & $+0.39$ & $+0.05$ & $+0.07$ & $+0.05$ \\
 &  & SFRA (ours) & $+0.66$ & $+0.60$ & $+0.69$ & $+0.85$ & $+0.28$ & $+0.83$ & $+0.41$ & $+0.80$ & $+0.26$ & $+0.61$ \\
\midrule
\multirow{11}{*}{SalUn \cite{fan2023salun}} & \multirow{3}{*}{$\mathcal{A}^{t}_{r}(\%)$} & Unlearned & $87.50$ & $87.67$ & $87.70$ & $87.72$ & $87.08$ & $87.42$ & $87.44$ & $87.73$ & $87.55$ & $87.22$ \\
 &  & PRA \cite{ha2025unlearning} & $87.43$ & $87.49$ & $87.53$ & $86.75$ & $86.99$ & $86.27$ & $87.39$ & $87.47$ & $87.41$ & $87.18$ \\
 &  & SFRA (ours) & $87.15$ & $87.53$ & $87.70$ & $87.66$ & $87.12$ & $87.15$ & $87.35$ & $87.65$ & $87.31$ & $86.97$ \\
\cmidrule(lr){2-13}
 & \multirow{3}{*}{$\mathcal{A}^{t}_{f}(\%)$} & Unlearned & $0.00$ & $0.00$ & $0.00$ & $0.00$ & $0.00$ & $0.00$ & $0.00$ & $0.00$ & $0.00$ & $0.00$ \\
 &  & PRA \cite{ha2025unlearning} & $96.00$ & $74.00$ & $97.00$ & $93.00$ & $80.00$ & $90.00$ & $90.00$ & $90.00$ & $89.00$ & $78.00$ \\
 &  & SFRA (ours) & $94.00$ & $70.00$ & $94.00$ & $28.00$ & $81.00$ & $27.00$ & $85.00$ & $68.00$ & $88.00$ & $59.00$ \\
\cmidrule(lr){2-13}
 & \multirow{1}{*}{$\mathcal{A}^{LP}_{f}(\%)$} & Linear Probe & $98.00$ & $82.00$ & $93.00$ & $84.00$ & $92.00$ & $66.00$ & $91.00$ & $82.00$ & $89.00$ & $92.00$ \\
\cmidrule(lr){2-13}
 & \multirow{2}{*}{$\mathrm{RS}$} & PRA \cite{ha2025unlearning} & $0.98$ & $0.85$ & $0.98$ & $0.96$ & $0.89$ & $0.94$ & $0.95$ & $0.95$ & $0.94$ & $0.88$ \\
 &  & SFRA (ours) & $0.97$ & $0.82$ & $0.97$ & $0.44$ & $0.90$ & $0.42$ & $0.92$ & $0.81$ & $0.94$ & $0.74$ \\
\cmidrule(lr){2-13}
 & \multirow{2}{*}{$\Delta\mathrm{RS}$} & PRA \cite{ha2025unlearning} & $+0.06$ & $+0.13$ & $+0.07$ & $+0.22$ & $+0.14$ & $+0.13$ & $+0.39$ & $+0.06$ & $+0.06$ & $-0.04$ \\
 &  & SFRA (ours) & $+0.63$ & $+0.52$ & $+0.69$ & $+0.36$ & $+0.20$ & $+0.35$ & $+0.39$ & $+0.68$ & $+0.24$ & $+0.39$ \\
\midrule
\multirow{11}{*}{DELETE \cite{zhou2025decoupled}} & \multirow{3}{*}{$\mathcal{A}^{t}_{r}(\%)$} & Unlearned & $85.16$ & $83.27$ & $82.02$ & $84.45$ & $82.43$ & $86.95$ & $83.89$ & $82.40$ & $80.85$ & $86.32$ \\
 &  & PRA \cite{ha2025unlearning} & $83.48$ & $83.17$ & $81.80$ & $83.63$ & $82.32$ & $86.03$ & $82.59$ & $81.80$ & $80.60$ & $84.77$ \\
 &  & SFRA (ours) & $76.91$ & $76.81$ & $81.38$ & $79.10$ & $74.43$ & $81.36$ & $81.36$ & $76.50$ & $73.30$ & $83.29$ \\
\cmidrule(lr){2-13}
 & \multirow{3}{*}{$\mathcal{A}^{t}_{f}(\%)$} & Unlearned & $0.00$ & $1.00$ & $0.00$ & $0.00$ & $1.00$ & $0.00$ & $0.00$ & $1.00$ & $0.00$ & $0.00$ \\
 &  & PRA \cite{ha2025unlearning} & $100.00$ & $38.00$ & $95.00$ & $91.00$ & $43.00$ & $89.00$ & $95.00$ & $80.00$ & $59.00$ & $94.00$ \\
 &  & SFRA (ours) & $100.00$ & $70.00$ & $95.00$ & $96.00$ & $86.00$ & $99.00$ & $95.00$ & $94.00$ & $97.00$ & $95.00$ \\
\cmidrule(lr){2-13}
 & \multirow{1}{*}{$\mathcal{A}^{LP}_{f}(\%)$} & Linear Probe & $98.00$ & $71.00$ & $89.00$ & $83.00$ & $87.00$ & $83.00$ & $90.00$ & $76.00$ & $83.00$ & $90.00$ \\
\cmidrule(lr){2-13}
 & \multirow{2}{*}{$\mathrm{RS}$} & PRA \cite{ha2025unlearning} & $0.99$ & $0.54$ & $0.97$ & $0.95$ & $0.59$ & $0.94$ & $0.97$ & $0.88$ & $0.74$ & $0.96$ \\
 &  & SFRA (ours) & $0.96$ & $0.79$ & $0.97$ & $0.95$ & $0.88$ & $0.97$ & $0.96$ & $0.94$ & $0.95$ & $0.96$ \\
\cmidrule(lr){2-13}
 & \multirow{2}{*}{$\Delta\mathrm{RS}$} & PRA \cite{ha2025unlearning} & $+0.07$ & $-0.18$ & $+0.06$ & $+0.21$ & $-0.16$ & $+0.12$ & $+0.41$ & $-0.00$ & $-0.14$ & $+0.04$ \\
 &  & SFRA (ours) & $+0.62$ & $+0.49$ & $+0.70$ & $+0.88$ & $+0.18$ & $+0.89$ & $+0.43$ & $+0.80$ & $+0.26$ & $+0.61$ \\
\bottomrule
\end{tabular}
}
\end{minipage}%

\end{table*}

\begin{table*}[t]
\centering
\caption{Per-forget-class single-class unlearning and relearning results on TinyImageNet using ViT-B/16. We report the unlearned checkpoint, the source-dependent PRA baseline, our proposed SFRA, and frozen-encoder linear probing. Each forget class column corresponds to a separate unlearned checkpoint in which that class is designated for forgetting.}
\label{tab:tiny_imagenet_vit-b-16_variant_by_forget}
\begin{minipage}[t]{0.49\textwidth}
\centering
\scriptsize
\setlength{\tabcolsep}{2pt}
\renewcommand{\arraystretch}{0.80}
\resizebox{\columnwidth}{!}{%
\begin{tabular}{l|l|l|cccccccccc}
\toprule
\multirow{2}{*}{Unlearning Method} & \multirow{2}{*}{Metric} & \multirow{2}{*}{Variant} & \multicolumn{10}{c}{Forget Class} \\
 & & & 0 & 20 & 40 & 60 & 80 & 100 & 120 & 140 & 160 & 180 \\
\midrule
\multirow{2}{*}{Original} & \multirow{1}{*}{$\mathcal{A}^{t}_{r}(\%)$} & Original & $89.00$ & $89.03$ & $89.08$ & $89.03$ & $89.06$ & $89.08$ & $89.11$ & $89.06$ & $89.06$ & $89.08$ \\
\cmidrule(lr){2-13}
 & \multirow{1}{*}{$\mathcal{A}^{t}_{f}(\%)$} & Original & $100.00$ & $94.00$ & $84.00$ & $94.00$ & $88.00$ & $84.00$ & $78.00$ & $88.00$ & $88.00$ & $82.00$ \\
\midrule
\multirow{9}{*}{Retrained} & \multirow{3}{*}{$\mathcal{A}^{t}_{r}(\%)$} & Unlearned & $88.04$ & $88.29$ & $88.00$ & $88.30$ & $88.06$ & $88.02$ & $88.39$ & $88.09$ & $87.98$ & $88.31$ \\
 &  & PRA \cite{ha2025unlearning} & $89.14$ & $89.12$ & $88.66$ & $89.15$ & $89.05$ & $89.24$ & $89.08$ & $89.28$ & $89.14$ & $88.96$ \\
 &  & SFRA (ours) & $82.05$ & $83.82$ & $84.72$ & $82.00$ & $80.46$ & $80.59$ & $82.02$ & $83.00$ & $80.55$ & $81.95$ \\
\cmidrule(lr){2-13}
 & \multirow{3}{*}{$\mathcal{A}^{t}_{f}(\%)$} & Unlearned & $0.00$ & $0.00$ & $0.00$ & $0.00$ & $0.00$ & $0.00$ & $0.00$ & $0.00$ & $0.00$ & $0.00$ \\
 &  & PRA \cite{ha2025unlearning} & $100.00$ & $90.00$ & $80.00$ & $94.00$ & $78.00$ & $78.00$ & $88.00$ & $74.00$ & $92.00$ & $72.00$ \\
 &  & SFRA (ours) & $100.00$ & $82.00$ & $58.00$ & $88.00$ & $84.00$ & $86.00$ & $84.00$ & $88.00$ & $94.00$ & $82.00$ \\
\cmidrule(lr){2-13}
 & \multirow{1}{*}{$\mathcal{A}^{LP}_{f}(\%)$} & Linear Probe & $100.00$ & $90.00$ & $80.00$ & $94.00$ & $86.00$ & $90.00$ & $84.00$ & $88.00$ & $94.00$ & $84.00$ \\
\cmidrule(lr){2-13}
 & \multirow{2}{*}{$\mathrm{RS}$} & PRA \cite{ha2025unlearning} & $1.00$ & $0.95$ & $0.89$ & $0.97$ & $0.88$ & $0.88$ & $0.94$ & $0.85$ & $0.96$ & $0.84$ \\
 &  & SFRA (ours) & $0.97$ & $0.88$ & $0.73$ & $0.91$ & $0.88$ & $0.89$ & $0.89$ & $0.91$ & $0.93$ & $0.87$ \\
\midrule
\multirow{11}{*}{Finetune \cite{golatkar2020eternal}} & \multirow{3}{*}{$\mathcal{A}^{t}_{r}(\%)$} & Unlearned & $78.20$ & $74.77$ & $78.28$ & $76.67$ & $75.74$ & $78.05$ & $74.92$ & $77.92$ & $74.39$ & $74.86$ \\
 &  & PRA \cite{ha2025unlearning} & $78.20$ & $74.77$ & $78.28$ & $76.67$ & $75.74$ & $78.05$ & $74.91$ & $77.92$ & $74.39$ & $74.86$ \\
 &  & SFRA (ours) & $70.75$ & $68.00$ & $70.67$ & $69.25$ & $68.20$ & $70.65$ & $67.88$ & $71.34$ & $67.66$ & $67.70$ \\
\cmidrule(lr){2-13}
 & \multirow{3}{*}{$\mathcal{A}^{t}_{f}(\%)$} & Unlearned & $0.00$ & $0.00$ & $0.00$ & $0.00$ & $0.00$ & $0.00$ & $0.00$ & $0.00$ & $0.00$ & $0.00$ \\
 &  & PRA \cite{ha2025unlearning} & $18.00$ & $6.00$ & $0.00$ & $4.00$ & $0.00$ & $0.00$ & $8.00$ & $10.00$ & $2.00$ & $2.00$ \\
 &  & SFRA (ours) & $88.00$ & $60.00$ & $48.00$ & $58.00$ & $76.00$ & $68.00$ & $46.00$ & $70.00$ & $50.00$ & $76.00$ \\
\cmidrule(lr){2-13}
 & \multirow{1}{*}{$\mathcal{A}^{LP}_{f}(\%)$} & Linear Probe & $96.00$ & $86.00$ & $76.00$ & $86.00$ & $88.00$ & $82.00$ & $84.00$ & $94.00$ & $78.00$ & $78.00$ \\
\cmidrule(lr){2-13}
 & \multirow{2}{*}{$\mathrm{RS}$} & PRA \cite{ha2025unlearning} & $0.31$ & $0.11$ & $0.00$ & $0.08$ & $0.00$ & $0.00$ & $0.15$ & $0.18$ & $0.04$ & $0.04$ \\
 &  & SFRA (ours) & $0.90$ & $0.73$ & $0.63$ & $0.71$ & $0.83$ & $0.78$ & $0.62$ & $0.80$ & $0.65$ & $0.84$ \\
\cmidrule(lr){2-13}
 & \multirow{2}{*}{$\Delta\mathrm{RS}$} & PRA \cite{ha2025unlearning} & $-0.69$ & $-0.83$ & $-0.89$ & $-0.89$ & $-0.88$ & $-0.88$ & $-0.79$ & $-0.67$ & $-0.92$ & $-0.80$ \\
 &  & SFRA (ours) & $-0.07$ & $-0.15$ & $-0.09$ & $-0.19$ & $-0.05$ & $-0.11$ & $-0.27$ & $-0.11$ & $-0.28$ & $-0.04$ \\
\midrule
\multirow{11}{*}{Negative Gradient \cite{golatkar2020eternal}} & \multirow{3}{*}{$\mathcal{A}^{t}_{r}(\%)$} & Unlearned & $86.92$ & $86.91$ & $87.93$ & $87.77$ & $85.72$ & $87.63$ & $87.47$ & $87.26$ & $87.47$ & $86.41$ \\
 &  & PRA \cite{ha2025unlearning} & $85.79$ & $85.93$ & $86.61$ & $86.68$ & $84.58$ & $86.26$ & $85.61$ & $86.66$ & $86.13$ & $85.14$ \\
 &  & SFRA (ours) & $80.44$ & $80.21$ & $81.00$ & $79.30$ & $77.60$ & $79.15$ & $79.93$ & $78.84$ & $79.83$ & $78.47$ \\
\cmidrule(lr){2-13}
 & \multirow{3}{*}{$\mathcal{A}^{t}_{f}(\%)$} & Unlearned & $0.00$ & $0.00$ & $2.00$ & $0.00$ & $0.00$ & $0.00$ & $0.00$ & $0.00$ & $0.00$ & $0.00$ \\
 &  & PRA \cite{ha2025unlearning} & $86.00$ & $88.00$ & $92.00$ & $86.00$ & $84.00$ & $84.00$ & $82.00$ & $78.00$ & $94.00$ & $58.00$ \\
 &  & SFRA (ours) & $96.00$ & $84.00$ & $88.00$ & $90.00$ & $68.00$ & $90.00$ & $90.00$ & $96.00$ & $88.00$ & $70.00$ \\
\cmidrule(lr){2-13}
 & \multirow{1}{*}{$\mathcal{A}^{LP}_{f}(\%)$} & Linear Probe & $94.00$ & $88.00$ & $80.00$ & $84.00$ & $74.00$ & $70.00$ & $78.00$ & $78.00$ & $88.00$ & $70.00$ \\
\cmidrule(lr){2-13}
 & \multirow{2}{*}{$\mathrm{RS}$} & PRA \cite{ha2025unlearning} & $0.92$ & $0.93$ & $0.94$ & $0.92$ & $0.91$ & $0.91$ & $0.89$ & $0.87$ & $0.96$ & $0.73$ \\
 &  & SFRA (ours) & $0.95$ & $0.88$ & $0.89$ & $0.91$ & $0.78$ & $0.91$ & $0.91$ & $0.94$ & $0.90$ & $0.80$ \\
\cmidrule(lr){2-13}
 & \multirow{2}{*}{$\Delta\mathrm{RS}$} & PRA \cite{ha2025unlearning} & $-0.08$ & $-0.02$ & $+0.05$ & $-0.05$ & $+0.03$ & $+0.03$ & $-0.04$ & $+0.02$ & $+0.01$ & $-0.11$ \\
 &  & SFRA (ours) & $-0.02$ & $+0.00$ & $+0.17$ & $-0.00$ & $-0.10$ & $+0.02$ & $+0.03$ & $+0.02$ & $-0.03$ & $-0.08$ \\
\midrule
\multirow{11}{*}{Negative Gradient+ \cite{kurmanji2023towards}} & \multirow{3}{*}{$\mathcal{A}^{t}_{r}(\%)$} & Unlearned & $88.35$ & $88.70$ & $87.24$ & $87.31$ & $86.62$ & $87.28$ & $87.62$ & $88.28$ & $87.21$ & $87.29$ \\
 &  & PRA \cite{ha2025unlearning} & $87.95$ & $87.58$ & $86.50$ & $87.28$ & $86.43$ & $86.88$ & $85.51$ & $87.17$ & $85.95$ & $85.87$ \\
 &  & SFRA (ours) & $86.54$ & $87.30$ & $84.50$ & $85.26$ & $80.57$ & $83.94$ & $80.42$ & $85.72$ & $81.32$ & $82.07$ \\
\cmidrule(lr){2-13}
 & \multirow{3}{*}{$\mathcal{A}^{t}_{f}(\%)$} & Unlearned & $0.00$ & $0.00$ & $0.00$ & $0.00$ & $0.00$ & $0.00$ & $0.00$ & $0.00$ & $0.00$ & $0.00$ \\
 &  & PRA \cite{ha2025unlearning} & $100.00$ & $100.00$ & $100.00$ & $8.00$ & $84.00$ & $2.00$ & $86.00$ & $96.00$ & $94.00$ & $62.00$ \\
 &  & SFRA (ours) & $100.00$ & $100.00$ & $14.00$ & $16.00$ & $32.00$ & $14.00$ & $84.00$ & $98.00$ & $12.00$ & $62.00$ \\
\cmidrule(lr){2-13}
 & \multirow{1}{*}{$\mathcal{A}^{LP}_{f}(\%)$} & Linear Probe & $98.00$ & $94.00$ & $82.00$ & $56.00$ & $84.00$ & $54.00$ & $78.00$ & $88.00$ & $92.00$ & $66.00$ \\
\cmidrule(lr){2-13}
 & \multirow{2}{*}{$\mathrm{RS}$} & PRA \cite{ha2025unlearning} & $1.00$ & $0.99$ & $1.00$ & $0.15$ & $0.91$ & $0.04$ & $0.92$ & $0.97$ & $0.96$ & $0.76$ \\
 &  & SFRA (ours) & $0.99$ & $0.99$ & $0.24$ & $0.28$ & $0.48$ & $0.24$ & $0.88$ & $0.98$ & $0.21$ & $0.75$ \\
\cmidrule(lr){2-13}
 & \multirow{2}{*}{$\Delta\mathrm{RS}$} & PRA \cite{ha2025unlearning} & $-0.00$ & $+0.05$ & $+0.11$ & $-0.82$ & $+0.04$ & $-0.84$ & $-0.02$ & $+0.12$ & $+0.01$ & $-0.08$ \\
 &  & SFRA (ours) & $+0.02$ & $+0.11$ & $-0.48$ & $-0.63$ & $-0.40$ & $-0.65$ & $-0.00$ & $+0.06$ & $-0.72$ & $-0.12$ \\
\midrule
\multirow{11}{*}{Random Label \cite{hayase2020selective}} & \multirow{3}{*}{$\mathcal{A}^{t}_{r}(\%)$} & Unlearned & $88.17$ & $88.34$ & $87.99$ & $88.10$ & $87.62$ & $88.36$ & $86.25$ & $88.02$ & $88.24$ & $87.66$ \\
 &  & PRA \cite{ha2025unlearning} & $86.47$ & $85.66$ & $86.09$ & $87.39$ & $86.30$ & $87.29$ & $86.24$ & $86.77$ & $86.37$ & $86.42$ \\
 &  & SFRA (ours) & $87.12$ & $81.75$ & $83.73$ & $85.05$ & $83.61$ & $86.95$ & $80.89$ & $83.69$ & $85.90$ & $83.34$ \\
\cmidrule(lr){2-13}
 & \multirow{3}{*}{$\mathcal{A}^{t}_{f}(\%)$} & Unlearned & $0.00$ & $0.00$ & $0.00$ & $0.00$ & $0.00$ & $0.00$ & $0.00$ & $0.00$ & $0.00$ & $0.00$ \\
 &  & PRA \cite{ha2025unlearning} & $100.00$ & $96.00$ & $90.00$ & $98.00$ & $90.00$ & $100.00$ & $72.00$ & $98.00$ & $98.00$ & $92.00$ \\
 &  & SFRA (ours) & $98.00$ & $98.00$ & $98.00$ & $100.00$ & $98.00$ & $100.00$ & $92.00$ & $98.00$ & $96.00$ & $94.00$ \\
\cmidrule(lr){2-13}
 & \multirow{1}{*}{$\mathcal{A}^{LP}_{f}(\%)$} & Linear Probe & $100.00$ & $94.00$ & $86.00$ & $92.00$ & $82.00$ & $92.00$ & $84.00$ & $88.00$ & $88.00$ & $84.00$ \\
\cmidrule(lr){2-13}
 & \multirow{2}{*}{$\mathrm{RS}$} & PRA \cite{ha2025unlearning} & $0.99$ & $0.97$ & $0.94$ & $0.99$ & $0.94$ & $0.99$ & $0.84$ & $0.98$ & $0.98$ & $0.95$ \\
 &  & SFRA (ours) & $0.98$ & $0.96$ & $0.97$ & $0.98$ & $0.97$ & $0.99$ & $0.93$ & $0.97$ & $0.97$ & $0.95$ \\
\cmidrule(lr){2-13}
 & \multirow{2}{*}{$\Delta\mathrm{RS}$} & PRA \cite{ha2025unlearning} & $-0.01$ & $+0.02$ & $+0.05$ & $+0.02$ & $+0.07$ & $+0.12$ & $-0.10$ & $+0.13$ & $+0.02$ & $+0.12$ \\
 &  & SFRA (ours) & $+0.02$ & $+0.07$ & $+0.24$ & $+0.08$ & $+0.09$ & $+0.10$ & $+0.05$ & $+0.05$ & $+0.04$ & $+0.07$ \\
\bottomrule
\end{tabular}
}
\end{minipage}%
\hfill
\begin{minipage}[t]{0.49\textwidth}
\centering
\scriptsize
\setlength{\tabcolsep}{2pt}
\renewcommand{\arraystretch}{0.80}
\resizebox{\columnwidth}{!}{%
\begin{tabular}{l|l|l|cccccccccc}
\toprule
\multirow{2}{*}{Unlearning Method} & \multirow{2}{*}{Metric} & \multirow{2}{*}{Variant} & \multicolumn{10}{c}{Forget Class} \\
 & & & 0 & 20 & 40 & 60 & 80 & 100 & 120 & 140 & 160 & 180 \\
\midrule
\multirow{11}{*}{Learn to Unlearn \cite{cha2024learning}} & \multirow{3}{*}{$\mathcal{A}^{t}_{r}(\%)$} & Unlearned & $86.84$ & $87.30$ & $87.63$ & $87.59$ & $86.73$ & $87.64$ & $87.52$ & $86.08$ & $87.70$ & $87.35$ \\
 &  & PRA \cite{ha2025unlearning} & $84.85$ & $86.37$ & $86.48$ & $86.05$ & $85.48$ & $86.09$ & $86.04$ & $85.73$ & $86.85$ & $85.69$ \\
 &  & SFRA (ours) & $78.42$ & $83.15$ & $84.50$ & $79.20$ & $83.46$ & $80.07$ & $80.03$ & $79.51$ & $81.66$ & $82.29$ \\
\cmidrule(lr){2-13}
 & \multirow{3}{*}{$\mathcal{A}^{t}_{f}(\%)$} & Unlearned & $0.00$ & $0.00$ & $0.00$ & $0.00$ & $0.00$ & $0.00$ & $0.00$ & $0.00$ & $0.00$ & $0.00$ \\
 &  & PRA \cite{ha2025unlearning} & $88.00$ & $90.00$ & $94.00$ & $88.00$ & $88.00$ & $84.00$ & $80.00$ & $86.00$ & $94.00$ & $72.00$ \\
 &  & SFRA (ours) & $96.00$ & $94.00$ & $50.00$ & $92.00$ & $76.00$ & $88.00$ & $88.00$ & $94.00$ & $90.00$ & $72.00$ \\
\cmidrule(lr){2-13}
 & \multirow{1}{*}{$\mathcal{A}^{LP}_{f}(\%)$} & Linear Probe & $88.00$ & $86.00$ & $80.00$ & $86.00$ & $76.00$ & $68.00$ & $80.00$ & $82.00$ & $90.00$ & $66.00$ \\
\cmidrule(lr){2-13}
 & \multirow{2}{*}{$\mathrm{RS}$} & PRA \cite{ha2025unlearning} & $0.93$ & $0.94$ & $0.96$ & $0.93$ & $0.93$ & $0.91$ & $0.88$ & $0.92$ & $0.97$ & $0.83$ \\
 &  & SFRA (ours) & $0.94$ & $0.95$ & $0.66$ & $0.92$ & $0.85$ & $0.90$ & $0.90$ & $0.94$ & $0.92$ & $0.82$ \\
\cmidrule(lr){2-13}
 & \multirow{2}{*}{$\Delta\mathrm{RS}$} & PRA \cite{ha2025unlearning} & $-0.07$ & $-0.00$ & $+0.08$ & $-0.04$ & $+0.05$ & $+0.03$ & $-0.05$ & $+0.07$ & $+0.01$ & $-0.01$ \\
 &  & SFRA (ours) & $-0.03$ & $+0.07$ & $-0.07$ & $+0.01$ & $-0.03$ & $+0.01$ & $+0.02$ & $+0.02$ & $-0.01$ & $-0.06$ \\
\midrule
\multirow{11}{*}{SCRUB \cite{kurmanji2023towards}} & \multirow{3}{*}{$\mathcal{A}^{t}_{r}(\%)$} & Unlearned & $87.23$ & $86.03$ & $84.80$ & $86.14$ & $88.13$ & $86.56$ & $86.21$ & $84.91$ & $88.25$ & $86.17$ \\
 &  & PRA \cite{ha2025unlearning} & $85.36$ & $84.67$ & $84.45$ & $84.94$ & $86.86$ & $84.98$ & $84.88$ & $83.76$ & $87.00$ & $85.53$ \\
 &  & SFRA (ours) & $79.05$ & $79.91$ & $77.72$ & $78.20$ & $79.39$ & $78.87$ & $79.21$ & $78.30$ & $82.21$ & $79.41$ \\
\cmidrule(lr){2-13}
 & \multirow{3}{*}{$\mathcal{A}^{t}_{f}(\%)$} & Unlearned & $0.00$ & $0.00$ & $0.00$ & $0.00$ & $2.00$ & $0.00$ & $0.00$ & $0.00$ & $4.00$ & $0.00$ \\
 &  & PRA \cite{ha2025unlearning} & $88.00$ & $76.00$ & $68.00$ & $74.00$ & $90.00$ & $72.00$ & $84.00$ & $82.00$ & $94.00$ & $42.00$ \\
 &  & SFRA (ours) & $96.00$ & $82.00$ & $50.00$ & $42.00$ & $96.00$ & $78.00$ & $68.00$ & $80.00$ & $32.00$ & $64.00$ \\
\cmidrule(lr){2-13}
 & \multirow{1}{*}{$\mathcal{A}^{LP}_{f}(\%)$} & Linear Probe & $90.00$ & $82.00$ & $80.00$ & $80.00$ & $76.00$ & $72.00$ & $78.00$ & $80.00$ & $90.00$ & $60.00$ \\
\cmidrule(lr){2-13}
 & \multirow{2}{*}{$\mathrm{RS}$} & PRA \cite{ha2025unlearning} & $0.93$ & $0.86$ & $0.81$ & $0.85$ & $0.93$ & $0.83$ & $0.91$ & $0.90$ & $0.94$ & $0.59$ \\
 &  & SFRA (ours) & $0.94$ & $0.88$ & $0.65$ & $0.58$ & $0.93$ & $0.85$ & $0.79$ & $0.86$ & $0.43$ & $0.76$ \\
\cmidrule(lr){2-13}
 & \multirow{2}{*}{$\Delta\mathrm{RS}$} & PRA \cite{ha2025unlearning} & $-0.07$ & $-0.09$ & $-0.08$ & $-0.12$ & $+0.05$ & $-0.04$ & $-0.03$ & $+0.05$ & $-0.02$ & $-0.25$ \\
 &  & SFRA (ours) & $-0.03$ & $-0.01$ & $-0.08$ & $-0.33$ & $+0.05$ & $-0.05$ & $-0.10$ & $-0.05$ & $-0.50$ & $-0.12$ \\
\midrule
\multirow{11}{*}{Bad Teacher \cite{chundawat2023can}} & \multirow{3}{*}{$\mathcal{A}^{t}_{r}(\%)$} & Unlearned & $89.24$ & $89.28$ & $89.02$ & $88.94$ & $89.09$ & $89.26$ & $89.23$ & $89.42$ & $89.28$ & $89.29$ \\
 &  & PRA \cite{ha2025unlearning} & $89.23$ & $89.27$ & $88.95$ & $88.91$ & $89.05$ & $89.19$ & $89.11$ & $89.40$ & $89.26$ & $89.24$ \\
 &  & SFRA (ours) & $88.50$ & $88.32$ & $82.46$ & $86.46$ & $83.76$ & $87.30$ & $81.15$ & $87.00$ & $81.29$ & $87.33$ \\
\cmidrule(lr){2-13}
 & \multirow{3}{*}{$\mathcal{A}^{t}_{f}(\%)$} & Unlearned & $0.00$ & $0.00$ & $0.00$ & $0.00$ & $4.00$ & $0.00$ & $0.00$ & $0.00$ & $0.00$ & $0.00$ \\
 &  & PRA \cite{ha2025unlearning} & $100.00$ & $98.00$ & $82.00$ & $96.00$ & $84.00$ & $90.00$ & $84.00$ & $92.00$ & $92.00$ & $86.00$ \\
 &  & SFRA (ours) & $100.00$ & $100.00$ & $100.00$ & $100.00$ & $100.00$ & $96.00$ & $100.00$ & $98.00$ & $98.00$ & $96.00$ \\
\cmidrule(lr){2-13}
 & \multirow{1}{*}{$\mathcal{A}^{LP}_{f}(\%)$} & Linear Probe & $100.00$ & $96.00$ & $80.00$ & $96.00$ & $86.00$ & $88.00$ & $76.00$ & $90.00$ & $92.00$ & $84.00$ \\
\cmidrule(lr){2-13}
 & \multirow{2}{*}{$\mathrm{RS}$} & PRA \cite{ha2025unlearning} & $1.00$ & $0.99$ & $0.90$ & $0.98$ & $0.89$ & $0.95$ & $0.91$ & $0.96$ & $0.96$ & $0.92$ \\
 &  & SFRA (ours) & $1.00$ & $1.00$ & $0.97$ & $0.99$ & $0.95$ & $0.97$ & $0.96$ & $0.98$ & $0.95$ & $0.97$ \\
\cmidrule(lr){2-13}
 & \multirow{2}{*}{$\Delta\mathrm{RS}$} & PRA \cite{ha2025unlearning} & $-0.00$ & $+0.04$ & $+0.01$ & $+0.01$ & $+0.01$ & $+0.07$ & $-0.02$ & $+0.11$ & $+0.00$ & $+0.09$ \\
 &  & SFRA (ours) & $+0.03$ & $+0.11$ & $+0.24$ & $+0.08$ & $+0.07$ & $+0.08$ & $+0.07$ & $+0.06$ & $+0.02$ & $+0.10$ \\
\midrule
\multirow{11}{*}{SalUn \cite{fan2023salun}} & \multirow{3}{*}{$\mathcal{A}^{t}_{r}(\%)$} & Unlearned & $88.52$ & $88.52$ & $88.60$ & $88.26$ & $88.64$ & $88.60$ & $88.51$ & $88.55$ & $88.64$ & $88.47$ \\
 &  & PRA \cite{ha2025unlearning} & $88.48$ & $88.52$ & $88.53$ & $88.26$ & $88.58$ & $88.60$ & $88.51$ & $88.55$ & $88.64$ & $88.44$ \\
 &  & SFRA (ours) & $88.20$ & $88.04$ & $87.37$ & $86.91$ & $87.80$ & $87.47$ & $87.17$ & $86.87$ & $88.02$ & $87.42$ \\
\cmidrule(lr){2-13}
 & \multirow{3}{*}{$\mathcal{A}^{t}_{f}(\%)$} & Unlearned & $0.00$ & $0.00$ & $0.00$ & $0.00$ & $0.00$ & $0.00$ & $0.00$ & $0.00$ & $0.00$ & $0.00$ \\
 &  & PRA \cite{ha2025unlearning} & $100.00$ & $92.00$ & $80.00$ & $72.00$ & $80.00$ & $66.00$ & $64.00$ & $74.00$ & $82.00$ & $82.00$ \\
 &  & SFRA (ours) & $100.00$ & $96.00$ & $84.00$ & $90.00$ & $86.00$ & $80.00$ & $80.00$ & $90.00$ & $88.00$ & $90.00$ \\
\cmidrule(lr){2-13}
 & \multirow{1}{*}{$\mathcal{A}^{LP}_{f}(\%)$} & Linear Probe & $100.00$ & $94.00$ & $80.00$ & $86.00$ & $84.00$ & $72.00$ & $82.00$ & $82.00$ & $86.00$ & $84.00$ \\
\cmidrule(lr){2-13}
 & \multirow{2}{*}{$\mathrm{RS}$} & PRA \cite{ha2025unlearning} & $1.00$ & $0.96$ & $0.89$ & $0.84$ & $0.89$ & $0.80$ & $0.78$ & $0.85$ & $0.90$ & $0.90$ \\
 &  & SFRA (ours) & $1.00$ & $0.98$ & $0.91$ & $0.94$ & $0.92$ & $0.88$ & $0.88$ & $0.94$ & $0.93$ & $0.94$ \\
\cmidrule(lr){2-13}
 & \multirow{2}{*}{$\Delta\mathrm{RS}$} & PRA \cite{ha2025unlearning} & $-0.00$ & $+0.01$ & $+0.00$ & $-0.13$ & $+0.01$ & $-0.08$ & $-0.16$ & $+0.00$ & $-0.06$ & $+0.06$ \\
 &  & SFRA (ours) & $+0.03$ & $+0.09$ & $+0.18$ & $+0.03$ & $+0.04$ & $-0.01$ & $-0.00$ & $+0.03$ & $+0.00$ & $+0.07$ \\
\midrule
\multirow{11}{*}{DELETE \cite{zhou2025decoupled}} & \multirow{3}{*}{$\mathcal{A}^{t}_{r}(\%)$} & Unlearned & $88.13$ & $88.45$ & $88.75$ & $88.52$ & $88.78$ & $88.69$ & $88.65$ & $88.48$ & $88.59$ & $88.49$ \\
 &  & PRA \cite{ha2025unlearning} & $86.38$ & $87.31$ & $86.95$ & $87.80$ & $87.65$ & $88.45$ & $86.55$ & $87.65$ & $86.53$ & $87.90$ \\
 &  & SFRA (ours) & $84.14$ & $82.07$ & $84.22$ & $84.92$ & $82.40$ & $86.21$ & $82.22$ & $82.91$ & $81.98$ & $82.72$ \\
\cmidrule(lr){2-13}
 & \multirow{3}{*}{$\mathcal{A}^{t}_{f}(\%)$} & Unlearned & $0.00$ & $0.00$ & $0.00$ & $0.00$ & $0.00$ & $0.00$ & $0.00$ & $0.00$ & $0.00$ & $0.00$ \\
 &  & PRA \cite{ha2025unlearning} & $100.00$ & $96.00$ & $92.00$ & $92.00$ & $94.00$ & $86.00$ & $90.00$ & $94.00$ & $96.00$ & $78.00$ \\
 &  & SFRA (ours) & $100.00$ & $100.00$ & $98.00$ & $96.00$ & $98.00$ & $94.00$ & $92.00$ & $98.00$ & $98.00$ & $96.00$ \\
\cmidrule(lr){2-13}
 & \multirow{1}{*}{$\mathcal{A}^{LP}_{f}(\%)$} & Linear Probe & $100.00$ & $94.00$ & $84.00$ & $92.00$ & $82.00$ & $84.00$ & $80.00$ & $88.00$ & $90.00$ & $80.00$ \\
\cmidrule(lr){2-13}
 & \multirow{2}{*}{$\mathrm{RS}$} & PRA \cite{ha2025unlearning} & $0.99$ & $0.97$ & $0.95$ & $0.95$ & $0.96$ & $0.92$ & $0.94$ & $0.97$ & $0.97$ & $0.87$ \\
 &  & SFRA (ours) & $0.98$ & $0.97$ & $0.97$ & $0.96$ & $0.96$ & $0.96$ & $0.93$ & $0.96$ & $0.96$ & $0.95$ \\
\cmidrule(lr){2-13}
 & \multirow{2}{*}{$\Delta\mathrm{RS}$} & PRA \cite{ha2025unlearning} & $-0.01$ & $+0.03$ & $+0.06$ & $-0.01$ & $+0.09$ & $+0.05$ & $+0.00$ & $+0.11$ & $+0.01$ & $+0.04$ \\
 &  & SFRA (ours) & $+0.01$ & $+0.08$ & $+0.24$ & $+0.05$ & $+0.08$ & $+0.07$ & $+0.04$ & $+0.05$ & $+0.02$ & $+0.08$ \\
\bottomrule
\end{tabular}
}
\end{minipage}%

\end{table*}

\begin{table*}[t]
\centering
\caption{Per-forget-class single-class unlearning and relearning results on CIFAR-10 using Swin-T. We report the unlearned checkpoint, the source-dependent PRA baseline, our proposed SFRA, and frozen-encoder linear probing. Each forget class column corresponds to a separate unlearned checkpoint in which that class is designated for forgetting.}
\label{tab:cifar10_swin-t_variant_by_forget}
\begin{minipage}[t]{0.49\textwidth}
\centering
\scriptsize
\setlength{\tabcolsep}{2pt}
\renewcommand{\arraystretch}{0.80}
\resizebox{\columnwidth}{!}{%
\begin{tabular}{l|l|l|cccccccccc}
\toprule
\multirow{2}{*}{Unlearning Method} & \multirow{2}{*}{Metric} & \multirow{2}{*}{Variant} & \multicolumn{10}{c}{Forget Class} \\
 & & & 0 & 1 & 2 & 3 & 4 & 5 & 6 & 7 & 8 & 9 \\
\midrule
\multirow{2}{*}{Original} & \multirow{1}{*}{$\mathcal{A}^{t}_{r}(\%)$} & Original & $82.73$ & $81.78$ & $83.44$ & $84.77$ & $82.68$ & $83.22$ & $82.01$ & $82.54$ & $81.77$ & $81.96$ \\
\cmidrule(lr){2-13}
 & \multirow{1}{*}{$\mathcal{A}^{t}_{f}(\%)$} & Original & $82.30$ & $90.90$ & $75.90$ & $64.00$ & $82.80$ & $77.90$ & $88.80$ & $84.00$ & $91.00$ & $89.30$ \\
\midrule
\multirow{9}{*}{Retrained} & \multirow{3}{*}{$\mathcal{A}^{t}_{r}(\%)$} & Unlearned & $82.66$ & $80.93$ & $84.06$ & $85.31$ & $82.58$ & $84.23$ & $81.42$ & $81.97$ & $81.80$ & $81.11$ \\
 &  & PRA \cite{ha2025unlearning} & $82.74$ & $80.83$ & $84.17$ & $85.38$ & $82.71$ & $84.16$ & $81.64$ & $82.19$ & $81.97$ & $81.21$ \\
 &  & SFRA (ours) & $78.74$ & $76.97$ & $80.10$ & $81.32$ & $78.64$ & $80.07$ & $77.57$ & $78.13$ & $78.11$ & $77.26$ \\
\cmidrule(lr){2-13}
 & \multirow{3}{*}{$\mathcal{A}^{t}_{f}(\%)$} & Unlearned & $0.00$ & $0.00$ & $0.00$ & $0.00$ & $0.00$ & $0.00$ & $0.00$ & $0.00$ & $0.00$ & $0.00$ \\
 &  & PRA \cite{ha2025unlearning} & $7.30$ & $1.20$ & $3.40$ & $0.30$ & $2.30$ & $0.50$ & $0.30$ & $3.80$ & $2.10$ & $0.70$ \\
 &  & SFRA (ours) & $27.50$ & $15.90$ & $27.80$ & $34.50$ & $20.80$ & $13.60$ & $17.60$ & $32.70$ & $16.60$ & $16.50$ \\
\cmidrule(lr){2-13}
 & \multirow{1}{*}{$\mathcal{A}^{LP}_{f}(\%)$} & Linear Probe & $75.20$ & $78.80$ & $57.80$ & $46.60$ & $64.40$ & $54.00$ & $75.90$ & $59.80$ & $73.60$ & $70.40$ \\
\cmidrule(lr){2-13}
 & \multirow{2}{*}{$\mathrm{RS}$} & PRA \cite{ha2025unlearning} & $0.14$ & $0.02$ & $0.07$ & $0.01$ & $0.04$ & $0.01$ & $0.01$ & $0.07$ & $0.04$ & $0.01$ \\
 &  & SFRA (ours) & $0.43$ & $0.27$ & $0.43$ & $0.51$ & $0.34$ & $0.24$ & $0.30$ & $0.49$ & $0.28$ & $0.28$ \\
\midrule
\multirow{11}{*}{Finetune \cite{golatkar2020eternal}} & \multirow{3}{*}{$\mathcal{A}^{t}_{r}(\%)$} & Unlearned & $89.86$ & $90.07$ & $90.87$ & $92.14$ & $90.11$ & $91.44$ & $89.58$ & $89.90$ & $87.52$ & $90.20$ \\
 &  & PRA \cite{ha2025unlearning} & $89.86$ & $90.07$ & $90.87$ & $92.14$ & $90.11$ & $91.44$ & $89.58$ & $89.90$ & $87.52$ & $90.20$ \\
 &  & SFRA (ours) & $82.79$ & $82.70$ & $85.32$ & $86.34$ & $85.44$ & $85.79$ & $80.74$ & $83.97$ & $78.93$ & $84.11$ \\
\cmidrule(lr){2-13}
 & \multirow{3}{*}{$\mathcal{A}^{t}_{f}(\%)$} & Unlearned & $0.00$ & $0.00$ & $0.00$ & $0.00$ & $0.00$ & $0.00$ & $0.00$ & $0.00$ & $0.00$ & $0.00$ \\
 &  & PRA \cite{ha2025unlearning} & $0.20$ & $0.00$ & $0.10$ & $0.00$ & $0.00$ & $0.00$ & $0.10$ & $0.00$ & $0.10$ & $0.00$ \\
 &  & SFRA (ours) & $52.30$ & $26.40$ & $52.00$ & $50.40$ & $42.00$ & $26.10$ & $56.70$ & $43.60$ & $42.70$ & $28.50$ \\
\cmidrule(lr){2-13}
 & \multirow{1}{*}{$\mathcal{A}^{LP}_{f}(\%)$} & Linear Probe & $84.80$ & $87.10$ & $75.00$ & $67.80$ & $85.50$ & $70.60$ & $83.10$ & $81.90$ & $81.40$ & $88.70$ \\
\cmidrule(lr){2-13}
 & \multirow{2}{*}{$\mathrm{RS}$} & PRA \cite{ha2025unlearning} & $0.00$ & $0.00$ & $0.00$ & $0.00$ & $0.00$ & $0.00$ & $0.00$ & $0.00$ & $0.00$ & $0.00$ \\
 &  & SFRA (ours) & $0.67$ & $0.41$ & $0.67$ & $0.66$ & $0.58$ & $0.41$ & $0.70$ & $0.60$ & $0.58$ & $0.44$ \\
\cmidrule(lr){2-13}
 & \multirow{2}{*}{$\Delta\mathrm{RS}$} & PRA \cite{ha2025unlearning} & $-0.13$ & $-0.02$ & $-0.06$ & $-0.01$ & $-0.04$ & $-0.01$ & $-0.00$ & $-0.07$ & $-0.04$ & $-0.01$ \\
 &  & SFRA (ours) & $+0.24$ & $+0.14$ & $+0.24$ & $+0.15$ & $+0.24$ & $+0.17$ & $+0.40$ & $+0.11$ & $+0.30$ & $+0.16$ \\
\midrule
\multirow{11}{*}{Negative Gradient \cite{golatkar2020eternal}} & \multirow{3}{*}{$\mathcal{A}^{t}_{r}(\%)$} & Unlearned & $78.49$ & $77.40$ & $78.26$ & $85.39$ & $78.74$ & $81.67$ & $79.74$ & $78.46$ & $79.73$ & $76.22$ \\
 &  & PRA \cite{ha2025unlearning} & $78.48$ & $77.39$ & $78.23$ & $85.39$ & $78.56$ & $81.33$ & $79.69$ & $78.44$ & $79.69$ & $76.22$ \\
 &  & SFRA (ours) & $73.97$ & $73.88$ & $74.68$ & $79.22$ & $71.82$ & $76.39$ & $74.09$ & $74.72$ & $73.70$ & $71.04$ \\
\cmidrule(lr){2-13}
 & \multirow{3}{*}{$\mathcal{A}^{t}_{f}(\%)$} & Unlearned & $1.50$ & $1.60$ & $1.80$ & $0.50$ & $1.80$ & $0.60$ & $2.00$ & $2.00$ & $0.40$ & $1.90$ \\
 &  & PRA \cite{ha2025unlearning} & $4.30$ & $0.40$ & $1.90$ & $0.10$ & $1.10$ & $5.50$ & $0.50$ & $2.30$ & $0.60$ & $0.10$ \\
 &  & SFRA (ours) & $51.90$ & $18.90$ & $31.40$ & $48.10$ & $69.50$ & $43.40$ & $29.10$ & $36.90$ & $27.30$ & $23.40$ \\
\cmidrule(lr){2-13}
 & \multirow{1}{*}{$\mathcal{A}^{LP}_{f}(\%)$} & Linear Probe & $72.40$ & $80.20$ & $64.10$ & $60.40$ & $71.70$ & $64.90$ & $82.20$ & $75.20$ & $75.50$ & $78.60$ \\
\cmidrule(lr){2-13}
 & \multirow{2}{*}{$\mathrm{RS}$} & PRA \cite{ha2025unlearning} & $0.05$ & $0.00$ & $0.00$ & $0.00$ & $0.00$ & $0.09$ & $0.00$ & $0.01$ & $0.00$ & $0.00$ \\
 &  & SFRA (ours) & $0.66$ & $0.29$ & $0.45$ & $0.63$ & $0.78$ & $0.59$ & $0.42$ & $0.51$ & $0.42$ & $0.35$ \\
\cmidrule(lr){2-13}
 & \multirow{2}{*}{$\Delta\mathrm{RS}$} & PRA \cite{ha2025unlearning} & $-0.08$ & $-0.02$ & $-0.06$ & $-0.01$ & $-0.04$ & $+0.08$ & $-0.01$ & $-0.07$ & $-0.04$ & $-0.01$ \\
 &  & SFRA (ours) & $+0.23$ & $+0.02$ & $+0.02$ & $+0.12$ & $+0.44$ & $+0.35$ & $+0.12$ & $+0.02$ & $+0.14$ & $+0.07$ \\
\midrule
\multirow{11}{*}{Negative Gradient+ \cite{kurmanji2023towards}} & \multirow{3}{*}{$\mathcal{A}^{t}_{r}(\%)$} & Unlearned & $81.96$ & $80.33$ & $83.17$ & $85.91$ & $82.70$ & $84.27$ & $80.77$ & $81.89$ & $81.19$ & $80.53$ \\
 &  & PRA \cite{ha2025unlearning} & $81.93$ & $80.16$ & $83.01$ & $85.90$ & $82.32$ & $83.80$ & $80.18$ & $81.67$ & $81.06$ & $80.53$ \\
 &  & SFRA (ours) & $76.43$ & $74.44$ & $79.82$ & $80.81$ & $74.66$ & $79.17$ & $76.11$ & $77.47$ & $75.07$ & $73.20$ \\
\cmidrule(lr){2-13}
 & \multirow{3}{*}{$\mathcal{A}^{t}_{f}(\%)$} & Unlearned & $0.30$ & $0.40$ & $0.50$ & $0.20$ & $0.40$ & $0.20$ & $0.30$ & $0.50$ & $0.50$ & $0.40$ \\
 &  & PRA \cite{ha2025unlearning} & $4.80$ & $1.00$ & $4.20$ & $0.40$ & $4.00$ & $5.10$ & $3.40$ & $4.60$ & $0.70$ & $0.10$ \\
 &  & SFRA (ours) & $48.80$ & $14.50$ & $32.40$ & $40.70$ & $60.50$ & $27.90$ & $25.60$ & $34.40$ & $26.40$ & $20.20$ \\
\cmidrule(lr){2-13}
 & \multirow{1}{*}{$\mathcal{A}^{LP}_{f}(\%)$} & Linear Probe & $72.00$ & $78.90$ & $61.50$ & $58.20$ & $69.70$ & $66.90$ & $81.80$ & $73.40$ & $77.20$ & $71.50$ \\
\cmidrule(lr){2-13}
 & \multirow{2}{*}{$\mathrm{RS}$} & PRA \cite{ha2025unlearning} & $0.09$ & $0.01$ & $0.07$ & $0.00$ & $0.07$ & $0.09$ & $0.06$ & $0.08$ & $0.00$ & $0.00$ \\
 &  & SFRA (ours) & $0.64$ & $0.25$ & $0.48$ & $0.57$ & $0.73$ & $0.43$ & $0.40$ & $0.50$ & $0.41$ & $0.33$ \\
\cmidrule(lr){2-13}
 & \multirow{2}{*}{$\Delta\mathrm{RS}$} & PRA \cite{ha2025unlearning} & $-0.05$ & $-0.01$ & $+0.01$ & $-0.00$ & $+0.02$ & $+0.08$ & $+0.05$ & $+0.01$ & $-0.04$ & $-0.01$ \\
 &  & SFRA (ours) & $+0.21$ & $-0.03$ & $+0.05$ & $+0.06$ & $+0.38$ & $+0.19$ & $+0.10$ & $+0.01$ & $+0.12$ & $+0.04$ \\
\midrule
\multirow{11}{*}{Random Label \cite{hayase2020selective}} & \multirow{3}{*}{$\mathcal{A}^{t}_{r}(\%)$} & Unlearned & $76.73$ & $73.03$ & $72.77$ & $76.59$ & $71.82$ & $72.60$ & $74.57$ & $71.87$ & $75.39$ & $73.93$ \\
 &  & PRA \cite{ha2025unlearning} & $75.81$ & $72.76$ & $71.93$ & $75.66$ & $71.57$ & $71.59$ & $74.21$ & $71.80$ & $74.34$ & $73.82$ \\
 &  & SFRA (ours) & $70.30$ & $66.14$ & $67.87$ & $69.92$ & $66.77$ & $65.52$ & $67.31$ & $64.94$ & $67.98$ & $66.63$ \\
\cmidrule(lr){2-13}
 & \multirow{3}{*}{$\mathcal{A}^{t}_{f}(\%)$} & Unlearned & $2.20$ & $11.80$ & $8.90$ & $11.80$ & $7.00$ & $9.40$ & $9.00$ & $8.70$ & $12.40$ & $15.10$ \\
 &  & PRA \cite{ha2025unlearning} & $60.10$ & $26.50$ & $36.80$ & $29.10$ & $19.00$ & $23.20$ & $27.70$ & $15.40$ & $49.70$ & $25.90$ \\
 &  & SFRA (ours) & $59.70$ & $96.00$ & $58.30$ & $74.60$ & $53.80$ & $80.70$ & $88.40$ & $84.80$ & $85.80$ & $87.70$ \\
\cmidrule(lr){2-13}
 & \multirow{1}{*}{$\mathcal{A}^{LP}_{f}(\%)$} & Linear Probe & $78.50$ & $83.00$ & $66.80$ & $67.20$ & $74.80$ & $64.10$ & $84.80$ & $75.60$ & $83.90$ & $81.40$ \\
\cmidrule(lr){2-13}
 & \multirow{2}{*}{$\mathrm{RS}$} & PRA \cite{ha2025unlearning} & $0.73$ & $0.26$ & $0.44$ & $0.29$ & $0.21$ & $0.24$ & $0.31$ & $0.13$ & $0.54$ & $0.19$ \\
 &  & SFRA (ours) & $0.71$ & $0.88$ & $0.65$ & $0.75$ & $0.63$ & $0.81$ & $0.86$ & $0.84$ & $0.82$ & $0.81$ \\
\cmidrule(lr){2-13}
 & \multirow{2}{*}{$\Delta\mathrm{RS}$} & PRA \cite{ha2025unlearning} & $+0.59$ & $+0.23$ & $+0.37$ & $+0.29$ & $+0.17$ & $+0.23$ & $+0.31$ & $+0.05$ & $+0.50$ & $+0.18$ \\
 &  & SFRA (ours) & $+0.28$ & $+0.61$ & $+0.22$ & $+0.24$ & $+0.29$ & $+0.57$ & $+0.56$ & $+0.35$ & $+0.54$ & $+0.53$ \\
\bottomrule
\end{tabular}
}
\end{minipage}%
\hfill
\begin{minipage}[t]{0.49\textwidth}
\centering
\scriptsize
\setlength{\tabcolsep}{2pt}
\renewcommand{\arraystretch}{0.80}
\resizebox{\columnwidth}{!}{%
\begin{tabular}{l|l|l|cccccccccc}
\toprule
\multirow{2}{*}{Unlearning Method} & \multirow{2}{*}{Metric} & \multirow{2}{*}{Variant} & \multicolumn{10}{c}{Forget Class} \\
 & & & 0 & 1 & 2 & 3 & 4 & 5 & 6 & 7 & 8 & 9 \\
\midrule
\multirow{11}{*}{Learn to Unlearn \cite{cha2024learning}} & \multirow{3}{*}{$\mathcal{A}^{t}_{r}(\%)$} & Unlearned & $78.51$ & $77.21$ & $79.36$ & $84.51$ & $78.79$ & $80.54$ & $79.47$ & $79.07$ & $78.62$ & $75.31$ \\
 &  & PRA \cite{ha2025unlearning} & $78.49$ & $77.13$ & $79.33$ & $84.50$ & $78.60$ & $80.10$ & $79.36$ & $79.04$ & $77.90$ & $75.31$ \\
 &  & SFRA (ours) & $74.72$ & $74.64$ & $75.88$ & $78.88$ & $70.98$ & $76.59$ & $74.57$ & $74.99$ & $72.91$ & $68.66$ \\
\cmidrule(lr){2-13}
 & \multirow{3}{*}{$\mathcal{A}^{t}_{f}(\%)$} & Unlearned & $1.50$ & $2.70$ & $1.60$ & $0.30$ & $2.00$ & $0.80$ & $1.80$ & $2.30$ & $0.30$ & $1.00$ \\
 &  & PRA \cite{ha2025unlearning} & $5.20$ & $1.00$ & $1.40$ & $0.20$ & $1.20$ & $8.20$ & $1.50$ & $2.30$ & $4.90$ & $0.90$ \\
 &  & SFRA (ours) & $59.30$ & $34.40$ & $37.60$ & $53.70$ & $71.10$ & $56.10$ & $41.40$ & $46.60$ & $40.10$ & $40.60$ \\
\cmidrule(lr){2-13}
 & \multirow{1}{*}{$\mathcal{A}^{LP}_{f}(\%)$} & Linear Probe & $69.80$ & $80.60$ & $64.00$ & $61.80$ & $70.70$ & $65.30$ & $83.00$ & $74.50$ & $75.00$ & $76.20$ \\
\cmidrule(lr){2-13}
 & \multirow{2}{*}{$\mathrm{RS}$} & PRA \cite{ha2025unlearning} & $0.07$ & $0.00$ & $0.00$ & $0.00$ & $0.00$ & $0.14$ & $0.00$ & $0.00$ & $0.09$ & $0.00$ \\
 &  & SFRA (ours) & $0.72$ & $0.48$ & $0.52$ & $0.68$ & $0.79$ & $0.70$ & $0.56$ & $0.61$ & $0.56$ & $0.56$ \\
\cmidrule(lr){2-13}
 & \multirow{2}{*}{$\Delta\mathrm{RS}$} & PRA \cite{ha2025unlearning} & $-0.06$ & $-0.02$ & $-0.07$ & $-0.01$ & $-0.04$ & $+0.13$ & $-0.01$ & $-0.07$ & $+0.05$ & $-0.01$ \\
 &  & SFRA (ours) & $+0.29$ & $+0.21$ & $+0.09$ & $+0.17$ & $+0.45$ & $+0.46$ & $+0.26$ & $+0.12$ & $+0.28$ & $+0.27$ \\
\midrule
\multirow{11}{*}{SCRUB \cite{kurmanji2023towards}} & \multirow{3}{*}{$\mathcal{A}^{t}_{r}(\%)$} & Unlearned & $83.27$ & $81.89$ & $84.32$ & $86.30$ & $83.27$ & $84.64$ & $82.06$ & $82.03$ & $82.40$ & $82.29$ \\
 &  & PRA \cite{ha2025unlearning} & $83.27$ & $81.81$ & $84.21$ & $86.30$ & $82.60$ & $84.40$ & $82.06$ & $82.02$ & $82.32$ & $82.29$ \\
 &  & SFRA (ours) & $79.16$ & $77.90$ & $80.12$ & $82.07$ & $79.13$ & $80.56$ & $78.01$ & $78.04$ & $78.33$ & $78.27$ \\
\cmidrule(lr){2-13}
 & \multirow{3}{*}{$\mathcal{A}^{t}_{f}(\%)$} & Unlearned & $0.70$ & $0.00$ & $0.00$ & $0.00$ & $0.00$ & $0.10$ & $0.00$ & $0.00$ & $0.80$ & $0.20$ \\
 &  & PRA \cite{ha2025unlearning} & $3.10$ & $0.90$ & $2.30$ & $0.10$ & $9.70$ & $3.00$ & $0.30$ & $0.50$ & $2.10$ & $0.10$ \\
 &  & SFRA (ours) & $48.90$ & $9.80$ & $30.70$ & $37.10$ & $18.40$ & $15.90$ & $22.10$ & $27.40$ & $23.40$ & $22.70$ \\
\cmidrule(lr){2-13}
 & \multirow{1}{*}{$\mathcal{A}^{LP}_{f}(\%)$} & Linear Probe & $70.70$ & $67.80$ & $56.00$ & $62.50$ & $57.40$ & $58.70$ & $70.70$ & $70.30$ & $75.80$ & $71.40$ \\
\cmidrule(lr){2-13}
 & \multirow{2}{*}{$\mathrm{RS}$} & PRA \cite{ha2025unlearning} & $0.05$ & $0.02$ & $0.04$ & $0.00$ & $0.18$ & $0.06$ & $0.01$ & $0.01$ & $0.03$ & $0.00$ \\
 &  & SFRA (ours) & $0.64$ & $0.18$ & $0.46$ & $0.53$ & $0.31$ & $0.27$ & $0.36$ & $0.43$ & $0.37$ & $0.36$ \\
\cmidrule(lr){2-13}
 & \multirow{2}{*}{$\Delta\mathrm{RS}$} & PRA \cite{ha2025unlearning} & $-0.09$ & $-0.01$ & $-0.02$ & $-0.00$ & $+0.13$ & $+0.05$ & $+0.00$ & $-0.06$ & $-0.02$ & $-0.01$ \\
 &  & SFRA (ours) & $+0.21$ & $-0.09$ & $+0.03$ & $+0.03$ & $-0.03$ & $+0.03$ & $+0.06$ & $-0.06$ & $+0.08$ & $+0.08$ \\
\midrule
\multirow{11}{*}{Bad Teacher \cite{chundawat2023can}} & \multirow{3}{*}{$\mathcal{A}^{t}_{r}(\%)$} & Unlearned & $81.89$ & $79.17$ & $82.16$ & $83.86$ & $81.41$ & $81.77$ & $80.78$ & $79.69$ & $79.72$ & $81.00$ \\
 &  & PRA \cite{ha2025unlearning} & $81.22$ & $78.48$ & $81.62$ & $83.56$ & $81.18$ & $81.01$ & $80.30$ & $79.22$ & $79.10$ & $80.77$ \\
 &  & SFRA (ours) & $74.56$ & $73.42$ & $74.78$ & $78.46$ & $75.27$ & $76.48$ & $74.47$ & $75.36$ & $75.06$ & $76.18$ \\
\cmidrule(lr){2-13}
 & \multirow{3}{*}{$\mathcal{A}^{t}_{f}(\%)$} & Unlearned & $0.20$ & $10.70$ & $9.70$ & $6.30$ & $1.90$ & $5.60$ & $12.70$ & $16.00$ & $1.40$ & $1.10$ \\
 &  & PRA \cite{ha2025unlearning} & $73.30$ & $77.90$ & $43.80$ & $21.20$ & $23.70$ & $47.00$ & $70.90$ & $65.10$ & $76.60$ & $58.40$ \\
 &  & SFRA (ours) & $96.60$ & $97.20$ & $87.00$ & $71.20$ & $90.50$ & $88.70$ & $96.70$ & $88.10$ & $93.30$ & $97.50$ \\
\cmidrule(lr){2-13}
 & \multirow{1}{*}{$\mathcal{A}^{LP}_{f}(\%)$} & Linear Probe & $89.00$ & $92.60$ & $76.00$ & $69.20$ & $81.70$ & $80.70$ & $92.00$ & $84.10$ & $88.90$ & $94.00$ \\
\cmidrule(lr){2-13}
 & \multirow{2}{*}{$\mathrm{RS}$} & PRA \cite{ha2025unlearning} & $0.84$ & $0.80$ & $0.51$ & $0.26$ & $0.36$ & $0.58$ & $0.73$ & $0.66$ & $0.86$ & $0.73$ \\
 &  & SFRA (ours) & $0.94$ & $0.90$ & $0.84$ & $0.77$ & $0.91$ & $0.89$ & $0.89$ & $0.82$ & $0.94$ & $0.96$ \\
\cmidrule(lr){2-13}
 & \multirow{2}{*}{$\Delta\mathrm{RS}$} & PRA \cite{ha2025unlearning} & $+0.71$ & $+0.78$ & $+0.44$ & $+0.25$ & $+0.31$ & $+0.57$ & $+0.73$ & $+0.58$ & $+0.82$ & $+0.71$ \\
 &  & SFRA (ours) & $+0.52$ & $+0.63$ & $+0.41$ & $+0.26$ & $+0.57$ & $+0.65$ & $+0.59$ & $+0.33$ & $+0.65$ & $+0.68$ \\
\midrule
\multirow{11}{*}{SalUn \cite{fan2023salun}} & \multirow{3}{*}{$\mathcal{A}^{t}_{r}(\%)$} & Unlearned & $83.28$ & $81.59$ & $84.44$ & $86.83$ & $83.71$ & $84.97$ & $82.30$ & $82.78$ & $81.81$ & $82.18$ \\
 &  & PRA \cite{ha2025unlearning} & $83.04$ & $81.24$ & $83.87$ & $86.26$ & $83.19$ & $84.83$ & $82.18$ & $82.58$ & $81.27$ & $82.12$ \\
 &  & SFRA (ours) & $75.01$ & $73.51$ & $78.80$ & $78.64$ & $75.34$ & $76.59$ & $74.12$ & $74.58$ & $73.72$ & $73.98$ \\
\cmidrule(lr){2-13}
 & \multirow{3}{*}{$\mathcal{A}^{t}_{f}(\%)$} & Unlearned & $1.50$ & $1.70$ & $4.20$ & $0.80$ & $3.60$ & $3.00$ & $2.60$ & $1.30$ & $1.40$ & $3.20$ \\
 &  & PRA \cite{ha2025unlearning} & $24.00$ & $7.20$ & $52.60$ & $38.20$ & $6.20$ & $4.00$ & $4.50$ & $13.40$ & $20.90$ & $6.30$ \\
 &  & SFRA (ours) & $80.90$ & $67.20$ & $71.40$ & $81.20$ & $73.80$ & $61.20$ & $77.20$ & $76.20$ & $72.50$ & $75.40$ \\
\cmidrule(lr){2-13}
 & \multirow{1}{*}{$\mathcal{A}^{LP}_{f}(\%)$} & Linear Probe & $77.20$ & $80.20$ & $69.70$ & $67.00$ & $75.60$ & $64.30$ & $81.70$ & $77.00$ & $80.90$ & $80.00$ \\
\cmidrule(lr){2-13}
 & \multirow{2}{*}{$\mathrm{RS}$} & PRA \cite{ha2025unlearning} & $0.37$ & $0.10$ & $0.65$ & $0.54$ & $0.05$ & $0.02$ & $0.04$ & $0.22$ & $0.33$ & $0.06$ \\
 &  & SFRA (ours) & $0.85$ & $0.76$ & $0.78$ & $0.86$ & $0.79$ & $0.71$ & $0.82$ & $0.82$ & $0.80$ & $0.81$ \\
\cmidrule(lr){2-13}
 & \multirow{2}{*}{$\Delta\mathrm{RS}$} & PRA \cite{ha2025unlearning} & $+0.23$ & $+0.08$ & $+0.59$ & $+0.54$ & $+0.01$ & $+0.01$ & $+0.03$ & $+0.14$ & $+0.28$ & $+0.05$ \\
 &  & SFRA (ours) & $+0.42$ & $+0.49$ & $+0.35$ & $+0.35$ & $+0.45$ & $+0.47$ & $+0.53$ & $+0.34$ & $+0.52$ & $+0.53$ \\
\midrule
\multirow{11}{*}{DELETE \cite{zhou2025decoupled}} & \multirow{3}{*}{$\mathcal{A}^{t}_{r}(\%)$} & Unlearned & $83.20$ & $81.94$ & $84.54$ & $86.38$ & $83.92$ & $84.99$ & $82.60$ & $82.78$ & $82.02$ & $81.60$ \\
 &  & PRA \cite{ha2025unlearning} & $82.86$ & $81.28$ & $84.37$ & $86.02$ & $83.36$ & $84.66$ & $82.40$ & $82.54$ & $81.73$ & $80.88$ \\
 &  & SFRA (ours) & $74.96$ & $73.81$ & $77.11$ & $80.00$ & $77.49$ & $78.10$ & $75.44$ & $76.58$ & $73.82$ & $73.58$ \\
\cmidrule(lr){2-13}
 & \multirow{3}{*}{$\mathcal{A}^{t}_{f}(\%)$} & Unlearned & $0.00$ & $0.00$ & $0.00$ & $0.00$ & $0.00$ & $0.00$ & $0.00$ & $0.00$ & $0.00$ & $0.00$ \\
 &  & PRA \cite{ha2025unlearning} & $10.30$ & $6.50$ & $8.30$ & $4.60$ & $7.20$ & $2.90$ & $6.90$ & $6.70$ & $8.50$ & $13.80$ \\
 &  & SFRA (ours) & $53.20$ & $19.00$ & $49.40$ & $38.80$ & $31.70$ & $13.40$ & $33.40$ & $33.00$ & $31.20$ & $25.30$ \\
\cmidrule(lr){2-13}
 & \multirow{1}{*}{$\mathcal{A}^{LP}_{f}(\%)$} & Linear Probe & $79.30$ & $80.80$ & $65.60$ & $68.80$ & $73.50$ & $62.40$ & $83.20$ & $78.30$ & $83.10$ & $79.60$ \\
\cmidrule(lr){2-13}
 & \multirow{2}{*}{$\mathrm{RS}$} & PRA \cite{ha2025unlearning} & $0.19$ & $0.12$ & $0.15$ & $0.09$ & $0.13$ & $0.06$ & $0.13$ & $0.13$ & $0.16$ & $0.24$ \\
 &  & SFRA (ours) & $0.67$ & $0.31$ & $0.64$ & $0.55$ & $0.47$ & $0.23$ & $0.49$ & $0.49$ & $0.47$ & $0.40$ \\
\cmidrule(lr){2-13}
 & \multirow{2}{*}{$\Delta\mathrm{RS}$} & PRA \cite{ha2025unlearning} & $+0.05$ & $+0.10$ & $+0.09$ & $+0.08$ & $+0.09$ & $+0.05$ & $+0.12$ & $+0.05$ & $+0.12$ & $+0.23$ \\
 &  & SFRA (ours) & $+0.25$ & $+0.04$ & $+0.21$ & $+0.04$ & $+0.13$ & $-0.00$ & $+0.19$ & $+0.00$ & $+0.18$ & $+0.12$ \\
\bottomrule
\end{tabular}
}
\end{minipage}%

\end{table*}

\begin{table*}[t]
\centering
\caption{Per-forget-class single-class unlearning and relearning results on CIFAR-100 using Swin-T. We report the unlearned checkpoint, the source-dependent PRA baseline, our proposed SFRA, and frozen-encoder linear probing. Each forget class column corresponds to a separate unlearned checkpoint in which that class is designated for forgetting.}
\label{tab:cifar100_swin-t_variant_by_forget}
\begin{minipage}[t]{0.49\textwidth}
\centering
\scriptsize
\setlength{\tabcolsep}{2pt}
\renewcommand{\arraystretch}{0.80}
\resizebox{\columnwidth}{!}{%
\begin{tabular}{l|l|l|cccccccccc}
\toprule
\multirow{2}{*}{Unlearning Method} & \multirow{2}{*}{Metric} & \multirow{2}{*}{Variant} & \multicolumn{10}{c}{Forget Class} \\
 & & & 0 & 10 & 20 & 30 & 40 & 50 & 60 & 70 & 80 & 90 \\
\midrule
\multirow{2}{*}{Original} & \multirow{1}{*}{$\mathcal{A}^{t}_{r}(\%)$} & Original & $88.27$ & $88.48$ & $88.31$ & $88.34$ & $88.32$ & $88.41$ & $88.31$ & $88.29$ & $88.43$ & $88.27$ \\
\cmidrule(lr){2-13}
 & \multirow{1}{*}{$\mathcal{A}^{t}_{f}(\%)$} & Original & $95.00$ & $74.00$ & $91.00$ & $88.00$ & $90.00$ & $81.00$ & $91.00$ & $93.00$ & $79.00$ & $95.00$ \\
\midrule
\multirow{9}{*}{Retrained} & \multirow{3}{*}{$\mathcal{A}^{t}_{r}(\%)$} & Unlearned & $88.36$ & $88.67$ & $88.40$ & $88.52$ & $88.48$ & $88.40$ & $88.43$ & $88.25$ & $88.53$ & $88.30$ \\
 &  & PRA \cite{ha2025unlearning} & $87.35$ & $87.90$ & $88.13$ & $87.71$ & $87.34$ & $88.10$ & $88.04$ & $87.66$ & $87.55$ & $87.67$ \\
 &  & SFRA (ours) & $86.85$ & $85.81$ & $85.85$ & $86.30$ & $84.97$ & $85.31$ & $85.11$ & $86.38$ & $84.96$ & $85.50$ \\
\cmidrule(lr){2-13}
 & \multirow{3}{*}{$\mathcal{A}^{t}_{f}(\%)$} & Unlearned & $0.00$ & $0.00$ & $0.00$ & $0.00$ & $0.00$ & $0.00$ & $0.00$ & $0.00$ & $0.00$ & $0.00$ \\
 &  & PRA \cite{ha2025unlearning} & $95.00$ & $60.00$ & $81.00$ & $62.00$ & $85.00$ & $63.00$ & $47.00$ & $61.00$ & $80.00$ & $86.00$ \\
 &  & SFRA (ours) & $4.00$ & $29.00$ & $29.00$ & $6.00$ & $73.00$ & $36.00$ & $28.00$ & $23.00$ & $49.00$ & $11.00$ \\
\cmidrule(lr){2-13}
 & \multirow{1}{*}{$\mathcal{A}^{LP}_{f}(\%)$} & Linear Probe & $98.00$ & $79.00$ & $86.00$ & $76.00$ & $91.00$ & $80.00$ & $87.00$ & $84.00$ & $79.00$ & $87.00$ \\
\cmidrule(lr){2-13}
 & \multirow{2}{*}{$\mathrm{RS}$} & PRA \cite{ha2025unlearning} & $0.97$ & $0.75$ & $0.89$ & $0.76$ & $0.91$ & $0.77$ & $0.64$ & $0.76$ & $0.88$ & $0.92$ \\
 &  & SFRA (ours) & $0.08$ & $0.45$ & $0.45$ & $0.11$ & $0.83$ & $0.52$ & $0.43$ & $0.37$ & $0.65$ & $0.20$ \\
\midrule
\multirow{11}{*}{Finetune \cite{golatkar2020eternal}} & \multirow{3}{*}{$\mathcal{A}^{t}_{r}(\%)$} & Unlearned & $87.91$ & $85.60$ & $86.36$ & $86.86$ & $87.00$ & $87.24$ & $87.82$ & $87.18$ & $86.93$ & $86.91$ \\
 &  & PRA \cite{ha2025unlearning} & $87.88$ & $84.78$ & $86.26$ & $85.82$ & $86.98$ & $87.07$ & $87.54$ & $87.00$ & $86.91$ & $86.88$ \\
 &  & SFRA (ours) & $87.06$ & $84.87$ & $85.87$ & $86.86$ & $86.13$ & $84.74$ & $87.21$ & $86.17$ & $86.22$ & $86.39$ \\
\cmidrule(lr){2-13}
 & \multirow{3}{*}{$\mathcal{A}^{t}_{f}(\%)$} & Unlearned & $0.00$ & $0.00$ & $9.00$ & $0.00$ & $1.00$ & $0.00$ & $0.00$ & $0.00$ & $2.00$ & $12.00$ \\
 &  & PRA \cite{ha2025unlearning} & $79.00$ & $72.00$ & $87.00$ & $80.00$ & $51.00$ & $49.00$ & $72.00$ & $61.00$ & $46.00$ & $81.00$ \\
 &  & SFRA (ours) & $24.00$ & $11.00$ & $22.00$ & $0.00$ & $23.00$ & $2.00$ & $9.00$ & $2.00$ & $12.00$ & $41.00$ \\
\cmidrule(lr){2-13}
 & \multirow{1}{*}{$\mathcal{A}^{LP}_{f}(\%)$} & Linear Probe & $97.00$ & $74.00$ & $89.00$ & $84.00$ & $90.00$ & $79.00$ & $91.00$ & $87.00$ & $87.00$ & $95.00$ \\
\cmidrule(lr){2-13}
 & \multirow{2}{*}{$\mathrm{RS}$} & PRA \cite{ha2025unlearning} & $0.88$ & $0.83$ & $0.88$ & $0.88$ & $0.67$ & $0.66$ & $0.84$ & $0.76$ & $0.61$ & $0.82$ \\
 &  & SFRA (ours) & $0.39$ & $0.20$ & $0.23$ & $0.00$ & $0.36$ & $0.04$ & $0.17$ & $0.04$ & $0.18$ & $0.45$ \\
\cmidrule(lr){2-13}
 & \multirow{2}{*}{$\Delta\mathrm{RS}$} & PRA \cite{ha2025unlearning} & $-0.09$ & $+0.09$ & $-0.02$ & $+0.12$ & $-0.25$ & $-0.11$ & $+0.20$ & $+0.00$ & $-0.27$ & $-0.10$ \\
 &  & SFRA (ours) & $+0.31$ & $-0.25$ & $-0.22$ & $-0.11$ & $-0.47$ & $-0.49$ & $-0.27$ & $-0.33$ & $-0.47$ & $+0.25$ \\
\midrule
\multirow{11}{*}{Negative Gradient \cite{golatkar2020eternal}} & \multirow{3}{*}{$\mathcal{A}^{t}_{r}(\%)$} & Unlearned & $86.58$ & $86.61$ & $86.44$ & $86.11$ & $83.53$ & $84.64$ & $87.26$ & $82.72$ & $86.60$ & $87.44$ \\
 &  & PRA \cite{ha2025unlearning} & $86.53$ & $86.60$ & $86.14$ & $85.14$ & $82.98$ & $83.39$ & $87.10$ & $82.53$ & $85.15$ & $86.63$ \\
 &  & SFRA (ours) & $78.53$ & $78.11$ & $78.53$ & $78.11$ & $75.83$ & $76.38$ & $78.57$ & $77.26$ & $78.29$ & $80.75$ \\
\cmidrule(lr){2-13}
 & \multirow{3}{*}{$\mathcal{A}^{t}_{f}(\%)$} & Unlearned & $0.00$ & $0.00$ & $2.00$ & $0.00$ & $0.00$ & $0.00$ & $1.00$ & $0.00$ & $2.00$ & $2.00$ \\
 &  & PRA \cite{ha2025unlearning} & $48.00$ & $16.00$ & $82.00$ & $70.00$ & $51.00$ & $72.00$ & $48.00$ & $17.00$ & $79.00$ & $96.00$ \\
 &  & SFRA (ours) & $36.00$ & $55.00$ & $71.00$ & $41.00$ & $75.00$ & $90.00$ & $69.00$ & $7.00$ & $92.00$ & $49.00$ \\
\cmidrule(lr){2-13}
 & \multirow{1}{*}{$\mathcal{A}^{LP}_{f}(\%)$} & Linear Probe & $91.00$ & $67.00$ & $84.00$ & $74.00$ & $80.00$ & $73.00$ & $89.00$ & $82.00$ & $73.00$ & $84.00$ \\
\cmidrule(lr){2-13}
 & \multirow{2}{*}{$\mathrm{RS}$} & PRA \cite{ha2025unlearning} & $0.65$ & $0.28$ & $0.89$ & $0.82$ & $0.67$ & $0.83$ & $0.64$ & $0.29$ & $0.86$ & $0.97$ \\
 &  & SFRA (ours) & $0.52$ & $0.69$ & $0.79$ & $0.57$ & $0.83$ & $0.91$ & $0.78$ & $0.13$ & $0.91$ & $0.63$ \\
\cmidrule(lr){2-13}
 & \multirow{2}{*}{$\Delta\mathrm{RS}$} & PRA \cite{ha2025unlearning} & $-0.32$ & $-0.47$ & $-0.01$ & $+0.06$ & $-0.24$ & $+0.06$ & $+0.00$ & $-0.47$ & $-0.02$ & $+0.04$ \\
 &  & SFRA (ours) & $+0.44$ & $+0.24$ & $+0.34$ & $+0.45$ & $-0.00$ & $+0.38$ & $+0.35$ & $-0.24$ & $+0.26$ & $+0.43$ \\
\midrule
\multirow{11}{*}{Negative Gradient+ \cite{kurmanji2023towards}} & \multirow{3}{*}{$\mathcal{A}^{t}_{r}(\%)$} & Unlearned & $87.38$ & $86.72$ & $86.38$ & $87.81$ & $85.09$ & $85.65$ & $87.03$ & $85.62$ & $85.34$ & $86.77$ \\
 &  & PRA \cite{ha2025unlearning} & $87.37$ & $86.71$ & $85.95$ & $86.85$ & $84.87$ & $85.60$ & $86.86$ & $84.73$ & $85.29$ & $86.68$ \\
 &  & SFRA (ours) & $87.38$ & $86.28$ & $85.78$ & $87.41$ & $85.08$ & $85.68$ & $86.68$ & $85.55$ & $85.34$ & $86.84$ \\
\cmidrule(lr){2-13}
 & \multirow{3}{*}{$\mathcal{A}^{t}_{f}(\%)$} & Unlearned & $0.00$ & $0.00$ & $0.00$ & $0.00$ & $0.00$ & $0.00$ & $0.00$ & $0.00$ & $0.00$ & $1.00$ \\
 &  & PRA \cite{ha2025unlearning} & $35.00$ & $6.00$ & $87.00$ & $77.00$ & $55.00$ & $14.00$ & $63.00$ & $75.00$ & $14.00$ & $25.00$ \\
 &  & SFRA (ours) & $0.00$ & $3.00$ & $2.00$ & $2.00$ & $5.00$ & $1.00$ & $3.00$ & $1.00$ & $3.00$ & $1.00$ \\
\cmidrule(lr){2-13}
 & \multirow{1}{*}{$\mathcal{A}^{LP}_{f}(\%)$} & Linear Probe & $91.00$ & $68.00$ & $85.00$ & $75.00$ & $82.00$ & $74.00$ & $89.00$ & $79.00$ & $72.00$ & $88.00$ \\
\cmidrule(lr){2-13}
 & \multirow{2}{*}{$\mathrm{RS}$} & PRA \cite{ha2025unlearning} & $0.52$ & $0.11$ & $0.93$ & $0.87$ & $0.71$ & $0.25$ & $0.77$ & $0.85$ & $0.25$ & $0.39$ \\
 &  & SFRA (ours) & $0.00$ & $0.06$ & $0.04$ & $0.04$ & $0.10$ & $0.02$ & $0.06$ & $0.02$ & $0.06$ & $0.00$ \\
\cmidrule(lr){2-13}
 & \multirow{2}{*}{$\Delta\mathrm{RS}$} & PRA \cite{ha2025unlearning} & $-0.45$ & $-0.63$ & $+0.04$ & $+0.10$ & $-0.20$ & $-0.53$ & $+0.13$ & $+0.10$ & $-0.64$ & $-0.53$ \\
 &  & SFRA (ours) & $-0.08$ & $-0.39$ & $-0.41$ & $-0.07$ & $-0.74$ & $-0.51$ & $-0.38$ & $-0.35$ & $-0.59$ & $-0.20$ \\
\midrule
\multirow{11}{*}{Random Label \cite{hayase2020selective}} & \multirow{3}{*}{$\mathcal{A}^{t}_{r}(\%)$} & Unlearned & $87.79$ & $87.50$ & $87.03$ & $86.92$ & $86.98$ & $87.28$ & $87.82$ & $87.07$ & $86.57$ & $86.92$ \\
 &  & PRA \cite{ha2025unlearning} & $87.19$ & $86.72$ & $86.09$ & $85.60$ & $85.16$ & $85.90$ & $86.96$ & $86.58$ & $85.34$ & $86.11$ \\
 &  & SFRA (ours) & $83.87$ & $79.20$ & $83.36$ & $79.27$ & $81.08$ & $81.75$ & $80.85$ & $84.59$ & $79.08$ & $82.77$ \\
\cmidrule(lr){2-13}
 & \multirow{3}{*}{$\mathcal{A}^{t}_{f}(\%)$} & Unlearned & $0.00$ & $0.00$ & $0.00$ & $0.00$ & $1.00$ & $0.00$ & $0.00$ & $0.00$ & $0.00$ & $0.00$ \\
 &  & PRA \cite{ha2025unlearning} & $99.00$ & $85.00$ & $97.00$ & $89.00$ & $99.00$ & $90.00$ & $95.00$ & $98.00$ & $84.00$ & $95.00$ \\
 &  & SFRA (ours) & $100.00$ & $90.00$ & $100.00$ & $89.00$ & $100.00$ & $98.00$ & $96.00$ & $98.00$ & $95.00$ & $98.00$ \\
\cmidrule(lr){2-13}
 & \multirow{1}{*}{$\mathcal{A}^{LP}_{f}(\%)$} & Linear Probe & $92.00$ & $72.00$ & $89.00$ & $78.00$ & $90.00$ & $69.00$ & $90.00$ & $91.00$ & $80.00$ & $89.00$ \\
\cmidrule(lr){2-13}
 & \multirow{2}{*}{$\mathrm{RS}$} & PRA \cite{ha2025unlearning} & $0.99$ & $0.92$ & $0.98$ & $0.94$ & $0.98$ & $0.94$ & $0.97$ & $0.99$ & $0.91$ & $0.97$ \\
 &  & SFRA (ours) & $0.98$ & $0.91$ & $0.98$ & $0.91$ & $0.96$ & $0.96$ & $0.94$ & $0.98$ & $0.94$ & $0.97$ \\
\cmidrule(lr){2-13}
 & \multirow{2}{*}{$\Delta\mathrm{RS}$} & PRA \cite{ha2025unlearning} & $+0.02$ & $+0.17$ & $+0.09$ & $+0.17$ & $+0.07$ & $+0.17$ & $+0.33$ & $+0.23$ & $+0.02$ & $+0.05$ \\
 &  & SFRA (ours) & $+0.90$ & $+0.46$ & $+0.53$ & $+0.79$ & $+0.13$ & $+0.44$ & $+0.51$ & $+0.60$ & $+0.29$ & $+0.77$ \\
\bottomrule
\end{tabular}
}
\end{minipage}%
\hfill
\begin{minipage}[t]{0.49\textwidth}
\centering
\scriptsize
\setlength{\tabcolsep}{2pt}
\renewcommand{\arraystretch}{0.80}
\resizebox{\columnwidth}{!}{%
\begin{tabular}{l|l|l|cccccccccc}
\toprule
\multirow{2}{*}{Unlearning Method} & \multirow{2}{*}{Metric} & \multirow{2}{*}{Variant} & \multicolumn{10}{c}{Forget Class} \\
 & & & 0 & 10 & 20 & 30 & 40 & 50 & 60 & 70 & 80 & 90 \\
\midrule
\multirow{11}{*}{Learn to Unlearn \cite{cha2024learning}} & \multirow{3}{*}{$\mathcal{A}^{t}_{r}(\%)$} & Unlearned & $87.36$ & $86.29$ & $85.71$ & $87.69$ & $84.59$ & $86.58$ & $86.86$ & $85.29$ & $85.80$ & $86.38$ \\
 &  & PRA \cite{ha2025unlearning} & $87.36$ & $86.27$ & $85.34$ & $86.70$ & $84.21$ & $85.99$ & $86.01$ & $85.11$ & $84.18$ & $86.23$ \\
 &  & SFRA (ours) & $85.77$ & $80.59$ & $77.27$ & $81.36$ & $77.53$ & $78.27$ & $83.22$ & $81.14$ & $77.69$ & $81.17$ \\
\cmidrule(lr){2-13}
 & \multirow{3}{*}{$\mathcal{A}^{t}_{f}(\%)$} & Unlearned & $0.00$ & $0.00$ & $0.00$ & $0.00$ & $0.00$ & $0.00$ & $0.00$ & $0.00$ & $0.00$ & $0.00$ \\
 &  & PRA \cite{ha2025unlearning} & $59.00$ & $16.00$ & $82.00$ & $74.00$ & $55.00$ & $69.00$ & $78.00$ & $25.00$ & $77.00$ & $31.00$ \\
 &  & SFRA (ours) & $5.00$ & $41.00$ & $50.00$ & $31.00$ & $65.00$ & $92.00$ & $49.00$ & $9.00$ & $89.00$ & $13.00$ \\
\cmidrule(lr){2-13}
 & \multirow{1}{*}{$\mathcal{A}^{LP}_{f}(\%)$} & Linear Probe & $91.00$ & $68.00$ & $83.00$ & $69.00$ & $82.00$ & $68.00$ & $87.00$ & $82.00$ & $73.00$ & $85.00$ \\
\cmidrule(lr){2-13}
 & \multirow{2}{*}{$\mathrm{RS}$} & PRA \cite{ha2025unlearning} & $0.74$ & $0.28$ & $0.90$ & $0.85$ & $0.71$ & $0.81$ & $0.87$ & $0.40$ & $0.86$ & $0.47$ \\
 &  & SFRA (ours) & $0.10$ & $0.57$ & $0.65$ & $0.47$ & $0.76$ & $0.92$ & $0.65$ & $0.16$ & $0.90$ & $0.23$ \\
\cmidrule(lr){2-13}
 & \multirow{2}{*}{$\Delta\mathrm{RS}$} & PRA \cite{ha2025unlearning} & $-0.23$ & $-0.47$ & $+0.01$ & $+0.08$ & $-0.20$ & $+0.04$ & $+0.23$ & $-0.36$ & $-0.02$ & $-0.45$ \\
 &  & SFRA (ours) & $+0.02$ & $+0.12$ & $+0.20$ & $+0.35$ & $-0.07$ & $+0.39$ & $+0.22$ & $-0.21$ & $+0.25$ & $+0.03$ \\
\midrule
\multirow{11}{*}{SCRUB \cite{kurmanji2023towards}} & \multirow{3}{*}{$\mathcal{A}^{t}_{r}(\%)$} & Unlearned & $87.14$ & $85.96$ & $85.43$ & $87.04$ & $83.79$ & $85.64$ & $86.40$ & $84.69$ & $86.33$ & $86.66$ \\
 &  & PRA \cite{ha2025unlearning} & $86.75$ & $85.95$ & $84.98$ & $87.03$ & $82.82$ & $85.02$ & $86.21$ & $84.53$ & $84.94$ & $86.55$ \\
 &  & SFRA (ours) & $87.14$ & $81.32$ & $77.01$ & $82.66$ & $77.03$ & $78.04$ & $82.56$ & $83.73$ & $79.61$ & $81.03$ \\
\cmidrule(lr){2-13}
 & \multirow{3}{*}{$\mathcal{A}^{t}_{f}(\%)$} & Unlearned & $0.00$ & $0.00$ & $0.00$ & $0.00$ & $0.00$ & $0.00$ & $0.00$ & $0.00$ & $0.00$ & $0.00$ \\
 &  & PRA \cite{ha2025unlearning} & $78.00$ & $17.00$ & $80.00$ & $17.00$ & $53.00$ & $74.00$ & $55.00$ & $23.00$ & $72.00$ & $24.00$ \\
 &  & SFRA (ours) & $0.00$ & $36.00$ & $37.00$ & $9.00$ & $59.00$ & $66.00$ & $18.00$ & $3.00$ & $76.00$ & $6.00$ \\
\cmidrule(lr){2-13}
 & \multirow{1}{*}{$\mathcal{A}^{LP}_{f}(\%)$} & Linear Probe & $94.00$ & $63.00$ & $84.00$ & $74.00$ & $79.00$ & $73.00$ & $79.00$ & $77.00$ & $72.00$ & $85.00$ \\
\cmidrule(lr){2-13}
 & \multirow{2}{*}{$\mathrm{RS}$} & PRA \cite{ha2025unlearning} & $0.87$ & $0.29$ & $0.89$ & $0.29$ & $0.69$ & $0.85$ & $0.71$ & $0.37$ & $0.83$ & $0.39$ \\
 &  & SFRA (ours) & $0.00$ & $0.52$ & $0.53$ & $0.16$ & $0.72$ & $0.77$ & $0.30$ & $0.06$ & $0.84$ & $0.11$ \\
\cmidrule(lr){2-13}
 & \multirow{2}{*}{$\Delta\mathrm{RS}$} & PRA \cite{ha2025unlearning} & $-0.09$ & $-0.46$ & $-0.01$ & $-0.47$ & $-0.22$ & $+0.08$ & $+0.07$ & $-0.38$ & $-0.05$ & $-0.53$ \\
 &  & SFRA (ours) & $-0.08$ & $+0.08$ & $+0.08$ & $+0.05$ & $-0.11$ & $+0.25$ & $-0.13$ & $-0.31$ & $+0.19$ & $-0.08$ \\
\midrule
\multirow{11}{*}{Bad Teacher \cite{chundawat2023can}} & \multirow{3}{*}{$\mathcal{A}^{t}_{r}(\%)$} & Unlearned & $88.36$ & $88.85$ & $88.48$ & $88.68$ & $88.42$ & $88.63$ & $88.61$ & $88.48$ & $88.67$ & $88.55$ \\
 &  & PRA \cite{ha2025unlearning} & $88.18$ & $88.05$ & $88.06$ & $88.01$ & $87.87$ & $87.35$ & $88.06$ & $87.88$ & $88.28$ & $88.09$ \\
 &  & SFRA (ours) & $87.60$ & $85.81$ & $87.14$ & $87.33$ & $85.30$ & $86.25$ & $86.61$ & $87.66$ & $86.47$ & $87.37$ \\
\cmidrule(lr){2-13}
 & \multirow{3}{*}{$\mathcal{A}^{t}_{f}(\%)$} & Unlearned & $0.00$ & $2.00$ & $0.00$ & $0.00$ & $0.00$ & $0.00$ & $0.00$ & $0.00$ & $0.00$ & $1.00$ \\
 &  & PRA \cite{ha2025unlearning} & $98.00$ & $93.00$ & $98.00$ & $96.00$ & $98.00$ & $96.00$ & $98.00$ & $98.00$ & $95.00$ & $100.00$ \\
 &  & SFRA (ours) & $98.00$ & $86.00$ & $96.00$ & $91.00$ & $99.00$ & $94.00$ & $93.00$ & $95.00$ & $90.00$ & $98.00$ \\
\cmidrule(lr){2-13}
 & \multirow{1}{*}{$\mathcal{A}^{LP}_{f}(\%)$} & Linear Probe & $97.00$ & $76.00$ & $91.00$ & $87.00$ & $95.00$ & $82.00$ & $92.00$ & $93.00$ & $86.00$ & $95.00$ \\
\cmidrule(lr){2-13}
 & \multirow{2}{*}{$\mathrm{RS}$} & PRA \cite{ha2025unlearning} & $0.99$ & $0.95$ & $0.99$ & $0.98$ & $0.99$ & $0.97$ & $0.99$ & $0.99$ & $0.97$ & $0.99$ \\
 &  & SFRA (ours) & $0.99$ & $0.90$ & $0.97$ & $0.95$ & $0.98$ & $0.96$ & $0.95$ & $0.97$ & $0.94$ & $0.98$ \\
\cmidrule(lr){2-13}
 & \multirow{2}{*}{$\Delta\mathrm{RS}$} & PRA \cite{ha2025unlearning} & $+0.02$ & $+0.20$ & $+0.09$ & $+0.21$ & $+0.07$ & $+0.20$ & $+0.35$ & $+0.23$ & $+0.09$ & $+0.07$ \\
 &  & SFRA (ours) & $+0.91$ & $+0.45$ & $+0.53$ & $+0.83$ & $+0.15$ & $+0.43$ & $+0.52$ & $+0.60$ & $+0.29$ & $+0.78$ \\
\midrule
\multirow{11}{*}{SalUn \cite{fan2023salun}} & \multirow{3}{*}{$\mathcal{A}^{t}_{r}(\%)$} & Unlearned & $87.88$ & $87.66$ & $88.55$ & $87.77$ & $88.27$ & $87.27$ & $88.52$ & $87.05$ & $88.36$ & $87.57$ \\
 &  & PRA \cite{ha2025unlearning} & $87.88$ & $87.62$ & $88.55$ & $87.75$ & $88.24$ & $87.10$ & $88.51$ & $86.94$ & $88.35$ & $87.52$ \\
 &  & SFRA (ours) & $87.34$ & $87.12$ & $88.00$ & $87.27$ & $87.39$ & $86.63$ & $88.10$ & $86.37$ & $87.56$ & $87.13$ \\
\cmidrule(lr){2-13}
 & \multirow{3}{*}{$\mathcal{A}^{t}_{f}(\%)$} & Unlearned & $0.00$ & $0.00$ & $0.00$ & $0.00$ & $0.00$ & $0.00$ & $0.00$ & $0.00$ & $0.00$ & $0.00$ \\
 &  & PRA \cite{ha2025unlearning} & $48.00$ & $54.00$ & $36.00$ & $57.00$ & $25.00$ & $65.00$ & $27.00$ & $64.00$ & $25.00$ & $79.00$ \\
 &  & SFRA (ours) & $47.00$ & $44.00$ & $18.00$ & $68.00$ & $8.00$ & $49.00$ & $15.00$ & $69.00$ & $17.00$ & $62.00$ \\
\cmidrule(lr){2-13}
 & \multirow{1}{*}{$\mathcal{A}^{LP}_{f}(\%)$} & Linear Probe & $97.00$ & $76.00$ & $91.00$ & $81.00$ & $89.00$ & $79.00$ & $94.00$ & $84.00$ & $93.00$ & $93.00$ \\
\cmidrule(lr){2-13}
 & \multirow{2}{*}{$\mathrm{RS}$} & PRA \cite{ha2025unlearning} & $0.65$ & $0.70$ & $0.53$ & $0.73$ & $0.40$ & $0.79$ & $0.43$ & $0.78$ & $0.40$ & $0.88$ \\
 &  & SFRA (ours) & $0.64$ & $0.61$ & $0.30$ & $0.81$ & $0.15$ & $0.66$ & $0.26$ & $0.81$ & $0.29$ & $0.76$ \\
\cmidrule(lr){2-13}
 & \multirow{2}{*}{$\Delta\mathrm{RS}$} & PRA \cite{ha2025unlearning} & $-0.32$ & $-0.05$ & $-0.36$ & $-0.04$ & $-0.51$ & $+0.02$ & $-0.21$ & $+0.02$ & $-0.48$ & $-0.04$ \\
 &  & SFRA (ours) & $+0.56$ & $+0.16$ & $-0.14$ & $+0.69$ & $-0.68$ & $+0.13$ & $-0.17$ & $+0.44$ & $-0.36$ & $+0.57$ \\
\midrule
\multirow{11}{*}{DELETE \cite{zhou2025decoupled}} & \multirow{3}{*}{$\mathcal{A}^{t}_{r}(\%)$} & Unlearned & $87.83$ & $88.05$ & $87.65$ & $86.89$ & $87.94$ & $87.23$ & $87.92$ & $87.74$ & $87.13$ & $88.03$ \\
 &  & PRA \cite{ha2025unlearning} & $87.06$ & $87.17$ & $87.11$ & $85.21$ & $87.34$ & $86.31$ & $86.92$ & $86.75$ & $86.04$ & $86.93$ \\
 &  & SFRA (ours) & $84.30$ & $79.99$ & $80.81$ & $82.17$ & $82.39$ & $79.03$ & $81.68$ & $84.66$ & $78.66$ & $86.03$ \\
\cmidrule(lr){2-13}
 & \multirow{3}{*}{$\mathcal{A}^{t}_{f}(\%)$} & Unlearned & $0.00$ & $0.00$ & $0.00$ & $0.00$ & $0.00$ & $0.00$ & $0.00$ & $0.00$ & $0.00$ & $0.00$ \\
 &  & PRA \cite{ha2025unlearning} & $99.00$ & $80.00$ & $94.00$ & $90.00$ & $96.00$ & $78.00$ & $94.00$ & $98.00$ & $85.00$ & $100.00$ \\
 &  & SFRA (ours) & $100.00$ & $90.00$ & $100.00$ & $89.00$ & $100.00$ & $99.00$ & $96.00$ & $97.00$ & $99.00$ & $99.00$ \\
\cmidrule(lr){2-13}
 & \multirow{1}{*}{$\mathcal{A}^{LP}_{f}(\%)$} & Linear Probe & $93.00$ & $65.00$ & $85.00$ & $80.00$ & $85.00$ & $66.00$ & $90.00$ & $86.00$ & $75.00$ & $94.00$ \\
\cmidrule(lr){2-13}
 & \multirow{2}{*}{$\mathrm{RS}$} & PRA \cite{ha2025unlearning} & $0.99$ & $0.89$ & $0.97$ & $0.94$ & $0.98$ & $0.87$ & $0.96$ & $0.99$ & $0.91$ & $0.99$ \\
 &  & SFRA (ours) & $0.98$ & $0.91$ & $0.96$ & $0.92$ & $0.97$ & $0.95$ & $0.95$ & $0.97$ & $0.95$ & $0.98$ \\
\cmidrule(lr){2-13}
 & \multirow{2}{*}{$\Delta\mathrm{RS}$} & PRA \cite{ha2025unlearning} & $+0.02$ & $+0.14$ & $+0.07$ & $+0.18$ & $+0.06$ & $+0.10$ & $+0.33$ & $+0.23$ & $+0.03$ & $+0.07$ \\
 &  & SFRA (ours) & $+0.91$ & $+0.46$ & $+0.52$ & $+0.81$ & $+0.14$ & $+0.43$ & $+0.51$ & $+0.60$ & $+0.30$ & $+0.79$ \\
\bottomrule
\end{tabular}
}
\end{minipage}%

\end{table*}

\begin{table*}[t]
\centering
\caption{Per-forget-class single-class unlearning and relearning results on TinyImageNet using Swin-T. We report the unlearned checkpoint, the source-dependent PRA baseline, our proposed SFRA, and frozen-encoder linear probing. Each forget class column corresponds to a separate unlearned checkpoint in which that class is designated for forgetting.}
\label{tab:tiny_imagenet_swin-t_variant_by_forget}
\begin{minipage}[t]{0.49\textwidth}
\centering
\scriptsize
\setlength{\tabcolsep}{2pt}
\renewcommand{\arraystretch}{0.80}
\resizebox{\columnwidth}{!}{%
\begin{tabular}{l|l|l|cccccccccc}
\toprule
\multirow{2}{*}{Unlearning Method} & \multirow{2}{*}{Metric} & \multirow{2}{*}{Variant} & \multicolumn{10}{c}{Forget Class} \\
 & & & 0 & 20 & 40 & 60 & 80 & 100 & 120 & 140 & 160 & 180 \\
\midrule
\multirow{2}{*}{Original} & \multirow{1}{*}{$\mathcal{A}^{t}_{r}(\%)$} & Original & $86.26$ & $86.30$ & $86.36$ & $86.31$ & $86.34$ & $86.34$ & $86.33$ & $86.34$ & $86.29$ & $86.34$ \\
\cmidrule(lr){2-13}
 & \multirow{1}{*}{$\mathcal{A}^{t}_{f}(\%)$} & Original & $100.00$ & $92.00$ & $80.00$ & $90.00$ & $84.00$ & $84.00$ & $86.00$ & $84.00$ & $94.00$ & $84.00$ \\
\midrule
\multirow{9}{*}{Retrained} & \multirow{3}{*}{$\mathcal{A}^{t}_{r}(\%)$} & Unlearned & $86.08$ & $86.14$ & $86.28$ & $86.10$ & $85.96$ & $86.13$ & $86.37$ & $86.19$ & $86.23$ & $86.27$ \\
 &  & PRA \cite{ha2025unlearning} & $86.18$ & $86.33$ & $85.77$ & $86.50$ & $86.09$ & $86.27$ & $86.17$ & $86.35$ & $86.40$ & $86.46$ \\
 &  & SFRA (ours) & $84.19$ & $78.88$ & $79.45$ & $78.63$ & $77.85$ & $79.45$ & $79.20$ & $78.37$ & $78.87$ & $78.65$ \\
\cmidrule(lr){2-13}
 & \multirow{3}{*}{$\mathcal{A}^{t}_{f}(\%)$} & Unlearned & $0.00$ & $0.00$ & $0.00$ & $0.00$ & $0.00$ & $0.00$ & $0.00$ & $0.00$ & $0.00$ & $0.00$ \\
 &  & PRA \cite{ha2025unlearning} & $100.00$ & $96.00$ & $82.00$ & $96.00$ & $78.00$ & $88.00$ & $82.00$ & $80.00$ & $96.00$ & $72.00$ \\
 &  & SFRA (ours) & $84.00$ & $80.00$ & $50.00$ & $76.00$ & $80.00$ & $86.00$ & $82.00$ & $78.00$ & $92.00$ & $74.00$ \\
\cmidrule(lr){2-13}
 & \multirow{1}{*}{$\mathcal{A}^{LP}_{f}(\%)$} & Linear Probe & $96.00$ & $90.00$ & $84.00$ & $90.00$ & $86.00$ & $86.00$ & $82.00$ & $88.00$ & $92.00$ & $88.00$ \\
\cmidrule(lr){2-13}
 & \multirow{2}{*}{$\mathrm{RS}$} & PRA \cite{ha2025unlearning} & $1.00$ & $0.98$ & $0.90$ & $0.98$ & $0.88$ & $0.94$ & $0.90$ & $0.89$ & $0.98$ & $0.84$ \\
 &  & SFRA (ours) & $0.91$ & $0.86$ & $0.65$ & $0.83$ & $0.86$ & $0.90$ & $0.87$ & $0.84$ & $0.92$ & $0.82$ \\
\midrule
\multirow{11}{*}{Finetune \cite{golatkar2020eternal}} & \multirow{3}{*}{$\mathcal{A}^{t}_{r}(\%)$} & Unlearned & $76.50$ & $77.16$ & $75.90$ & $77.58$ & $76.94$ & $77.16$ & $76.92$ & $77.82$ & $76.92$ & $77.24$ \\
 &  & PRA \cite{ha2025unlearning} & $76.50$ & $77.16$ & $75.90$ & $77.58$ & $76.94$ & $77.16$ & $76.90$ & $77.82$ & $76.91$ & $76.68$ \\
 &  & SFRA (ours) & $69.67$ & $70.98$ & $68.36$ & $70.67$ & $69.59$ & $70.45$ & $70.29$ & $71.51$ & $69.59$ & $70.70$ \\
\cmidrule(lr){2-13}
 & \multirow{3}{*}{$\mathcal{A}^{t}_{f}(\%)$} & Unlearned & $0.00$ & $0.00$ & $0.00$ & $0.00$ & $0.00$ & $0.00$ & $0.00$ & $0.00$ & $0.00$ & $0.00$ \\
 &  & PRA \cite{ha2025unlearning} & $18.00$ & $14.00$ & $4.00$ & $22.00$ & $6.00$ & $0.00$ & $32.00$ & $18.00$ & $4.00$ & $6.00$ \\
 &  & SFRA (ours) & $62.00$ & $56.00$ & $58.00$ & $56.00$ & $64.00$ & $36.00$ & $32.00$ & $68.00$ & $60.00$ & $64.00$ \\
\cmidrule(lr){2-13}
 & \multirow{1}{*}{$\mathcal{A}^{LP}_{f}(\%)$} & Linear Probe & $100.00$ & $86.00$ & $80.00$ & $92.00$ & $78.00$ & $84.00$ & $82.00$ & $86.00$ & $84.00$ & $76.00$ \\
\cmidrule(lr){2-13}
 & \multirow{2}{*}{$\mathrm{RS}$} & PRA \cite{ha2025unlearning} & $0.31$ & $0.25$ & $0.08$ & $0.36$ & $0.11$ & $0.00$ & $0.48$ & $0.31$ & $0.08$ & $0.11$ \\
 &  & SFRA (ours) & $0.74$ & $0.70$ & $0.71$ & $0.70$ & $0.76$ & $0.52$ & $0.48$ & $0.79$ & $0.73$ & $0.76$ \\
\cmidrule(lr){2-13}
 & \multirow{2}{*}{$\Delta\mathrm{RS}$} & PRA \cite{ha2025unlearning} & $-0.69$ & $-0.73$ & $-0.82$ & $-0.62$ & $-0.76$ & $-0.94$ & $-0.42$ & $-0.58$ & $-0.90$ & $-0.72$ \\
 &  & SFRA (ours) & $-0.16$ & $-0.16$ & $+0.06$ & $-0.14$ & $-0.10$ & $-0.38$ & $-0.39$ & $-0.06$ & $-0.19$ & $-0.06$ \\
\midrule
\multirow{11}{*}{Negative Gradient \cite{golatkar2020eternal}} & \multirow{3}{*}{$\mathcal{A}^{t}_{r}(\%)$} & Unlearned & $82.42$ & $83.76$ & $85.53$ & $84.97$ & $85.06$ & $84.49$ & $84.15$ & $84.95$ & $85.12$ & $84.97$ \\
 &  & PRA \cite{ha2025unlearning} & $81.55$ & $83.12$ & $84.79$ & $84.70$ & $84.04$ & $83.53$ & $82.88$ & $84.69$ & $84.56$ & $84.69$ \\
 &  & SFRA (ours) & $76.70$ & $76.20$ & $77.90$ & $76.59$ & $77.38$ & $76.73$ & $76.46$ & $78.11$ & $79.03$ & $78.46$ \\
\cmidrule(lr){2-13}
 & \multirow{3}{*}{$\mathcal{A}^{t}_{f}(\%)$} & Unlearned & $0.00$ & $0.00$ & $0.00$ & $2.00$ & $2.00$ & $2.00$ & $0.00$ & $8.00$ & $0.00$ & $2.00$ \\
 &  & PRA \cite{ha2025unlearning} & $84.00$ & $88.00$ & $74.00$ & $76.00$ & $72.00$ & $66.00$ & $84.00$ & $68.00$ & $94.00$ & $52.00$ \\
 &  & SFRA (ours) & $86.00$ & $76.00$ & $86.00$ & $86.00$ & $86.00$ & $92.00$ & $80.00$ & $92.00$ & $96.00$ & $84.00$ \\
\cmidrule(lr){2-13}
 & \multirow{1}{*}{$\mathcal{A}^{LP}_{f}(\%)$} & Linear Probe & $84.00$ & $84.00$ & $74.00$ & $78.00$ & $80.00$ & $76.00$ & $76.00$ & $78.00$ & $80.00$ & $62.00$ \\
\cmidrule(lr){2-13}
 & \multirow{2}{*}{$\mathrm{RS}$} & PRA \cite{ha2025unlearning} & $0.91$ & $0.93$ & $0.85$ & $0.85$ & $0.82$ & $0.78$ & $0.91$ & $0.75$ & $0.97$ & $0.67$ \\
 &  & SFRA (ours) & $0.90$ & $0.83$ & $0.89$ & $0.88$ & $0.88$ & $0.91$ & $0.86$ & $0.88$ & $0.95$ & $0.87$ \\
\cmidrule(lr){2-13}
 & \multirow{2}{*}{$\Delta\mathrm{RS}$} & PRA \cite{ha2025unlearning} & $-0.09$ & $-0.05$ & $-0.05$ & $-0.13$ & $-0.06$ & $-0.16$ & $+0.01$ & $-0.14$ & $-0.01$ & $-0.17$ \\
 &  & SFRA (ours) & $-0.01$ & $-0.02$ & $+0.24$ & $+0.04$ & $+0.02$ & $+0.02$ & $-0.01$ & $+0.04$ & $+0.03$ & $+0.05$ \\
\midrule
\multirow{11}{*}{Negative Gradient+ \cite{kurmanji2023towards}} & \multirow{3}{*}{$\mathcal{A}^{t}_{r}(\%)$} & Unlearned & $84.92$ & $83.16$ & $85.23$ & $84.83$ & $84.40$ & $84.65$ & $85.14$ & $84.56$ & $85.54$ & $84.91$ \\
 &  & PRA \cite{ha2025unlearning} & $84.04$ & $83.16$ & $84.23$ & $83.46$ & $83.23$ & $82.40$ & $84.32$ & $83.09$ & $85.24$ & $83.30$ \\
 &  & SFRA (ours) & $83.46$ & $83.16$ & $84.26$ & $83.15$ & $82.11$ & $82.27$ & $80.67$ & $82.23$ & $79.88$ & $82.08$ \\
\cmidrule(lr){2-13}
 & \multirow{3}{*}{$\mathcal{A}^{t}_{f}(\%)$} & Unlearned & $0.00$ & $0.00$ & $0.00$ & $0.00$ & $0.00$ & $2.00$ & $0.00$ & $0.00$ & $0.00$ & $0.00$ \\
 &  & PRA \cite{ha2025unlearning} & $82.00$ & $0.00$ & $90.00$ & $82.00$ & $80.00$ & $66.00$ & $84.00$ & $74.00$ & $96.00$ & $78.00$ \\
 &  & SFRA (ours) & $38.00$ & $0.00$ & $2.00$ & $22.00$ & $28.00$ & $62.00$ & $68.00$ & $32.00$ & $88.00$ & $60.00$ \\
\cmidrule(lr){2-13}
 & \multirow{1}{*}{$\mathcal{A}^{LP}_{f}(\%)$} & Linear Probe & $84.00$ & $92.00$ & $82.00$ & $84.00$ & $76.00$ & $82.00$ & $78.00$ & $74.00$ & $82.00$ & $68.00$ \\
\cmidrule(lr){2-13}
 & \multirow{2}{*}{$\mathrm{RS}$} & PRA \cite{ha2025unlearning} & $0.90$ & $0.00$ & $0.94$ & $0.90$ & $0.88$ & $0.77$ & $0.91$ & $0.85$ & $0.98$ & $0.87$ \\
 &  & SFRA (ours) & $0.55$ & $0.00$ & $0.04$ & $0.36$ & $0.44$ & $0.74$ & $0.79$ & $0.48$ & $0.91$ & $0.74$ \\
\cmidrule(lr){2-13}
 & \multirow{2}{*}{$\Delta\mathrm{RS}$} & PRA \cite{ha2025unlearning} & $-0.10$ & $-0.98$ & $+0.04$ & $-0.08$ & $+0.01$ & $-0.16$ & $+0.01$ & $-0.04$ & $-0.00$ & $+0.03$ \\
 &  & SFRA (ours) & $-0.36$ & $-0.86$ & $-0.61$ & $-0.48$ & $-0.42$ & $-0.15$ & $-0.08$ & $-0.36$ & $-0.01$ & $-0.08$ \\
\midrule
\multirow{11}{*}{Random Label \cite{hayase2020selective}} & \multirow{3}{*}{$\mathcal{A}^{t}_{r}(\%)$} & Unlearned & $85.41$ & $85.59$ & $85.52$ & $85.05$ & $84.57$ & $84.44$ & $84.49$ & $83.71$ & $85.28$ & $83.47$ \\
 &  & PRA \cite{ha2025unlearning} & $84.45$ & $85.33$ & $84.88$ & $83.37$ & $82.98$ & $82.96$ & $83.15$ & $82.82$ & $84.59$ & $82.06$ \\
 &  & SFRA (ours) & $82.41$ & $80.47$ & $81.50$ & $77.66$ & $77.25$ & $76.49$ & $77.74$ & $77.52$ & $78.87$ & $77.14$ \\
\cmidrule(lr){2-13}
 & \multirow{3}{*}{$\mathcal{A}^{t}_{f}(\%)$} & Unlearned & $0.00$ & $0.00$ & $0.00$ & $0.00$ & $0.00$ & $0.00$ & $0.00$ & $0.00$ & $0.00$ & $0.00$ \\
 &  & PRA \cite{ha2025unlearning} & $100.00$ & $94.00$ & $92.00$ & $94.00$ & $90.00$ & $84.00$ & $84.00$ & $92.00$ & $98.00$ & $70.00$ \\
 &  & SFRA (ours) & $96.00$ & $100.00$ & $98.00$ & $86.00$ & $90.00$ & $92.00$ & $60.00$ & $92.00$ & $100.00$ & $58.00$ \\
\cmidrule(lr){2-13}
 & \multirow{1}{*}{$\mathcal{A}^{LP}_{f}(\%)$} & Linear Probe & $98.00$ & $90.00$ & $84.00$ & $88.00$ & $78.00$ & $74.00$ & $74.00$ & $84.00$ & $82.00$ & $68.00$ \\
\cmidrule(lr){2-13}
 & \multirow{2}{*}{$\mathrm{RS}$} & PRA \cite{ha2025unlearning} & $1.00$ & $0.97$ & $0.96$ & $0.96$ & $0.94$ & $0.91$ & $0.91$ & $0.95$ & $0.99$ & $0.82$ \\
 &  & SFRA (ours) & $0.96$ & $0.97$ & $0.97$ & $0.89$ & $0.91$ & $0.92$ & $0.73$ & $0.93$ & $0.97$ & $0.72$ \\
\cmidrule(lr){2-13}
 & \multirow{2}{*}{$\Delta\mathrm{RS}$} & PRA \cite{ha2025unlearning} & $-0.00$ & $-0.01$ & $+0.06$ & $-0.02$ & $+0.07$ & $-0.03$ & $+0.01$ & $+0.07$ & $+0.01$ & $-0.02$ \\
 &  & SFRA (ours) & $+0.06$ & $+0.11$ & $+0.32$ & $+0.06$ & $+0.06$ & $+0.03$ & $-0.14$ & $+0.08$ & $+0.04$ & $-0.11$ \\
\bottomrule
\end{tabular}
}
\end{minipage}%
\hfill
\begin{minipage}[t]{0.49\textwidth}
\centering
\scriptsize
\setlength{\tabcolsep}{2pt}
\renewcommand{\arraystretch}{0.80}
\resizebox{\columnwidth}{!}{%
\begin{tabular}{l|l|l|cccccccccc}
\toprule
\multirow{2}{*}{Unlearning Method} & \multirow{2}{*}{Metric} & \multirow{2}{*}{Variant} & \multicolumn{10}{c}{Forget Class} \\
 & & & 0 & 20 & 40 & 60 & 80 & 100 & 120 & 140 & 160 & 180 \\
\midrule
\multirow{11}{*}{Learn to Unlearn \cite{cha2024learning}} & \multirow{3}{*}{$\mathcal{A}^{t}_{r}(\%)$} & Unlearned & $85.20$ & $84.20$ & $84.74$ & $84.38$ & $84.57$ & $83.47$ & $82.65$ & $83.00$ & $84.09$ & $83.82$ \\
 &  & PRA \cite{ha2025unlearning} & $83.74$ & $83.26$ & $83.80$ & $83.38$ & $83.44$ & $81.88$ & $81.83$ & $81.62$ & $82.71$ & $81.86$ \\
 &  & SFRA (ours) & $85.08$ & $85.20$ & $84.84$ & $74.60$ & $84.05$ & $85.40$ & $82.44$ & $84.40$ & $83.94$ & $88.40$ \\
\cmidrule(lr){2-13}
 & \multirow{3}{*}{$\mathcal{A}^{t}_{f}(\%)$} & Unlearned & $0.00$ & $0.00$ & $0.00$ & $0.00$ & $0.00$ & $0.00$ & $0.00$ & $0.00$ & $0.00$ & $0.00$ \\
 &  & PRA \cite{ha2025unlearning} & $94.00$ & $98.00$ & $82.00$ & $82.00$ & $74.00$ & $70.00$ & $74.00$ & $82.00$ & $92.00$ & $68.00$ \\
 &  & SFRA (ours) & $54.00$ & $77.94$ & $0.00$ & $75.99$ & $18.00$ & $76.09$ & $32.00$ & $75.32$ & $6.00$ & $75.97$ \\
\cmidrule(lr){2-13}
 & \multirow{1}{*}{$\mathcal{A}^{LP}_{f}(\%)$} & Linear Probe & $86.00$ & $84.00$ & $78.00$ & $76.00$ & $78.00$ & $76.00$ & $72.00$ & $80.00$ & $84.00$ & $60.00$ \\
\cmidrule(lr){2-13}
 & \multirow{2}{*}{$\mathrm{RS}$} & PRA \cite{ha2025unlearning} & $0.96$ & $0.99$ & $0.90$ & $0.90$ & $0.85$ & $0.82$ & $0.85$ & $0.90$ & $0.95$ & $0.80$ \\
 &  & SFRA (ours) & $0.70$ & $0.88$ & $0.00$ & $0.82$ & $0.30$ & $0.86$ & $0.48$ & $0.86$ & $0.11$ & $0.86$ \\
\cmidrule(lr){2-13}
 & \multirow{2}{*}{$\Delta\mathrm{RS}$} & PRA \cite{ha2025unlearning} & $-0.04$ & $+0.01$ & $-0.00$ & $-0.08$ & $-0.03$ & $-0.12$ & $-0.05$ & $+0.01$ & $-0.03$ & $-0.03$ \\
 &  & SFRA (ours) & $-0.20$ & $+0.02$ & $-0.65$ & $-0.01$ & $-0.55$ & $-0.03$ & $-0.39$ & $+0.01$ & $-0.81$ & $+0.04$ \\
\midrule
\multirow{11}{*}{SCRUB \cite{kurmanji2023towards}} & \multirow{3}{*}{$\mathcal{A}^{t}_{r}(\%)$} & Unlearned & $82.88$ & $83.98$ & $84.09$ & $83.62$ & $83.49$ & $82.29$ & $83.47$ & $82.98$ & $84.43$ & $84.52$ \\
 &  & PRA \cite{ha2025unlearning} & $81.80$ & $83.12$ & $83.11$ & $82.58$ & $82.12$ & $81.15$ & $82.78$ & $82.50$ & $83.46$ & $84.30$ \\
 &  & SFRA (ours) & $76.50$ & $76.56$ & $76.00$ & $75.51$ & $75.73$ & $74.10$ & $75.34$ & $76.76$ & $77.94$ & $78.92$ \\
\cmidrule(lr){2-13}
 & \multirow{3}{*}{$\mathcal{A}^{t}_{f}(\%)$} & Unlearned & $0.00$ & $2.00$ & $0.00$ & $0.00$ & $0.00$ & $0.00$ & $0.00$ & $0.00$ & $0.00$ & $0.00$ \\
 &  & PRA \cite{ha2025unlearning} & $80.00$ & $94.00$ & $86.00$ & $74.00$ & $68.00$ & $50.00$ & $70.00$ & $54.00$ & $80.00$ & $38.00$ \\
 &  & SFRA (ours) & $62.00$ & $18.00$ & $14.00$ & $58.00$ & $80.00$ & $86.00$ & $78.00$ & $76.00$ & $78.00$ & $82.00$ \\
\cmidrule(lr){2-13}
 & \multirow{1}{*}{$\mathcal{A}^{LP}_{f}(\%)$} & Linear Probe & $72.00$ & $86.00$ & $76.00$ & $74.00$ & $70.00$ & $66.00$ & $76.00$ & $76.00$ & $78.00$ & $64.00$ \\
\cmidrule(lr){2-13}
 & \multirow{2}{*}{$\mathrm{RS}$} & PRA \cite{ha2025unlearning} & $0.88$ & $0.95$ & $0.92$ & $0.85$ & $0.81$ & $0.66$ & $0.82$ & $0.70$ & $0.89$ & $0.55$ \\
 &  & SFRA (ours) & $0.75$ & $0.27$ & $0.24$ & $0.71$ & $0.86$ & $0.89$ & $0.84$ & $0.84$ & $0.85$ & $0.88$ \\
\cmidrule(lr){2-13}
 & \multirow{2}{*}{$\Delta\mathrm{RS}$} & PRA \cite{ha2025unlearning} & $-0.12$ & $-0.03$ & $+0.02$ & $-0.13$ & $-0.07$ & $-0.27$ & $-0.08$ & $-0.19$ & $-0.09$ & $-0.29$ \\
 &  & SFRA (ours) & $-0.16$ & $-0.59$ & $-0.41$ & $-0.12$ & $+0.00$ & $-0.01$ & $-0.03$ & $-0.01$ & $-0.07$ & $+0.06$ \\
\midrule
\multirow{11}{*}{Bad Teacher \cite{chundawat2023can}} & \multirow{3}{*}{$\mathcal{A}^{t}_{r}(\%)$} & Unlearned & $86.02$ & $85.77$ & $85.96$ & $85.93$ & $85.86$ & $85.74$ & $86.00$ & $85.87$ & $85.96$ & $86.01$ \\
 &  & PRA \cite{ha2025unlearning} & $85.99$ & $85.72$ & $85.66$ & $85.81$ & $85.59$ & $85.59$ & $85.71$ & $85.65$ & $85.82$ & $85.84$ \\
 &  & SFRA (ours) & $83.56$ & $84.42$ & $79.18$ & $82.65$ & $78.76$ & $82.85$ & $78.02$ & $82.51$ & $79.12$ & $83.17$ \\
\cmidrule(lr){2-13}
 & \multirow{3}{*}{$\mathcal{A}^{t}_{f}(\%)$} & Unlearned & $0.00$ & $0.00$ & $0.00$ & $0.00$ & $0.00$ & $4.00$ & $0.00$ & $0.00$ & $0.00$ & $0.00$ \\
 &  & PRA \cite{ha2025unlearning} & $100.00$ & $98.00$ & $94.00$ & $96.00$ & $94.00$ & $90.00$ & $96.00$ & $90.00$ & $96.00$ & $88.00$ \\
 &  & SFRA (ours) & $100.00$ & $98.00$ & $98.00$ & $100.00$ & $100.00$ & $96.00$ & $98.00$ & $98.00$ & $100.00$ & $96.00$ \\
\cmidrule(lr){2-13}
 & \multirow{1}{*}{$\mathcal{A}^{LP}_{f}(\%)$} & Linear Probe & $98.00$ & $94.00$ & $84.00$ & $92.00$ & $86.00$ & $84.00$ & $84.00$ & $90.00$ & $92.00$ & $86.00$ \\
\cmidrule(lr){2-13}
 & \multirow{2}{*}{$\mathrm{RS}$} & PRA \cite{ha2025unlearning} & $1.00$ & $0.99$ & $0.97$ & $0.98$ & $0.97$ & $0.92$ & $0.98$ & $0.95$ & $0.98$ & $0.94$ \\
 &  & SFRA (ours) & $0.99$ & $0.98$ & $0.96$ & $0.98$ & $0.96$ & $0.94$ & $0.95$ & $0.97$ & $0.96$ & $0.97$ \\
\cmidrule(lr){2-13}
 & \multirow{2}{*}{$\Delta\mathrm{RS}$} & PRA \cite{ha2025unlearning} & $+0.00$ & $+0.01$ & $+0.07$ & $+0.00$ & $+0.09$ & $-0.01$ & $+0.08$ & $+0.06$ & $-0.00$ & $+0.10$ \\
 &  & SFRA (ours) & $+0.08$ & $+0.12$ & $+0.30$ & $+0.15$ & $+0.11$ & $+0.05$ & $+0.08$ & $+0.13$ & $+0.04$ & $+0.14$ \\
\midrule
\multirow{11}{*}{SalUn \cite{fan2023salun}} & \multirow{3}{*}{$\mathcal{A}^{t}_{r}(\%)$} & Unlearned & $86.02$ & $82.44$ & $85.91$ & $84.29$ & $86.23$ & $84.97$ & $86.19$ & $83.08$ & $86.08$ & $83.80$ \\
 &  & PRA \cite{ha2025unlearning} & $85.98$ & $81.08$ & $85.79$ & $82.28$ & $86.09$ & $83.08$ & $85.98$ & $81.51$ & $86.03$ & $81.20$ \\
 &  & SFRA (ours) & $85.40$ & $75.09$ & $85.02$ & $76.94$ & $85.58$ & $78.80$ & $85.69$ & $75.16$ & $85.41$ & $76.75$ \\
\cmidrule(lr){2-13}
 & \multirow{3}{*}{$\mathcal{A}^{t}_{f}(\%)$} & Unlearned & $0.00$ & $2.00$ & $0.00$ & $2.00$ & $0.00$ & $0.00$ & $0.00$ & $6.00$ & $0.00$ & $0.00$ \\
 &  & PRA \cite{ha2025unlearning} & $100.00$ & $98.00$ & $82.00$ & $94.00$ & $76.00$ & $88.00$ & $92.00$ & $92.00$ & $94.00$ & $86.00$ \\
 &  & SFRA (ours) & $94.00$ & $90.00$ & $74.00$ & $88.00$ & $54.00$ & $94.00$ & $70.00$ & $94.00$ & $84.00$ & $86.00$ \\
\cmidrule(lr){2-13}
 & \multirow{1}{*}{$\mathcal{A}^{LP}_{f}(\%)$} & Linear Probe & $100.00$ & $90.00$ & $86.00$ & $84.00$ & $82.00$ & $76.00$ & $84.00$ & $82.00$ & $92.00$ & $68.00$ \\
\cmidrule(lr){2-13}
 & \multirow{2}{*}{$\mathrm{RS}$} & PRA \cite{ha2025unlearning} & $1.00$ & $0.97$ & $0.90$ & $0.95$ & $0.86$ & $0.93$ & $0.96$ & $0.92$ & $0.97$ & $0.91$ \\
 &  & SFRA (ours) & $0.97$ & $0.90$ & $0.85$ & $0.89$ & $0.70$ & $0.94$ & $0.82$ & $0.90$ & $0.91$ & $0.89$ \\
\cmidrule(lr){2-13}
 & \multirow{2}{*}{$\Delta\mathrm{RS}$} & PRA \cite{ha2025unlearning} & $+0.00$ & $-0.01$ & $+0.00$ & $-0.03$ & $-0.01$ & $-0.01$ & $+0.06$ & $+0.03$ & $-0.01$ & $+0.08$ \\
 &  & SFRA (ours) & $+0.06$ & $+0.04$ & $+0.20$ & $+0.06$ & $-0.16$ & $+0.04$ & $-0.05$ & $+0.05$ & $-0.01$ & $+0.07$ \\
\midrule
\multirow{11}{*}{DELETE \cite{zhou2025decoupled}} & \multirow{3}{*}{$\mathcal{A}^{t}_{r}(\%)$} & Unlearned & $85.77$ & $85.99$ & $86.00$ & $85.79$ & $85.89$ & $85.06$ & $85.93$ & $85.52$ & $85.75$ & $85.91$ \\
 &  & PRA \cite{ha2025unlearning} & $85.69$ & $85.95$ & $85.41$ & $85.72$ & $85.25$ & $82.86$ & $85.25$ & $84.65$ & $85.32$ & $85.77$ \\
 &  & SFRA (ours) & $83.77$ & $82.13$ & $78.20$ & $78.57$ & $79.40$ & $78.13$ & $77.96$ & $77.75$ & $81.41$ & $77.97$ \\
\cmidrule(lr){2-13}
 & \multirow{3}{*}{$\mathcal{A}^{t}_{f}(\%)$} & Unlearned & $0.00$ & $0.00$ & $0.00$ & $26.00$ & $2.00$ & $0.00$ & $0.00$ & $28.00$ & $0.00$ & $0.00$ \\
 &  & PRA \cite{ha2025unlearning} & $100.00$ & $96.00$ & $88.00$ & $88.00$ & $84.00$ & $92.00$ & $90.00$ & $92.00$ & $98.00$ & $80.00$ \\
 &  & SFRA (ours) & $100.00$ & $100.00$ & $98.00$ & $98.00$ & $92.00$ & $94.00$ & $96.00$ & $98.00$ & $98.00$ & $92.00$ \\
\cmidrule(lr){2-13}
 & \multirow{1}{*}{$\mathcal{A}^{LP}_{f}(\%)$} & Linear Probe & $98.00$ & $92.00$ & $82.00$ & $82.00$ & $84.00$ & $76.00$ & $82.00$ & $84.00$ & $86.00$ & $78.00$ \\
\cmidrule(lr){2-13}
 & \multirow{2}{*}{$\mathrm{RS}$} & PRA \cite{ha2025unlearning} & $1.00$ & $0.98$ & $0.93$ & $0.77$ & $0.90$ & $0.95$ & $0.94$ & $0.78$ & $0.99$ & $0.89$ \\
 &  & SFRA (ours) & $0.99$ & $0.98$ & $0.95$ & $0.81$ & $0.92$ & $0.94$ & $0.94$ & $0.80$ & $0.97$ & $0.92$ \\
\cmidrule(lr){2-13}
 & \multirow{2}{*}{$\Delta\mathrm{RS}$} & PRA \cite{ha2025unlearning} & $-0.00$ & $-0.00$ & $+0.03$ & $-0.21$ & $+0.02$ & $+0.01$ & $+0.04$ & $-0.11$ & $+0.01$ & $+0.05$ \\
 &  & SFRA (ours) & $+0.08$ & $+0.12$ & $+0.30$ & $-0.02$ & $+0.06$ & $+0.04$ & $+0.07$ & $-0.05$ & $+0.04$ & $+0.10$ \\
\bottomrule
\end{tabular}
}
\end{minipage}%

\end{table*}

%\clearpage
\begin{comment}
{
    \small
    \bibliographystyle{ieeenat_fullname}
    \bibliography{main}
}
\end{comment}

\end{document}